\documentclass[format=a4-sdq]{sdqdiss} 

\usepackage[T1]{fontenc}
\usepackage[utf8]{inputenc}
\usepackage{todonotes}
\usepackage{epigraph} 
\usepackage{booktabs,color} 
\usepackage{array}
\usepackage{multirow}
\usepackage{xcolor}
\usepackage{tikz}
\usetikzlibrary{bayesnet}
\usepackage[boxed]{algorithm2e}
\usepackage{wrapfig}
\usepackage{bm,lipsum,booktabs}
\usepackage[bookmarks=false]{hyperref}
\usetikzlibrary{arrows.meta}
\usepackage{caption}
\usepackage{float}
\usepackage{pdfpages}
\usepackage{subcaption}
\usepackage{adjustbox}
\usepackage{longtable}
\usepackage{geometry}   
\usepackage{pifont}
\usepackage{enumitem} 
\usepackage{tabu}
\makeatletter
\renewcommand{\@chapapp}{}
\newenvironment{chapquote}[2][2em]
  {\setlength{\@tempdima}{#1}%
   \def\chapquote@author{#2}%
   \parshape 1 \@tempdima \dimexpr\textwidth-2\@tempdima\relax%
   \itshape}
  {\par\normalfont\hfill--\ \chapquote@author\hspace*{\@tempdima}\par\bigskip}
\makeatother

\newcommand{\xmark}{\text{\ding{55}}}

\usepackage[ngerman,english]{babel}
\usepackage[german=guillemets]{csquotes}

\newcommand{\cvec}[1]{\boldsymbol{#1}}
\newcommand{\cmat}[1]{\mathbf{#1}}
\newcommand{\tikzTaskinfer}
{ \begin{tikzpicture}[
thick,
node distance = 0.3cm and 0.3cm,
invisible_node/.style={draw=none,
                       minimum size=0.6cm},
]

\node[obs,minimum size=22pt] (r) {$r_{i}$}; 
\node[latent,right=of r, yshift=0cm,xshift=.5cm, minimum size=22pt] (l) { $\ell$};
     
\plate {}{(r)}{$N$};
\edge {l}{r}

     \end{tikzpicture}
}

\newcommand{\tikzsimpleSSM}{
\begin{tikzpicture}[
 thick,
node distance = 0.5cm and 0.5cm,
minimum size=0.8cm,
invisible_node/.style={draw=none},
]

\node[latent]                                (z0) {${z_{1}}$};
\node[latent, right=of z0, xshift=0.7cm]                   (z1) {${z_{2}}$};
\node[latent, right=of z1, xshift=0.7cm]                   (z2) {${z_{3}}$};
\node[invisible_node, right=of z2, xshift=0.3cm]  (zf) {};

\node[obs, below=of z0]                      (o0) {${w_{1}}$};
\node[obs, below=of z1]                      (o1) {${w_{2}}$};
\node[obs, below=of z2]                      (o2) {${w_{3}}$};

\edge{z0}{z1}
\edge{z1}{z2}
\edge{z0}{o0}
\edge{z1}{o1}
\edge{z2}{o2}

\edge[dotted]{z2}{zf}
\end{tikzpicture}
 }

 \newcommand{\tikzBA}{
\begin{tikzpicture}[
thick,
node distance = 0.3cm and 0.3cm,
invisible_node/.style={draw=none,
                       minimum size=0.6cm},
]

\node[obs, minimum size=1cm]                      (r)      {$ r_{n}$};
\node[obs, minimum size=1cm, left = of r, xshift=0.3cm]      (o_t)  {$o_{n}$};

\node[latent, minimum size=1cm, right= of r, xshift=0.5cm]         (l)      {$l$};

\edge[-{Triangle[open]}] {o_t}{r};

\edge{l}{r};

\node[invisible_node] (inv)    [above = of o_t, xshift=2.1cm, yshift=-0.15cm]                {$N$};
\draw[draw=black, rounded corners] (-2.5, 1.3) rectangle ++(3.5,-3);

\end{tikzpicture}
}
\tikzset{%
  dashedarrow/.style = {%
    draw, > = {Latex[width = 2.5mm, length = 3.5mm, open, fill = white]}, 
  }%
}%

\newcommand{\tikzACRKN}{
\begin{tikzpicture}[thick]
		\tikzset{invisible/.style={draw=none},
	}

    \draw[orange] (0.25, 2) -- (1.75, 2) -- (2.25, 3.0) -- (-0.25, 3.0) -- (0.25, 2.0)  node[black, pos=.5, xshift=27.5]{Encoder};
    \draw[->, >=stealth] (1.0, 2.0) -- (1.0, 0.85) node[pos=.35, xshift=5]{$\vec{w}_t ~~ \vec{\sigma}^\textrm{obs}_t$};
    \draw[->, >=stealth] (1.0, 3.5) -- (1.0, 3.0) node[pos=.2, xshift=5]{$~~\vec{o}_t$};
    
    \draw[->, >=stealth] (4.5, 2.5) -- (4.5, 0.85) node[pos=.2, xshift=-10, align=left]{$\vec{a}_t$};
    
	\draw[densely dotted, black!60!green] (-0.25, -0.4)   rectangle  (5.75, 1.3) node[black, pos=0.1, xshift=10]{RKN Cell};
	\draw (3.5, 0.15) rectangle (5.5, 0.85) node[pos=.5, align=center]{Predict};
	\draw (0.0, 0.15) rectangle (2.0, 0.85) node[pos=.5, align=center]{Update}; 
    \draw[->, >=stealth] (-1.5, 0.5) -- (0.0, 0.5) node[pos=.25, yshift=11]{$\vec{z}_t^-$, $\vec{\Sigma}_t^-$};
    \draw[->, >=stealth] (-1.5, 0.5) -- (0.0, 0.5) node[pos=.25, yshift=-8]{\small{Latent State}};
    \draw[->, >=stealth] (2.0, 0.5) -- (3.5, 0.5) node[pos=.5, yshift=11]{$\vec{z}_t^+$, $\vec{\Sigma}_t^+$};
    \draw[->, >=stealth] (5.5, 0.5) -- (7.0, 0.5) node[pos=.75, yshift=11]{$\vec{z}_{t+1}^-$, $\vec{\Sigma}_{t+1}^-$};

    \draw[->, >=stealth] (6.5, 0.5) -- (6.5, -1.0); 
    
    \draw[blue] (5.75, -1.0) -- (7.25, -1.0) -- (7.75, -2.0) -- (5.25, -2.0) -- (5.75, -1.0) node[black, pos=.5, xshift=27.5]{Decoder};

    \draw[->, >=stealth] (6.5, -2.0) -- (6.5, -2.5) node[pos=0.6, xshift=13]{$\hat{\vec{o}}_{t+1}$}; 
  \draw[orange, decorate, decoration={brace, amplitude=1ex, raise=1ex}]
  (-0.2, 1.5) -- (-0.2, 3.2) node[pos=.5, xshift=-75.5] {\shortstack{Bottom Up\\Sensory Signals}};
  \draw[decoration={brace, mirror}, decorate, blue]
  (7.5, -1.0) -- (7.5, 1.0) node[pos=.5, xshift=60] {\shortstack{Top Down\\Predictions}};
\end{tikzpicture}
}

\newcommand{\tikzRKNGM}{ \begin{tikzpicture}[very thick,scale=1.0, every node/.style={scale=1.0}]
    \node[latent, fill, minimum size=42pt] (s1) {\LARGE$\vec{z}_t$}; %
    \node[latent,right=of s1, minimum size=42pt, xshift=2cm] (s2) {\LARGE$\vec{z}_{t+1}$};
    \coordinate[left of=s1,xshift=-1cm] (c1);
    \coordinate[right of=s2,xshift=1cm] (c2);

     \node[latent,below=of s1, fill,minimum size=42pt] (h1) {\LARGE$\vec{w}_t$}; 
     \node[latent,below=of s2, fill,minimum size=42pt] (h2) {\LARGE$\vec{w}_{t+1}$};

     \node[obs,below=of h1, xshift=-1cm,yshift=0cm,minimum size=42pt] (o1) {\LARGE$\vec{o}_{t}$};
     \node[obs,below=of h2,xshift=-1cm,yshift=0cm,minimum size=42pt] (o2) {\LARGE$\vec{o}_{t+1}$};
    
     \node[obs,below=of h1, minimum size=42pt, fill={rgb:red,0;green,1;white,3}, xshift=1cm
     ] (a1) {\LARGE$\vec{a}_{t}$};%
     \node[obs,below=of h2, minimum size=42pt,xshift=1cm, fill={rgb:red,0;green,1;white,3}
     ] (a2) {\LARGE$\vec{a}_{t+1}$};%

     \edge{s1} {h1}
     \edge[-{Triangle[open]}]{o1}{h1};
     \edge[-{Triangle[open]}]{a1}{h1};
     \edge {s1} {s2}
     \edge {s2} {h2}
     \edge[-{Triangle[open]}]{o2}{h2};
     \edge[-{Triangle[open]}]{a2}{h2};
     \edge[dashed] {s2} {c2}
     \edge[dashed] {c1} {s1}
     \end{tikzpicture}
}

\newcommand{\tikzACRKNGM}{ \begin{tikzpicture}[very thick,scale=1.0, every node/.style={scale=1.0}]
    \node[latent, fill, minimum size=42pt] (s1) {\LARGE$\vec{z}_t$}; %
    \node[latent,right=of s1, minimum size=42pt, xshift=2cm] (s2) {\LARGE$\vec{z}_{t+1}$};
    \coordinate[left of=s1,xshift=-1cm] (c1);
    \coordinate[right of=s2,xshift=1cm] (c2);

     \node[latent,below=of s1, fill,minimum size=42pt] (h1) {\LARGE$\vec{w}_t$}; 
     \node[latent,below=of s2, fill,minimum size=42pt] (h2) {\LARGE$\vec{w}_{t+1}$};

     \node[obs,below=of h1, minimum size=42pt] (o1) {\LARGE$\vec{o}_{t}$};
     \node[obs,below=of h2, minimum size=42pt] (o2) {\LARGE$\vec{o}_{t+1}$};
    
    \coordinate[right of=s2,xshift=1cm,yshift=1.0cm] (c3);
    
     \node[obs,above=of s1,yshift=-0.3cm,fill={rgb:red,0;green,1;white,3},minimum size=42pt] (a1) {\LARGE $\vec{a}_t$};
     \node[obs,above=of s2,yshift=-0.3cm,fill={rgb:red,0;green,1;white,3},minimum size=42pt] (a2) {\LARGE $\vec{a}_{t+1}$};
    
     \edge {s1} {h1}
     \edge[-{Triangle[open]}]{o1}{h1};
     \edge {a1} {s2}
     \edge {s1} {s2}
     \edge {s2} {h2}
     \edge[-{Triangle[open]}]{o2}{h2};
     \edge[dashed] {s2} {c2}
     \edge[dashed] {a2} {c3}
 \edge[dashed] {c1} {s1}
 \end{tikzpicture}
 }

 \newcommand{\tikzACRKNGMForward}{ \begin{tikzpicture}[very thick,scale=1.0, every node/.style={scale=1.0}]
    \node[latent, fill, minimum size=42pt] (s1) {\LARGE$\vec{z}_1$}; %
    \node[latent,right=of s1, minimum size=42pt, xshift=1cm] (s2) {\LARGE$\vec{z}_{2}$};
    \node[latent, right=of s2,fill, minimum size=42pt, xshift=1cm] (s3) {\LARGE$\vec{z}_3$}; %
    \node[latent,right=of s3, minimum size=42pt, xshift=1cm] (s4) {\LARGE$\vec{z}_{4}$};
    \node[latent, fill, right=of s4,minimum size=42pt, xshift=1cm] (s5) {\LARGE$\vec{z}_5$}; %
    \node[latent,right=of s5, minimum size=42pt, xshift=1cm] (s6) {\LARGE$\vec{z}_{6}$};

    \coordinate[left of=s1,xshift=-1cm] (c1);
    \coordinate[right of=s2,xshift=1cm] (c2);

     \node[obs,below=of s1, fill,minimum size=42pt] (h1) {\LARGE$\vec{w}_1$}; 
     \node[obs,below=of s2, fill,minimum size=42pt] (h2) {\LARGE$\vec{w}_{2}$};

    \coordinate[right of=s2,xshift=1cm,yshift=1.0cm] (c3);
    
     \node[obs,above=of s1,yshift=-0.3cm,fill={rgb:red,0;green,1;white,3},minimum size=42pt] (a1) {\LARGE $\vec{a}_1$};
     \node[obs,above=of s2,yshift=-0.3cm,fill={rgb:red,0;green,1;white,3},minimum size=42pt] (a2) {\LARGE $\vec{a}_{2}$};
     \node[obs,above=of s3,yshift=-0.3cm,fill={rgb:red,0;green,1;white,3},minimum size=42pt] (a3) {\LARGE $\vec{a}_{3}$};
     \node[obs,above=of s4,yshift=-0.3cm,fill={rgb:red,0;green,1;white,3},minimum size=42pt] (a4) {\LARGE $\vec{a}_{4}$};
     \node[obs,above=of s5,yshift=-0.3cm,fill={rgb:red,0;green,1;white,3},minimum size=42pt] (a5) {\LARGE $\vec{a}_{5}$};

     \edge {s1} {h1}
     \edge {a1} {s2}
     \edge {a2} {s3}
     \edge {a3} {s4}
     \edge {a4} {s5}
     \edge {a5} {s6}
     \edge {s1} {s2}
     \edge {s2} {s3}
     \edge {s3} {s4}
     \edge {s4} {s5}
     \edge {s5} {s6}
     \edge {s2} {h2}

 \end{tikzpicture}
 }

 \newcommand{\tikzACRKNINVDYN}{
\begin{tikzpicture}[thick]
		\tikzset{invisible/.style={draw=none},
	}

    \draw[orange] (0.25, 2) -- (1.75, 2) -- (2.25, 3.0) -- (-0.25, 3.0) -- (0.25, 2.0)  node[black, pos=.5, xshift=27.5]{Encoder};
    \draw[->, >=stealth] (1.0, 2.0) -- (1.0, 0.85) node[pos=.35, xshift=5]{$\vec{w}_t ~~ \vec{\sigma}^\textrm{obs}_t$};
    \draw[->, >=stealth] (1.0, 3.5) -- (1.0, 3.0) node[pos=.2, xshift=5]{$~~\vec{o}_t$};
    
    \draw[->, >=stealth] (4.5, 2.5) -- (4.5, 0.85) node[pos=.2, xshift=25, align=left, black!30!red]{\small{Executed}\\\small{Action}};
    \draw[->, >=stealth] (4.5, 2.5) -- (4.5, 0.85) node[pos=.2, xshift=-10, align=left]{$\vec{a}_t$};
    
	\draw[densely dotted, black!60!green] (-0.25, -0.4)   rectangle  (5.75, 1.3) node[black, pos=0.1, xshift=10]{RKN Cell};
	\draw (3.5, 0.15) rectangle (5.5, 0.85) node[pos=.5, align=center]{Predict};
	\draw (0.0, 0.15) rectangle (2.0, 0.85) node[pos=.5, align=center]{Update}; 
    \draw[->, >=stealth] (-1.5, 0.5) -- (0.0, 0.5) node[pos=.25, yshift=11]{$\vec{z}_t^-$, $\vec{\Sigma}_t^-$};
    \draw[->, >=stealth] (-1.5, 0.5) -- (0.0, 0.5) node[pos=.25, yshift=-8]{\small{Latent State}};
    \draw[->, >=stealth] (2.0, 0.5) -- (3.5, 0.5) node[pos=.5, yshift=11]{$\vec{z}_t^+$, $\vec{\Sigma}_t^+$};
    \draw[->, >=stealth] (5.5, 0.5) -- (7.0, 0.5) node[pos=.75, yshift=11]{$\vec{z}_{t+1}^-$, $\vec{\Sigma}_{t+1}^-$};

    \draw[->, >=stealth] (2.75, 0.5) -- (2.75, -1.0);
    \draw[->, >=stealth] (6.5, 0.5) -- (6.5, -1.0); 
    
    \draw[blue] (5.75, -1.0) -- (7.25, -1.0) -- (7.75, -2.0) -- (5.25, -2.0) -- (5.75, -1.0) node[black, pos=.5, xshift=27.5]{Decoder};
    \draw[black!30!red] (2.0, -1.0) -- (3.5, -1.0) -- (4.0, -2.0) -- (1.5, -2.0) -- (2.0, -1.0) node[black, pos=.5, xshift=27.5, align=center]{Action\\Decoder};
    
    \draw[->, >=stealth] (2.75, -2.) -- (2.75, -2.5) node[pos=0.6, xshift=10]{$\hat{\vec{a}}_t$};
    
    \draw[->, >=stealth] (0, -1.5) -- (1.75, -1.5) node[black!30!red, pos=0.0, align=center]{\small{(Desired) Next}\\ \small{Observation}};
    
    \draw[->, >=stealth] (6.5, -2.0) -- (6.5, -2.5) node[pos=0.6, xshift=13]{$\hat{\vec{o}}_{t+1}$}; 
\end{tikzpicture}
}

\newcommand{\tikzHiPRSSM}{
\begin{tikzpicture}[thick]
		\tikzset{invisible/.style={draw=none},
	}

    \draw[->, >=stealth] (4.5, 2.0) -- (4.5, 1.0) node[pos=.6, xshift=0, align=left]{${\mu}_l ~~ {\sigma}_l$};
    \draw (3.5, 2.0) rectangle (5.7,2.85) node[pos=0.5, align=center, scale=0.8]{Context\\Update};
    \draw[->, >=stealth] (4.5, 3.4) -- (4.5, 2.85) node[pos=0.5,  xshift=37, align=right]{$\{{r}_n^l, ({\sigma}_{r_n}^l)^2\}_{n=1}^N$};
    \draw[black!30!red] (3.2, 4.2)--(5.8,4.2)--(5.3, 3.4)--(3.7,3.4)--cycle node[black, pos=0.5, align=center, xshift=30, scale=0.8]{Context\\Encoder};
    
    \draw[->,>=stealth] (4.5,4.7) -- (4.5,4.2) node[pos=0.1, align=right, xshift=10]{$\cmat{C}_l$};
    
	\draw[densely dotted, black!60!green] (-0.25, -0.1)   rectangle  (5.9, 1.8) node[ pos=0.9, xshift=-100]{HiP-RSSM Cell};
	
	\draw (3.5, 0.15) rectangle (5.7, 1.0) node[scale=0.8,pos=.5, align=center]{Time\\Update};
	\draw (0.0, 0.15) rectangle (2.2, 1.0) node[pos=.5, align=center, scale=0.8]{Observation\\Update}; 
    \draw[->, >=stealth] (-1.5, 0.57) -- (0.0, 0.57) node[pos=.25, yshift=11]{${z}_t^-$, $\cmat{\Sigma}_t^-$};
    \draw[->, >=stealth] (2.2, 0.57) -- (3.5, 0.57) node[pos=.5, yshift=11]{${z}_t^+$, $\cmat{\Sigma}_t^+$};
    \draw[->, >=stealth] (5.7, 0.57) -- (7.2, 0.57) node[pos=.75, yshift=11]{${z}_{t+1}^-$, $\cmat{\Sigma}_{t+1}^-$};

    \draw[->, >=stealth] (1.1, -0.6)--(1.1, 0.15)node[pos=.35, xshift=3]{${w}_t ~~ {\sigma}^\textrm{obs}_t$};
    \draw[black!30!red] (0.3, -0.6) -- (1.9, -0.6) -- (2.4, -1.4) -- (-0.2, -1.4) -- cycle node[black, pos=.5, xshift=30, align=center, scale=0.8]{Observation\\Encoder};
    \draw[->, >=stealth] (1.1, -1.9) --(1.1, -1.4) node[pos=0.5, xshift=10]{${o}_t$};
    
    \draw[->, >=stealth] (6.3, 0.57) -- (6.3, -0.6); 
    \draw[blue] (5.0, -0.6) -- (7.6, -0.6) -- (7.1, -1.4) -- (5.5, -1.4) -- cycle node[black, pos=.5, xshift=30, align=center, scale=0.8]{Decoder\\Output};
    \draw[->, >=stealth] (6.3, -1.4) -- (6.3, -1.9) node[pos=0.6, xshift=13]{$\hat{{o}}_{t+1}$}; 
    
    \draw[->, >=stealth] (4.5,-0.6) -- (4.5,0.15)node[pos=0.1, xshift=-8, align=left]{${a}_t$};

\end{tikzpicture}
}

\newcommand{\tikzHiPGMForward}{ \begin{tikzpicture}[very thick,scale=1.0, every node/.style={scale=1.0}]
    \node[latent, fill, minimum size=42pt] (s1) {\LARGE$\vec{z}_1$}; %
    \node[latent,right=of s1, minimum size=42pt, xshift=1cm] (s2) {\LARGE$\vec{z}_{2}$};
    \node[latent, right=of s2,fill, minimum size=42pt, xshift=1cm] (s3) {\LARGE$\vec{z}_3$}; %
    \node[latent,right=of s3, minimum size=42pt, xshift=1cm] (s4) {\LARGE$\vec{z}_{4}$};
    \node[latent, fill, right=of s4,minimum size=42pt, xshift=1cm] (s5) {\LARGE$\vec{z}_5$}; %
    \node[latent,right=of s5, minimum size=42pt, xshift=1cm] (s6) {\LARGE$\vec{z}_{6}$};

    \node[latent,above=of s1, xshift=1cm,yshift=1cm, fill,minimum size=42pt] (l) {\LARGE$\vec{l}$};

    \node[obs,above=of l, yshift=0.6cm, fill,minimum size=42pt] (Cl) {\LARGE$\vec{C_l}$};

    \coordinate[left of=s1,xshift=-1cm] (c1);
    \coordinate[right of=s2,xshift=1cm] (c2);

     \node[obs,below=of s1, yshift=-0.3cm, fill,minimum size=42pt] (h1) {\LARGE$\vec{w}_1$}; 
     \node[obs,below=of s2, yshift=-0.3cm,fill,minimum size=42pt] (h2) {\LARGE$\vec{w}_{2}$};

    \coordinate[right of=s2,xshift=1cm,yshift=1.0cm] (c3);
    
     \node[obs,below=of s1,xshift=1.7cm,yshift=0.5cm,fill={rgb:red,0;green,1;white,3},minimum size=42pt] (a1) {\LARGE $\vec{a}_1$};
     \node[obs,below=of s2,xshift=1.7cm,yshift=0.5cm,fill={rgb:red,0;green,1;white,3},minimum size=42pt] (a2) {\LARGE $\vec{a}_{2}$};
     \node[obs,below=of s3,xshift=1.7cm,yshift=0.5cm,fill={rgb:red,0;green,1;white,3},minimum size=42pt] (a3) {\LARGE $\vec{a}_{3}$};
     \node[obs,below=of s4,xshift=1.7cm,yshift=0.5cm,fill={rgb:red,0;green,1;white,3},minimum size=42pt] (a4) {\LARGE $\vec{a}_{4}$};
     \node[obs,below=of s5,xshift=1.7cm,yshift=0.5cm,fill={rgb:red,0;green,1;white,3},minimum size=42pt] (a5) {\LARGE $\vec{a}_{5}$};

     \edge {s1} {h1}
     \edge {a1} {s2}
     \edge {a2} {s3}
     \edge {a3} {s4}
     \edge {a4} {s5}
     \edge {a5} {s6}
     \edge {s1} {s2}
     \edge {s2} {s3}
     \edge {s3} {s4}
     \edge {s4} {s5}
     \edge {s5} {s6}
     \edge {s2} {h2}
     \edge {l} {s1}
     \edge {l} {s2}
     \edge {l} {s3}
     \edge {l} {s4}
     \edge {l} {s5}
     \edge {l} {s6}
     \edge {l}  {Cl}

 \end{tikzpicture}
 }

 \newcommand{\tikzPhilippsHiPSSM}{
\begin{tikzpicture}[
 thick,
node distance = 0.5cm and 0.5cm,
minimum size=0.8cm,
invisible_node/.style={draw=none},
]

\node[latent]                                (z0) {${z_0}$};
\node[latent, right=of z0]                   (z1) {${z_1}$};
\node[latent, right=of z1]                   (z2) {${z_2}$};
\node[invisible_node, right=of z2, xshift=0.3cm]  (zf) {};

\node[obs, below=of z0]                      (o0) {${w_0}$};
\node[obs, below=of z1]                      (o1) {${w_1}$};
\node[obs, below=of z2]                      (o2) {${w_2}$};

\node[latent, above=of z1]                   (l)  {${l}$};
\node[obs,    above=of l]                    (c)  {$\mathcal{C}_{{l}}$};

\edge{z0}{z1}
\edge{z1}{z2}
\edge{z0}{o0}
\edge{z1}{o1}
\edge{z2}{o2}

\edge{l}{z0}
\edge{l}{z1}
\edge{l}{z2}
\edge{l}{c}
\edge[dotted]{z2}{zf}
\edge[dotted]{l}{zf}

\end{tikzpicture}
 }

 \newcommand{\tikzPhilippsHipRSSM}{
\begin{tikzpicture}[
thick,
node distance = 0.3cm and 0.3cm,
invisible_node/.style={draw=none,
                       minimum size=0.6cm},
]

\node[obs, minimum size=0.9cm]                      (r)      {${r^l}$};     
\node[obs, minimum size=0.9cm, below = of r]        (a_t)    {${a}_{t}^{{l}}$};
\node[obs, minimum size=0.9cm, right = of a_t, xshift=-0.5cm]      (o_tp1)  {${o}_{t+1}^{{l}}$};
\node[obs, minimum size=0.9cm, left = of a_t, xshift=0.5cm]       (o_t)    {${o}_{t}^{{l}}$};

\node[latent, right = of o_tp1, yshift=0.1cm]       (z_0)    {${z}_0$};
\node[latent, right = of z_0, xshift=0.8cm]         (z_1)    {${z}_1$};
\node[latent, right = of z_1, xshift=0.8cm]         (z_2)    {${z}_2$};
\node[invisible_node, right = of z_2, xshift=0.8cm] (z_f)    {};

\node[obs, below = of z_0, xshift=-0.6cm]           (w_0)    {${w}_0$};
\node[obs, below = of z_1, xshift=-0.6cm]           (w_1)    {${w}_1$};
\node[obs, below = of z_2, xshift=-0.6cm]           (w_2)    {${w}_2$};

\node[obs, below = of z_0, xshift=0.6cm]            (a_0)    {${a}_0$};
\node[obs, below = of z_1, xshift=0.6cm]            (a_1)    {${a}_1$};
\node[obs, below = of z_2, xshift=0.6cm]            (a_2)    {${a}_2$};

\node[obs, below = of w_0, yshift=0.4cm]            (o_0)    {${o}_0$};
\node[obs, below = of w_1, yshift=0.4cm]            (o_1)    {${o}_1$};
\node[obs, below = of w_2, yshift=0.4cm]            (o_2)    {${o}_2$};

\node[latent, above = of z_0, yshift=0.1cm]         (l)      {${l}$};

\edge[-{Triangle[open]}] {o_t}{r};
\edge[-{Triangle[open]}] {a_t}{r};
\edge[-{Triangle[open]}] {o_tp1}{r};

\edge[-{Triangle[open]}] {o_0}{w_0};
\edge[-{Triangle[open]}] {o_1}{w_1};
\edge[-{Triangle[open]}] {o_2}{w_2};

\edge{z_0}{w_0};
\edge{z_1}{w_1};
\edge{z_2}{w_2};

\edge{a_0}{z_0};
\edge{a_1}{z_1};
\edge{a_2}{z_2};

\edge{z_0}{z_1};
\edge{z_1}{z_2};
\edge[dotted]{z_2}{z_f};

\edge{l}{z_0};
\edge{l}{z_1};
\edge{l}{z_2};
\edge{l}{r};
\edge{l}{z_f};

\node[invisible_node] (inv)    [above = of o_tp1, yshift=1.1cm]                {$N$};
\draw[draw=black, rounded corners] (-2, 0.75) rectangle ++(4.0,-3.3);

\end{tikzpicture}
}

\newcommand{\tikzVaisakhMTS}{
\begin{tikzpicture}[
 thick,
node distance = 0.5cm and 0.5cm,
minimum size=0.8cm,
invisible_node/.style={draw=none},
]

\node[latent]                                (z0) {${z_{1,1}}$};
\node[latent, right=of z0, xshift=0.7cm]                   (z1) {${z_{1,2}}$};
\node[latent, right=of z1, xshift=0.7cm]                   (z2) {${z_{1,3}}$};
\node[invisible_node, right=of z2, xshift=0.3cm]  (zf) {};

\node[obs, below=of z0]                      (o0) {${w_{1,1}}$};
\node[obs, below=of z1]                      (o1) {${w_{1,2}}$};
\node[obs, below=of z2]                      (o2) {${w_{1,3}}$};
\node[obs, right=of o0,yshift=0.5cm, xshift=-0.5cm, fill={rgb:red,0;green,1;white,3}]                      (a1) {${a_{1,1}}$};
\node[obs, right=of o1,yshift=0.5cm, xshift=-0.5cm, fill={rgb:red,0;green,1;white,3}]                      (a2) {${a_{1,2}}$};

\node[latent, above=of z1]                   (l0)  {${l_1}$};
\node[obs,    above=of l0, yshift=.5cm]                    (c0)  {$\beta_{{1,t}}$};
\node[invisible_node] (inv)    [above = of c0, xshift=.2cm, yshift=-0.7cm]                {$t=1..H$};
\node[latent,    above=of l0,  xshift=-1.5cm, yshift=-0.45cm,]                    (al0)  {$\alpha_{{1}}$};

\edge{z0}{z1}
\edge{z1}{z2}
\edge{z0}{o0}
\edge{a1}{z1}
\edge{a2}{z2}
\edge{z1}{o1}
\edge{z2}{o2}

\edge{l0}{z0}
\edge{l0}{z1}
\edge{l0}{z2}
\edge{l0}{c0}
\edge{al0}{l0}
\edge[dotted]{z2}{zf}

\node[latent, right=of z2,  xshift=2.5cm]                                (z0) {${z_{2,1}}$};
\node[latent, right=of z0, xshift=0.7cm]                   (z1) {${z_{2,2}}$};
\node[latent, right=of z1, xshift=0.7cm]                   (z2) {${z_{2,3}}$};
\node[invisible_node, right=of z2, xshift=0.3cm]  (zf) {};
\node[invisible_node, left=of z0, xshift=-0.3cm]  (zs) {};

\node[obs, below=of z0]                      (o0) {${w_{2,1}}$};
\node[obs, below=of z1]                      (o1) {${w_{2,2}}$};
\node[obs, below=of z2]                      (o2) {${w_{2,3}}$};
\node[obs, right=of o0,yshift=0.5cm, xshift=-0.5cm, fill={rgb:red,0;green,1;white,3}]                      (a1) {${a_{2,1}}$};
\node[obs, right=of o1,yshift=0.5cm, xshift=-0.5cm, fill={rgb:red,0;green,1;white,3}]                      (a2) {${a_{2,2}}$};

\node[latent, above=of z1]                   (l1)  {${l_2}$};
\node[obs,    above=of l1, yshift=.5cm]                    (c1)  {$\beta_{{2,t}}$};
\node[invisible_node] (inv)    [above = of c1, xshift=0.2cm, yshift=-0.7cm]                {$t=1..H$};
\draw[draw=black, rounded corners] (1.41, 4.8) rectangle ++(1.93,-1.8);
\draw[draw=black, rounded corners] (10.51, 4.8) rectangle ++(1.93,-1.8);
\draw[draw=black, rounded corners] (19.61, 4.8) rectangle ++(1.93,-1.8);
\node[latent,    above=of l1,  xshift=-1.5cm, yshift=-0.45cm,]                      (al1)  {$\alpha_{{2}}$};

\edge{z0}{z1}
\edge{z1}{z2}
\edge{z0}{o0}
\edge{a1}{z1}
\edge{a2}{z2}
\edge{z1}{o1}
\edge{z2}{o2}

\edge{l1}{z0}
\edge{l1}{z1}
\edge{l1}{z2}
\edge{l1}{c1}
\edge{al1}{l1}
\edge[dotted]{z2}{zf}
\edge[dotted]{zs}{z0}

\node[latent, right=of z2,  xshift=2.5cm]                                (z0) {${z_{3,1}}$};
\node[latent, right=of z0, xshift=0.7cm]                   (z1) {${z_{3,2}}$};
\node[latent, right=of z1, xshift=0.7cm]                   (z2) {${z_{3,3}}$};
\node[invisible_node, right=of z2, xshift=0.3cm]  (zf) {};
\node[invisible_node, left=of z0, xshift=-0.3cm]  (zs) {};

\node[obs, below=of z0]                      (o0) {${w_{3,1}}$};
\node[obs, below=of z1]                      (o1) {${w_{3,2}}$};
\node[obs, below=of z2]                      (o2) {${w_{3,3}}$};
\node[obs, right=of o0,yshift=0.5cm, xshift=-0.5cm, fill={rgb:red,0;green,1;white,3}]                      (a1) {${a_{3,1}}$};
\node[obs, right=of o1,yshift=0.5cm, xshift=-0.5cm, fill={rgb:red,0;green,1;white,3}]                      (a2) {${a_{3,2}}$};

\node[latent, above=of z1]                   (l2)  {${l_3}$};
\node[obs,    above=of l2, yshift=.5cm]                    (c2)  {$\beta_{{3,t}}$};
\node[latent,    above=of l2,  xshift=-1.5cm, yshift=-0.45cm,]                      (al2)  {$\alpha_{{3}}$};
\node[invisible_node] (inv)    [above = of c2, xshift=0.2cm, yshift=-0.7cm]                {$ t=1..H$};

\edge{z0}{z1}
\edge{z1}{z2}
\edge{z0}{o0}
\edge{a1}{z1}
\edge{a2}{z2}
\edge{z1}{o1}
\edge{z2}{o2}

\edge{l2}{z0}
\edge{l2}{z1}
\edge{l2}{z2}
\edge{l2}{c2}
\edge{al2}{l2}
\edge[dotted]{z2}{zf}
\edge[dotted]{zs}{z0}

\node[invisible_node, right=of l2, xshift=1.5cm]  (lf) {};
\edge{l0}{l1}
\edge{l1}{l2}
\edge[dotted]{l2}{lf}
\end{tikzpicture}
 }

\newcommand{\tikzActAb}{
\begin{tikzpicture}[
thick,
node distance = 0.3cm and 0.3cm,
invisible_node/.style={draw=none,
                       minimum size=0.6cm},
]

\node[obs, minimum size=1cm]                      (r)      {$\alpha_{k,t}$};
\node[obs, minimum size=1cm, below = of r, yshift=2.7cm, xshift=-1.5cm ]        (t)    {$t$};
\node[obs, minimum size=1cm, below = of r, yshift=1cm, xshift=-1.5cm]      (a_t)  {$a_{t,k}$};

\node[latent, minimum size=1cm, right= of r, xshift=0.5cm]         (l)      {$\alpha_k$};

\edge[-{Triangle[open]}] {a_t}{r};
\edge[-{Triangle[open]}] {t}{r};

\edge{l}{r};

\node[invisible_node] (inv)    [above = of a_t, xshift=1.7cm, yshift=1cm]                {$N$};
\draw[draw=black, rounded corners] (-2.2, 1.3) rectangle ++(3.2,-3);

\end{tikzpicture}
}
\newcommand{\Abstract}[1][Abstract]{\chapter*{#1}\addcontentsline{toc}{chapter}{#1}\markboth{#1}{#1}} 

\usepackage{blindtext}
\usepackage{subfiles}
\usepackage{awesomebox}
\usepackage{tcolorbox}

\usepackage{amsthm}
\usepackage{tikz}
\usepackage{thmtools}
\usepackage{thm-restate}
\usetikzlibrary{decorations.pathreplacing}

\usepackage{booktabs}

\newtheorem{definition}{Definition}[section]
\newtheorem{coro}{Corollary}

\newtheorem{minv}{Inversion Of Block Diagonal Matrix}

\usepackage[citestyle=authoryear,style=authoryear,backend=bibtex]{biblatex}
\title{Learning World Models With Hierarchical Temporal Abstractions: A Probabilistic Perspective}
\author{Vaisakh Shaj}
\subject{}

\subtitle{
\vskip1.5em
Zur Erlangung des akademischen Grades eines\\[1em]
{\Large Doktors/Doktorin der Ingenieurwissenschaften}\\[1em]
von der KIT-Fakultät für Informatik des\\
Karlsruher Instituts für Technologie (KIT)\\[.5em]
{vorgelegte}\\[.3em]
{\Large Dissertation}
}

\author{\normalsize{von}\\
{\LARGE Vaisakh Shaj Kumar}\\
\normalsize{aus Kerala (Indien)}
}

\vspace{3cm}
\publishers{%
  \flushleft
  \normalsize Tag der mündlichen Prüfung: 19.~November~2024\\[1em]
  \begin{tabular}{@{}l@{\hspace{1em}}l@{}}
    \textbf{Referent:} & Prof.\ Dr.\ Gerhard Neumann\\[-0.2em]
                       & {\small Karlsruhe Institute of Technology (KIT)}\\[0.6em]
    \textbf{Referent:} & Prof.\ Dr.\ Kristian Kersting\\[-0.2em]
                       & {\small Technische Universität Darmstadt (TU~Darmstadt)}
  \end{tabular}
}

\date{}

\begin{document}
\selectlanguage{english}
\maketitle

\frontmatter

\selectlanguage{english}
\Abstract{Machines that can replicate human intelligence with type 2~\parencite{daniel2017thinking} reasoning capabilities should be able to reason at multiple levels of spatio-temporal abstractions and scales using internal world models~\parencite{friston2008hierarchical,lecun2022path}. Devising formalisms to develop such internal world models, which accurately reflect the causal hierarchies inherent in the dynamics of the real world, is a critical research challenge in the domains of artificial intelligence and machine learning. This thesis identifies several limitations with the prevalent use of state space models (SSMs) as internal world models and propose two new probabilistic formalisms namely Hidden-Parameter SSMs and Multi-Time Scale SSMs to address these drawbacks. The structure of graphical models in both formalisms facilitates scalable exact probabilistic inference using belief propagation, as well as end-to-end learning via backpropagation through time. This approach permits the development of scalable, adaptive hierarchical world models capable of representing nonstationary dynamics across multiple temporal abstractions and scales. Moreover, these probabilistic formalisms integrate the concept of uncertainty in world states, thus improving the system's capacity to emulate the stochastic nature of the real world and quantify the confidence in its predictions. The thesis also discuss how these formalisms are in line with related neuroscience literature on Bayesian brain hypothesis and predicitive processing. Our experiments on various real and simulated robots demonstrate that our formalisms can match and in many cases exceed the performance of contemporary transformer variants in making long-range future predictions. We conclude the thesis by reflecting on the limitations of our current models and suggesting directions for future research.}

\selectlanguage{ngerman}
\Abstract{Maschinen, die menschliche Intelligenz mit Fähigkeiten zur Schlussfolgerung des Typs 2~\parencite{daniel2017thinking} replizieren können, sollten in der Lage sein, auf mehreren Ebenen räumlich-zeitlicher Abstraktionen und Maßstäbe zu schlussfolgern, indem sie interne Weltmodelle verwenden~\parencite{friston2008hierarchical,lecun2022path}. Die Entwicklung von Formalismen zur Entwicklung solcher internen Weltmodelle, die die in der Dynamik der realen Welt inhärenten kausalen Hierarchien genau widerspiegeln, stellt eine entscheidende Forschungsherausforderung in den Bereichen Künstliche Intelligenz und maschinelles Lernen dar. Diese Dissertation identifiziert mehrere Einschränkungen bei der verbreiteten Verwendung von Zustandsraummodellen (SSMs) als interne Weltmodelle und schlägt zwei neue probabilistische Formalismen vor, nämlich Versteckte-Parameter-SSMs und Mehr-Zeitskalen-SSMs, um diese Nachteile anzugehen. Die Struktur der grafischen Modelle in beiden Formalismen ermöglicht skalierbare exakte probabilistische Inferenz mittels Belief Propagation sowie End-to-End-Lernen durch Backpropagation durch die Zeit. Dieser Ansatz ermöglicht die Entwicklung skalierbarer, adaptiver hierarchischer Weltmodelle, die nichtstationäre Dynamiken über mehrere zeitliche Abstraktionen und Maßstäbe darstellen können. Darüber hinaus integrieren diese probabilistischen Formalismen das Konzept der Unsicherheit in Weltzuständen und verbessern somit die Fähigkeit des Systems, die stochastische Natur der realen Welt nachzuahmen und das Vertrauen in seine Vorhersagen zu quantifizieren. Die Arbeit diskutiert auch, wie diese Formalismen mit der verwandten neurowissenschaftlichen Literatur zur Hypothese des Bayesianischen Gehirns und zur prädiktiven Verarbeitung übereinstimmen. Unsere Experimente mit verschiedenen realen und simulierten Robotern zeigen, dass unsere Formalismen in vielen Fällen die Leistung zeitgenössischer Transformer-Varianten bei der Vorhersage langfristiger Zukünfte übertreffen können. Wir schließen die Dissertation mit einer Reflexion über die Grenzen unserer aktuellen Modelle und Vorschlägen für zukünftige Forschungsrichtungen ab.}

\selectlanguage{english}
\chapter{Acknowledgement}
This thesis is the result of the support of several people, to whom I am extremely grateful. I would first like to express my thanks and gratitude to my supervisor, Prof. Gerhard Neumann, for providing me the opportunity to work with him and his consistent mentoring over the last 5 years. His mentorship has helped shape my approach to research, introducing me to a probabilistic way of thinking and emphasizing the importance of rigor. He is one of the most approachable people I have met in my life and carries himself with a deep sense of humility and patience, qualities that continue to inspire me both professionally and personally.
\par
I am also thankful for the guidance and support from several other mentors across various projects during my PhD. I extend my gratitude to Professor Marc Hanheide, Dr. Dieter Buchler, Harit Pandya, Philipp Becker, Aravind Srinivas, Ozan Demir and Niels van Duijkeren, who each played a crucial role at different stages of my doctoral journey. 
\par
My heartfelt thanks go to our administrative secretary, Christine Brand, for her kind support over the PhD, which made navigating daily challenges as a non-German speaking foreigner much simpler.
\par
One of the highlights of my PhD journey was being able to be part of two separate research groups, at KIT in Germany and the University of Lincoln in the UK. This provided me an incredible opportunity to immerse myself in two distinct cultures. I am indebted to the several brilliant, kind, and inclusive colleagues at both these research groups with whom I had the privilege of sharing many hikes, retreats, ski trips, dinners, coffee breaks, and gossip sessions. The technical debates I had with them have significantly contributed to molding my thought process as a researcher. I would like to extend special thanks to Harit Pandya, Riccardo, Philipp Becker, Onur, Bruce (Ge Li), and Max, whose friendship and support made the initial phases of my PhD journey and the transition between two countries significantly easier.
\par
I would like to thank Denis Blessing and Emiliyan Gospodinov for proofreading the thesis and giving valuable feedback. Special thanks to my friend Sharvin Raphael, who assisted with the Photoshop-based graphics used in this thesis, enhancing the visual presentation of my research.
\par
It was also a privilege to work with and mentor several brilliant students (Andreas Boltres, Martin Herault, Moritz Reuss, Rohit Sonker, Ruben Jacob, Stefan Geyer, Shuheng Zhang, Emiliyan Gospodinov, Thorben Comes, among others). The discussions with them were invaluable, significantly influencing my thought process. Additionally, working with them helped me discover my passion for mentoring and teaching, for which I am immensely grateful.
\par
I would like to thank my parents, brother, friends and family in India and abroad, whose constant support and encouragement were crucial to my pursuit and perseverance in this ambitious endeavor of moving to Europe for my PhD. Finally, special thanks to Kevin for making my experience in Germany rich and colourful.

\selectlanguage{english}

\tableofcontents

\listoffigures
\listoftables

\mainmatter
\chapter{Introduction}
The human brain maintains an intricate internal mental model of the world, a sophisticated representation that allows us to navigate complex environments, predict outcomes, and learn from interactions with minimal explicit instruction. This concept of "World Models"~\parencite{sutton1995td,ha2018world,lecun2022path} in machine learning research draws direct inspiration from our cognitive processes~\parencite{friston2005theory,friston2008hierarchical}, aiming to endow AI systems with a similar ability to abstract, understand, and anticipate the dynamics of their surroundings, a process that remains challenging for current AI systems. Research on foundational world models is pivotal for advancing beyond the current limitations of machine learning, where systems often require vast datasets and extensive training to perform tasks that humans manage with ease.
 
Recent advances in transformer~\parencite{vaswani2017attention} based large language models (LLMs) like BERT~\parencite{devlin2018bert}, GPT-3,4~\parencite{brown2020language,kocon2023chatgpt} etc have impressed the AI community due to their remarkable ability to generate coherent text, translate languages, and even produce code. However, the perception that LLMs might be the definitive pathway to AGI overlooks critical nuances. At their core, LLMs excel in pattern recognition and statistical inference from vast datasets, but lack an understanding of causality, physical principles, and reasoning, essential components of human-like intelligence. Recent research~\parencite{zevcevic2023causal} on the causal nature of LLMs argues that they are heavily based on correlational statistics and do not capture the causal dynamics essential for understanding and interacting with the real world. By focusing on learning the causal structure of the world and enabling machines to simulate and predict future states, researchers can create more autonomous, efficient, and adaptable AI systems. Such a research path represents a measured step towards developing AI systems with a deeper, more intuitive grasp of the world, aligning machine learning processes more closely with natural intelligence. Advancements in learning a foundational world model promise profound implications in several fields, including autonomous vehicles, robotics, healthcare, environmental modeling, and beyond, marking a crucial step towards achieving AI that can truly understand and navigate the world as humans do~\parencite{pezzulo2018hierarchical, lecun2022path}.  

When discussing the foundational world model (FWM), I refer to a model (or a set of models) that embodies the following characteristics expanding on the definition of ~\cite{gupta2024essential}. 
\begin{enumerate}[label=(\roman*)]
    \item (Representation) Conceptually understand the components, structures, and interaction dynamics within a given system at different levels of abstractions~\parencite{lee2003hierarchical,friston2008hierarchical,pezzulo2018hierarchical,lecun2022path};
    \item (Veridicality) Quantitatively model the underlying laws of such a system, that enables accurate predictions of counterfactual consequences of interventions/actions;
    \item (Probabilistic) Model uncertainties (aleatoric  and epistemic) via probabilistic representations. This ensures that the FWM can handle the stochastic nature of the real world, quantify confidence in its predictions and imagine diverse plausible futures.
    \item (Foundational) Being able to generalize and scale (i), (ii) and (iii) across diverse systems or domains encountered in the world. 
\end{enumerate}

The definition of FWM in \cite{gupta2024essential} is expanded to explicitly include the "probabilistic" property. The reason for this is elaborated further in Section \ref{sec:intro-probl}. Furthermore, Section \ref{sec:comp-neuro} provides a reasoning for this from a computational neuroscience point of view. 

\section{Thesis Problem Statement}
\label{sec:intro-probl}
In this dissertation, I explore the difficulties associated with existing world model formalisms to fit into the notion of FWM and introduce the formalism(s) designed to tackle three challenges in developing foundational models of the world, as outlined below. \\
\begin{enumerate}
\item \textbf{\textit{How can we model the world with a principled probabilistic formalism that also quantifies uncertainties in predictions in a scalable manner?}}\\
The real world involves a significant amount of uncertainty. It is intrinsically stochastic (aleatoric uncertainty), and we are often uncertain about the true state of the system because our observations about it are partial (epistemic uncertainty), or the prediction accuracy of the world model is imperfect due to limited training data, representational power, or computational constraints. This necessitates probabilistic world models that can imagine multiple plausible future scenarios and the associated uncertainty. Furthermore, the inference algorithms on these probabilistic models/representations should scale well for large datasets and long sequences. \\
\item \textbf{\textit{How can we make sure the model adapts to changing dynamics / non-stationary situations?}} \\ In various real-world control scenarios, agents are presented with tasks that may exhibit different dynamics due to changing environmental factors or task requirements. For instance, a robot engaged in playing table tennis might use rackets of differing weights or sizes, or an agent tasked with bottle manipulation might deal with varying fluid levels. Traditional models, operating on a singular time scale or with flat world representations, struggle to accommodate these dynamic shifts. Therefore, it's essential to develop formalisms capable of inferring task abstractions for seamless adaptability in multi-task environments.\\

\item \textbf{\textit{Can we model the world at multiple temporal abstractions and time scales?}} \\
One important dimension of world models is the level of temporal granularity/time scale at which they operate. Existing literature on world models operates at a single level of temporal abstraction, typically at a fine-grained level such as milliseconds. One drawback of single-time scale world models is that they may not capture longer-term trends and patterns in the data. For efficient long-horizon prediction, reasoning, and planning, the model needs to predict at multiple levels of temporal abstractions.
\end{enumerate}

\notebox{The challenges outlined here are not exhaustive of all the difficulties encountered in developing foundational world models (FWMs). Instead, they specifically target three central aspects that are vital for the successful creation of FWMs.}
\section{Thesis Contribution}
\label{subsec:intro-contri}
This thesis introduces a systematic approach to address the said challenges in a principled probabilistic framework through the development of three formalisms, each building on the other. Each formalism exhibits increasing complexity and expressiveness, with the incorporation of one utility at a time, culminating in a final model that combines the favorable properties of all three. The sequential presentation also ensures a comprehensive understanding of the progression from the foundational state space models (SSMs) to the sophisticated multi-time scale SSMs, showcasing a systematic evolution in complexity and capability. 

 \begin{figure*}[h]
 \hspace{-0.1cm}
\includegraphics[scale=.54]{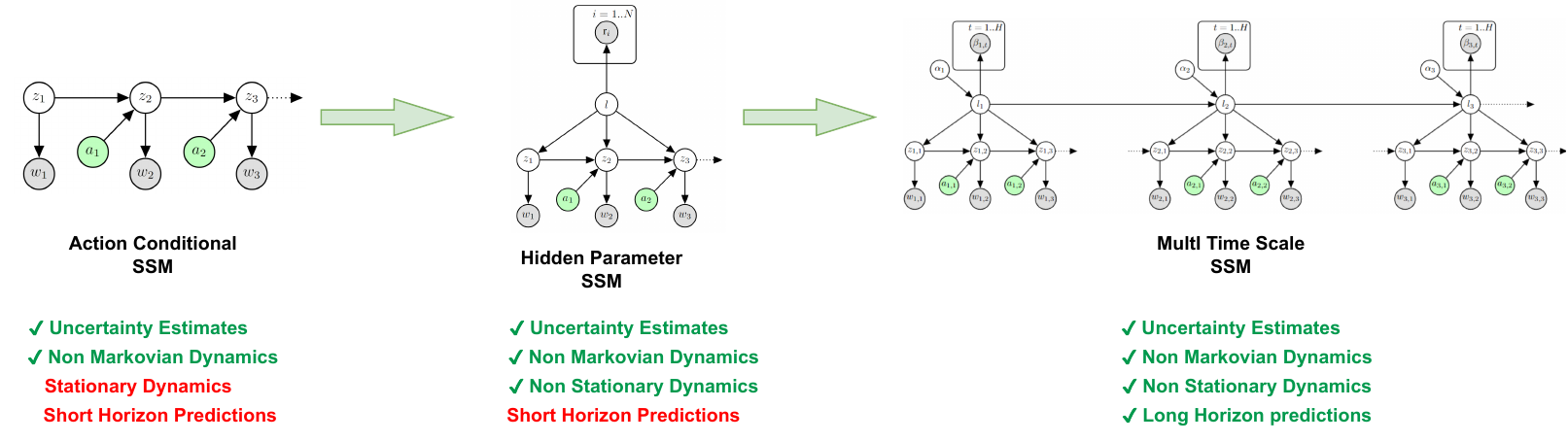}
\caption{Evolution of thesis represented in terms of the evolution of the probabilistic graphical models for each world model formalism. All formalisms infer the latent states via exact inference resulting in closed-form update rules. } 
 \label{fig:schematic}
\end{figure*}

The main contributions that culminate from this thesis are (i) identifying the drawbacks of the widely used SSM formalism and (ii) introducing two new formalisms / declarative representations~\parencite{koller2009probabilistic} called HiP-SSM and Multi-Time Scale SSM to address these drawbacks. The thesis also proposes algorithms to perform scalable exact inference via message passing for forward predictions in these models. Finally, it is demonstrated that end-to-end learning is possible in all of the proposed models using a single loss function and backpropagation through time.


\section{Thesis Outline}
\begin{figure}[H]
\begin{center}
\includegraphics[scale=.73]{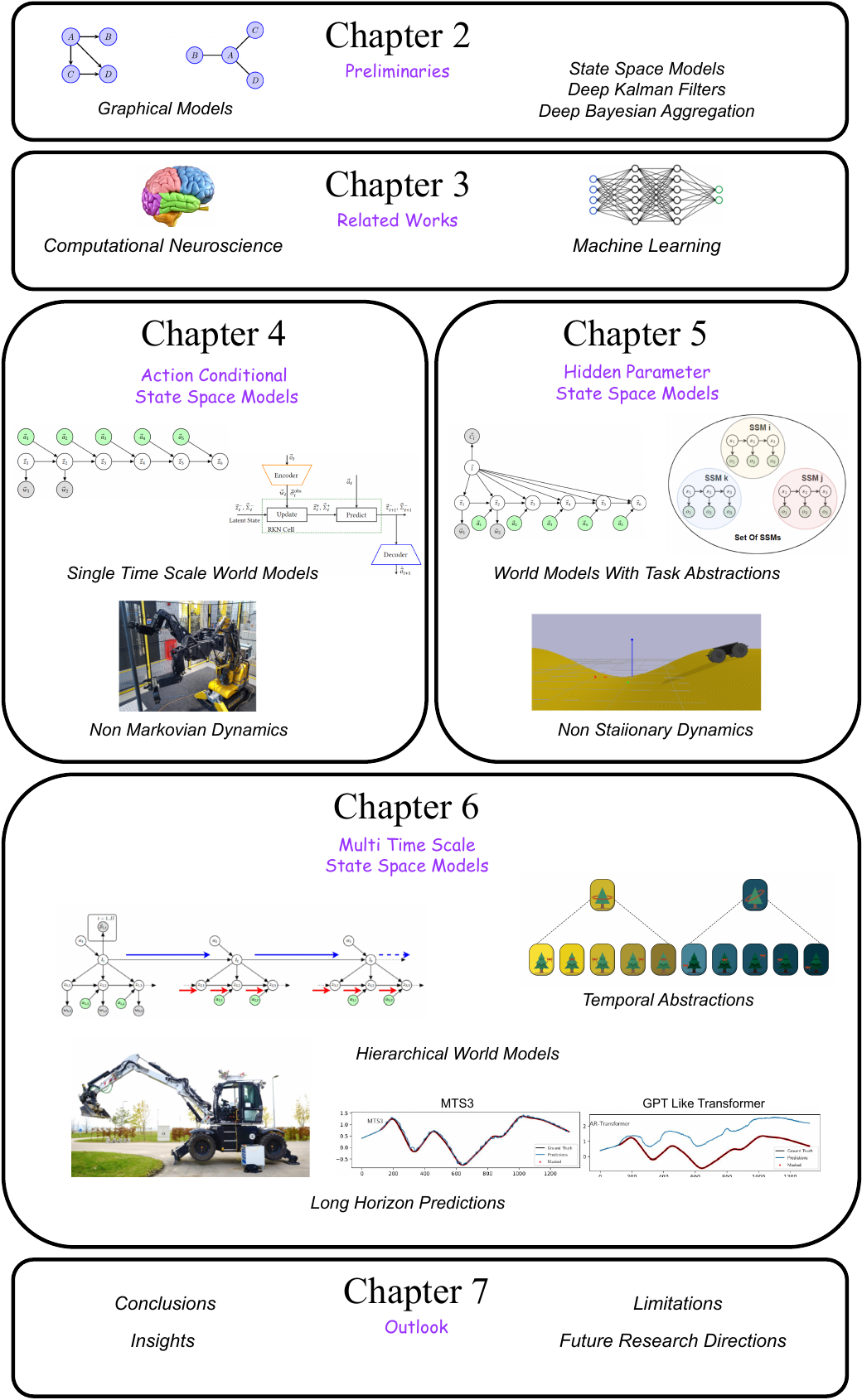}
\end{center}
\caption{A visual overview of the thesis.} 
 \label{fig:visual}
\end{figure}

The remainder of this thesis is structured as follows. A visual overview of the thesis is given in Figure \ref{fig:visual}.

In Chapter 2, we give a list of preliminaries that will assist readers in navigating the rest of the document. This chapter begins with a concise introduction to Probabilistic Graphical Models (PGMs), a core framework utilized throughout this thesis for formalizing representations of the world dynamics and performing learning and inference in these. Additionally, we introduce fundamental concepts, including deep state space models (SSMs) and Bayesian aggregation schemes.

Chapter 3 reviews the existing literature on world models, drawing from two distinct areas: computational neuroscience and machine learning. The segment on computational neuroscience aims to highlight connections between the representations explored in this thesis and the conceptualizations of world models in the human brain set forth by the neuroscience community. Additionally, we succinctly review various methodologies adopted by the machine learning community for building world models, providing a backdrop for the thesis's further discussions.

In Chapter 4, we use the well-established formalism of state space models (SSMs) for learning single time-scale World Models. We critically analyze and extend a recent deep Kalman model (deep Gaussian SSM) known as recurrent Kalman networks (RKNs)~\parencite{becker2019recurrent} for the specific objective of modeling world dynamics. A significant limitation identified in the original models was their inability to systematically include control or action input in the latent dynamics. This chapter suggests a methodologically sound approach to embed action signals into state space models that respect the causal relation with the inferred latent states. Conditioning actions in a manner that adheres to causal relationships is essential, particularly for decision-making and control tasks. This approach enables precise execution of interventions ("doing") and the simulation of counterfactual scenarios ("imagining") using actions or control signals. The hypothesis is validated by modelling the dynamics of a variety of real robots with non-markovian but stationary dynamics.
In Chapter 5, we focus on the aspect of "adaptability" of world models within dynamic, non-stationary environments. Traditional State Space Models (SSMs) typically operate under the assumption of static dynamics, a presumption that falls short in the complexity of real-world applications. Recognizing the challenge of developing distinct SSMs for all possible variation in dynamics, we introduce the formalism of Hidden Parameter State Space Models (HiP-SSMs). HiP-SSMs leverage hierarchical latent task abstractions to infer the causal factors behind environmental non-stationarity, facilitating seamless adaptation to changing conditions. Moreover, we detail a simple and computationally efficient method for learning and inference in this Gaussian graphical model that avoids approximations like variational inference.

In Chapter 6, the thesis explores the notion of "temporal depth" in the world dynamics via causal hierarchies. Learning world models with hierarchical temporal abstractions and multiple time scales is a largely unaddressed problem in existing literature for world models (deterministic or stochastic). We come up with a principled probabilistic formalism for learning such multi-time scale world models as a hierarchical sequential latent variable model called multi-time scale state space models (MTS3). MTS3 maintains all the advantages of the previous formalisms (SSM and HiP-SSM)  for learning world models but can additionally make hierarchical long-horizon predictions. We also derive computationally efficient inference schemes on multiple time scales for highly accurate long-horizon predictions and uncertainty estimates spanning several seconds into the future. We show these lightweight hierarchical linear models can compete with state of the art transformers on long-horizon predictions on continuous dynamical systems.

Chapter 7 concludes the thesis by summarizing the findings, outlining the limitations, and directions for future research.

\chapter{Preliminaries}
\section{Probabilistic Graphical Models}

Probabilistic Graphical Models (PGMs)~\parencite{jordan2004graphical,koller2007graphical} use a graph-based representation as the basis for compactly encoding a complex distribution over a high-dimensional space. In various applied fields including bioinformatics, speech processing, image processing and control theory, statistical models have long been formulated in terms of graphs, and algorithms for computing basic statistical quantities such as likelihoods and score functions have often been expressed in terms of recursions operating on these graphs; examples include phylogenies, pedigrees, hidden Markov models, Markov random fields, and Kalman filters. These ideas can be understood, unified, and generalized within the formalism of graphical models. Indeed, graphical models provide a natural tool for formulating variations on these classical architectures, as well as for exploring entirely new families of statistical models. 

The intuitive formalisms via PGMs enable effective communication between scientists across the mathematical divide by fostering substantive debate in the context of a scientific problem and ultimately facilitate the joint development of statistical and computational tools for quantitative data analysis. Consequently, in fields that involve the study of large numbers of interacting variables, graphical models are increasingly in evidence.

\subsection{Structure Of Graph and Independencies}
\label{subsec:graphind}
There is a dual perspective that can be used to interpret the structure of the graph in PGMs. 

\begin{enumerate}
    \item \textbf{The graph as a representation of a set of
independencies:} From one perspective, the graph is a compact representation of a set of independencies that hold in the distribution; these properties take the form $X$ is independent of $\boldsymbol{Y}$ given $Z$, denoted $\boldsymbol{X} \perp \boldsymbol{Y} \mid \boldsymbol{Z})$, for some subsets of variables $\boldsymbol{X}, \boldsymbol{Y}, \boldsymbol{Z}$.

 \item \textbf{The graph as a skeleton for factorizing a distribution:} The other perspective is that the graph defines a skeleton for compactly representing a high-dimensional distribution: Rather than encode the probability of every possible assignment to all of the variables in our domain, we can "break up" the distribution into smaller factors, each over a much smaller space of possibilities.
\end{enumerate}
 
\notebox{The two perspectives are, in a
deep sense, equivalent. The independence properties of the distribution are precisely what allow it to be represented compactly in a factorized form. Conversely, a particular
factorization of the distribution guarantees that certain independencies hold.}
\begin{figure}[t]
    \begin{minipage}{0.45\textwidth}
        \centering
        \includegraphics[width=0.45\textwidth]{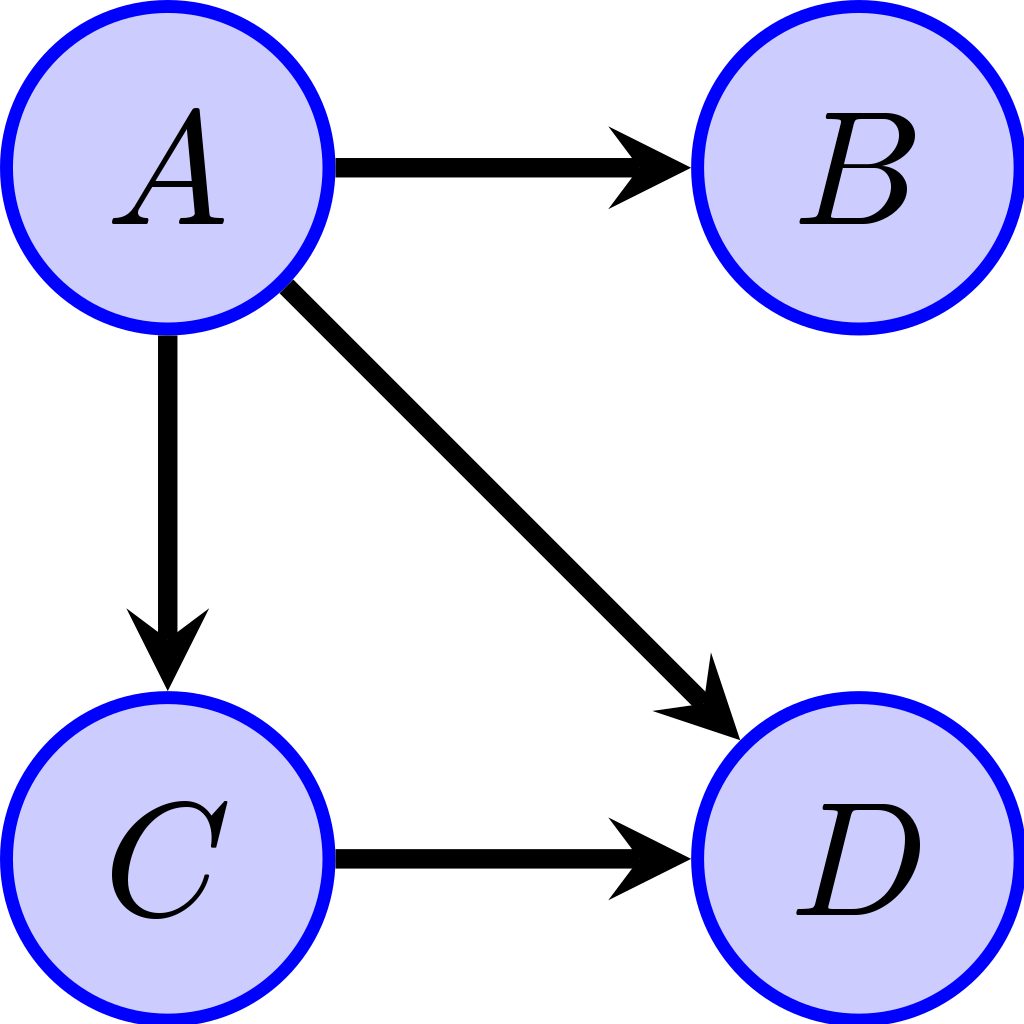}
    \end{minipage}
    \hfill 
    \begin{minipage}{0.45\textwidth}
        \centering
        \includegraphics[width=0.65\textwidth]{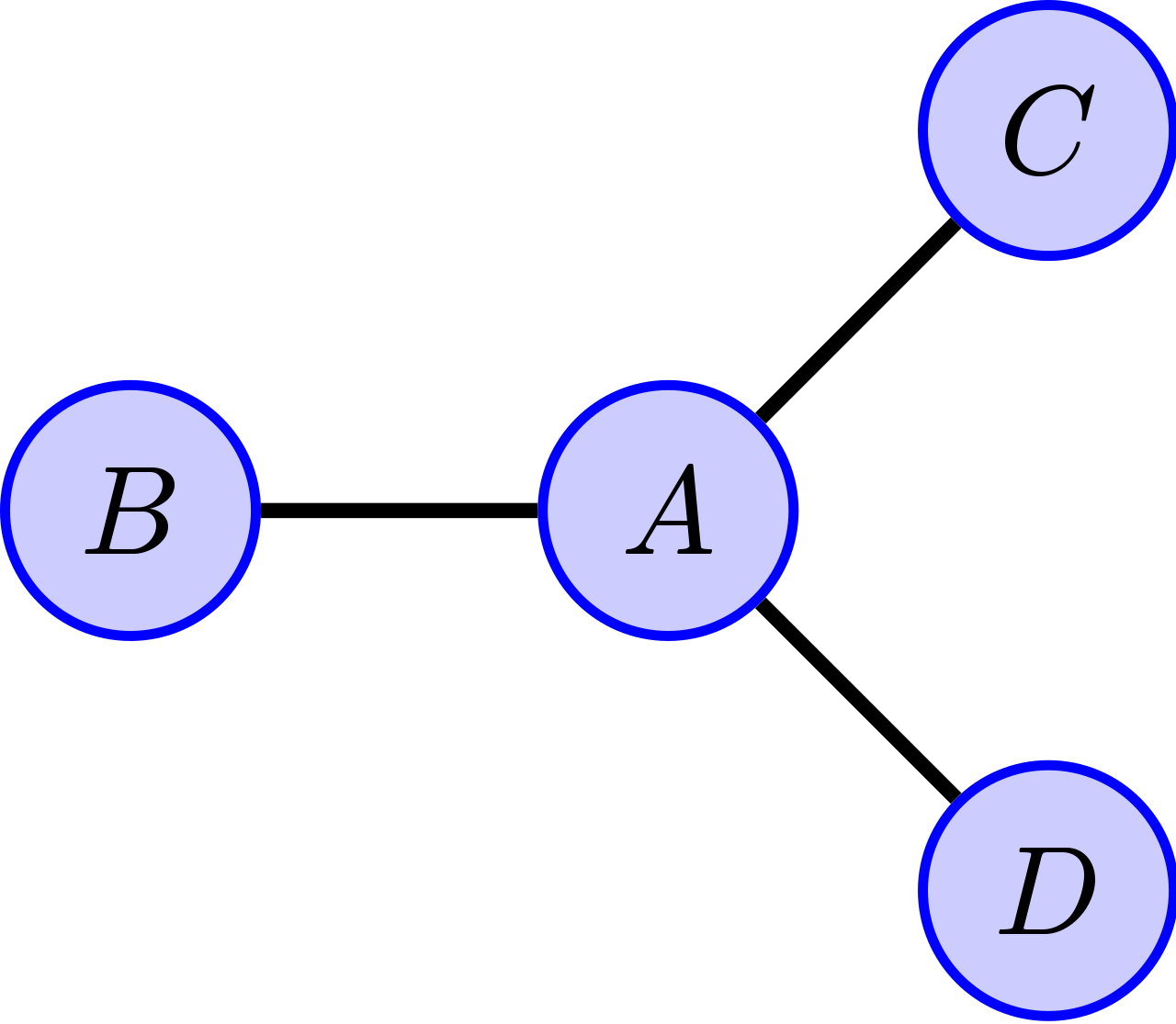}
    \end{minipage}
            \caption{(left) A Bayesian Network, where the directed edges indicate causal relationships. (right) A Markov network, where the undirected edges give a notion of correlation/affinity}
\end{figure}

\subsection{Representation, Inference and Learning in PGMs}
\subsubsection{Representation} As detailed in section \ref{subsec:graphind}, the representation of distributions in graphical language exploits structure that appears in many distributions that we want to encode in practice. The graph structure allows us to factorize a high-dimensional joint distribution into a product of lower-dimensional CPDs or factors that exploit conditional independencies. This framework has many advantages, as listed below. 

\begin{enumerate}
    \item It often allows the distribution to be written down tractably, even in cases where the explicit representation of the joint distribution is astronomically large.
    \item A human expert can understand and evaluate its semantics and properties. This property is important for constructing models that provide an accurate reflection of our understanding of a domain.
\end{enumerate} 

PGMs can be broadly classified into 2 families based on the types of their graphical structure, Bayesian and Markov networks. \textbf{Bayesian Networks} uses a directed graph that captures causal relationships between variables. \textbf{Markov Networks} uses an undirected graph and captures correlations between variables. Both representations provide the duality of independencies and factorization, but they differ in the set of independencies they can encode and in the factorization of the distribution that they induce.

\notebox{In this thesis, we come up with formalisms for internal mental models of the world/environment dynamics using these graph-based representations, specifically Bayesian Networks.}

\subsubsection{Inference} The same structure often also allows the distribution to be used effectively for inference, answering queries using the distribution as our model of the world. In particular, there exist algorithms for computing the posterior probability of some variables given evidence of others. These inference algorithms work directly on the graph structure and are generally orders of magnitude faster than manipulating the joint distribution explicitly.

\begin{figure}[H]
    \centering
    \includegraphics[width=14cm]{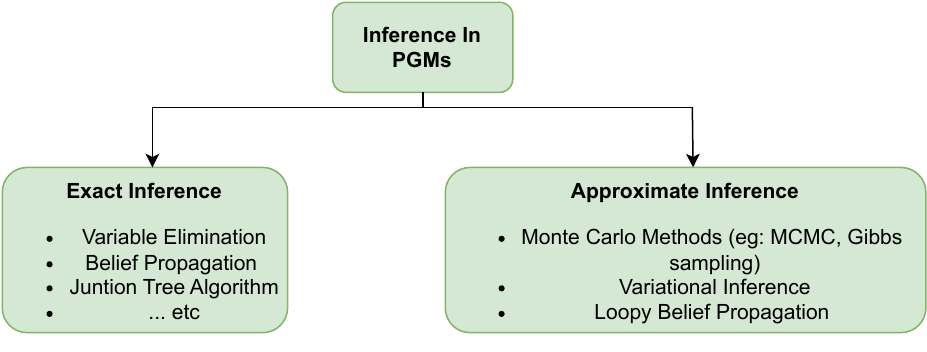}
    \caption{A summary of standard inference routines on PGMs.}
    \label{fig:galaxy}
\end{figure}

\notebox{This thesis perform exact inference via closed form messages (belief propagation) to answer several such queries on the proposed graphical representation of the world model.}

\subsubsection{Learning} PGM framework facilitates the effective construction of approximate models by learning from past experience (in the form of observed data). In this approach, a human expert provides some rough guidelines on how to model a given domain. For example, the human usually specifies the attributes that the model should contain, often some of the main dependencies that it should encode, and perhaps other aspects. However, the details are usually filled in automatically by fitting the model to the data. The models produced by this process are usually much better reflections of the domain than models that are purely hand-constructed. In addition, they can sometimes reveal surprising connections between variables and provide novel insights into a domain. Learning in PGMs can be broadly classified into (i) structure learning and (ii) parameter learning. \textbf{Structure learning} focuses on discovering the underlying causal or dependence structure between the variables in the model. It essentially answers the question "What variables influence each other?". \textbf{Parameter learning}, on the other hand, focuses on estimating the quantitative relationships between variables based on the discovered structure. It essentially answers the question "How strong are these relationships?".

\notebox{This thesis proposes end-to-end parameter learning strategies for the Bayesian Network representations from data via backpropagation.}

\subsection{Predictive World Models As Inference In PGMs}

In this thesis, we use PGM as a tool to model the causal dynamics of the world in a probabilistically principled manner. In fact, if a particular learning problem can be set up as a probabilistic graphical model, this can often serve as the first and most important step in solving it. The PGM framework offers us the flexibility that once we write down the model and pose the question, the objectives for learning and inference emerge automatically. 

To set up models that reflect the causal structure of our world/environment dynamics, we choose directed graphical models (Bayesian Networks) as declarative graphical \textbf{representations}. We can then effectively use the representation to answer a wide range of questions that are of interest using well-studied and efficient \textbf{inference} algorithms. These queries may include (i) "state queries" that seek answers about the past, present, and future world states, and (ii) "causal queries" such as intervention and counterfactual queries. In this thesis, we perform both "state queries" and "causal queries". The "causal queries" primarily take the form of action-conditional future predictions. Finally, for those parameters that are unknown to the human expert, the principled \textbf{learning} objectives can be used to learn the parameters from the observed data. A summary of the schemes used in the thesis is given in Table \ref{tab:summarythesistable}.
\newcommand{\highlight}[1]{\textcolor{green}{#1}} 
\begin{table}[H]
\centering
\caption{The summary of the representation, inference and learning schemes used in the thesis for learning probabilistic world models.}
\label{tab:summarythesistable} 
\begin{tabular}{|c|c|c|}
\hline
& Types & Used In Thesis \\ \hline
\hline
Representation & Bayesian Network & \textcolor{green!50!black}{$\checkmark$}\\ & Markov Network &\\ \hline
Inference & Exact Inference & \textcolor{green!50!black}{$\checkmark$} \\
& Approximate Inference & \\ \hline
Learning & Parameter Learning &\textcolor{green!50!black}{$\checkmark$}\\
&Structural Learning &\\ 
&Backpropagation (Gradient Descent) &\textcolor{green!50!black}{$\checkmark$}\\
& Expectation Maximization &\\ \hline
\end{tabular}
\end{table}
We start with the most widely used and studied PGM for modelling the world, commonly referred to as State Space Models (SSMs). This formalism will form the basis for all other formalisms developed in this thesis.

\section{State Space Models}
\label{sec: ssm}
\begin{wrapfigure}[10]{r}{0.45\textwidth}
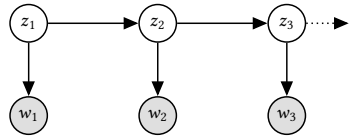

  \centering
  \begin{center}
 \resizebox{\linewidth}{!}{\tikzsimpleSSM}
\end{center}
\caption{PGM for a State Space Model (SSM).} 
 \label{fig:ssm}
\end{wrapfigure}
State space models (SSMs) are Bayesian probabilistic graphical models \parencite{koller2009probabilistic, jordan2004graphical} that are popular for learning patterns and predicting behaviour in sequential data and dynamical systems as shown in Figure \ref{fig:ssm}.
Formally, we define a state space model as a tuple $( \mathcal{Z}, \mathcal{A}, \mathcal{O}, f, h, \Delta t)$, where $\mathcal{Z}$ is the state space, $\mathcal{A}$ the action space and $\mathcal{O}$ the observation space of the SSM. The parameter $\Delta t$ denotes the discretization time step and $f$ and $h$ the dynamics and observation models, respectively.

We will consider the Gaussian state space model that is represented using the following equations
\begin{equation*}
  \begin{aligned}[c]
  \cvec{z}_t &= f(\cvec{z}_{t-1},\cvec{a}_{t-1}) + \cvec{\epsilon}_t, \quad \cvec{\epsilon}_t \sim \mathcal{N}(\cvec{0}, \cmat{\Sigma^{trans}}), \\
   \cvec{o}_t &= h(\cvec{z}_t) + \cvec{v}_t, \quad \cvec{v}_t \sim \mathcal{N}(\cvec{0}, \cmat{\Sigma^{obs}}).
  \end{aligned}
\end{equation*}
Here $\cvec{z}_t \in \mathcal{Z}$, $\cvec{a}_t \in \mathcal{A}$ and $\cvec{o}_t \in \mathcal{O}$ are the latent states, actions, and observations at time t. The vectors $\cvec{\epsilon}_t$ and $\cvec{v}_t$ denote zero-mean Gaussian transition and observation noise, respectively.

\subsection{Linear Gaussian SSMs (Kalman Filters).}
\label{subsec: Kalman}
For SSMs defined in \ref{sec: ssm}, when $f$ and $h$ are linear/locally linear, inference can be performed efficiently via exact inference. Such SSMs are referred to as Kalman Filters, which were developed independently of PGM Literature. The Kalman filter \parencite{kalman1961new,srkk2013bayesian} works by iteratively answering two queries, (i) estimating the prior marginal (predict step) and (ii) updating the posterior given current observations (update step), which are detailed below:
\subsubsection{Predict Step}
During the prediction step the transition model $\cmat{A}$ is used to infer the current prior state marginal $p(\cvec{z}_t|\cvec{w}_{1:t-1}) = \mathcal{N}(\cvec{z}_t^-,\cmat{\Sigma}_t^-)$, that is, a priori to the observation, from the previous posterior estimate $p(\cvec{z}_{t-1}|\cvec{w}_{1:t-1}) = \mathcal{N}\left(\cvec{z}_{t-1}^+, \cmat{\Sigma}_{t-1}^+ \right)$. The marginal computation is given as follows: 
\begin{equation}
\begin{aligned}
p(\cvec{z}_t|\cvec{w}_{1:t-1}) &= \mathcal{N}(\cvec{z}_t^-,\cmat{\Sigma}_t^-) \\ &= \int p(\cvec{z}_t|\cvec{z}_{t-1})p(\cvec{z}_{t-1}|\cvec{w}_{1:t-1})d\cvec{z}_{t-1}.
\end{aligned}
\end{equation}
This is an instance of forward belief propagation or variable elimination in the chain-structured graphical model~\parencite{koller2007graphical}. Due to the Gaussian linear assumption, the parameters of this marginal can be estimated in closed form as follows:
\begin{equation}
\begin{aligned}
\cvec{z}^-_{t} &= \cmat{A} \cvec{z}^+_{t-1} \\ \text{and} \quad \cmat{\Sigma}^-_{t} &= \cmat{A} \cmat{\Sigma}_{t-1}^+ \cmat{A}^T + \cmat{\Sigma}^\mathrm{trans}.
\end{aligned}
\end{equation}
\subsubsection{Update Step}
The prior estimate is then updated using the current observation $\cvec{w_t}$ and the linear observation model $\cmat{H}$ to obtain the posterior estimate $\left(\cvec{z}_t^+, \cmat{\Sigma}_t^+ \right)$ using the Bayes rule.

 Since we have Gaussian assumptions on all random variables and a linear observation model, the posterior estimate $\left(\cvec{z}_t^+, \cmat{\Sigma}_t^+\right)$ can be obtained in closed form as:

\begin{equation}
    \begin{aligned}
\cvec{z}_t^+ &= \cvec{z}_t^- + \cmat{Q}_t \left(\cvec{w}_t - \cmat{H} \cvec{z}_t^-  \right), \\ 
\cmat{\Sigma}^+_t &= \left(\cmat{I} - \cmat{Q}_t \cmat{H} \right) \cmat{\Sigma}^-_t, \\
\text{with} \quad \cmat{Q}_t &= \cmat{\Sigma}^-_t \cmat{H}^T\left(\cmat{H} \cmat{\Sigma}^-_t \cmat{H}^T  + \cmat{\Sigma}^\mathrm{obs} \right)^{-1},
    \end{aligned}
    \label{eq:KalmUp}
\end{equation}

where $\cmat{I}$ denotes the identity matrix. The matrix $\cmat{Q}_t$ is called the Kalman gain.

The whole update step can be interpreted as a weighted average between the state and the observation estimate, where the weighting, that is, $\cmat{Q}_t$, depends on the uncertainty about those estimates. \par We refer to \cite{srkk2013bayesian} for detailed derivations of the Kalman filtering equations. The derivations can also be obtained using the sum-product algorithm in a factor graph of the state-space model in Figure \ref{fig:ssm}.  

\subsection{Deep Kalman Filters}

To model dynamics under partially observable scenarios, state-space models, in particular the Kalman filter~\parencite{kalman1960new}, have recently been integrated with deep learning, by modeling dynamics in learned latent space. We discuss two such works~\parencite{haarnoja2016backprop,becker2019recurrent}, that borrow concepts from deep learning and graphical model communities, where the architecture of the network is informed by the structure of the probabilistic state estimator. The various exact inference schemes in the learned latent spaces derived and executed in the thesis are motivated by this work.

\subsubsection{Backprop Kalman Filter}
The BackpropKF~\parencite{haarnoja2016backprop} uses a deep convolutional encoder to encode a high-dimensional observation $\cvec o$ to a latent observation $\cvec w_{t} = \textrm{enc}_w(\cvec o_{t})$. In addition to the latent observation, the encoder also learns to output the uncertainty of this observation, i.e., $\cvec \sigma_{o_t} = \textrm{enc}_{\sigma}(\cvec o_{t}).$ Due to the non-linear observation encoder, a simplified linear Gaussian state space model / (extended) Kalman filter with a known transition model is used in the latent space for state estimation. The parameters of the latent state
distribution are directly optimized as a deterministic computation graph via backpropagation through time. This scheme can also be motivated using the Koopman theory~\parencite{koopman1931hamiltonian,mondal2023efficient}.

\subsubsection{Recurrent Kalman Networks (RKN)}
\label{sec:prelimrkn}
\begin{figure}[h]
    \centering
    \includegraphics[width=11cm]{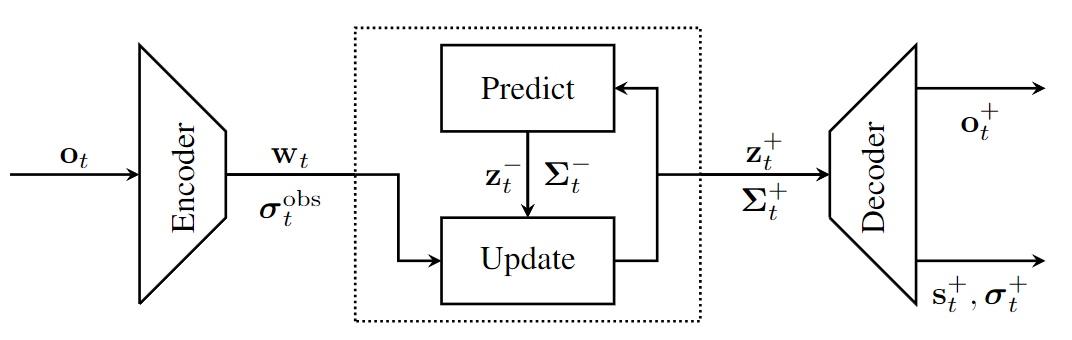}
    \caption{Recurrent Kalman Network (RKN) from \cite{becker2019recurrent}. It embeds Kalman filter update equations into the latent space of a deep encoder-decoder network. The Kalman predict step gives the current latent prior $\left(\mathbf{z}_t^{-}, \boldsymbol{\Sigma}_t^{-}\right)$using the last posterior $\left(\mathbf{z}_{t-1}^{+}, \boldsymbol{\Sigma}_{t-1}^{+}\right)$and subsequently update the prior using the latent observation $\left(\mathbf{w}_t, \boldsymbol{\sigma}_t^{\mathrm{obs}}\right)$. The factorized representation as shown in Figure \ref{fig:meancov} of $\Sigma_t$ converts matrix inversions to scalar operations. Further it allows splitting the latent state $\mathbf{z}_t$ to the observable units $\mathbf{p}_t$ as well as the corresponding memory units $\mathbf{m}_t$ as discussed in Section \ref{sec:prelimrkn}. Finally, a decoder is tasked to reconstruct the sensory information.}
    \label{fig:galaxy}
\end{figure}
\sloppy Recurrent Kalman Network (RKN)~\parencite{becker2019recurrent}, builds upon this idea but learns a locally linear transition model as opposed to a known model as in BackpropKF. Most importantly, the authors come up with clever factorized latent state representation as shown in Figure \ref{fig:meancov}, that avoids expensive matrix inversions in the Kalman Update Step. We discuss two important implications of this factorized representation below:
\begin{figure}[h]
    \centering
        \includegraphics[width=5cm]{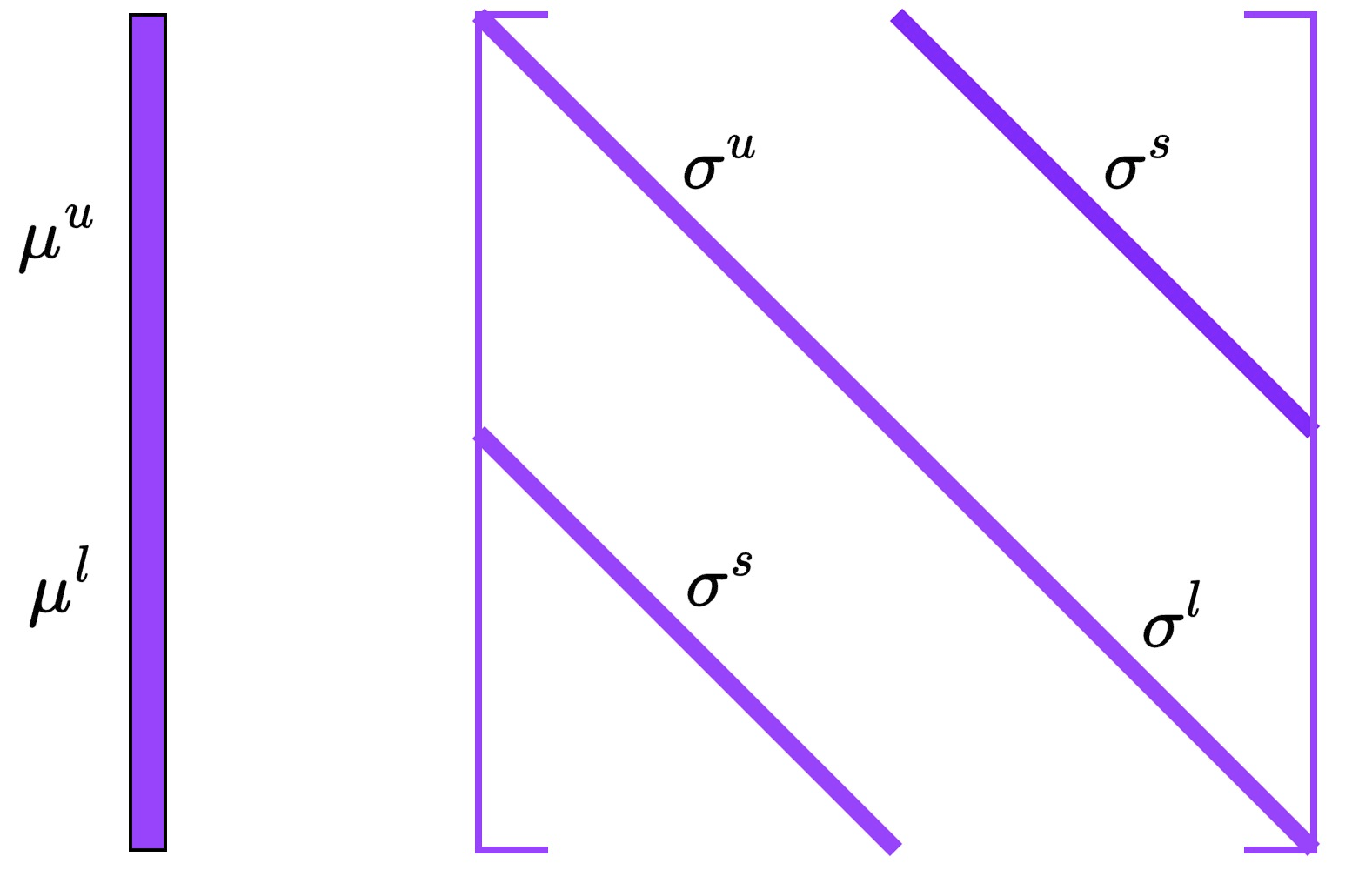}
    \caption{The latent state beliefs (mean $\mu$ and covariance $\Sigma$) in \cite{becker2019recurrent} is factorized into an upper (observed) and lower (unobserved) part as shown in the figure. The lower part can be dubbed as "memory states" and is assumed to keep a memory of information accumulated over time, because of the specific structure of the observation model. The side covariances ($\sigma^s$) are learnt to capture the correlation between the two halves.}
    \label{fig:meancov}
\end{figure}
\begin{enumerate}
    \item \textbf{Matrix inversions as scalar operations:} Kalman observation update, which is an instance of Bayesian inversion via conditioning involves high dimensional matrix inversions that are expensive to evaluate and hard to backpropagate for end-to-end learning. Hence, \cite{becker2019recurrent} introduce a factorization of the belief $p(\cvec z_t|\cvec o_{1:t}, \cvec a_{1:t-1}) = \mathcal{N}(\cvec z_t^+, \cvec \Sigma_t^+)$ such that only the diagonal and one off-diagonal vector of the covariance need to be computed, i.e. 
$$\cvec \Sigma_t^+ = \left[\begin{array}{cc}\cvec \Sigma_t^{u,+} & \cvec \Sigma^{s,+}_t \\ \cvec \Sigma^{s,+}_t & \cvec \Sigma^{l,+}_t  \end{array} \right], \textrm{ with } \cvec \Sigma_u^+ = \textrm{diag}(\cvec \sigma_t^{u,+}), \; \cvec \Sigma_l^+ = \textrm{diag}(\cvec \sigma_t^{l,+}) $$
and $\cvec \Sigma_{s}^+ = \textrm{diag}(\cvec \sigma_{t}^{s,+})$. Using this factorization, the Kalman filter update step described in Equation \ref{eq:KalmUp} can be executed using only scalar divisions, which are computationally efficient and simpler to implement in back-propagation. Consequently, the closed-form equations for updating the mean can be succinctly expressed using the following scalar equations:
\begin{align*}
\boldsymbol{z}_t^+ = \boldsymbol{z}_t^- +
\left[\begin{array}{c} \boldsymbol{\sigma}^\mathrm{u,-}_t \\
\boldsymbol{\sigma}^\mathrm{l,-}_t \end{array}\right]
\odot
\left[\begin{array}{c}\boldsymbol{w}_t - \boldsymbol{z}^{\mathrm{u},-}_t \\
\boldsymbol{w}_t - \boldsymbol{z}^{\mathrm{u},-}_t  \end{array}\right]
\oslash 
\left[\begin{array}{c}  \boldsymbol{\sigma}_t^{\mathrm{u},-} + \boldsymbol{\sigma}_t^\mathrm{obs} 
 \\  \boldsymbol{\sigma}_t^{\mathrm{u},-} + \boldsymbol{\sigma}_t^\mathrm{obs} \end{array}\right],
\end{align*}

The corresponding equations for the variance update can be expressed as the following scalar operations,
\begin{align*}
\boldsymbol{\sigma}^{\mathrm{u},+}_t &= \boldsymbol{\sigma}^{\mathrm{u},-}_t \odot \boldsymbol{\sigma}^{\mathrm{u},-}_t \oslash \left( \boldsymbol{\sigma}_t^{\mathrm{u},-} + \boldsymbol{\sigma}_t^\mathrm{obs} \right),  \\
\boldsymbol{\sigma}^{\mathrm{s},+}_t &= \boldsymbol{\sigma}^{\mathrm{u},-}_t \odot \boldsymbol{\sigma}^{\mathrm{s},-}_t \oslash \left( \boldsymbol{\sigma}_t^{\mathrm{u},-} + \boldsymbol{\sigma}_t^\mathrm{obs} \right), \\
\boldsymbol{\sigma}^{\mathrm{l},+}_t &= \boldsymbol{\sigma}^{\mathrm{l}, -}_t - \boldsymbol{\sigma}^{\mathrm{s},-}_t \odot \boldsymbol{\sigma}^{\mathrm{s},-}_t \oslash \left( \boldsymbol{\sigma}_t^{\mathrm{u},-} + \boldsymbol{\sigma}_t^\mathrm{obs} \right),
\end{align*}
where $\odot$ denotes the elementwise vector product and  $\oslash$ denotes an elementwise vector division. 
These factorization assumptions form the basis for the derivations of exact inference routines for all proposed models in this dissertation.
    \item \label{subsec: memoryunit}\textbf{Effective Estimation with Correlated Memory and Observation in Latent Variables:} Choosing the observation model as $\cvec H = [\cvec I,  \cvec 0]$ allows for the division of the latent state vector into two distinct components. The latent state $\cvec z_t = [\cvec p_t^T, \cvec d_t^T]^T$ has twice the dimensionality of the latent observation $\cvec w_t$ and only the first half of the latent state, i.e., $\cvec p_t$, can be observed. The upper part utilizes the identity matrix in $\cvec H = [\cvec I,  \cvec 0]$ to directly extract information from the observations. Meanwhile, the second lower part remains unobservable and is meant to hold information inferred over time, such as velocities in ordinary dynamical systems or images. The key aspect that contributes to the effectiveness of this choice is the selection of the covariance matrix structure discussed previously. The covariance matrix is designed to incorporate both diagonal and off-diagonal elements, ensuring that the correlation between the memory and the observation parts is effectively learned in the off-diagonal part. On the contrary, if we were to utilize a pure diagonal covariance structure, it would not update the memory units (the later half) or their variance adequately during the Observation/Task/Kalman Update step. Thus, the second half of the latent state, i.e., $\cvec d_t$, serves as derivative or velocity units that can be used by the model to estimate the change of the observable part of the latent state.
\end{enumerate}

\notebox{In this thesis, we rely on the same factorization assumption for inference in all proposed formalisms to account for "memory" accumulated over time.}


\section{Deep Bayesian Aggregation.} 
To aggregate information from a set of observations into a consistent representation, \cite{volpp2020bayesian} introduce Bayesian aggregation in the context of Meta-Learning and Neural Process~\parencite{garnelo2018neural} literature. \begin{wrapfigure}[12]{r}{.35\linewidth}
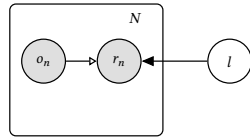

\begin{subfigure}[b]{0.31\textwidth}
   \centering
   \begin{adjustbox}{width=\textwidth}
         \tikzBA
     \end{adjustbox}
     \end{subfigure}
\caption{Generative model for the latent task variable ($l$) inference. The hollow arrows are deterministic transformations leading to an implicit distribution $r_{n}$ using a set encoder~\parencite{zaheer2017deep}.}
\label{fig:pgmba}
\end{wrapfigure} Since the aggregation scheme was introduced in the context of multitask latent variable models, we call this latent variable $l$ the task variable in this section. The generative model for the aggregation of observations $n$ is shown in Figure \ref{fig:pgmba}. \cite{volpp2020bayesian} derive a closed-form update rule for the posterior $p(\cvec l| \cvec r_{1:N})$ using the Bayes rule, given an observation model of the form, $$p(\cvec{r}_{n}|\cvec{l}) = \mathcal{N}\left(\cvec{r}_n|\cvec H \cvec{l},\textrm{diag}\left(\cvec{\sigma}_{n}\right)\right)$$ with $\cvec H = \cvec I$ and a prior $p(\cvec l) = \mathcal{N}(\cvec \mu_0, \textrm{diag}(\cvec \sigma_0))$. 
\par The Gaussian assumption allows us to obtain a closed-form solution for the posterior estimate of the latent task variable $p(\cvec{l}|\mathcal{\cmat{C}}_{\cvec{l}})$, based on Gaussian conditioning. The factorization assumption further simplifies this update rule by avoiding computationally expensive matrix inversions into a simpler update rule, as follows,
\begin{equation}
   \begin{aligned}
        \cvec{\sigma_{l}} = \left( \left(\cvec{\sigma}_{0}\right)^\ominus+ \sum_{n=1}^N \left(\cvec{\sigma}_n\right)^\ominus\right)^\ominus, \quad \cvec{\mu}_{\cvec{l}} = \cvec{\mu}_{0} + \cvec{\sigma}_{\cvec{l}} \odot \sum_{n=1}^N \left(\cvec{r}_n - \cvec{\mu}_{0}\right) \oslash \cvec{\sigma}_n.
    \end{aligned} 
    \label{eq:baeq}
\end{equation}

Here, $\ominus$, $\odot$, and $\oslash$ denote element-wise inversion, product, and division, respectively.

\subsection{Using Deep Set Encoders To Learn Observation Uncertainties} Note that the posterior calculation is based on the assumption that we have access to information on the latent (compact) observation $\cvec r_n$ and its corresponding uncertainty $\cvec \sigma_{o_n}$. We learn these using a Deep Set encoder as follows,
\begin{align}
    p(\cvec{r}_n|\cvec{l}) = \mathcal{N}\left(\cvec{r}_n|\cvec{l},\textrm{diag}\left(\cvec{\sigma}_{n}\right)\right), \quad \cvec{r}_n = \textrm{enc}_{\cvec{r}}\left(\cvec{x}_n\right), \quad \cvec{\sigma}_n = \textrm{enc}_{\cvec{\sigma}}\left(\cvec{x}_n\right).
\end{align}

\begin{figure}[h]
        \includegraphics[width=14cm]{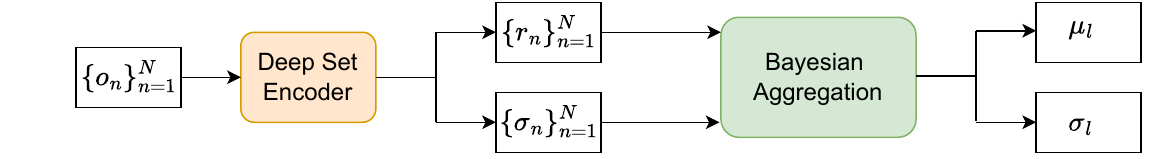}
    \caption{Given a set of N observations, the deep set encoder emits a latent representation for each of the observations and their corresponding uncertainty. The set of latent representations are then aggregated via Bayesian aggregation using update rules \ref{eq:baeq} to obtain a posterior over the abstract latent task variable $l$, $p(\cvec{l}|\mathcal{\cmat{o}}_{1:N})$. The architecture of the deep set encoder can vary depending on the type of sensory signals aggregated.}
    \label{fig:baarch}
\end{figure}

The architecture of the Bayesian Aggregation Scheme is shown in \ref{fig:baarch}.

\subsection{Bayesian Aggregation As Probabilistic Attention} Intuitively the mean of the latent task variable $\cvec{\mu}_{\cvec{l}}$ is a uncertainty weighted sum of the individual latent observations $\cvec{r}_n$, while the variance of the latent task variable $\cvec{\sigma}^2_{\cvec{l}}$ gives the uncertainty of the abstract representation of the task.  The derived aggregation rules can be thought of as simple "probabilistic attention," where the learned uncertainty about the individual sensory observations in the set encoder gives the attention weights. This is in line with ideas from computational neuroscience that treat cognitive attention to different sensory signals as a form of "precision weighting" \parencite{hohwy2013predictive,seth2014cybernetic}. The derived update equations have only a linear computational complexity of $O(N)$, while similar deep set operations (self-attention) in transformers~\cite{vaswani2017attention} have a complexity of $O(N^2)$.

Note that computing this posterior is a simplified case of the Kalman update rule used in Gaussian SSMs \parencite{becker2019recurrent}, with no memory units, $\cvec H 
 = \cvec I$ and no dynamics. 
\notebox{In this thesis, we explore the concept of Bayesian Aggregation to synthesize temporally abstract representations within generative world models. We also derive, in Chapter \ref{chap:mts3}, generic closed-form update equations for Bayesian aggregation that avoid the restrictive assumptions of observation models being identity matrices, as seen in \cite{volpp2020bayesian}.}

\chapter{Related Works}
In this section, I explore the existing literature on internal-world models from two primary fields: computational neuroscience and machine learning. This dissertation introduces scalable, end-to-end deep learning-based approaches for generative internal-world models, along with their respective inference and learning methods. These approaches aim to implement versions of various internal world models that computational neuroscientists have studied, adapting them within the framework of modern deep learning technologies. The discussions on computational neuroscience also offer insight into why certain representations were chosen for the Bayesian networks outlined in this dissertation from a neuroscientific perspective. 

\section{Computational Neuroscience Literature}
\label{sec:comp-neuro}

\subsection{The Predictive Bayesian Brain}
An increasingly popular theory in cognitive science claims that the brains are essentially prediction machines~\parencite{hohwy2013predictive}. The theory is commonly known as the Bayesian brain~\parencite{knill2004bayesian,polydoros2015real}, predictive processing~\parencite{clark2013whatever}, and predictive mind~\parencite{hohwy2013predictive}, among others. At its most fundamental, predictive brain theory says that perception is the result of the brain inferring the most likely causes of its sensory inputs by minimizing the difference between actual sensory signals and the signals expected on the basis of continuously updated predictive models. Arguably, predictive processing provides the most complete framework to date for explaining perception, cognition, and action in terms of fundamental theoretical principles and neurocognitive architectures~\parencite{seth2014cybernetic,jiang2021predictive}.
\paragraph{Helmholtz's Unconscious Inference}
\begin{wrapfigure}[26]{r}{0.35\textwidth}
  \centering
  \includegraphics[width=0.35\textwidth]{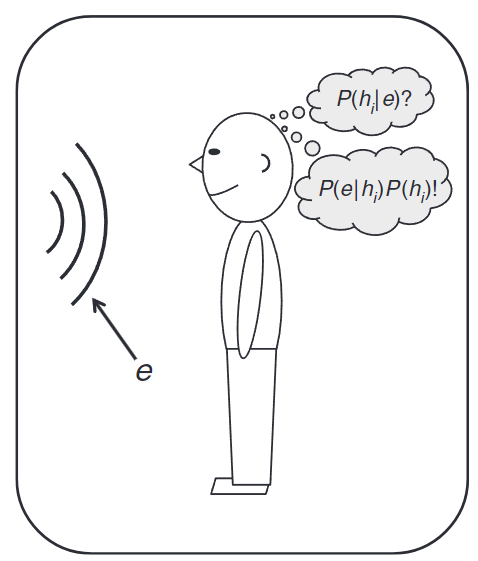}
  \caption{Shows a basic perceptual inference problem of figuring out what caused a sensory signal (sound e). The mechanism is analogous to the situation for the brain\parencite{hohwy2013predictive,seth2014cybernetic}. Given that the prior probability of hypothesis (cause) is $h_i: P\left(h_i\right)$ and the likelihood that the evidence e would occur, given $h_i$ is $P\left(e \mid h_i\right)$, the brain computes the posterior probability of hypothesis $h_i$, given the evidence $e: P\left(h_i \mid e\right)$ using internal generative models. The figure shows simplified version of Bayes' rule that puts it together: $P\left(h_i \mid e\right)=P\left(e \mid h_i\right) P\left(h_i\right)$.}
  \label{fig:predmind}
\end{wrapfigure}
\sloppy 
Bayesian Brain Hypothesis has its origins in Hermann von Helmholtz's notion of unconscious inference~\parencite{helmholtz1948concerning}, proposing that perception is a result of inferential processes by the brain, rather than direct imprinting of the external world onto our sensory apparatus. The observer is typically not aware of such prior assumptions but rather, they are incorporated by the neural circuits subconsciously to compute beliefs over hidden causes through the dynamics of neural activities (thereby implementing perception as “unconscious inference”). As part of the internal model, such priors can be expected to be adapted to the environment that the organism lives in. This idea laid the groundwork for the Bayesian brain hypothesis, which posits that the brain computes probabilities to infer the state of the world from ambiguous sensory inputs and prior beliefs, effectively engaging in a form of statistical reasoning or Bayesian inference.

\paragraph{Bayesian Brain Hypothesis}
The Bayesian brain hypothesis \parencite{knill2004bayesian} uses Bayesian probability
theory to formulate perception as a constructive process based on internal or generative models. The underlying idea is that the brain has a model of the world~\parencite{von2013treatise,mackay1956epistemological,neisser2014cognitive,gregory1968perceptual} that it tries to optimize using sensory inputs~\parencite{ballard1983parallel,kawato1993forward,kersten2004object,lee2003hierarchical,friston2005theory}. This idea is related to analysis by synthesis~\parencite{neisser2014cognitive} and epistemological automata~\parencite{mackay1956epistemological}. In this view, the brain is an inference machine that
actively predicts and explains its sensations\parencite{von2013treatise,gregory1980perceptions,dayan1995helmholtz}. Central to this hypothesis is a probabilistic model that can generate predictions against which sensory samples are tested to update beliefs about their causes. This generative model is decomposed into a likelihood (the probability of sensory data given their causes) and a 
prior (the a priori probability of those causes). Perception then becomes the process of inverting the likelihood model (mapping from causes to sensations) to access the posterior probability of the causes, given sensory data (mapping from sensations to causes).

\subsection{Kalman Filters As Internal World Models / Spatiotemporal Predictive Coding}

The world is dynamic; most of the time, animals receive time-varying stimuli either due to their own movement or due to other moving objects in the environment. This makes the ability to predict future stimuli essential for survival (e.g. predicting the location of predators). Rao's seminal work~\parencite{rao1999optimal} from the computational neuroscience literature introduces a foundational mathematical framework that underpins how organisms perceive their environment in a dynamic setting as opposed to static settings discussed previously. This framework posits that visual perception is a stochastic, dynamic process and the task of perception is one of optimally estimating (in a Bayesian sense) the causes of visual events and on a longer time scale, learning efﬁcient spatio-temporal internal models of the visual environment. The spatiotemporal internal generative model is elegantly represented as a stochastic linear dynamical system similar to a Kalman Filter with transition and observation matrices. By employing inference techniques rooted in Kalman filter update equations, Rao's approach provides a Bayesian optimal method for estimating the internal states that govern the dynamics of the mental world model.

\begin{figure}[h]
    \centering
    \includegraphics[width=11cm]{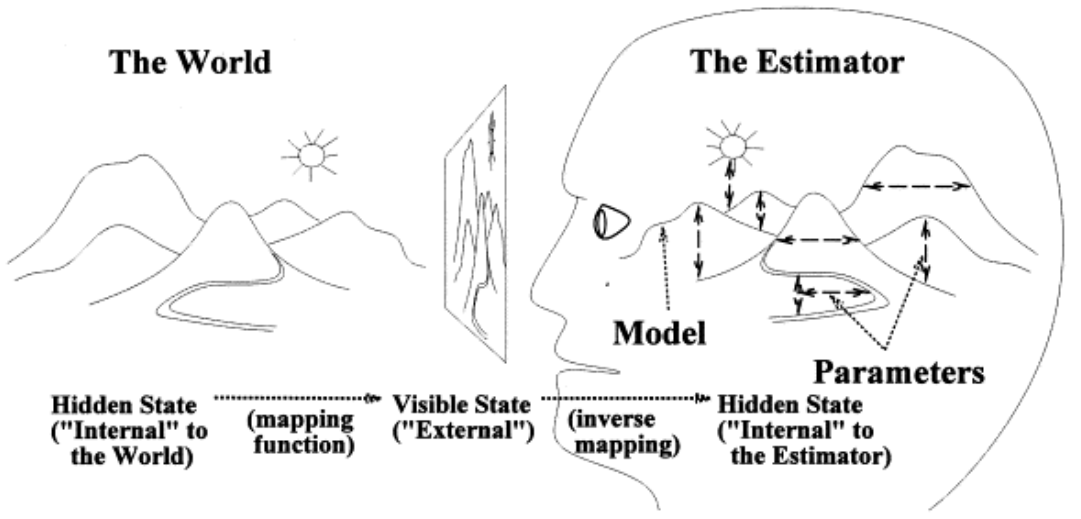}
    \caption{Figure from \cite{rao1999optimal} presents the challenge organisms face in perceiving the external world through internal models. Without direct access to the world's hidden internal states, organisms rely on sensory measurements to estimate these states, solving the 'inverse' problem to interpret and understand their environment. Though the paper focused on visual systems, the authors argue that the concept extends beyond visual and auditory senses to include internal models for motor systems, emphasizing the broader application of understanding and interacting with the external world through internal representations.}
    \label{fig:galaxy}
\end{figure}

\paragraph{Combining Top-Down Predictions With Bottom-Up Sensory Signals Using Kalman Filter Updates}
The proposed Kalman-filter model in \cite{rao1999optimal} provides a mathematical framework that demonstrates how top-down expectations (informed by the internal generative model) interact with bottom-up signals (the current sensory input) to produce a robust estimation of the visual environment. The Kalman Filter combines these two sources of information, weighted according to their reliability (inverse variances or “precisions”), to compute the posterior beliefs about the world in a Bayesian optimal manner, which is interpreted as a form of attention. The proposed model allows for the possibility that the organism or agent might want to perform internal simulations of the dynamics of the external world (e.g., for planning) by predicting how future states evolve given a starting state (and possibly actions).

\paragraph{Learning Of Internal Models} The work also derived update rules for learning the transition and precision matrices online based on incoming input data, as opposed to hand-coded dynamics models.

\cite{rao1999optimal} further provides experimental results that support the model's capability for learning dynamic and static visual stimuli, recognizing objects under various conditions, and segmenting scenes into constituent elements. The work from \cite{rao1999optimal}, also referred to as "spatio-temporal predictive coding" in literature, thus sets a foundational stone for exploring the synergies between computational neuroscience and the development of popular machine learning architectures like Deep Kalman Models. 

\notebox{The neuroscientific insights discussed above are exploited in Chapter \ref{chap:Acssm} of the thesis using modern deep learning techniques.}

\subsection{Adaptive Behavior In Brain With Multiple Internal Models}
 \begin{wrapfigure}[28]{r}{0.35\textwidth}
  \centering
  \includegraphics[width=0.35\textwidth]{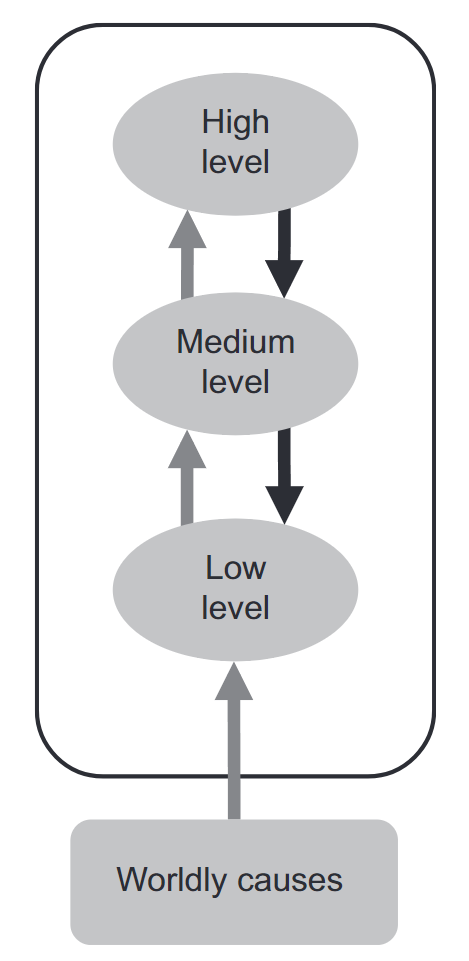}
  \caption{\textbf{The perceptual hierarchy}~\parencite{hohwy2013predictive}: Processing of causal regularities at different time scales influence each other in a bottom-up–top-down fashion. Sensory input (dark grey arrows pointing up) is met with prior expectations (black arrows pointing down) and perceptual inference is determined in multiple layers of the hierarchy simultaneously, building up a structured representation of the world. This figure simplifies greatly as there are of course not just three well-defined hierarchical levels in the brain. }
  \label{fig:hiermind}
\end{wrapfigure} 
One of the most notable abilities of biological creatures is their capacity to adapt their behavior to different contexts and environments (i.e., cognitive
flexibility) through learning.People can learn to call on various responses depending on the situation—for example, independently move the right and left hands when playing an instrument and speak several different languages. Such multitasking abilities are particularly crucial in communication with several people, who each demand subtly different forms of interaction~\parencite{taborsky2012social,parkinson2015repurposed} and is a key component of exhibiting social intelligence. The work from \cite{Isomura2019} attempts to understand how the brain attains cognitive flexibility in different contexts by entertaining the possibility of distinct generative models in a context-sensitive setting. \cite{Isomura2019} further present the results of perceptual learning and inference using this form of model selection or structure learning, predicated on an ensemble of generative models.  Using this setup, it is shown that Bayesian model averaging provides a plausible account of how multiple hypotheses can be combined to predict the sensorium, while Bayesian model selection enables perceptual categorization and selective learning. 

\notebox{The generative model proposed in Chapter \ref{chap:Hipssm} of the dissertation have a similar motivation of using a "set of SSMs/world models" for adaptive behaviour to changing tasks.}

\subsection{Hierarchical Generative Models In The Brain}
Researchers in computational neuroscience argue that the previously discussed internal predictive world models are hierarchical in nature~\parencite{friston2008hierarchical,hohwy2013predictive,seth2014cybernetic}. The hierarchical nature adeptly mirrors the complex and layered structure of the world itself, capturing the essence of how causes and effects interrelate across different spatial and temporal scales~\parencite{hohwy2013predictive}.
\par
These hierarchical generative models overturn classical notions of perception that describe a largely 'bottom-up' process of evidence accumulation or feature detection. Instead, predictive processing proposes that perceptual content is determined by top-down predictive signals emerging from multilayered and hierarchically organized generative models of the causes of sensory signals~\parencite{lee2003hierarchical}. These models are continually refined by mismatches (prediction errors) between predicted signals and actual signals across hierarchical levels, which iteratively update predictive models via approximations to Bayesian inference. This means that the brain can induce accurate generative models of hidden environmental causes by operating only on signals to which it has direct access: predictions and prediction errors. It also means that even low-level perceptual content is determined via cascades of predictions flowing from very general abstract expectations, which constrain successively more fine-grained predictions. 
\notebox{The hierarchical generative model proposed in Chapter \ref{chap:mts3} takes inspiration from these insights.}

\section{Machine Learning Literature}

We give a brief overview of notable related works in learning computational mental models of the world in the field of machine learning. A detailed discussion on several of these are provided in Sections \ref{rw:acssm}, \ref{rw:hipssm} and \ref{rw:mts3}.

\subsection{World Models Based On Gaussain Processes}

Bayesian nonparametric models, such as Gaussian processes (GPs), are often the dynamics model of choice in Model-Based RL, especially in low-dimensional problems where data efficiency is critical due their robustness in handling uncertainty~\parencite{kocijan2004gaussian,ko2007gaussian,nguyen2008local,grancharova2008explicit,deisenroth2013gaussian,kamthe2018data}.
\begin{figure}[h]
    \centering
    \includegraphics[width=12cm]{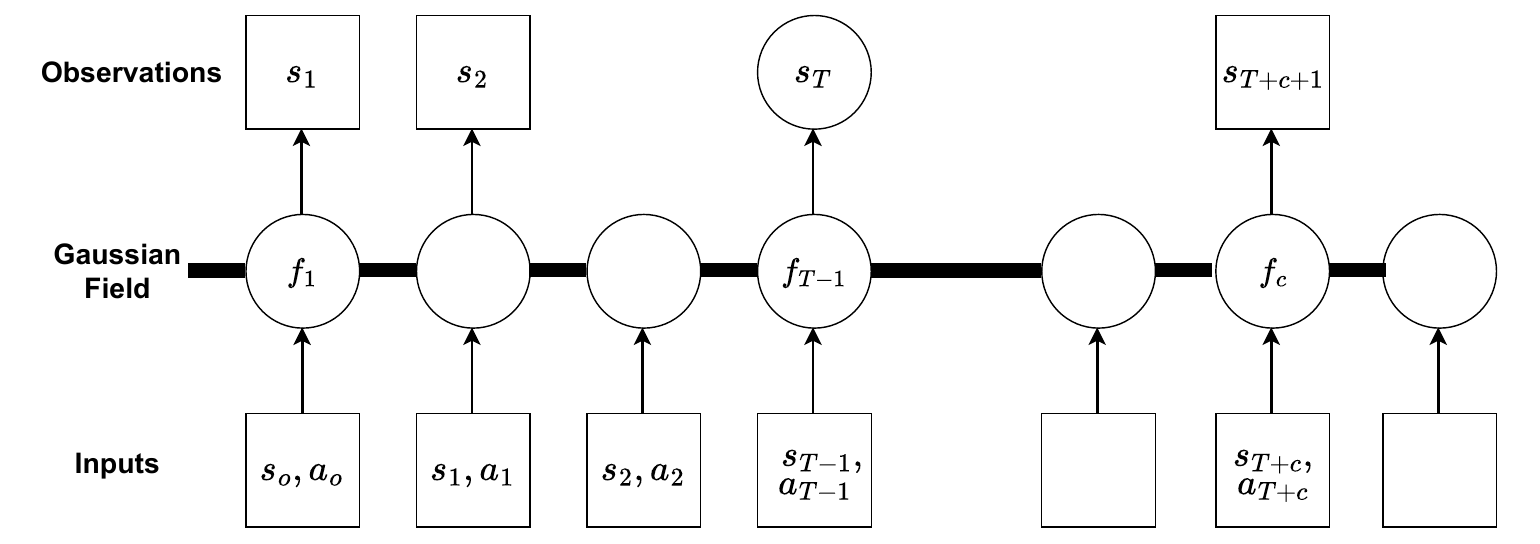}
    \caption{PGM representation adapted from \cite{rasmussen2003gaussian} of a GP regression for learning dynamics. Squares represent observed variables, and circles represent unknowns. The thick horizontal bar represents a set of fully connected nodes and indicates that the forward dynamic function $f_i$ belongs to a Gaussian field. During posterior inference $f_i$'s act as random variables and are integrated out, which means that every prediction $y_*$ depends on all the other inputs and observations (including the training data) making GPs often computationally challenging.}
    \label{fig:gp}
\end{figure}

However, such models introduce additional assumptions on the system, such as the smoothness assumption inherent in GPs with squared-exponential kernels~\parencite{rasmussen2003gaussian}. Also, these methods do not scale to high-dimensional environments. Furthermore, these models are inherently single-time scale and are not widely used in tasks where long-horizon predictions and planning are required.

\subsection{World Models Based On Neural Networks}
The advancements in parametric function approximators, such as neural networks (NNs), brought about by the advent of deep learning, have led to the development of a series of world models employing multi-layer perceptrons. These models are cited in several works, including \cite{baranes2013active}, \cite{fu2016one}, \cite{punjani2015deep}, \cite{lenz2015deepmpc}, \cite{gal2016improving}, \cite{depeweg2016learning}, \cite{williams2017information}, and \cite{nagabandi2018neural}. In contrast to Gaussian processes, these deep learning-based models offer constant-time inference and manageable training with large datasets. They also possess the capability to model more intricate functions, including the non-smooth dynamics frequently encountered in robotics~\parencite{fu2016one,nagabandi2018neural}.\begin{wrapfigure}[22]{r}{0.5\textwidth}
\hspace{-1cm}
  \includegraphics[width=0.65\textwidth]{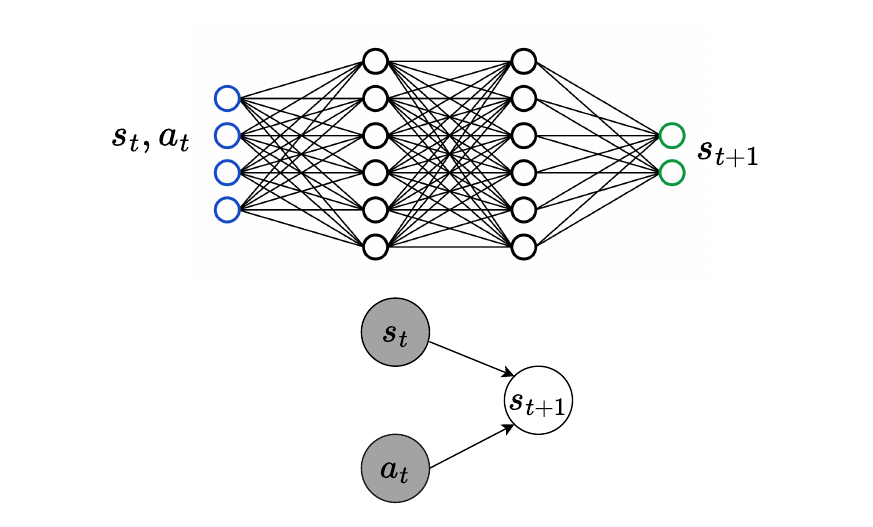}
  \caption{(Top) A dynamics model based on deep neural networks. (Bottom) A graphical representation using PGM of the same model, where the shaded nodes indicate observed variables. These models operate under the assumption that the current states of the internal model of the world are already known and don't need to be inferred. This assumption is limiting, particularly when these models need to be derived from sensory information that is both high-dimensional and fraught with noise.}
  \label{fig:nndynamics}
\end{wrapfigure} However, the majority of these studies utilizing NNs have concentrated on deterministic models, which do not align well with the Bayesian perspective of best guesses discussed earlier. \cite{chua2018deep} proposed utilizing an ensemble of neural network models to learn probabilistic dynamics, capturing both the aleatoric uncertainty inherent in the environment and the epistemic uncertainty of the data. This approach allows for more data-efficient control and planning algorithms. Nonetheless, irrespective of being deterministic or stochastic, these NN-based dynamics models face challenges in learning with high-dimensional sensory data or in partially observable domains due to the presumption of known internal world states. Additionally, for long-term predictions, these auto-regressive models are less effective because of the accumulation of prediction errors inherent in autoregressive processes. They also presume static world dynamics and operate at a singular, fine-grained temporal scale, thereby lacking in their ability to serve as a general-purpose world model capable of addressing partial observability, non-stationarity, and hierarchical predictions across multiple time scales.

\subsection{World Models Based On State Space Models}
Classical State-Space Models (SSMs) are well-regarded for their clear inference processes and interpretable outcomes, but they often struggle with scalability and efficiency when applied to high-dimensional data or large datasets. To overcome these limitations, recent advances have introduced deep SSMs, which integrate neural networks to provide efficient and scalable inference for complex datasets. The works of \cite{haarnoja2016backprop, becker2019recurrent} have leveraged neural network architectures to derive closed-form solutions for the forward inference algorithms in SSMs, leading to state-of-the-art performance in tasks like state estimation and dynamics prediction. Other studies, such as those of \cite{krishnan2017structured, karl2016deep, fraccaro2017disentangled, hafner2019learning, becker2022uncertainty}, have focused on applying variational approximations to facilitate learning and inference within SSMs.

Despite these advances, many recurrent state-space models are based on the assumption of fixed dynamics, which does not align with the changing conditions observed in real-world scenarios, such as in robotics. To address the limitations of fixed dynamics, research by \cite{linderman2017bayesian, becker2019switching, dong2020collapsed} has introduced models that incorporate additional discrete switching latent variables to account for changing or multi-modal dynamics. However, these models typically operate on a single time scale, with both continuous and discrete levels operating at a fine-grained temporal resolution. They also do not adhere to a strictly top-down approach, as the top-layer latent variables depend on the predictions from the lower layer, rendering these models not suitable for long-horizon predictions and reasoning. Moreover, the introduction of discrete states complicates the learning and inference processes.
\label{chap:related}
\chapter{Action Conditional Deep SSMs: Probabilistic World Models on a Single Time Scale}
\label{chap:Acssm}
\begin{chapquote}{}
This chapter is based on "Action Conditional Recurrent Kalman Networks for Forward and Inverse Dynamics Learning"~\parencite{shaj2020action}.
\end{chapquote}
In our quest to formalize a computational mental model of the world, we start with a simple and widely used Bayesian network called the State Space Models (SSMs). Though it operates at a single time scale (often the time scale at which observations are recorded), and hence may not satisfy the expressiveness and hierarchical structure of the mental models that the Neuroscience community advocates (Section \ref{sec:comp-neuro}), SSMs have several computational advantages. The chain structure of SSMs allows for tractable exact inference via message passing allowing for closed-form update rules under linear and Gaussian assumptions. When these updates are performed in the learnt latent space of a Deep Network, these linear assumptions are no longer limiting. These deep models learn a latent space where these linear assumptions work in practice. The Deep SSMs with exact inference are known in the literature as Deep Kalman models and we use one such model namely Recurrent Kalman Networks (RKN)~\parencite{becker2019recurrent} discussed in the preliminaries Section \ref{sec:prelimrkn} as our foundation for learning a world model.

\textbf{Research Objective} Deep Kalman Models including RKN were primarily designed for state estimation from sensory observations in simple unactuated dynamical systems. In the real world, an agent's actions actively shape its environment, and ignoring this causal relationship in the mental models of the world can lead to inaccurate predictions and suboptimal decision-making. Hence we try to answer these questions in this chapter of the thesis:

\begin{enumerate}
    \item Can deep Kalman models be adapted for the task of world modelling?
    \item How to incorporate control action conditioning in the latent dynamics in a principled manner such that causal relations are respected?
    \item How good are these models in modelling the dynamics of complex robotic systems when faced with non-markovian, partially observable and non-stationary settings?
\end{enumerate}

\section{Related Works}
\label{rw:acssm}
\subsection{Action Conditional Predictive Models.} In model-based control and reinforcement learning (RL) problems, learning to predict future states and observations conditioned on actions is a key component.
Approaches such as~\cite{deisenroth2011pilco}, \cite{nagabandi2018neural}, \cite{lenz2015deepmpc}, \cite{oh2015action}, \cite{finn2016unsupervised} try to achieve this by using traditional models like GPs, feed forward neural networks, or LSTMs. 
The action conditioning is realized by concatenating the input observations and action signals~\parencite{nagabandi2018neural,deisenroth2011pilco} or via factored conditional units~\parencite{taylor2009factored}, where features from some number of inputs modulate multiplicatively and are then weighted to form network outputs. 
In each of these approaches action conditioning happens outside of the recurrent neural network cell which leads to sub-optimal performance as observations and actions are treated similarly.

\subsection{Predictive Models Of Robotic Agents.} Due to the increasing complexity of robot systems, analytical models are more difficult to obtain. This problem leads to a variety of model estimation techniques which allow the roboticist to acquire models from data.
When learning the model, mostly standard regression techniques with Markov assumptions on the states and actions are applied to fit either the forward or inverse dynamics model to the training data.
Authors used Linear Regression~\parencite{schaal2002scalable,haruno2001mosaic}, Gaussian Mixture Regression~\parencite{calinon2010learning,khansari2011learning}, Gaussian Process Regression~\parencite{nguyen2009model,nguyen2010using}, Support Vector Regression~\parencite{ferreira2007simulation}, and feed forward neural networks~\parencite{polydoros2015real}. 
However the Markov assumptions are violated in many cases, e.g., for learning the dynamics of complex robots like hydraulically and pneumatically actuated robots or soft robots.
Here, the Markov assumptions for states and actions no longer hold due to hysteresis effects and unobserved dependencies. 
Similarly, learning the dynamics of robots in frictional contacts with unknown objects also violates the Markov assumption and requires reliance on recurrent models which can take into account the temporal dynamic behavior of input sequences while making predictions. 
Recently, using RNNs such as LSTMs or GRUs has been attempted. 
Those scale easily to big data, available due to the high data frequencies, and allow learning in $O(n)$ time~\parencite{rueckert2017learning}. 
However, the lack of a principled probabilistic modelling and action conditioning makes them less reliable to be applied for robotic control applications.   

\section{Action Conditional Recurrent Kalman Networks}

We extend the Recurrent Kalman Network approach to the task of world modelling / forward dynamics learning with the following innovations: \textbf{(i)} We use an \textbf{action-conditional prediction} update which provides a natural way to incorporate control inputs into the latent dynamical system.  \textbf{(ii)} We modify the architecture to perform prediction instead of filtering and learn \textbf{accurate forward dynamics models} for multiple robots with complex actuator dynamics. We refer to the resulting approach as action-conditional RKN (ac-RKN).

\subsection{Action Conditioning}
\label{sec:action-con}

\begin{figure}
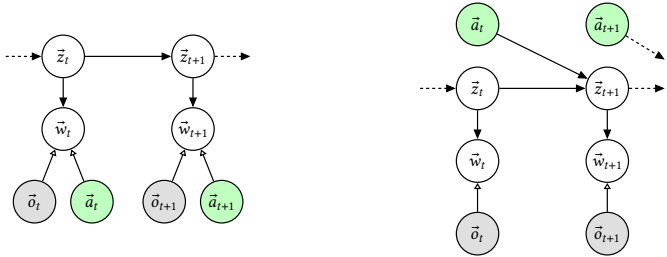

\hspace{0.1\textwidth}
  \begin{minipage}{0.32\textwidth} 
    \centering
     \resizebox{\textwidth}{!}{\tikzRKNGM}
  \end{minipage}
  \hspace{0.2\textwidth}%
  \begin{minipage}{0.32\textwidth} 
     \resizebox{\textwidth}{!}{\tikzACRKNGM}
  \end{minipage}
  \caption{Graphical models for actions (green). (Left) Treated as fake observations as in \cite{becker2019recurrent}. (Right) ac-RKN treating actions in a principled manner by capturing its causal effect on the state transition via additive interactions with the latent state. The hollow arrows denote deterministic transformation leading to implicit distributions.}
  \label{fig:acVsrk}
\end{figure}

To achieve action conditioning within the recurrent cell, we modify the transition model proposed in \cite{becker2019recurrent} to include  a control model $\vec{b}(\vec a_t)$ in addition to the locally linear transition model $\vec{A}_t$. 
The control model $\vec{b}(\vec a_t)$ is added to the locally
linear state transition, i.e., $\vec{z}^-_{t+1} = \vec{A}_t \vec{z}^+_{t} + \vec b(\vec a_t)$. 
As illustrated in \autoref{fig:acVsrk}, this formulation of latent dynamics captures the causal effect of the actions variables $\vec{a}_t$ on the state transitions in a more principled manner than including the action as an observation, which is the common approach for action conditioning in RNNs \parencite{becker2019recurrent}. 
The control model $\vec{b}(\vec a_t)$ can be represented in several ways, i.e.:\begin{itemize}[leftmargin=0.5cm]
\item[\textbf{(i)}] \textbf{Linear:} $\vec{b}_{l}(\vec a_t) =\vec{B} \vec a_t$, where $\vec{B} $ is a linear transformation matrix.
\item[\textbf{(ii)}] \textbf{Locally-Linear:} $\vec{b}_{m}(\vec a_t) = \vec{B}_t \vec a_t$, where $\vec{B}_t = \sum_{k=0}^K \beta^{(k)}(\vec{z_t}) \vec{B}^{(k)}$ is a linear combination of k linear control models $\vec{B}^{(k)}$. A small neural network with softmax output is used to learn $\beta^{(k)}$.  
\item[\textbf{(iii)}] \textbf{Non-Linear:} $\vec{b}_n(\vec a_t) = \vec{f}(\vec a_t)$, where $\vec{f}(.)$ can be any non-linear function approximator. We use a multilayer neural network regressor with ReLU activations.
\end{itemize} 

Note that the prediction step for the variance is not affected by this choice of action conditioning, i.e., $\vec{\Sigma}^-_{t+1} = \vec{A}_t \vec{\Sigma}_{t}^+ \vec{A}_t^T + \vec{I} \cdot \vec{\sigma^\mathrm{trans}}$ as the action is known and not uncertain.
Thus, unlike the state transition model $\vec{A}_t$, we do not need to constrain the model $ \vec{b}$ to be linear or locally linear, as it neither affects the Kalman gain nor how the covariances are updated. 
Hence, we use the nonlinear approach, $\vec{b}_n(\vec a_t)$, as it provides the most flexibility and achieves the best performance, as shown in Section \ref{sec:acssmExp}.
The Kalman update step is also unaffected by the new action-conditioned prediction step and therefore remains as presented in~\cite{becker2019recurrent}.


Our principled treatment of control signals is crucial for learning accurate forward dynamics models, as detailed in Section \ref{subsec:ac-world}. Moreover, the disentangled representation of actions in the latent space gives us more flexibility in manipulating the control actions for different applications including inverse dynamics learning (Appendix \ref{sec:inverseDyn}), extensions to model-based reinforcement learning, and planning. For long-term prediction using different action sequences, we can still apply the latent transition dynamics without using observations for future time steps (skipping the observation update step).

\subsection{Ac-RKN as Single Time Scale Probabilistic World Models}
\label{subsec:ac-world}
\begin{figure}[h]
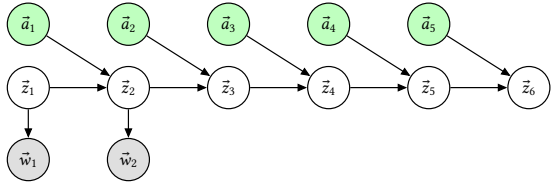

    \centering
    \resizebox{0.7\textwidth}{!}{\tikzACRKNGMForward}
    \caption{Using internal world models to envision the future: With past and present observations ($w_{1:2}$) as a basis, Ac-RKN projects future world states ($z_{3:6})$ by considering a sequence of action signals ($a_{1:5}$).}
    \label{fig:galaxy}
\end{figure}
World models attempt to learn a compact and expressive representation of the environment dynamics from observed data. These models can predict possible future world states and associated uncertainties as a function of an imagined action sequence. More formally, we want to learn a forward dynamics model of the world $f: \vec o_{1:t}, \vec a_{1:t} \mapsto \vec o_{t+1}$ that predicts the next observation $\vec o_{t+1}$ given the histories of observation $\vec o_{1:t}$ and actions $\vec a_{1:t}$. 

Formulating action-conditional future predictions as inference tasks within a Gaussian State Space Model (SSM) in deep latent spaces enables the development of probabilistic world models that exhibit the following properties.
\begin{enumerate}
    \item \textbf{The propagation of uncertainties} ($\cvec \Sigma_t$) about the belief of our future world states in addition to the mean estimates ($\cvec{z}_t$).
    \item \textbf{Learning under partial observability} and missing observations. The RKN cell is well equipped to handle missing observations, as the update step can just be omitted in this case, i.e. the posterior is equal to the prior if no observation is available. In this case, we repeatedly apply the ac-RKN prediction step while omitting the Kalman update. 

    \item \textbf{Capturing non-markovian dynamics} The model captures non-markovian dynamics in a principled way, since we predict the future belief states $p(\vec z_{t+1}| \vec w_{1:t}, \vec a_{1:t})$ conditioned on all the past observations and the control actions given so far. These beliefs are obtained in closed form using the forward inference algorithm (action conditional Kalman predict steps).
\end{enumerate}  

In line with theories from neuroscience discussed in Section \ref{sec:comp-neuro}, this results in a generative model that makes Bayesian best guesses about the world states based on the action conditional prior beliefs (top-down predictions) and bottom-up sensory signals/ observations when available. Neuroscience research~\parencite{hohwy2013predictive,seth2014cybernetic,seth2021being} similarly posits that the Brain is not only tasked with the challenge of figuring out the most likely causes of its sensory signals but also figuring out how relevant the sensory inputs are. Similarly, the Ac-RKN model learns to estimate the reliability of the incoming sensory information by predicting uncertainties of the sensory signals via the observation encoder (bottom-up processing). For a visual representation of these top-down and bottom-up predictions in the Ac-RKN, please see Figure \ref{fig:acrkn-block}.

\subsection{End To End Learning Via Backpropagation}
\label{subsec:trainingac}
The network is tasked to minimize the prediction errors by maximizing the posterior predictive log-likelihood, which is given below for a single trajectory, i.e., 
\begin{align}
\label{eq:objective}
     L & =\sum_{t=1}^H \log p(\cvec{o}_{t+1}|\cvec{w}_{1:t}, \cvec{a}_{1:t}) \nonumber \\ & = \sum_{t=1}^H \log \int p(\cvec{o}_{t+1}|\cvec{z}_{t+1})  p(\cvec{z}_{t+1}|\cvec{w}_{1:t}, \cvec a_{1:t}) d\cvec{z}_{t+1} \nonumber \\  
\end{align}
The extension to multiple trajectories is straightforward and omitted to keep the notation uncluttered. Here, $\cvec{o}_{t+1}$ is the ground truth observations at the time step $t+1$ which needs to be predicted from all observations up to time step $t$. 
\par
\subsubsection{Approximating the likelihood} \sloppy We employ a Gaussian approximation of the posterior predictive log-likelihood of the form $ p(\cvec{o}_{t+1}|\cvec{w}_{1:t}, \cvec{a}_{1:t}) \approx \mathcal{N}(\cvec{\mu}_{\cvec{o}_{t+1}},\textrm{diag}(\cvec{\sigma}_{\cvec{o}_{t+1}}))$ where we use the mean of the prior belief $\cvec{\mu}_{z_{t+1}}^-$ to decode the predictive mean, i.e, $\cvec{\mu}_{\cvec{o}_{t+1}} =  \textrm{dec}_{\cvec{\mu}}(\cvec{\mu}_{z_{t+1}}^{-})$ and the variance estimate of the prior belief to decode the observation variance, i.e., $\cvec{\sigma}_{o_{t+1}} = \textrm{dec}_{\sigma}(\cmat{\Sigma}_{z_{t+1}}^{-})$. This approximation can be motivated by a moment-matching perspective and allows for end-to-end optimization of the log-likelihood without using auxiliary objectives such as the ELBO \parencite{becker2019recurrent}. Thus the approximate Gaussian predictive log-likelihood for a single sequence is then computed as  
\begin{align}
\label{eq: likeli-acrkn}
&\mathcal{L}\left(\cvec{o}_{(1:T)}\right) = \\&
\dfrac{1}{T} \sum_{t=1}^T  \log \mathcal{N}\left(\cvec{o}_t \bigg| \textrm{dec}_{\mu}(\cvec{z}_t^+), \textrm{dec}_{\Sigma} (\cvec{\sigma}^{\mathrm{u},+}_t,
\cvec{\sigma}^{\mathrm{s},+}_t,
\cvec{\sigma}^{\mathrm{l},+}_t )\right), \nonumber
\end{align}
where $\textrm{dec}_{\mu}(\cdot)$ and $\textrm{dec}_{\Sigma}(\cdot)$ denote the parts of the decoder that are responsible for decoding the latent mean and latent variance respectively.
\begin{figure}[h]
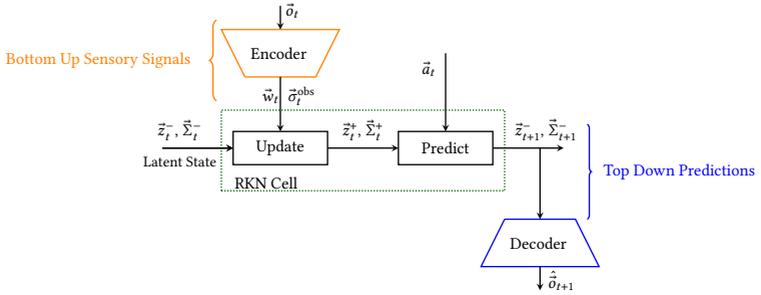

    \centering
    \resizebox{0.98\textwidth}{!}{\tikzACRKN}
    \caption{Schematic diagram of ac-RKN for forward dynamics learning. Action conditioning is implemented by adding a latent control vector $\vec b(\vec a_t)$ to the RKN dynamics model. The output of the prediction stage, which forms the prior for the next time step $\left(\vec{z}_{t+1}^-, \vec{\Sigma}_{t+1}^- \right)$ is decoded to obtain the prediction of the next observation. \label{fig:forward_model}}
    \label{fig:acrkn-block}
\end{figure}
\subsubsection{Variations Of The Learning Objective for Learning Real Robot Dynamics}
In this part of the thesis, we specifically look at the learning challenges associated with modeling the dynamics of real robot systems and propose different learning objectives. We explicitly focus on non-Markovian systems, where the observations are typically joint angles and, if available, joint velocities of all degrees of freedom of the robot. 
In our experiments, we aim to predict the joint angles at the next time step given a sequence of joint angles and control actions. 
Due to unmodeled effects such as hysteresis or unknown contacts, the state transitions are non-Markovian even if joint angles and velocities are known. We can assume that the observations are almost noise-free, since the measurement errors for our observations (joint angles) are minimal. Nevertheless, as the observations do not contain the full state information of the system, we still have to model uncertainty in our latent state using the RKN.
\begin{enumerate}
    \item [1.] \textbf{Mean Predictions} To ensure a fair comparison with deterministic world models such as LSTM and GRU, we train the model with a variation of the objective in Equaton \ref{eq: likeli-acrkn}, where only the mean of the latent states are decoded, while the variance units are kept fixed, resulting in an RMSE loss.
    \begin{align*}
    \mathcal{L}_{\textrm{fwd}} = \sqrt{
    \dfrac{1}{T}\sum_{t=1}^T  \parallel \left(\vec{o}_{t+1} -  \vec{o}_{t}\right) - \textrm{dec}\left(\vec{z}^-_{t+1}\right) \parallel^2}.
    \end{align*} 
    We also observed that in use cases where only mean predictions are required, this results in slightly better predictions. 

    \item [2.] \textbf{High-frequency Predictions} The function $f$ can be challenging to learn when the observations, i.e. joint positions and velocities, of two subsequent time steps are too similar (for example, our Franka Robot's operating frequency in 1000Hz). In this case, the action has seemingly little effect on the output, and the learned strategy is often to copy the previous state for the next-state prediction. This difficulty becomes more pronounced as the time step between states becomes smaller, e.g., $1$ms, as minor errors in absolute states estimates can already create unrealistic dynamics.  Therefore, a standard method for model learning is to predict the normalized differences between subsequent observations or states instead of the absolute values~\cite{deisenroth2011pilco}, that is, during training, the next predicted observation is $\hat{\vec o}_{t+1} = \vec o_t + \textrm{dec}\left(\vec{z}_{t+1}^-\right)$, where $\vec{o}_t$ is the true observation in $t$ and $\textrm{dec}\left(\vec{z}_{t+1}^-\right)$ is the output of the actual decoder network. During inference, we introduce an additional memory $\vec{m}_t$ that stores the last prediction and uses it as an observation in case of a missing observation, that is, $\vec{\hat{o}}_{t+1} = \vec{m}_t + \textrm{dec}\left(\vec{z}_{t+1}^-\right)$ where we use the true observation, $\vec{m}_t= \vec{o}_t$ if available, and the predicted observation, $\vec{m}_t= \hat{\vec{o}}_t$, otherwise. 
\end{enumerate}
\notebox{Irrespective of the learning objectives/loss functions used, we still have to model uncertainty in our latent state using the RKN as the observations do not contain the full state information of the system.}
The architecture of the ac-RKN is summarized in Figure \ref{fig:forward_model}.

Note that we decode the prior mean $\vec{z}^-_{t+1}$ to predict $\vec o_{t+1}$. 
The prior mean $\vec{z}^-_{t+1}$ integrates all information up to time step $t$, including $\vec a_t$, but already denotes the belief for the next time step. 
Hence, as we are interested in prediction, we work with this prior.
On the contrary, \cite{becker2019recurrent} used posterior belief, as their goal was filtering rather than prediction. 

\subsection{Self Supervised Training For Multi-Step Prediction \label{sec:selfsup}} 
\begin{figure}[H]
\centering
\includegraphics[width=0.7\linewidth]{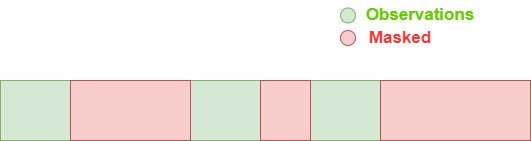}
\caption{Self-supervised training for multi-step ahead predictions by tasking the model to "fill in" the masked observations.}
\label{fig:self-super-ac}
\end{figure}
Using the training objective(s) discussed in Section \ref{subsec:trainingac} results in models that are good in one-time step prediction, but typically perform poorly in long-term predictions as the loss assumes that observations are always available up to time step $t$. To increase the performance of the long-term prediction, we can treat the long-term prediction problem as a case of the problem of ``missing value'', where the missing observations occur in future time steps. Thus, to train our model for long-term prediction, we randomly mask a fraction of observations as shown in Figure \ref{fig:self-super-ac} and explicitly task the network to impute the missing observations, resulting in a strong self-supervised learning signal for long-term prediction with varying prediction horizon length.

\section{Experiments}
\label{sec:acssmExp}
In this section, we discuss the experimental evaluation of ac-RKN on learning forward dynamics models on robots with different actuator dynamics. A full listing of hyperparameters can be found in the Supplementary Material. We compare ac-RKN to both standard deep recurrent neural network baselines (LSTMs, GRUs and standard RKN), analytical baselines based on classical rigid body dynamics, and nonrecurrent baselines like FFNN. For the RKN and LSTM baselines, we replaced the ac-RKN transition layer with generic LSTM and RKN layers. For recurrent models, the parameters of the recurrent cell are tuned through hyperparameter optimization using GPyOpt~\parencite{gpyopt2016}, but the size of the encoder and decoder are similar. The observations and actions are concatenated as in \cite{nagabandi2018neural} and \cite{deisenroth2011pilco} for all baseline experiments during the forward dynamics learning, that is, we treat actions as extended observations. The data, i.e., the states or observations, actions, and targets, are normalized (zero mean and unit variance) before training. We denormalize the predicted values during inference and evaluate their performance in the test set. We evaluate the model using RMSE to ensure a fair comparison with deterministic models such as FFNNs and LSTMs. 

\begin{figure*}[h!t!]
\centering
\includegraphics[width=4.1cm,height=3.5cm]{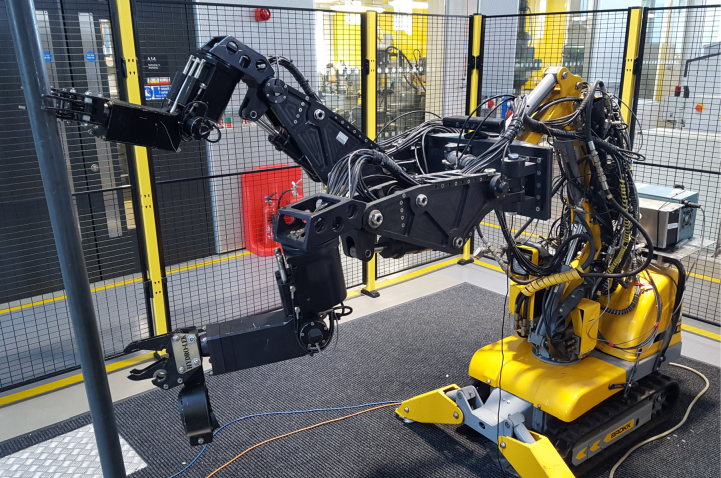}
\includegraphics[width=4.1cm,height=3.5cm]{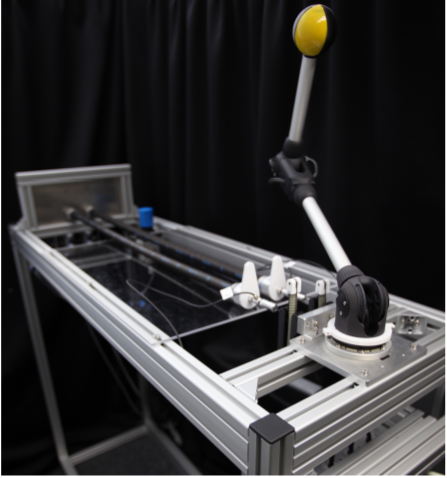}
\includegraphics[width=4.1cm,height=3.5cm]{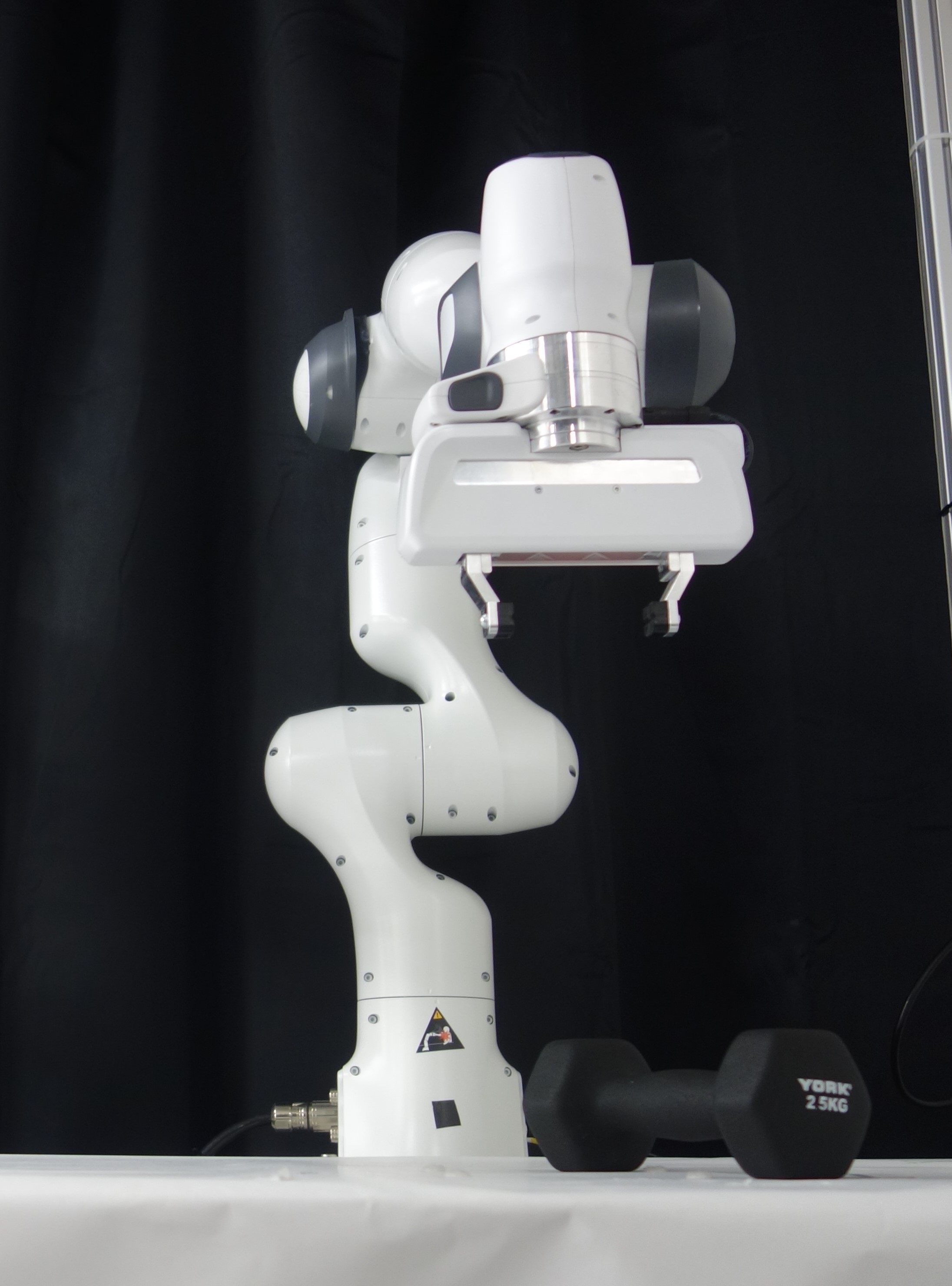}
\vspace{0.1cm}  
\caption{The experiments are performed on data from robots with different actuator dynamics. From left to right, these include: Hydraulically actuated BROKK-40 \parencite{taylor2013state}, pneumatically actuated artificial muscles \parencite{buchler2016lightweight}, Franka Emika Panda Robotic Arm}
\label{fig:robots}
\vspace{-0.5cm}
\end{figure*}
\par
We evaluate the ac-RKN on three robotic systems with different actuator dynamics each of which poses unique learning challenges.
\begin{enumerate}
    \item \textbf{Hydraulically Actuated BROKK-40 Robot Arm}. The data consists of measured joint positions and the input current to the controller sampled at 100Hz from a hydraulic BROKK-40 demolition robot~\parencite{taylor2013state}. The position angle sensors are rotary linear potentiometers. We chose a similar experimental setup and metrics as in \cite{becker2019recurrent} to ensure a fair comparison. We trained the model to predict the joint position 2 seconds, i.e.  200 time-steps, into the future, given only control inputs. Afterwards, the model receives the next observation and the prediction process repeated. Learning the forward model here is difficult due to inherent hysteresis associated with hydraulic control. 
    \item \textbf{Pneumatically Actuated Musculoskeletal Robot Arm.} This four DoF robotic arm is actuated by Pneumatic Artificial Muscles (PAMs)~\parencite{buchler2016lightweight}. Each DoF is actuated by an antagonistic pair of PAMs, yielding a total of eight actuators. The robot arm reaches high joint angle accelerations of up to 28,000 $\textrm{deg}/s^2$ while avoiding dangerous joint limits thanks to the antagonistic actuation and limits on the air pressure ranges.  The data consists of trajectories of hitting movements with varying speeds while playing table tennis~\parencite{buchler2020learning} and is recorded at 100Hz. The fast motions with high accelerations of this robot are complicated to model due to hysteresis.
    \item \textbf{Franka Emika Panda Arm.}  We collected the data from a 7 DoF Franka Emika Panda manipulator during free motion at a sampling frequency of 1kHz. It involved a mix of movements of different velocities from slow to swift motions. The high frequency, together with the abruptly changing movements results in complex dynamics which are interesting to analyze.
\end{enumerate}

\begin{figure}
\centering
\begin{subfigure}{.47\textwidth}
\vspace{0.5cm}
\begin{adjustbox}{width=0.72\columnwidth,center}
\begin{tabular}{|l|c|c|} \hline
         & RMSE  & NLL \\ \hline
ac-RKN &\textbf{0.0144} & \textbf{6.247}    \\ \hline
RKN    & 0.0282 & 4.930     \\ \hline
LSTM &0.2067  & 0.952\\ \hline
GRU &0.2015  & 1.186\\ \hline
\end{tabular}
\end{adjustbox}
\vspace{0.5cm}
\caption{Hydraulic Brokk 40}

  \end{subfigure} 
     \hfill
     \begin{subfigure}{.47\textwidth}
     \hspace*{-0.2cm}
    \includegraphics[width=\linewidth]{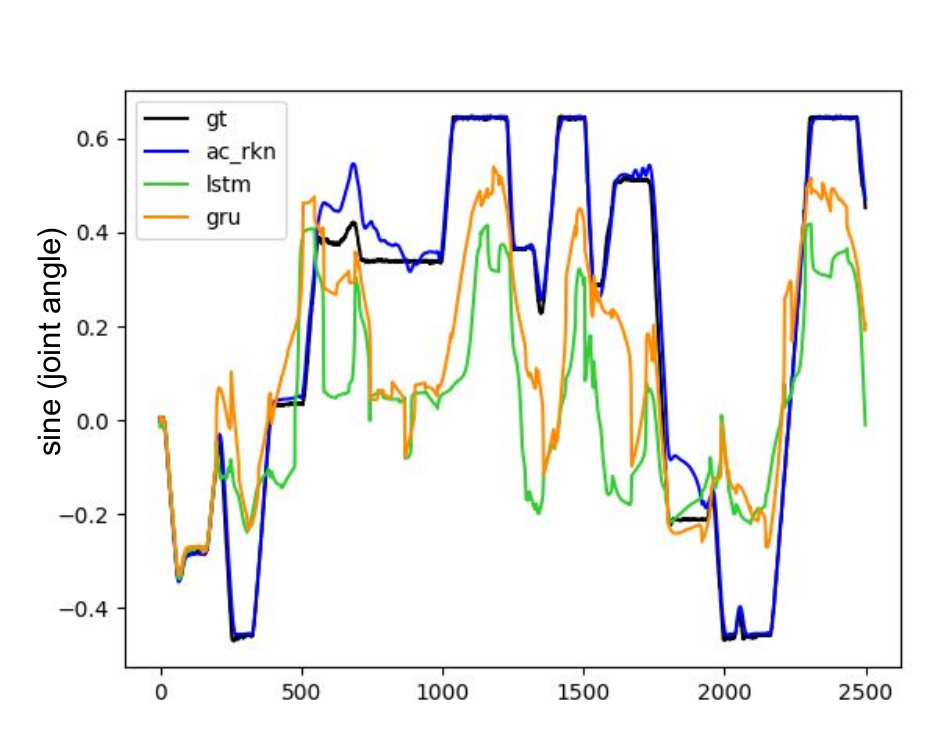}
    \vspace{0.2cm}
    \caption{Hydraulic Arm Predictions}
    \label{fig:hyd_traj}
  \end{subfigure}
        \vspace{0.002\textwidth}
\caption{\textbf{(a)} Performance comparison of various recurrent models for the BROKK-40 hydraulically actuated robot arm. \textbf{(b)} Visualization of the predicted trajectories of the hydraulic arm for 200 step ahead predictions. \label{fig:forwardModelPerformance}}
\end{figure}

\subsection{Multi Step Ahead Prediction} 
\begin{figure*}[t]
  \centering
  \hspace*{-0.1cm}
  \begin{minipage}[b]{.47\textwidth} 
    \centering
    \includegraphics[width=\linewidth]{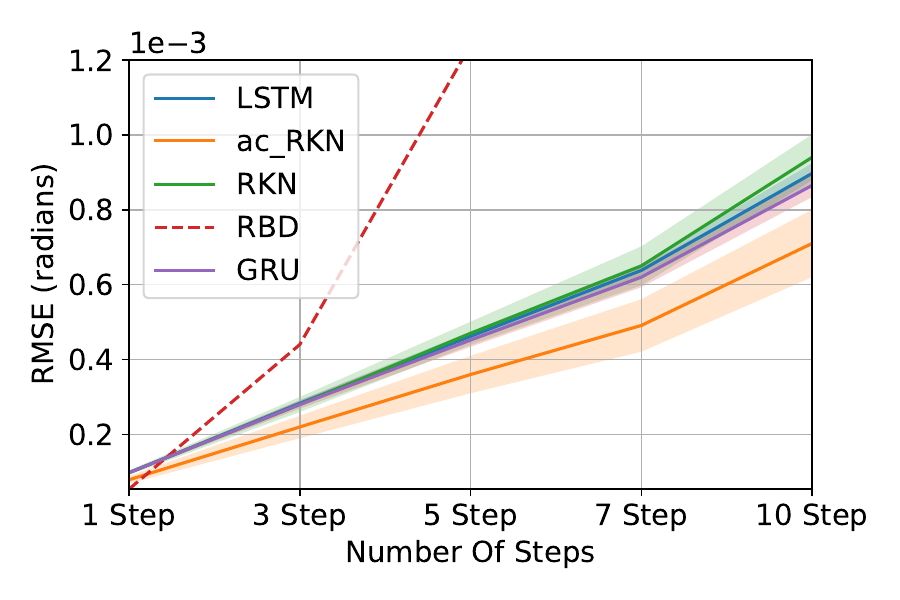}
    \caption{}
    \label{fig:franka_control}
  \end{minipage}\hfill
  \begin{minipage}[b]{.47\textwidth} 
    \centering
    \includegraphics[width=\linewidth]{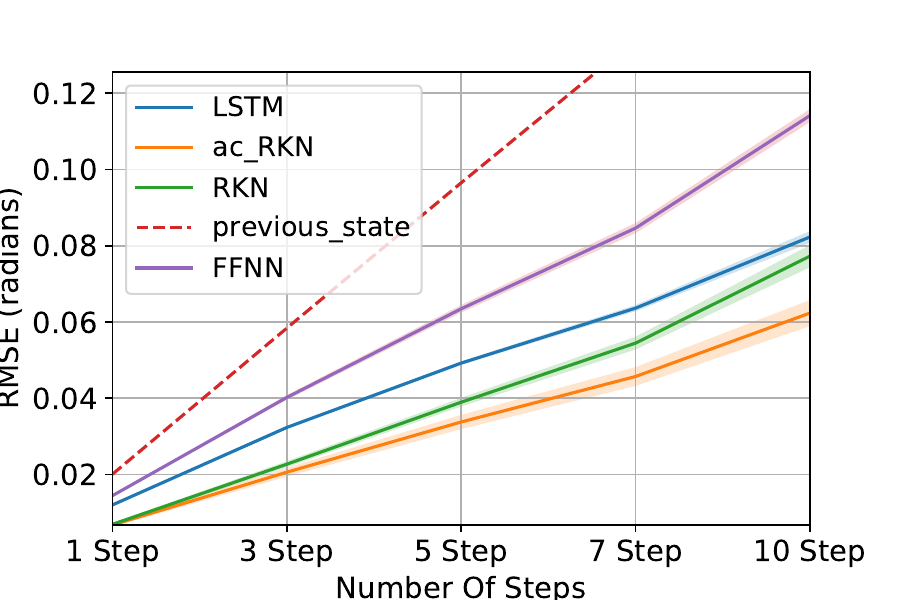} 
    \caption{}
    \label{fig:pam_traj}
  \end{minipage}
  \caption{Plots show the comparison of different action-conditioning models discussed in Section~\ref{sec:action-con} for (left) Franka Emika Panda Robot and (right) Pneumatic Muscular Robot Arm. The plots clearly show that the predictions given by our approach are by far the most accurate.}
  \label{fig:multiac}
\end{figure*}

Figures \ref{fig:forwardModelPerformance} and \ref{fig:multiac} summarize the test set performance for each of these robots for a multi-step ahead prediction task.  We benchmarked the performance of ac-RKN with state-of-the-art deterministic deep recurrent models (LSTM and GRU) and non-recurrent models like feed-forward neural network (FFNN). In all three robots, the ac-RKN gave much better multi-step ahead prediction performance than these deterministic deep models. The performance improvement was more significant for robots with hydraulic and pneumatic actuator dynamics due to explicit non-markovian dynamics and hysteresis, which is often difficult to model via analytical models. We also validated the performance improvement due to our principled action-conditioning in the latent transition dynamics by comparing it with RKN~\parencite{becker2019recurrent}, which treats actions as part of the observations by `concatenation'. In all three robots, our principled treatment brought significant improvement in performance. Figures \ref{fig:hyd_traj} and \ref{fig:forwardModelTrajectories} show the predicted trajectories for the BROKK and the pneumatically actuated robot arm.

\par
\subsection{Comparison with Analytical Model} We were also interested in comparing these with analytical models of Franka. For the analytical model, in addition to the inertia properties of the links, Coulomb friction was also identified for each joint. A detailed description of the same can be found in Appendix C. As seen in Figure \ref{fig:multiac}, the performance of the analytical model outperformed ac-RKN for step-by-step predictions, but for multistep forward predictions, the data-driven models had a clear advantage over ac-RKN that provides the most accurate results.
\subsection{Ablation Study for Action-Conditioning.}
\begin{wrapfigure}[9]{r}{0.45\textwidth} 
  \centering
  \vspace{-.7cm}
  \includegraphics[width=\linewidth]{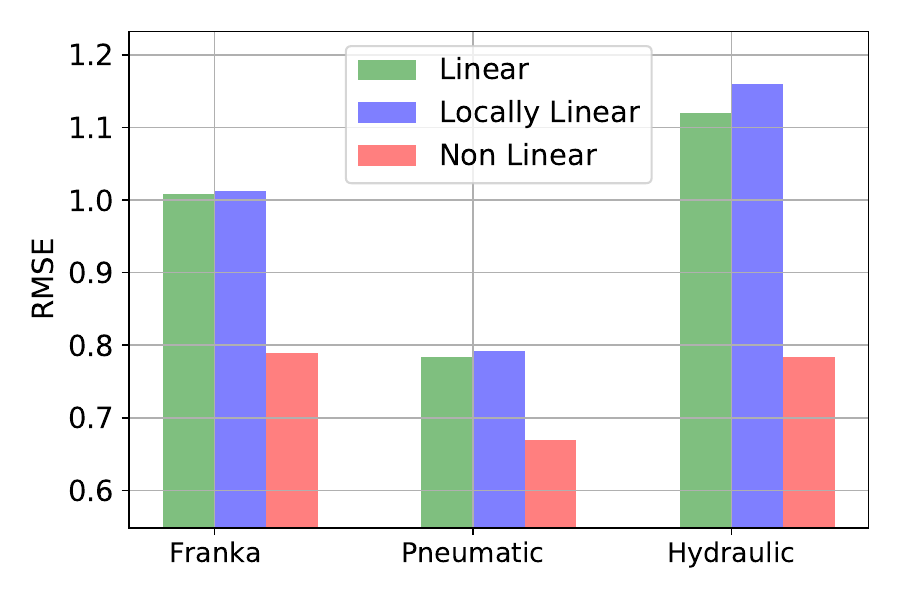}
  \caption{\small Impact of control input models}
  \label{fig:control}
\end{wrapfigure}
We evaluated the performance of the model for different action conditioning schemes (linear, locally-linear and non-linear) discussed in Section \ref{sec:action-con}. The resulting evaluation can be seen in Figure \ref{fig:control} and shows the advantage of using nonlinear models for the additive action conditioning of the latent dynamics.
\par
\subsection{Inverse Dynamics Learning}
In addition, we also adapted ac-RKN for inverse dynamics learning tasks. Since this is out of scope of the main topic of the thesis of World modeling, we report the details in Appendix \ref{sec:inverseDyn}.

\begin{figure}[h]
  \centering
  \hspace*{-0.1cm}
  \includegraphics[width=0.7\linewidth]{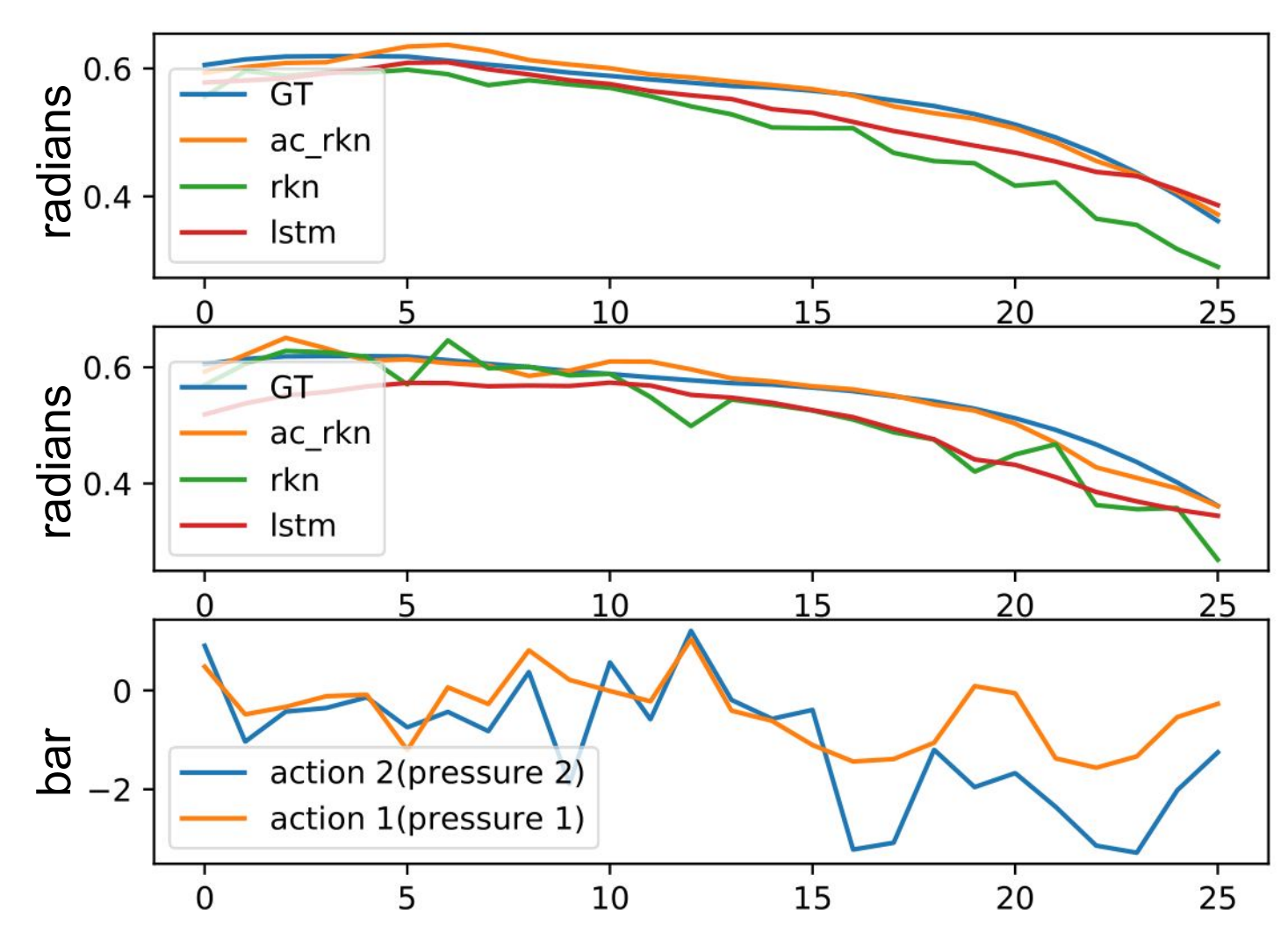}
  \caption{Visualization of the predicted trajectories of the pneumatic arm for 5-step ahead (top) and 10-step ahead (middle) predictions for one of the joints. The bottom plot shows the corresponding pressure (actions) applied in bars. The figure clearly shows that the predictions given by our principled action-conditional scheme are by far the most accurate.}
  \label{fig:forwardModelTrajectories}
\end{figure}

\section{CONCLUSION}

In this section, we modified a contemporary deep Kalman filter approach for world modeling by incorporating action/control signals into the generative process giving them the capability to predict the sensory consequences of actions. By conditioning the actions on the latent state while maintaining the causal relationships between these entities, we enable the execution of interventions and counterfactual analyses with these signals. Such capabilities are essential for decision-making and planning processes using these models. This design facilitates the learning of precise forward dynamics models for complex robots and scenarios lacking analytical models. We validated the efficacy of our method on robots equipped with hydraulic, pneumatic, and electric actuators in settings with stationary dynamics. Initial experiments indicated that the current generative model, the ac-SSM, falls short in handling non-stationary dynamics. We plan to address this limitation in the subsequent chapter.

\chapter{Hidden Parameter SSMs: Probabilistic World Models with Task Abstractions}
\begin{chapquote}{}
This chapter is based on "Hidden-Parameter Recurrent State-Space Models for Changing Dynamics Scenarios"~\parencite{shaj2022hidden}.
\end{chapquote}

One of the most notable abilities of intelligent biological agents is their ability to adapt their behavior to different contexts and environments (i.e., cognitive flexibility) through learning. \cite{Isomura2019} postulates that the brain maintains a collection of generative internal models of the world, and consequential flexibility arises from contextualization and selection on the basis of higher-order beliefs about the most plausible hypothesis, task, or context in play. 
\begin{figure}[h]
\centering
\includegraphics[width=0.8\linewidth]{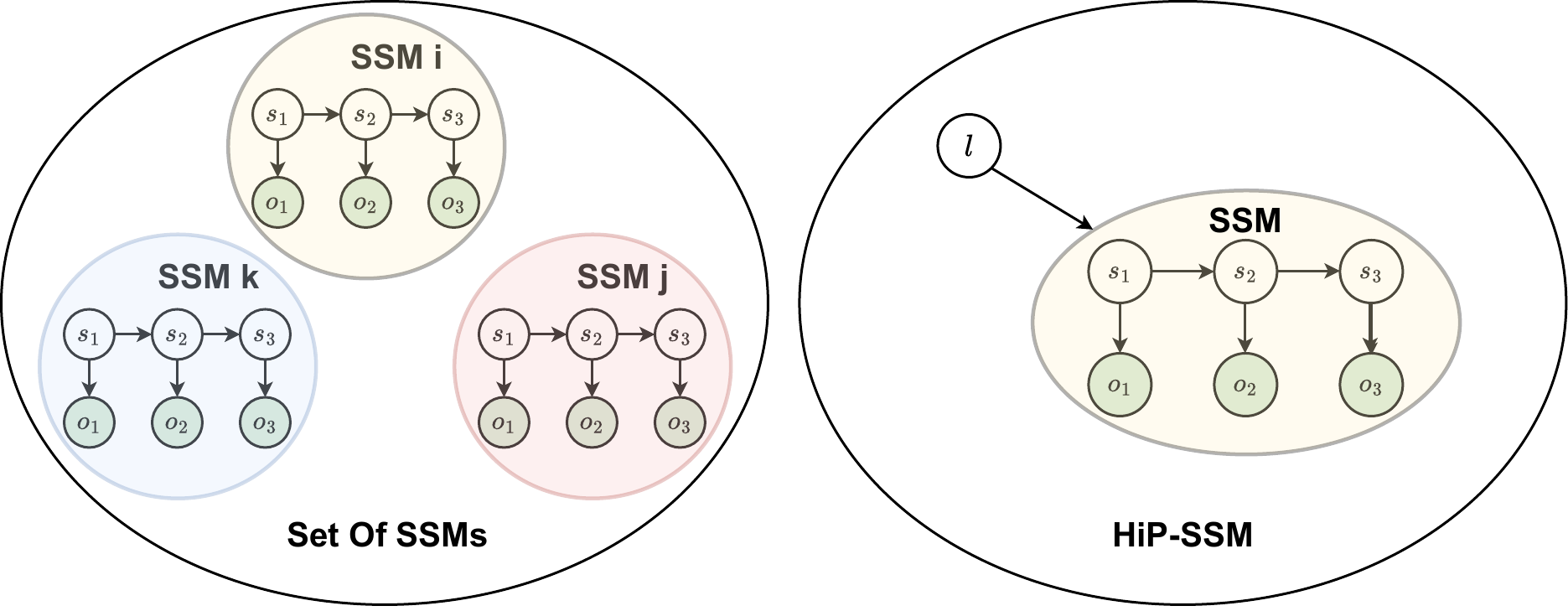}
    \caption{(left) We visualize a set of internal models of the world, each depicted as an individual State-Space Model (SSM) as detailed in Chapter \ref{chap:Acssm}. These models cater to diverse dynamics and tasks. (right) The proposed HiP-SSM framework enables modeling of multitask dynamics using a singular, overarching model. It achieves this through a hierarchical latent task variable, denoted as $l$ that parametrize the latent dynamics.}
    \label{fig:ssmsvship}
\end{figure}
\par
In this part of the thesis, we focus on the adaptability aspects of world models. For many real-world control applications, an intelligent agent must learn to solve tasks with similar, but not identical, dynamics. A robot playing table tennis may encounter several bats with different weights or lengths, while an agent manipulating a bottle may encounter bottles with varying amounts of liquid. An agent driving a car may encounter many different cars, each with unique handling characteristics. Humanoids learning to walk may face different terrains with varying slopes or friction coefficients. Any real-world dynamical system might change over time due to multiple reasons, some of which might not be directly observable or understood. For example, in soft robotics small variations in temperature or changes in friction coefficients of the cable drives of a robot can significantly change the dynamics. Similarly, a robot may undergo wear and tear over time, which can change its dynamics. Assuming a global model as discussed in Chapter \ref{chap:Acssm} that is accurate throughout the entire state space or duration of use is a limiting factor for using such models in real-world applications. 
\par
  
\paragraph{Research Objective} Given the limitations of existing literature on recurrent models, particularly in modeling the dynamics of nonstationary situations, and the impracticality of employing separate State Space Models (SSMs) for each task in continuously changing systems, this research attempts to answer the following question:
\begin{enumerate}
  \item Is it feasible to develop a formalism that enables the learning of a unified global dynamics model, applicable across multiple tasks and environments, while incorporating a hierarchical latent task variable for task-specific specialization?
  \item Can we perform learning and inference in a scalable manner with such a generative model ?
  \item How do these models compare against traditional Recurrent State Space Models (RSSMs) and other contemporary state-of-the-art models in environments with evolving dynamics?
\end{enumerate}
We thus introduce hidden parameter state-space models (HiP-SSM), which allow capturing the variation in the dynamics of different instances through a set of hidden task parameters. We formalize the HiP-SSM and show how to perform inference in this graphical model. Under Gaussian assumptions, we obtain a probabilistically principled yet scalable approach. We name the resulting probabilistic recurrent neural network as Hidden Parameter Recurrent State Space Model (HiP-RSSM). HiP-RSSM achieves state-of-the-art performance for several systems whose dynamics change over time. Interestingly, we also observe that HiP-RSSM often exceeds traditional RSSMs in performance for dynamical systems previously assumed to have unchanging global dynamics due to the identification of unobserved causal effects in the dynamics.
 
\section{Related Work}
\label{rw:hipssm}
\paragraph{Deep State Space Models. } Classical State-space models (SSMs) are popular due to their tractable inference and interpretable predictions. However, inference and learning in SSMs with high dimensional and large datasets are often not scalable. Recently, there have been several works on deep SSMs that offer tractable inference and scalability to high dimensional and large datasets. \cite{haarnoja2016backprop, becker2019recurrent, shaj2020action} use neural network architectures based on closed-form solutions of the forward inference algorithm on SSMs and achieve state-of-the-art results in state estimation and dynamics prediction tasks. \cite{krishnan2017structured, karl2016deep, hafner2019learning} perform learning and inference in SSMs using variational approximations. However, most of these recurrent state-space models assume that the dynamics is fixed, which is a significant drawback, since this is rarely the case in real-world applications such as robotics.
\paragraph{Recurrent Switching Dynamical Systems.} \cite{linderman2017bayesian, becker2019switching, dong2020collapsed} tries to address the problem of changing/multimodal dynamics by incorporating additional discrete switching latent variables. However, these discrete states make learning and inference more involved.  \cite{linderman2017bayesian} uses auxiliary variable methods for approximate inference in a multi-stage training process, while \cite{becker2019switching, dong2020collapsed} uses variational approximations and relies on additional regularization/annealing to encourage discrete state transitions. On the other hand, \cite{fraccaro2017disentangled} uses “soft” switches, which can be interpreted as a switching linear dynamical system which interpolate linear regimes continuously rather than using truly discrete states. We take a rather different, simpler formalism for modeling changing dynamics by viewing it as a multi-task learning problem with a continuous hierarchical hidden parameter that model the distribution over these tasks. Further our experiments in appendix \ref{app:softswitch} show that our model significantly outperforms the soft switching baseline~\parencite{fraccaro2017disentangled}.
\paragraph{Hidden Parameter MDPs.} Hidden Parameter Markov Decision Process (HiP-MDP)~\parencite{doshi2016hidden,killian2016transfer} address the setting of learning to solve tasks with  similar, but not identical, dynamics. In HiP-MDP formalism, the states are assumed to be fully observed. However, we formalize the partially observable setting where we are interested in modelling the dynamics in a latent space under changing scenarios. The formalism is critical for learning from high dimensional observations and dealing with partial observability and missing data. The formalism can be easily extended to HiP-POMDPs by including rewards into the graphical model \ref{fig:hippgm}, for planning and control in the latent space~\parencite{hafner2019learning,sekar2020planning}. However this is left as a future work.
\paragraph{Meta Learning For Changing Dynamics.} There exists a family of approaches that attempt online model adaptation to changing dynamics scenarios via meta-learning \parencite{nagabandi2018learning, nagabandi2018deep}. They perform online adaptation on the parameters of the dynamics models through gradient descent~\parencite{finn2017model} from interaction histories. Our method fundamentally differs from these approaches in the sense that we do not perform a gradient descent step at every time step during test time, which is computationally impractical in modern robotics, where we often deal with high-frequency data. We also empirically show that our approach adapts better, especially in scenarios with non-markovian dynamics, a property that is often encountered in real-world robots due to stiction, slip, friction, pneumatic/hydraulic/cable-driven actuators etc. \cite{saemundsson2018meta,achterhold2021explore} on the other hand, learn context-conditioned hierarchical dynamics model for control in a formalism similar to HiP-MDPs.  The former meta-learn a context conditioned gaussian process while the later use a context conditioned determisitic GRU. Our method on the other hand focuses on a principled probabilistic formulation and inference scheme for context conditioned recurrent state space models, which is critical for learning under partial observabilty/high dimensional observations, with noisy and missing data.
\section{Hidden Parameter State Space Models (HiP-SSMs)}
\begin{wrapfigure}[19]{r}{.35\linewidth}
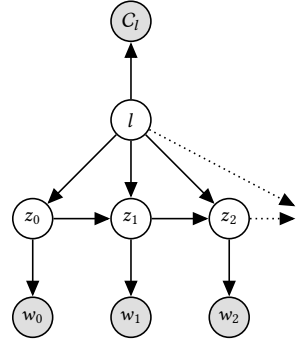

 \scalebox{0.75}{
\begin{subfigure}[b]{.42\textwidth}
    \centering
    \tikzPhilippsHiPSSM
     \end{subfigure}}
\caption{ The graphical model for a particular instance for the HiP-SSM. The transition dynamics between latent states is a function of the previous latent state, and a specific latent task parameter $ \cvec{l} $ which is inferred from a context set of observed past observations. Actions are omitted for brevity.} 
 \label{fig:hippgm}
 \end{wrapfigure}
We denote a set of SSMs with transition dynamics $ f_{\cvec{l}} $  that are fully described by hidden parameters $ \cvec{l} $ and observation model $ h $  as a Hidden Parameter SSM (HiP-SSM). In this definition we assume the observation model $h$ to be independent of the hidden parameter $\cvec{l}$ as we only focus on cases where the dynamics change. HiP-SSMs allow us to extend SSMs to multitask settings where dynamics can vary across tasks. Defining the changes in dynamics by a latent variable unifies the dynamics across tasks as a single global function. In dynamical systems, for example, parameters can be physical quantities like gravity, friction of a surface, or the strength of a robot actuator. Their effects influence the dynamics but are not directly observed, so $\cvec{l}$ is not part of the observation space and is treated as a latent task parameter vector. The Bayesian network corresponding to HiP-SSM is shown in Figure \ref{fig:hippgm} and a formal definition is given below. 

\begin{definition}
A HiP-SSM is represented by a tuple $\lbrace{\mathcal{L},\mathcal{C},\mathcal{Z},\mathcal{A},\mathcal{W},g,f,h\rbrace}$, where $\mathcal{Z}$, $\mathcal{A}$, $\mathcal{W}$, $\mathcal{L}$ and $\mathcal{C}$ denote the sets of hidden states $\cvec{z}$, actions $\cvec{a}$, observations $\cvec{w}$, latent tasks $\cvec{l}$ and contexts $C_l$ respectively.
Based on a task model $g$, transition model $f$ and observation model $g$, HiP-SSM have the following causal relationship:
\[
\begin{aligned}
\cvec{C}_l &= g(\cvec{l}) + \cvec{v}_l, \\
\cvec{z}_t &= f(\cvec{z}_{t-1},\cvec{a}_{t-1},\cvec{l}) + \cvec{\epsilon}_t, \\
\cvec{w}_t &= h(\cvec{z}_t) + \cvec{u}_t.
\end{aligned}
\]
Here, $\cvec{z}_t \in \mathcal{Z}$, $\cvec{a}_t \in \mathcal{A}$, and $\cvec{w}_t \in \mathcal{W}$ represent latent states, actions, and observations at time $t$, respectively. The vector $\cvec{\epsilon}_t$ denotes the process / transition noise, while $\cvec{u}_t$ and $\cvec{v}_l$ denote the measurement noise corresponding to observations and contexts.
\end{definition}
Thus, a HiP-SSM describes a class of dynamics, and a particular instance of that class is obtained by fixing the parameter vector $\cvec{l} \in \mathcal{L}$. The dynamics $f$ for each instance depends on the value of the hidden parameters $\cvec{l}.$ 
\par
Each instance of a HiP-SSM is an SSM conditioned on $\cvec{l}$. We also make the additional assumption that the parameter vector $\cvec{l}$ is fixed for the duration of the task (i.e., a local segment of a trajectory), and thus the latent task parameter has no dynamics. This assumption considerably simplifies the procedure to infer hidden parameterization and is reasonable, since dynamics can be assumed to be locally consistent over small trajectories in most applications~\parencite{nagabandi2018learning}.

\label{headings}
\par
The definition is based on the related literature on HiP-MDPs \parencite{doshi2016hidden}, where the only unobserved variable is the latent task variable. One can connect HiP-SSMs with HiP-MDPs by including rewards in the definition and formalizing HiP-POMDPs. However, this is left for future work.
\par
\section{Exact Inference In HiP-SSMs} 
\begin{figure}[h]
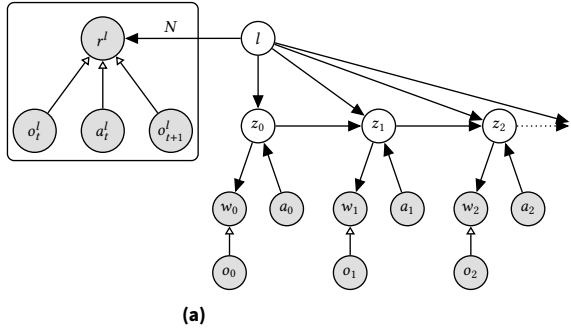

\centering
\begin{subfigure}{0.6\textwidth}
    \scalebox{0.8}{
    \tikzPhilippsHipRSSM}
    \caption{}
    \end{subfigure}%
    \caption{The Gaussian graphical model corresponding to a particular instance $\cvec{l} \in \mathcal{L}$ for the HiP-RSSM. The posterior of the latent task/context variable is inferred via a Bayesian aggregation procedure described in \ref{subsec:latentTask} based on a set of N interaction histories. The prior over the latent states $\cvec{z}_t^-$ is inferred via task conditional Kalman time update described in \ref{subsec:timeUpdate} and the posterior over the latent states $\cvec{z}_t^+$ is inferred via Kalman measurement update described in \ref{subsec:obsUpdate}. Here, the hollow arrows denote deterministic transformation leading to implicit distributions, using the context and observation encoders.}
    \label{fig:hiprssmpgm}
\end{figure}
We perform inference and learning in the HiP-SSM borrowing concepts from both deep learning and graphical model communities following recent work on recurrent neural network models \parencite{haarnoja2016backprop,becker2019recurrent}, where the architecture of the network is informed by the structure of the probabilistic state estimator. We denote the resulting probabilistic recurrent neural network architecture as the Hidden Parameter Recurrent State Space Model (HiP-RSSM).
\par

The structure of the Bayesian network shown in Figure \ref{fig:hiprssmpgm} allows a tractable inference of latent variables by the forward algorithm \parencite{jordan2004graphical, koller2009probabilistic}. Since we are dealing with continuous dynamical systems, we assume a Gaussian multivariate distribution over all variables (both observed and hidden) for the graph shown in Figure \ref{fig:hiprssmpgm}. This assumption has several advantages. Firstly, it makes the inference very similar to the well studied Kalman Filtering approaches. Secondly, the Gaussian assumptions and conditional in-dependencies allows us to have a closed form solution to each of these update procedures which are fully differentiable and can be backpropagated to the deep encoders. The update of beliefs about the hidden variables $\cvec{z}_t$ and $\cvec{l}$ takes place in three stages. Similar to Kalman filtering approaches, we have two recursive belief state update stages, the time update and observation update which calculate the prior and posterior belief over the latent states, respectively. However, we have an additional hierarchical latent variable $\cvec{l}$ which models the (unobserved) causal factors of variation in dynamics in order to achieve efficient generalization. Hence, we have a third belief update stage to calculate the posterior over the latent task variable based on the observed context set.  Each of these three stages is detailed in the following sections:

\subsection{Inferring The Latent Task Variable (Context Update)}
\label{subsec:latentTask}
 In this stage, we infer the posterior over the Gaussian latent task variable $\cvec{l}$ based on an observed context set $\mathcal{C}_{\cvec{l}}$. For any HiP-RSSM instance defined on a target sequence $\mathcal{T}=(\cvec{o}_t,\cvec{a}_t,\cvec{o}_{t+1},...,\cvec{o}_{t+N},\cvec{a}_{t+N})$, over which we intend to perform state estimation/prediction, we maintain a fixed context set $\mathcal{C}_{\cvec{l}}$. $\mathcal{C}_{\cvec{l}}$ in our HiP-SSM formalism can be obtained according to the algorithm designer's choice. We choose a $\mathcal{C}_{\cvec{l}}$ consisting of a set of tuples $\mathcal{C}_{\cvec{l}} = \lbrace \cvec{o}_{t-n}^{\cvec{l}},\cvec{a}_{t-n}^{\cvec{l}},\cvec{o}_{t-n+1}^{\cvec{l}}\rbrace_{n=1}^N$. Here each set element is a tuple consisting of the current state/observation, current action, and next state/observation for the previous N time steps.

Inferring a latent task variable $\cvec{l} \in \mathcal{L}$ based on an observed context set $\mathcal{C}_{\cvec{l}} = \{\cmat{x}_{n}^{\cvec{l}}\}_{n=1}^N$ has previously been explored by different neural process architectures \parencite{gordon2018meta,garnelo2018neural}.  Neural processes are multitask latent variable models that rely on deep set functions \parencite{zaheer2017deep} to generate a latent representation from a varying number of context points in a permutation invariant manner.
Similar to \cite{volpp2020bayesian} we formulate the aggregation of context data as a Bayesian inference problem, where the information contained in $\mathcal{\cmat{C}}_l$ is directly aggregated into the statistical description of $\cvec{l}$ based on a factorized Gaussian observation model of the form $p(\cvec{r}_n^{\cvec{l}}|\cvec{l})$, where
\begin{align*}
\cvec{r}_n^{\cvec{l}} & = \textrm{enc}_{\cvec{r}}(\cvec{o}_{t-n}^{\cvec{l}},\cvec{a}_{t-n}^{\cvec{l}},\cvec{o}_{t-n+1}^{\cvec{l}}), \\ \cvec{\sigma}_n^{\cvec{l}} & = \textrm{enc}_{\cvec{\sigma}}(\cvec{o}_{t-n}^{\cvec{l}},\cvec{a}_{t-n}^{\cvec{l}},\cvec{o}_{t-n+1}^{\cvec{l}}).
\end{align*}
Here $n$ is the index of an element from the context set $\mathcal{\cvec{C}}_{\cvec{l}}$. 
Given a prior $p_0(\cvec{l}) = \mathcal{N}(\cvec{l}|\cvec{\mu}_{0},\textrm{diag}(\cvec{\sigma}_{0}))$ we can compute the posterior $p(\cvec{l}|\mathcal{\cmat{C}}_{\cvec{l}})$ using the Bayes rule. 
The Gaussian assumption allows us to obtain a closed-form solution for the posterior estimate of the latent task variable, $p(\cvec{l}|\mathcal{\cmat{C}}_{\cvec{l}})$ based on Gaussian conditioning. The factorization assumption further simplifies this update rule by avoiding computationally expensive matrix inversions into a simpler update rule as
\begin{align*}
        \cvec{\sigma_{l}} & = \left( \left(\cvec{\sigma}_{0}\right)^\ominus+ \sum_{n=1}^N \left(\cvec{\sigma}_n^{\cvec{l}}\right)^\ominus\right)^\ominus, \\ \cvec{\mu}_{\cvec{l}} & = \cvec{\mu}_{0} + \cvec{\sigma}_{\cvec{l}} \odot \sum_{n=1}^N \left(\cvec{r}^{\cvec{l}}_n - \cvec{\mu}_{0}\right) \oslash \cvec{\sigma}_n^{\cvec{l}}
\end{align*}
where $\ominus$, $\odot$ and $\oslash$ denote element-wise inversion, product, and division, respectively. Intuitively, the mean of the latent task variable $\cvec{\mu}_{\cvec{l}}$ is a weighted sum of the individual latent observations $\cvec{r}^{\cvec{l}}_n$, while the variance of the latent task variable $\cvec{\sigma}_{\cvec{l}}$ gives the uncertainty of this task representation.

\subsection{Inferring Prior Latent States over the Next Time Step (Task Conditional Prediction)}
\label{subsec:timeUpdate}
The goal of this step is to compute the prior marginal 
\begin{align}
\label{eq:predict}
\begin{split}
p(\cvec{z}_t|\cvec{w}_{1:t-1},\cvec{a}_{1:t},\mathcal{C}_{\cvec{l}}) = \iint p(\cvec{z}_t|\cvec{z}_{t-1},\cvec{a}_{t},\cvec{l})&p(\cvec{z}_{t-1}|\cvec{w}_{1:t-1},\cvec{a}_{1:t-1},\mathcal{C}_{\cvec{l}}) \\ &p(\cvec{l}|\mathcal{C}_{\cvec{l}})d\cvec{z}_{t-1}d\cvec{l}.
\end{split}
\end{align} 

\sloppy
We assume a dynamical model of the following form: $p(\cvec{z}_t|\cvec{z}_{t-1},\cvec{a}_{t},\cvec{l}) = \mathcal{N}\left(\cmat{A}_{t-1}\cvec{z}_{t-1} + \cmat{B}\cvec{a}_t + \cmat{C}\cvec{l},  \cmat{\Sigma}_{\textrm{trans}}\right)$ and
denote the posterior from the previous time-step by $p(\cvec{z}_{t-1}|\cvec{w}_{1:t-1},\cvec{a}_{1:t-1},\mathcal{C}_{\cvec{l}}) = \mathcal{N}\left(\cvec{z}_{t-1}^+, \cmat{\Sigma}_{t-1}^+\right)$. 
Following \cite{shaj2020action}, we assume that the action $\cvec{a}_t$ is known and not subject to noise. 
\par
At any time t, the posterior over the belief state $\cvec{z}_{t-1}|\cvec{w}_{1:t-1},\cvec{a}_{1:t-1}$, posterior over the latent task variable $\cvec{l}|\mathcal{C}_{\cvec{l}}$ and the action $\cvec{a}_t$ are independent of each other since they form a ``common effect" / v structure \parencite{koller2007graphical} with the unobserved variable $\cvec{z}_{t}$.  With these independencies and Gaussian assumptions, according to the Gaussian identity \ref{theo:predict}, it can be shown that calculating the integral in equation \ref{eq:predict} has a closed form solution as follows,
$p(\cvec{z}_t|\cvec{w}_{1:t-1},\cvec{a}_{1:t},\mathcal{C}_{\cvec{l}}) = \mathcal{N}(\cvec{z}_t^-,\cmat{\Sigma}_t^-)$, where
\begin{align}
\label{eq:cond-predict}
\cvec{z}_t^- &=\cmat{A}_{t-1}\cvec{z}_{t-1} + \cmat{B}\cvec{a}_t + \cmat{C}\cvec{l}, \\
\cmat{\Sigma}_t^- &= \cmat{A}_{t-1}\cmat{\Sigma}_{t-1}^+\cmat{A}_{t-1}^T + \cmat{C}\cmat{\Sigma}_{\cvec{l}}\cmat{C}^T + \cmat{\Sigma}_{\textrm{trans}}.
\end{align}

\begin{restatable}[Linear Combination Gaussian Marginalization]{gi}{gausslinearmarg}
\label{theo:taskpredict}
\sloppy If $\cvec{u} \sim \mathcal{N}(\cvec{\mu}_{\cvec{u}} + \cvec{b},\cmat{\Sigma}_{\cvec{u}})$ and $\cvec{v} \sim \mathcal{N}(\cvec{\mu}_{\cvec{v}},\cmat{\Sigma}_{\cvec{v}})$  are normally distributed independent random variables and if the conditional distribution $p\left(\cvec{y}|\cvec{u}, \cvec{v}\right) = \mathcal{N}(\cmat{A}\cvec{u} + \cvec{b} + \cmat{B}\cvec{v}, \cmat{\Sigma})$, then marginal $p(\cvec{y}) = \int p(\cvec{y}|\cvec{u},\cvec{v})p(\cvec{u})p(\cvec{v})) d\cvec{u}d\cvec{v} = \mathcal{N}(\cmat{A}\cvec{\mu}_{\cvec{u}} + \cvec{b} + \cmat{B}\cvec{\mu}_{\cvec{v}}, \cmat{A}\cmat{\Sigma}_u\cmat{A}^T + \cmat{B}\cmat{\Sigma}_{\cvec{v}}\cmat{B}^T +  \cmat{\Sigma}).$
\end{restatable}

\begin{proof}
The proof for the above identity is given in Appendix.
\end{proof}

\paragraph{Modelling Choices} We use a locally linear transition model $\cmat{A}_{t-1}$ as in \cite{becker2019recurrent} (see Appendix \ref{sec:locally-linear}) and a nonlinear control model as in \cite{shaj2020action} (Chapter \ref{chap:Acssm}). The local linearization around the posterior mean, can be interpreted as equivalent to an EKF. For the latent task transformation we can either use a linear, locally-linear or non-linear transformation. More details on the latent task transformation model can be found in the Appendix \ref{app:taskmodel}. Our estimations (Figure \ref{fig:ablTask}) show that a nonlinear feedforward neural network $f(.)$ that outputs mean and variances and interacts additively gave the best performance in practice.  $f(.)$ transforms the latent task moments $\cvec{\mu_l}$ and $\cvec{\sigma_l}$ directly into the latent space of the state-space model through additive interactions. The corresponding time update equations are given below:
\begin{align*}
\cvec{z}_t^- &=\cmat{A}_{t-1}\cvec{z}_{t-1}^+ + \cmat{b}(\cvec{a}_t) + \cmat{f}(\cvec{\mu}_l), \\
\cmat{\Sigma}_t^- &= \cmat{A}_{t-1}\cmat{\Sigma}_{t-1}^+\cmat{A}_{t-1}^T + \cmat{f}(\cvec{\sigma}_{\cvec{l}})+ \cmat{\Sigma}_{\textrm{trans}}.
\end{align*}
\notebox{This chapter primarily aimed at accurately modeling mean predictions under non-stationary conditions, focusing on minimizing mean squared errors. We opted for locally linear transitions $\cvec A_t$ and non-linear transformations $f$ to enhance mean predictions, without prioritizing uncertainty quantification. In contrast, Chapter \ref{chap:mts3} equally emphasizes uncertainty quantification and mean predictions. Subsequent chapters adopted a simple linear model for both matrices $\cvec A$ and $\cvec C$, chosen for its robustness in both mean and uncertainty predictions and computational efficiency \parencite{gu2021efficiently,mondal2023efficient}.}
\subsection{Inferring posterior latent states / Observation Update}
\label{subsec:obsUpdate}

The goal of this step is to compute the posterior belief $p(\cvec{z}_t|\cvec{o}_{1:t},\cvec{a}_{1:t},\cvec{C_l})$. We first map the observations at each time step $\cvec{o}_t$ to a latent space using an observation encoder~\parencite{haarnoja2016backprop,becker2019recurrent} that emits latent features $\cvec{w}_t$ along with uncertainty in those features via a variance vector $\cvec{\sigma}_{obs}^t$. We then computed the posterior belief $p(\cvec{z}_t|\cvec{w}_{1:t},\cvec{a}_{1:t},\cvec{C_l})$, based on $p(\cvec{z}_t|\cvec{w}_{1:t-1},\cvec{a}_{1:t},\cvec{C_l}) $ obtained from the time update, the latent observation $(\cvec{w}_t,\cvec{\sigma}_t^{obs})$ and the observation model $\cmat{H}$. This is exactly the Kalman update step, which has a closed form solution as shown below for a time instant $t$, \\
\resizebox{\linewidth}{!}{
  \begin{minipage}{\linewidth}
\begin{align*}
&\text{Kalman Gain:}  &\cmat{Q}_t &= \cmat{\Sigma_t}^- \cmat{H}^T\left(\cmat{H} \cmat{\Sigma_t}^- \cmat{H}^T  + \cvec{I} \cdot \cmat{\sigma_t}^\mathrm{obs} \right)^{-1}, \\
&\text{Posterior Mean:}  &\cvec{z_t}^+ &= \cvec{z_t}^- + \cmat{Q}_t \left(\cvec{w_t} - \cmat{H} \cvec{z_t}^-  \right), \\
&\text{Posterior Covariance:}  &\cmat{\Sigma_t}^+ &= \left(\cmat{I} - \cmat{Q}_t \cmat{H} \right) \cmat{\Sigma_t}^-,
\end{align*}
  \end{minipage}
}
\newline
\newline
where $\cmat{I}$ denotes the identity matrix. This update is added as a layer in the computation graph (\cite{haarnoja2016backprop}; \cite{becker2019recurrent}).  However, the Kalman update involves computationally expensive matrix inversions of the order of $\mathcal{O}(L^3)$, where $L$ is the dimension of the latent state $\cvec{z}_t$. Thus, in order to make the approach scalable, we follow the same factorization assumptions as in \cite{becker2019recurrent}. This factorization provides a simple way to reduce the observation update equation to a set of scalar operations, reducing the computational complexity from $\mathcal{O}(L^3)$ to $\mathcal{O}(L)$.
More mathematical details on the simplified update equation can be found in Section \ref{sec:prelimrkn}.
\notebox{ From a computational perspective, this a Gaussian conditioning layer, similar to section \ref{subsec:latentTask}. Both output a posterior distribution on latent variables $\cvec{z}$, given a prior $p(\cvec{z})$ and an observation model $p(\boldsymbol{w}|\boldsymbol{z})$, using the Bayes rule: $p(\boldsymbol{z}|\boldsymbol{w}) = p(\boldsymbol{w}|\boldsymbol{z})p(\boldsymbol{z})/p(\boldsymbol{w})$. This has a closed form solution because of Gaussian assumptions, which is coded as a layer in the neural network. The observation model is assumed to have the following structure, $P(\boldsymbol{w}|\boldsymbol{z}) = \mathcal{N}(\boldsymbol{H}\boldsymbol{z}, \boldsymbol{\Sigma}_{obs})$.}

\section{HiP-SSM as Adaptive Multi-Task World Models}
\begin{figure}[h]
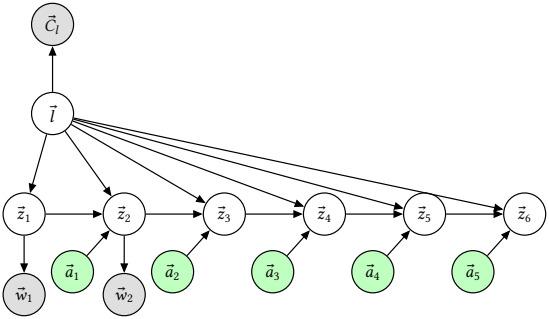

    \centering
    \resizebox{0.7\textwidth}{!}{\tikzHiPGMForward}
    \caption{Using adaptive internal world models to envision the future: Our predictions about future world states rely on more than just a sequence of action signals; they are also conditioned on latent task variables ($l$). These latent variables introduce additional causal factors that influence the dynamics, factors not considered in basic State Space Model (SSM) approaches introduced in Chapter \ref{chap:Acssm}. This inclusion enables a deeper and more nuanced ability to manipulate and foresee future states in diverse scenarios.}
    \label{fig:galaxy}
\end{figure}

\subsection{End To End Learning For Online Adaptation To Changes}

\paragraph{Multitask Data Creation}The latent task variable $l$ models a distribution over functions \parencite{garnelo2018neural}, rather than a single function. In our case, these functions are latent dynamics functions. In order to train such a model, we use a training procedure that reflects this objective, where we form datasets consisting of timeseries, each with different latent transition dynamics. Thus, we collect a set of trajectories over which the dynamics changes over time. These can be trajectories where a robot picks up objects of different weights or a robot traverses terrain of different slopes. Now, we introduce a multitask setting with a rather flexible definition of task, where each temporal segment of a trajectory can be considered to be a different ``task" and the observed context set based on interaction histories from the past $N$ time steps provides information about the current task setting. This definition allows us to have a potentially infinite number of tasks/local dynamical systems, and the distribution over these tasks/systems is modeled using a hierarchical latent task variable $l$. The formalism is based on the assumption that over these local temporal segments the dynamics is unchanging. This local consistency in dynamics holds for most real world scenarios~\parencite{nagabandi2018deep,nagabandi2018learning}. However, our formalism can model the global changes in dynamics at the test time, since we obtain a different instance of the HiP-RSSM for each temporal segment based on the observed context set. We also provide a detailed description of the multitask data set creation process in Table \ref{algo:data} and a pictorial illustration in Appendix \ref{algo:hipinfer}.

\RestyleAlgo{ruled}
\begin{algorithm}[H]
\caption{Multi Task Dataset Creation For Training HiP-RSSM}\label{alg:three}
\textbf{Required:}{A set $S$ of trajectories of changing dynamics }\\
$D \leftarrow \phi$ \\
\For{each trajectory $\tau \in S$ }{
\begin{enumerate}
\item  divide the trajectory $\tau$ into non-overlapping windows $T_l$ \\ of length N. Let $T = \{T_1, T_2, T_3, .,.,.\}$ be the list of all temporal segments/time-series. \\

\For{each time window $T_l \in T$}{
\begin{enumerate}
\item maintain a context set $C_l$ consisting of N previous interactions;
\item update $D \leftarrow D \cup \{C_l,T_l\}$ \\
\end{enumerate}}
\end{enumerate}
}
\label{algo:data}
\textbf{Output}: Output $D$ consisting of batch of context and target sets.
\end{algorithm}

\par
\paragraph{Batch Training.} Let $T  \in \mathcal{R}^{B \times N \times D}$ be the batch of local temporal segments with different dynamics that we intend to model with the HiP-RSSM formalism. Given a target batch $T$, HiP-RSSM can be trained in a supervised manner similar to popular recurrent architectures such as LSTMs or GRUs.  However, for each local temporal sequence $t \in T$, in which the dynamics is modeled, we also input a set of previous interactions $N$, which forms the context set $C \in \mathcal{R}^{B \times N \times D}$ to infer the latent task as explained in Section \ref{subsec:latentTask}. Processing the context set $C$ adds minimal additional computational / memory constraints, as we use a permutation-invariant set encoder. The set encoder allows for parallelization in processing the context set as opposed to recurrent processing of the target set.
\par

The learnable parameters in the computation graph include the locally linear transition model $\cvec{A}_t$, the nonlinear control factor $\cvec{b}$, the linear/nonlinear latent task transformation model $C$, the transition noise $\cvec{\Sigma}_{\textrm{trans}}$, along with the observation encoder, context encoder, and output decoder. 
\par

\textbf{Loss Function.} The network is tasked to minimize the prediction errors by maximizing the posterior predictive log-likelihood, which is given below for a single trajectory, i.e., 
\begin{align}
\label{eq:objective}
     L & =\sum_{t=1}^H \log p(\cvec{o}_{t+1}|\cvec{w}_{1:t}, \cvec{a}_{1:t}, \cvec C_l) \nonumber \\ & = \sum_{t=1}^H \log \int p(\cvec{o}_{t+1}|\cvec{z}_{t+1})  p(\cvec{z}_{t+1}|\cvec{w}_{1:t}, \cvec a_{1:t}, \cvec C_l) d\cvec{z}_{t+1} \nonumber \\  
\end{align}

The extension to multiple trajectories is straightforward and was omitted to keep the notation uncluttered. Here, $\cvec{o}_{t+1}$ are the ground truth observations at the time step $t+1$ that must be predicted from all observations up to the time step $t$.

\paragraph{Approximating the likelihood} \sloppy We employ a Gaussian approximation of the posterior predictive log-likelihood of the form $ p(\cvec{o}_{t+1}|\cvec{w}_{1:t}, \cvec{a}_{1:t}, \cvec C_l) \approx \mathcal{N}(\cvec{\mu}_{\cvec{o}_{t+1}},\textrm{diag}(\cvec{\sigma}_{\cvec{o}_{t+1}}))$ where we use the mean of the prior belief $\cvec{\mu}_{z_{t+1}}^-$ to decode the predictive mean, that is, $\cvec{\mu}_{\cvec{o}_{t+1}} =  \textrm{dec}_{\cvec{\mu}}(\cvec{\mu}_{z_{t+1}}^{-})$ and the variance estimate of the prior belief to decode the observation variance, that is, $\cvec{\sigma}_{o_{t+1}} = \textrm{dec}_{\sigma}(\cmat{\Sigma}_{z_{t+1}}^{-})$. This approximation can be motivated by a moment matching perspective and allows end-to-end optimization of logarithmic likelihood without using auxiliary objectives such as the ELBO \cite{becker2019recurrent}. Thus, the approximate Gaussian predictive log-likelihood for a single sequence is then computed as  
\begin{align}
\label{eq: likeli-acrkn}
&\mathcal{L}\left(\cvec{o}_{(1:T)}\right) = \\&
\dfrac{1}{T} \sum_{t=1}^T  \log \mathcal{N}\left(\cvec{o}_t \bigg| \textrm{dec}_{\mu}(\cvec{z}_t^+), \textrm{dec}_{\Sigma} (\cvec{\sigma}^{\mathrm{u},+}_t,
\cvec{\sigma}^{\mathrm{s},+}_t,
\cvec{\sigma}^{\mathrm{l},+}_t )\right), \nonumber
\end{align}
where $\textrm{dec}_{\mu}(\cdot)$ and $\textrm{dec}_{\Sigma}(\cdot)$ denote the parts of the decoder that are responsible for decoding the latent mean and latent variance respectively.

\paragraph{Variations Of The Learning Objectives}

We optimize the root mean square error (RMSE) between the decoder output and the ground-truth states. As in \cite{shaj2020action} we use the differences to the next state as our ground truth states, as this results in better performance for dynamic learning, especially at higher frequencies. In principle, we could train on the Gaussian log-likelihood instead of the RMSE and hence model the variances. Training in RMSE yields slightly better predictions and allows a fair comparison with deterministic baselines that use the same loss, such as feedforward neural networks, LSTMs and metalearning algorithms such as MAML~\parencite{finn2017model}. Therefore, we report results with the RMSE loss. 
\begin{figure}[h]
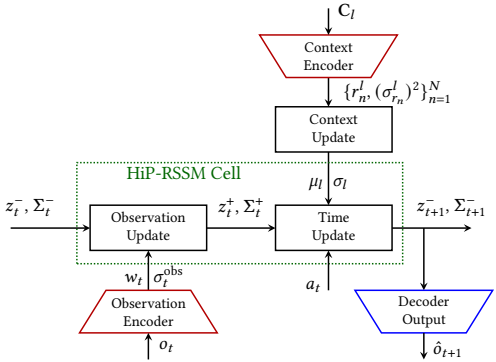

\centering
\scalebox{1.1}{\tikzHiPRSSM}
    \caption{Depicts the schematic of HiP-RSSM. The output of the task conditional `Time Update' stage, which forms the prior for the next time step $(z_{t+1}^-,\Sigma_{t+1}^-)$ is decoded to get the prediction of the next observation.}
    \label{fig:hip-schematic}
\end{figure}
\par
Gradients are computed using (truncated) backpropagation through time (BPTT) (\cite{werbos1990backpropagation}) and clipped. We optimize the objective using the Adam (\cite{kingma2014adam}) stochastic gradient descent optimizer with default parameters.

\notebox{Irrespective of the learning objectives/loss functions used, we still have to model uncertainty in HiP-SSM latent state as the context set and observations do not contain the full task and state information of the system.}
The architecture of the HiP-SSM is summarized in Figure \ref{fig:forward_model}.

\subsection{HiP-RSSM during Test Time / Inference}
\label{algo:hipinfer}
\begin{figure}[t]
\includegraphics[width=\linewidth]{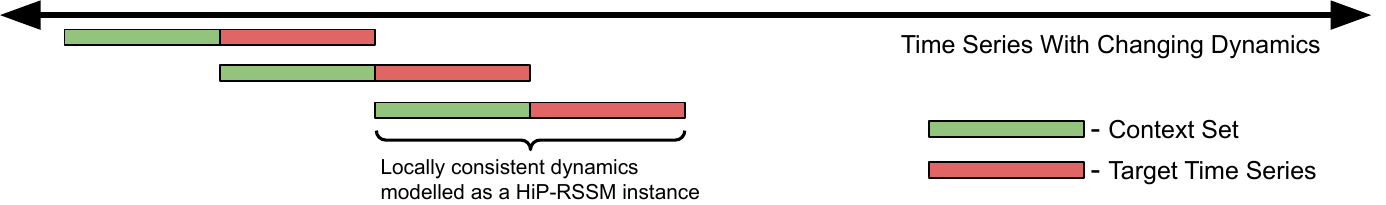}
    \caption{HiP-RSSM Inference Under Changing Dynamics Scenarios}
    \label{fig:changingPic}
\end{figure}

We perform inference using HiP-RSSM at test time on a trajectory with varying dynamics using algorithm \ref{alg:changing}. A pictorial representation of the same is given in the Figure \ref{fig:changingPic}. We use this inference scheme to visualize how the latent variable $l$, that describe different instances of a HiP-RSSM over different temporal segments, evolve at a global level. The visualizations are reported in Figures \ref{fig:frankalatent} and \ref{fig:wheeledlatent} in the experiments section discussed below. 


\RestyleAlgo{ruled}
\begin{algorithm}
\caption{HiP-RSSM Test Time Inference}
\label{alg:changing}
\textbf{Required:}{ Trained HiP-RSSM Model}\\
\textbf{Required:}{ A time series $\tau$ of length $K>>N$} \\
Divide the time series $\tau$ into non-overlapping windows $T_l$ of length N. Let $T = \{T_1, T_2, T_3, .,.,.\}$ be the ordered list of all temporal segments, sorted in the ascending order of time of occurrence.\\
\ForEach{each time window $T_l \in T$}{
\begin{enumerate}
\item maintain a context set $C_l$ consisting of N previous interactions;\\
\item infer posterior latent task variable $p(\cvec{l}|\mathcal{C}_{\cvec{l}})$ using context update stage as in section \ref{subsec:latentTask};\\
\item using the posterior over latent task variable $\cvec{l}|\mathcal{C}_{\cvec{l}}$ and observations in sequence $T_l$ to perform sequential Bayesian inference over the state space model using Kalman observation update (\ref{subsec:obsUpdate}) and task conditional Kalman time update; (\ref{subsec:timeUpdate})
\end{enumerate}

}
\end{algorithm}
\section{Experiments}
This section evaluates our approach on a diverse set of dynamical systems from the robotics domain in simulations and real systems. We show that HiP-RSSM outperforms contemporary recurrent state-space models (RSSMs) and recurrent neural networks (RNNs) by a significant margin under changing dynamics scenarios. Further, we show that HiP-RSSM outperforms these models even under situations with partial observability/ missing values. We also baseline our HiP-RSSM with contemporary multi-task models and improve performance, particularly in modelling non-Markovian dynamics and under partial observability. Finally, the visualizations of the Gaussian latent task variables in HiP-RSSM demonstrates that they learn meaningful representations of the causal factors of variations in dynamics in an unsupervised fashion.

We consider the following baselines:
\begin{itemize}
  \item {RNNs} - We compare our method to two widely used recurrent neural network architectures, LSTMs~\parencite{lstm} and GRUs~\parencite{gru}. 
  \item {RSSMs} - Among several RSSMs from the literature, we chose RKN~\cite{becker2019recurrent} as these have shown excellent performance for dynamics learning~\parencite{shaj2020action} and relies on exact inference as in our case.
  \item {Multi Task Models} - We also compare with state of the art multi-task learning models like Neural Processes (NPs) and MAML~\parencite{nagabandi2018learning}. Both models receive the same context information as in HiP-RSSM.
\end{itemize}

In case of recurrent models we replace the HiP-RSSM cell with a properly tuned LSTM, GRU and RKN Cell respectively, while fixing the encoder and decoder architectures. For the NP baseline we use the same context encoder and aggregation mechanism as in HiP-RSSM to ensure a fair comparison. We create partially observable settings by imputing $50\%$ of the observations during inference. More details regarding the baselines and hyperparameters can be found in the Appendix \ref{sec:dataHip}.

\subsection{Soft Robot Playing Table Tennis}
\label{subsec:pam}
We first evaluate our model on learning the dynamics of a pneumatically actuated muscular robot. This four Degree of Freedom (DoF) robotic arm is actuated by Pneumatic Artificial Muscles (PAMs)~\parencite{buchler2016lightweight}. The data consists of trajectories of hitting movements with varying speeds while playing table tennis~\parencite{buchler2020learning}. This robot's fast motions with high accelerations are complicated to model due to hysteresis and hence require recurrent models~\parencite{shaj2020action}.

We show the prediction accuracy in the RMSE in Table \ref{fig:pamtab}. We observe that the HiP-RSSM can outperform the previous state of the art predictions obtained by recurrent models. Based on domain knowledge, we hypothesize that the latent context variable captures multiple unobserved causal factors of variation that affect the dynamics in the latent space, which are not modelled in contemporary recurrent models. These causal factors could be, in particular, the temperature changes or the friction due to a different path that the Bowden trains take within the robot. Disentangling and interpreting these causal factors can be exciting and improve generalization, but it is out of scope for the current work. Further, we find that the multitask models like NPs and MAML fail to model these dynamics accurately compared to all the recurrent baselines because of the non-markovian dynamics resulting from the high accelerations in this pneumatically actuated robot.
\begin{table*}[t]
\begin{minipage}{.5\linewidth}
\centering
\scalebox{0.65}{
\begin{tabular}{lcc} 
\toprule
& No Imputation  & $50 \%$ Imputation \\ 
\midrule
HiP-RSSM & \bm{$2.30 \pm0.043$} & \bm{$2.47\pm 0.012$}    \\
RKN    & $3.088\pm 0.046$  &  $3.223\pm0.014$     \\ 
LSTM & $3.108\pm 0.041$  & $3.630\pm0.097$\\ 
GRU & $3.287\pm 0.013$ &  $3.621\pm0.047$ \\
FFNN &$8.150\pm 0.047$   & -\\ 
NP & $5.526\pm 0.030$  & -\\ 
MAML &$7.314 \pm 0.021$    & -\\ 
\bottomrule
\end{tabular}}
\subcaption{Pneumatic RMSE $\left(10^{-3}\right)$}
\end{minipage}%
\begin{minipage}{.5\linewidth}
\centering
\scalebox{0.65}{
\begin{tabular}{lcc} 
\toprule
 & No Imputation  & 50 $\%$ Imputation \\ \midrule
HiP-RSSM &\bm{$2.833 \pm 0.024$} & \bm{$2.843 \pm 0.024$}    \\ 
RKN    &  $3.392 \pm 0.062$  &   $ 3.398 \pm 0.062$\\ 
LSTM & $3.503 \pm0.006$ &  $ 3.736 \pm 0.062$ \\ 
GRU & $3.407 \pm 0.02$ & $ 3.642 \pm 0.153$\\ 
FFNN & $3.313 \pm 0.018 $  & -\\ 
NP & \bm{$2.765 \pm0.004 $}  & -\\ 
MAML & $3.202 \pm 0.006$ & -\\ 
\bottomrule
\end{tabular}}
\subcaption{Franka RMSE $\left(10^{-4}\right)$}
\end{minipage}%
\caption{Prediction Error in RMSE for (a) pneumatic muscular arm (\ref{subsec:pam}) and (b) Franka Arm manipulating varying loads (\ref{subsec:franka}) for both fully observable and partially observable scenarios.} 
\label{fig:frankatab}
\end{table*}
\subsection{Robot Manipulation With Varying Loads }
\label{subsec:franka}
We collected data from a $7$ DoF Franka Emika Panda manipulator carrying six different loads at its end effector. It involved a mix of movements of different velocities from slow to swift movements that covered the entire robot workspace. We chose trajectories with four different loads as training set and evaluated the performance on two unseen weights, which results in a scenario where the dynamics change over time. Here, the causal factor of variation in dynamics is the weights attached to the end-effector and assumed to be unobserved.
\begin{figure}[t]
\centering  
    \begin{subfigure}[t]{.25\textwidth}
        \centering
        \includegraphics[height=3.7cm,width=.9\linewidth]{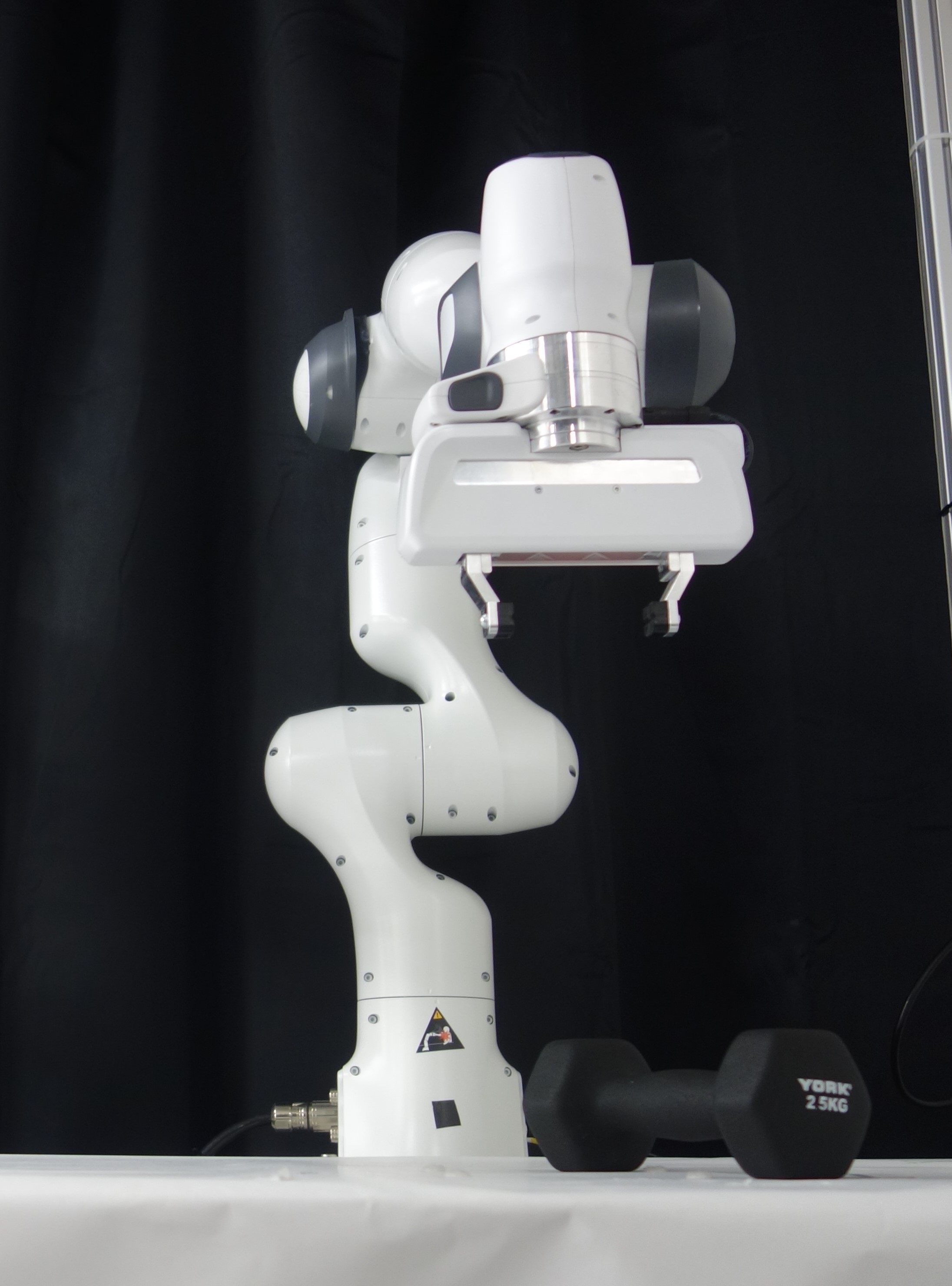}
        \caption{}
        \label{fig:franka}
    \end{subfigure}
    \hspace{1cm} 
    \begin{subfigure}[t]{.65\textwidth}
        \centering
        \includegraphics[height=3.7cm,width=\linewidth]{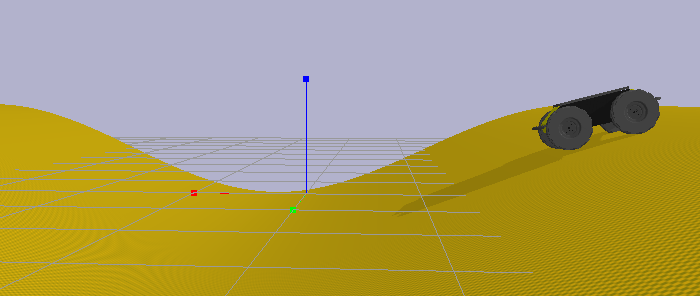}
        \caption{}
        \label{fig:wheeled}
    \end{subfigure}%
    \caption{ Figures of a subset of the agents used for collecting multi-task datasets. (left) Franka manipulator that was used to collect trajectories with different loads attached. (b) Mobile robot traversing terrain with varying slopes.}
    \label{fig:side-by-side}
\end{figure}
\par
We show the prediction errors in RMSE in Table \ref{fig:wheeltab}. HiP-RSSMs outperform all recurrent state-space models, including the RKN and deterministic RNNs, in modelling these dynamics under fully observable and partially observable conditions. The multi-task baselines of NPs and MAML perform equally well under full observability for this task because of the near Markovian dynamics of the Franka Manipulator, which often does not need recurrent models. However, HiP-RSSMs have an additional advantage in that they are naturally suited for partially observable scenarios and can predict ahead in a compact latent space, a critical component for recent success in model-based control and planning~\parencite{hafner2019learning}.
\begin{figure}[t]
    \begin{subfigure}[t]{.5\textwidth}
\includegraphics[width=\linewidth]{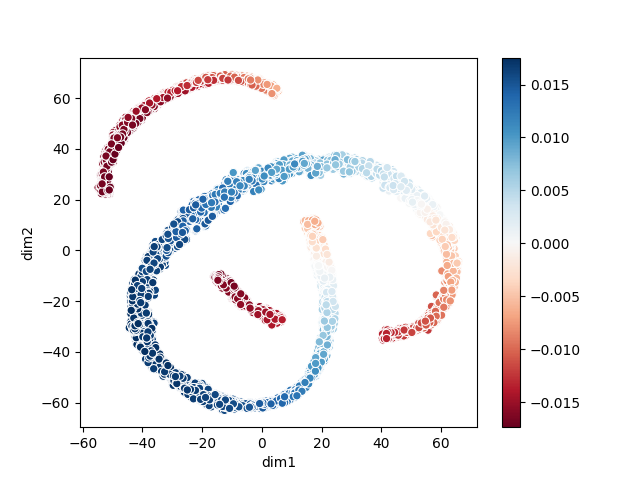}
\caption{}
\label{fig:tsne-mobile}
\end{subfigure} 
  \begin{subfigure}[t]{.5\textwidth}
\includegraphics[width=\linewidth]{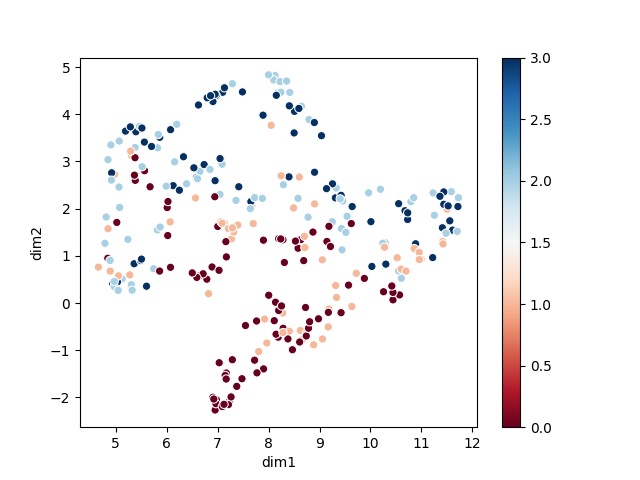}
\caption{}
\end{subfigure}
\caption{Figure shows the tSNE~\parencite{van2008visualizing} plots of the latent task embeddings produced from randomly sampled instances of HiP-RSSM for two different robots. (a) The wheeled robot discussed in section \ref{subsec:wheeled} traversing terrains of varying slopes. The color coding indicates the average gradients of the terrain for each of these instances. These can have either positive or negative values. (b) The Franka manipulator with loads of varying weights attached to the end-effector. The color coding indicated weights ranging from 0 to 3 kilograms.}
\end{figure}
\subsection{Robot Locomotion In Terrains Of Different Slopes}
\label{subsec:wheeled}
Wheeled mobile robots are the most common types of robots being used in exploration of unknown terrains where they may face uneven and challenging terrain. We set up an environment using a Pybullet simulator \parencite{coumans2016pybullet} where a four-wheeled mobile robot traverses an uneven terrain of varying steepness generated by sinusoidal functions~\parencite{sonker2020adding} as shown in \ref{fig:wheeled}. This problem is challenging due to the highly non-linear dynamics involving wheel-terrain interactions. In addition, the varying steepness levels of the terrain results in a changing dynamics scenario, which further increases the complexity of the task.
\begin{wraptable}[10]{r}{.5\linewidth}
\centering
\scalebox{0.7}{%
\begin{tabular}{lcc}
\toprule
& No Imputation & $50\%$ Imputation \\
\midrule
HiP-RSSM & \bm{$2.96 \pm 0.212$} & \bm{$6.15 \pm 0.327$} \\
RKN & $7.17\pm0.017$ & $14.66 \pm 0.224$ \\
LSTM & $9.14\pm0.026$ & $51.21 \pm 0.431$ \\
GRU & $9.216\pm0.073$ & $53.14 \pm 0.242$ \\
FFNN & $8.72\pm0.021$ & - \\
NP & $4.57\pm0.013$ & - \\
MAML & $5.04\pm0.051$ & - \\
\bottomrule
\end{tabular}
}
\caption{Prediction error for wheeled mobile robot trajectories in RMSE ($10^{-5}$) for both fully observable and partially observable scenarios.}
\label{fig:wheeltab}
\end{wraptable}

We show the prediction errors in RMSE in Table \ref{fig:wheeltab}. When most recurrent models, including RSSMs and deterministic RNNs, fail to model these dynamics, HiP-RSSMs are by far the most accurate in modelling these challenging dynamics in the latent space. Further, the HiP-RSSMs perform much better than state-of-the-art multitask models like NPs and MAML.

\par
We finally visualize the latent task representations using TSNE in Figure \ref{fig:tsne-mobile}. As seen in the plot, the HiP-SSM instances under similar terrain slopes cluster in the latent task space, indicating that the model correctly identifies causal effects in the dynamics in an unsupervised fashion. 
\subsection{Ablation Study for Latent Task Conditioning.}
\begin{wrapfigure}[10]{r}{.5\linewidth}
\vspace{-1cm}
 \scalebox{0.95}{\includegraphics[width=\linewidth]{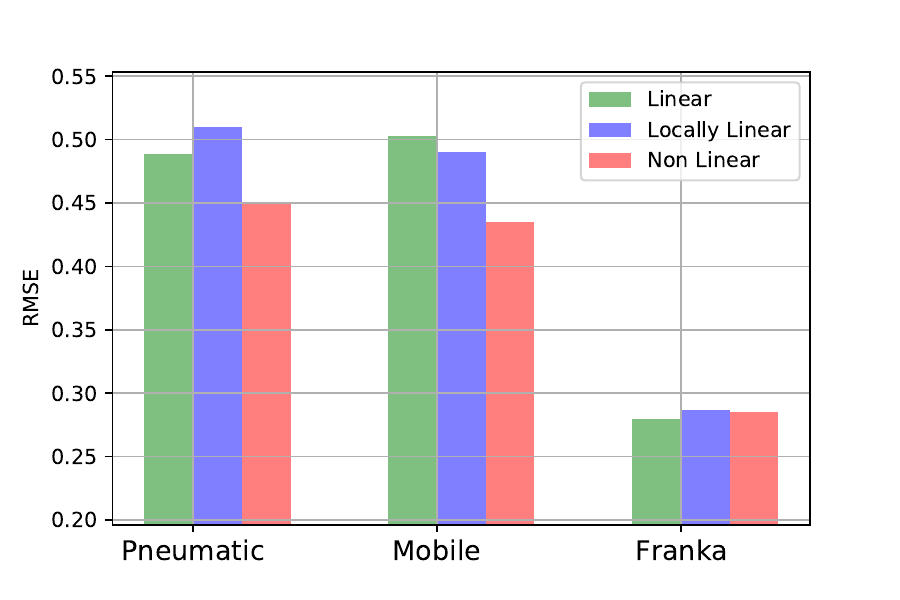}}
\caption{ An ablation on the performance of different task transformation models discussed in Sections \ref{subsec:timeUpdate} and \ref{app:taskmodel}.} 
 \label{fig:ablTask}
 \end{wrapfigure}
We evaluated the performance of the model for different action conditioning schemes (linear, locally-linear and non-linear) as discussed in Section \ref{app:taskmodel}. The resulting evaluation can be seen in Figure \ref{fig:ablTask} and shows the advantage of using nonlinear models for the additive task conditioning of latent dynamics to obtain accurate mean predictions under RMSE loss.

\subsection{Ablation On Context Encoders For Inferring Task Abstractions}
In table \ref{tab:rectab}, we report the details of evaluating our Bayesian aggregation-based context set encoder~(discussed in \ref{subsec:latentTask}) against a causal / recurrent encoder that takes into account the temporal structure of the context data due to its sequential nature. We used a probabilistic recurrent encoder~\parencite{becker2019recurrent}, whose mean and variance from the last time step is used to infer the posterior latent task distribution $p(\cvec{l}|\mathcal{C}_{\cvec{l}}) = \mathcal{N}(\cvec{\mu}_{\cvec{l}}, \textrm{diag}(\cvec{\sigma}_{\cvec{l}}^2))$. The dimensions of the latent task parameters obtained from both the set and the recurrent encoders remain the same (60). \\
The reported experiments are conducted on data from a wheeled mobile robot discussed in Section \ref{subsec:wheeled}. As reported in Table \ref{tab:rectab}, the permutation-invariant set encoder outperforms the recurrent encoder by a good margin in terms of prediction accuracy for both fully and partially observable scenarios. Additionally, the set encoder is far more efficient in terms of computational time required for training, as seen from the time taken per epoch for each of these cases, as it allows for efficient parallelization.
\begin{table}[H]
\caption{Comparison between the permutation invariant set encoder and recurrent encoder. The performance is measured in terms of prediction RMSE (10-5) and mean of the training time per epoch (in seconds) over 5 runs. }
\resizebox{\columnwidth}{!}{
\begin{tabular}{lccc} 
\toprule
& No Imputation RMSE  & $50 \%$ Imputation RMSE & Training Time Per Epoch  \\ 
\midrule
Set Encoder & \bm{$2.96 \pm 0.212$} & \bm{$6.15 \pm 0.327$}   & \bm{$6.71$} \\
Recurrent Encoder & $5.10 \pm0.041$ & $10.12\pm 0.112$  & 14.13  \\
\bottomrule
\end{tabular}
\label{tab:rectab}
}
\end{table}

\subsection{Visualizing Changing Hidden Parameters At Test Time Over Trajectories With Varying Dynamics}

\begin{figure}[H]
\begin{subfigure}{.5\textwidth}
\includegraphics[width=1.1\linewidth]{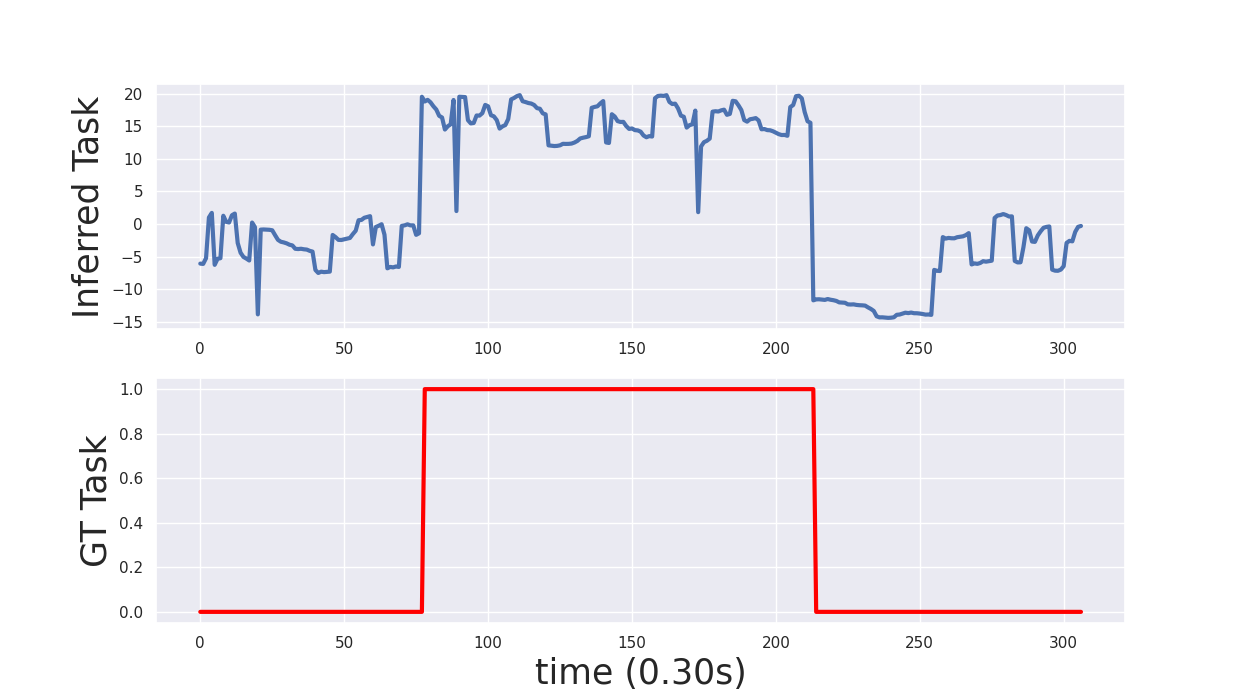}
\caption{}
\label{fig:frankalatent}
\end{subfigure}%
 \begin{subfigure}{.5\textwidth}
\includegraphics[width=1.1\linewidth]{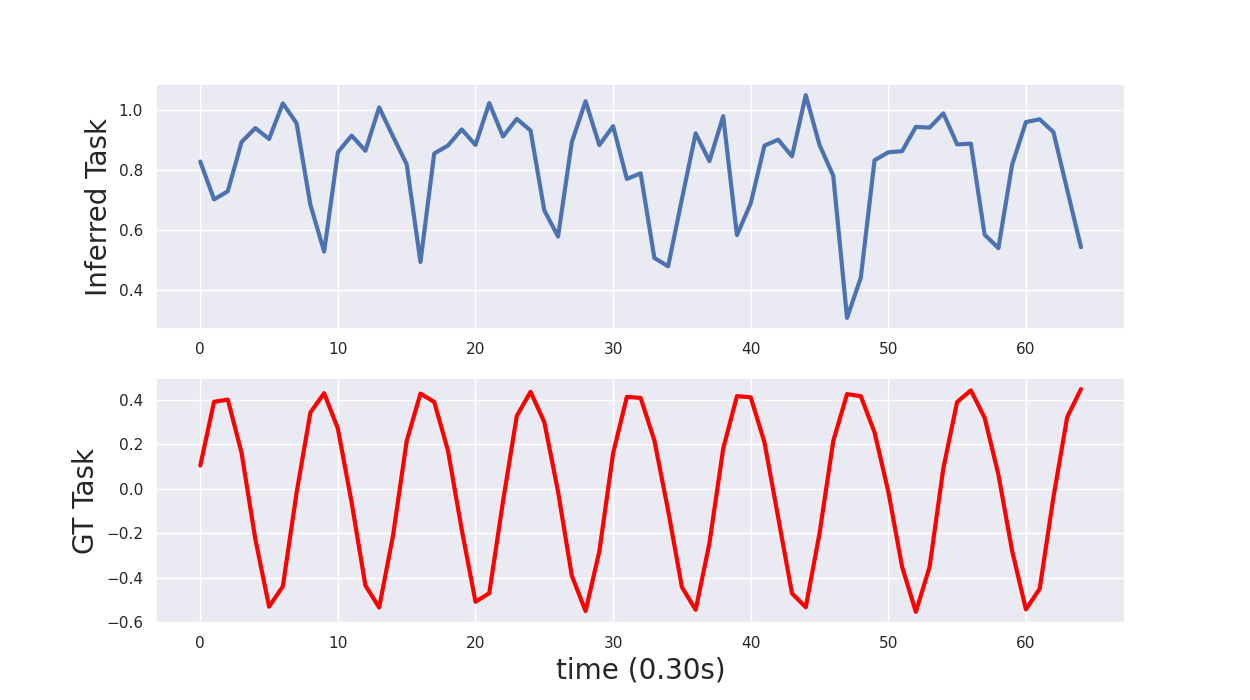}
\caption{}
\label{fig:wheeledlatent}
\end{subfigure} 
\caption{ (a) and (b) shows how the one dimensional UMAP~\parencite{mcinnes2018umap} embedding of the inferred latent task variable (top blue plots) changes at test time when an agent undergoes changes in its dynamics for the franka robot and mobile robot respectively. An indicator of the Ground Truth (GT) Task (bottom red plots) variables are also given. In case of the Franka Robot, the groundtruth (GT) tasks denotes the switching of dynamics between 0 kg (free motion) and 2.5 kg loads. In case of the mobile robot the groundtruth (GT) tasks denoted slopes of the terrain.}
\end{figure}

We finally perform inference using the trained HiP-RSSM in a multi-task / changing dynamics scenario where the dynamics continuously changes over time. We use the inference procedure described in appendix \ref{algo:hipinfer} based on a fluid definition for ``task'' as the local dynamics in a temporal segment. We plot the global variation in the latent task variable captured by each instance of the HiP-RSSM over these local temporal segments using the dimensionality reduction technique UMAP~\parencite{mcinnes2018umap}. As seen in Figures \ref{fig:frankalatent} and \ref{fig:wheeledlatent}, the latent task variable captures these causal factors of variations in an interpretable manner.

\section{Conclusion}
We proposed HiP-RSSM, a probabilistically principled recurrent neural network architecture for modelling the world under non-stationary dynamics. We start by formalizing a new framework called the Hidden Parameter State Space Model (HiP-SSM), to address the multi-task state-space modelling setting. HiP-SSM assumes a shared latent state and action space across tasks but additionally assumes latent structure in the dynamics.  We exploit the structure of the resulting Bayesian network to learn a universal dynamics model with latent parameter $l$ via exact inference and backpropagation through time. The resulting recurrent neural network, namely HiP-RSSM, learns to cluster SSM instances with similar dynamics together in an unsupervised fashion. Our experimental results on various robotic benchmarks show that HiP-RSSMs significantly outperform state-of-the-art recurrent neural network architectures on dynamics modelling tasks. We believe that modelling the dynamics in the latent space which disentangles the state, action and task representations can benefit multiple future applications including latent planning/control under non-stationary dynamics and causal factor identification.

\label{chap:Hipssm}
\chapter{Multi Time Scale SSMs: Hierarchical World Models at Multiple Temporal Abstractions}
\begin{chapquote}{}
This chapter is based on "Multi Time Scale World Models"~\parencite{shaj2023multi}.
\end{chapquote}

\begin{figure}[t]
\centering
\includegraphics[width=1\linewidth]{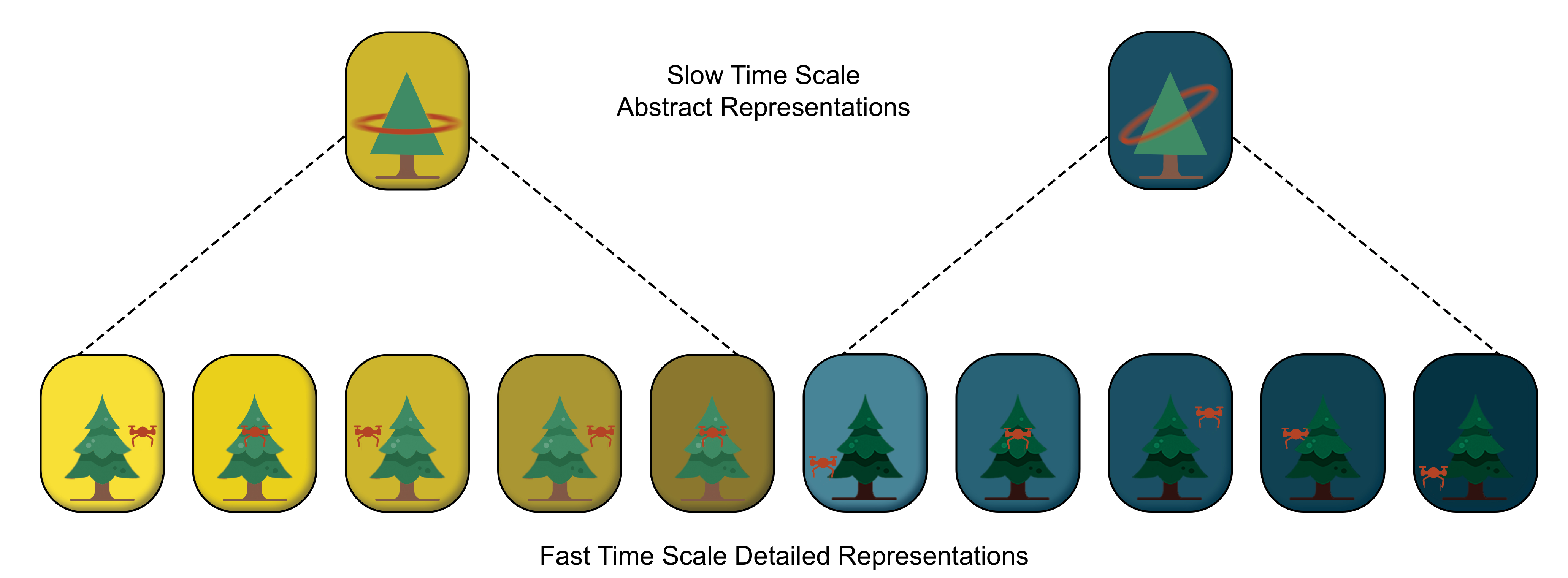}
    \caption{Conceptual Depiction of Hierarchical Temporal Abstraction in World Models. At the lower level, represented by the array of yellow and blue squares, we observe a drone's movement patterns around a central tree—circular during the day (indicated by a yellow background) and elliptical at night (indicated by a blue background). The upper tier conveys a more abstracted temporal perspective, encapsulating more invariant representations without the granular detail. This abstraction captures the cyclical day-to-night transition and the drone's corresponding behavioral patterns: circular in daylight, elliptical in darkness. This model illustrates how complex, time-variant dynamics are captured within multi-scale temporal frameworks. The intuition is formalized in Section \ref{sec:mts3}.}
    \label{fig:abstractions}
\end{figure}

This chapter of the thesis explores the concept of "hierarchical depth" within generative models of the world by structuring them through nested causal hierarchies that span across different spatio-temporal scales (see Figure \ref{fig:abstractions}). This approach addresses a significant limitation found in the prevailing literature on world models: the reliance on a singular hierarchical depth or time scale. Present formulations in the fields of machine learning and artificial intelligence typically operate with a single, finely detailed temporal scale, such as milliseconds. Singular time scale often limits accurate prediction and planning over long hoizons~\parencite{lecun2022path}. For efficient long-term prediction and planning, the model must predict at multiple levels of temporal abstractions~\parencite {sutton1995td,precup1997multi,lecun2022path}.

The world is rife with regularities. Day follows night, seasons follow one another, milk goes sour, faulty brakes are often followed by accidents, and so on. Regularities come at different time scales, ranging from tens of milliseconds to hundreds, to seconds, minutes, and upwards towards regularities or rules that are stable over weeks, months, and years.  For example, commuting to work involves planning at a higher level slow time scale (in minutes or hours) like which route to drive (or walk) and at a lower level fast time scale (in seconds or milliseconds) like how to brake/steer or choose precise skeletal muscle actions. Similarly, in robotic manipulation, the robot must be able to perform precise and coordinated movements to grasp and manipulate the object at a fast time scale while at a slower time scale the robot must also be able to recognize and utilize higher-level patterns and structures in the task, such as the shape, size, and location of objects, and the overall goal of the manipulation task. Generative models with "temporal depth" in the form of nested hierarchies offers several advantages in dealing with such scenarios~\parencite{lee2003hierarchical, hohwy2013predictive, lecun2022path} as listed below:

\begin{itemize}
    \item \textbf{Reflection of Causal Depth}: The world is inherently structured hierarchically~\parencite{hohwy2013predictive}, with causes and effects nested within one another on various scales of time and space (for example, in autonomous driving applications, slow regularities like traffic congestion trends including peak and off-peak hours, holiday traffic patterns, and seasonal changes in road conditions affect the fast regularities like real-time decisions on route navigation and speed adjustments). A hierarchical model allows the representation of this causal depth, enabling intelligent agents to understand not just immediate sensory inputs but also the broader and interconnected causal relationships that govern the external world.

    \item \textbf{Better Long Term Prediction}: By organizing regularities in the world from faster, detailed levels to slower, more abstract levels, hierarchical models enable predictions at multiple scales of precision and time. This flexibility is fundamental to adaptive behavior, allowing for both immediate, detailed responses and longer-term planning based on abstracted patterns of regularity~\parencite{friston2008hierarchical,hohwy2013predictive}. Fast changing regularities are good for detail; slower regularities are more general and abstract. This makes sense when we consider what regularities allow us to predict. If I want to predict something with great perceptual precision, then I cannot do it very far into the future, so I need to rely on a fast changing regularity. On the other hand, predictions further into the future come at a loss of precision and often detail.

    \item \textbf{Adaptability and Transferability}: As seen in Chapter \ref{chap:Hipssm} hierarchical temporal abstractions can capture relevant task structures across dynamical systems under non-stationary conditions, which can be used to identify similarities and differences between tasks. Learning dynamics of these nested hierarchical abstractions allow for making "predictions about predictions" and subsequent transfer of knowledge learned from one task to another~\parencite{shanahan2022abstraction, lecun2022path}. This adaptability is key to navigating a world that is constantly changing, allowing for the refinement of predictions and behaviors in response to new information.
\end{itemize}  
 \par

\par
This chapter of the dissertation attempt to come up with a principled probabilistic formalism for learning such multi-time scale world models as a hierarchical sequential latent variable model. We show that such models can better capture the complex, nonlinear dynamics of a system more efficiently and robustly than models that learn on a single timescale. This is exemplified in several challenging simulated and real-world prediction tasks such as the D4RL dataset, a simulated mobile robot, and real manipulators including data from heavy machinery excavators. 
\section{Related Work}
\label{rw:mts3}
\paragraph{Multi Time Scale World Models} One of the early works that enabled environment models at different temporal scales to be intermixed, producing temporally abstract world models was proposed by \cite{sutton1995td}. The work was limited to tabular settings but showed the importance of learning environment dynamics at multiple abstractions. However, there have been limited works that actually solve this problem at scale as discussed in \cite{lecun2022path}. A probabilistically principled formalism for these has been lacking in literature and this work is an early attempt to address this issue.
\par
\paragraph{Deep State Space Models.} Deep SSMs combine the benefits of deep neural nets and SSMs by offering tractable probabilistic inference and scalability to large and high-dimensional datasets. \cite{haarnoja2016backprop, becker2019recurrent, shaj2020action} use neural network architectures based on exact inference on SSMs and perform state estimation and dynamics prediction tasks. \cite{shaj2022hidden} extend these models to modelling non-stationary dynamics. \cite{krishnan2017structured, karl2016deep, hafner2019learning} perform learning and inference in SSMs using variational approximations. However, most of these recurrent state-space models have been evaluated on very short-term prediction tasks in the range of a few milliseconds and model the dynamics at a single time scale.
\par
\paragraph{Transformers} 
Recent advances in Transformers~\parencite{vaswani2017attention, radford2019language, brown2020language}, which rely on the attention mechanism, have demonstrated superior performance in capturing long-range dependency compared to RNN models in several domains, including time series forecasting~\parencite{zhou2021informer,liu2022nonstationary} and learning world models~\parencite{micheli2023transformers}. \cite{zhou2021informer,liu2022nonstationary,nie2023a} use transformer architectures based on a direct multistep loss~\parencite{zeng2022transformers} and show promising results for long-term forecasting since they avoid the accumulation of errors from autoregression. On the other hand \cite{micheli2023transformers} uses a GPT-like autoregressive version of transformers to learn world models. These deterministic models, however, do not deal with temporal abstractions and uncertainty estimation in a principled manner. Nevertheless, we think Transformers that operate at multiple timescales based on our formalism can be a promising alternative research direction.
\section{Multi Time Scale State Space Models}
\label{sec:mts3}
\begin{figure}[h]
\centering
\includegraphics[width=0.7\linewidth]{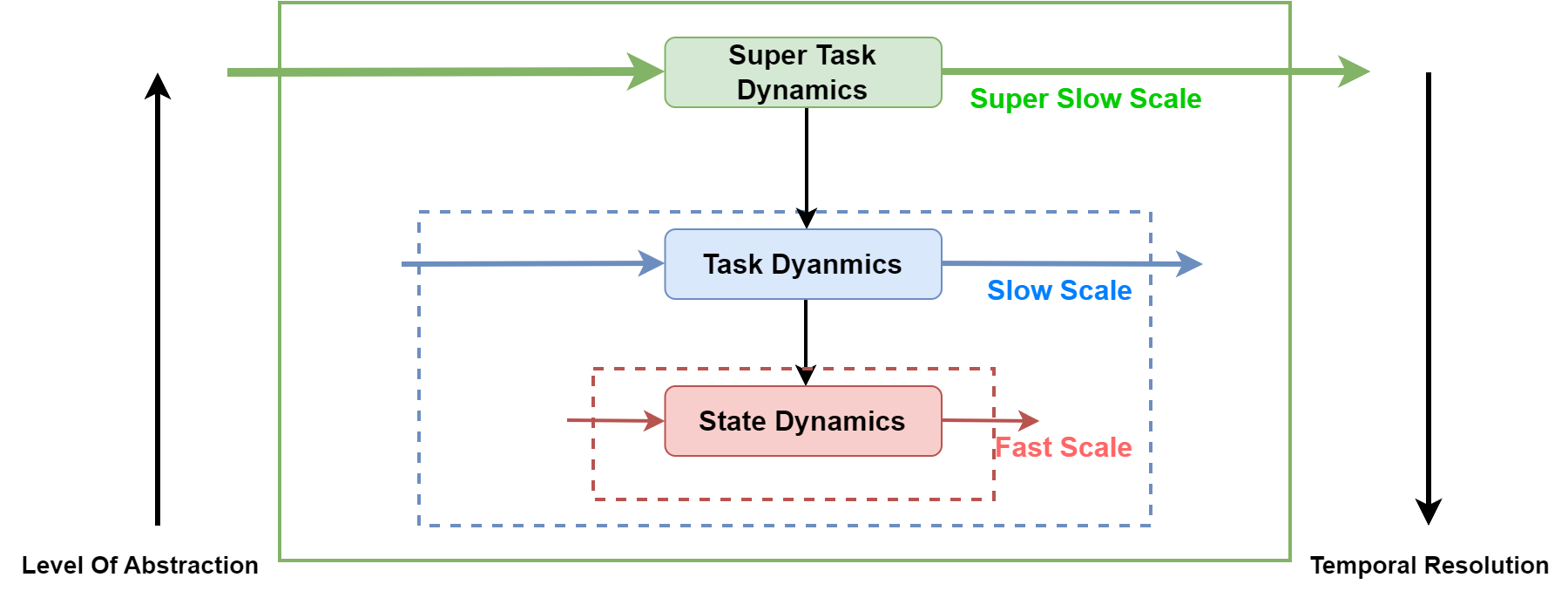}
    \caption{Regularities in our world can be ordered hierarchically, from faster to slower~\parencite{hohwy2013predictive}. Levels in the hierarchy can be connected such that certain slow regularities, at higher levels, pertain to relevant lower level, faster regularities. A complete such hierarchy would reveal the causal structure and depth of the world—the way causes interact and nest with each other across spatiotemporal scales.}
    \label{fig:ssmsvship}
\end{figure}
Our goal is to learn a principled sequential latent variable model that can model the dynamics of artificial embodied agents (specifically robotic systems) under multiple levels of temporal abstractions based on observed sensory data. To do so, we introduce a new formalism, called Multi-Time Scale State Space (MTS3) Model, with the following desiderata:
i) It is capable of modeling dynamics at multiple time scales. ii) It allows for a single global model to be learned that can be shared across changing configurations of the environments. iii) It can provide accurate long-term predictions and uncertainty estimates. iv) It is probabilistically principled, yet scalable during learning and inference.

We start by introducing a formal definition of a 2 time scale Multi-time scale SSM in Section \ref{def:2-mts3} and then proceed to a general definition of an MTS3 with an arbitrary number of hierarchies/timescales in Section \ref{subsec: N-level}.

\subsection{Formal Definition Of a 2-Level MTS3}
\label{def:2-mts3}
An MTS3 model with 2 timescales is defined by two SSMs on a fast and a slow time scale respectively. Both SSMs are coupled via the latent state of the slow time scale SSM, which parameterizes / ``reconfigures'' the system dynamics of the fast time scale SSM. While the fast time scale SSM runs at the original time step $\Delta t$ of the dynamical system, the slow time scale SSM is only updated every $H$ step, i.e., the slow time scale time step is given by $H \Delta t$. We will derive closed-form Gaussian inference for obtaining the beliefs for both time scales, resulting in variations of the Kalman update rule which are also fully differentiable and used to back-propagate the error signal \parencite{becker2019recurrent,haarnoja2016backprop}. The definition with a 2-level MTS3 along with the inference and learning schemes that we propose is directly extendable to an arbitrary number of temporal abstractions by introducing additional feudal~\parencite{dayan1992feudal} hierarchies with longer discretization steps and is further detailed in Section \ref{subsec: N-level}.

\subsubsection{Fast time-scale SSM}
\label{sec: fts}
  The fast time-scale (fts) SSM is given by $\mathcal{S}_{\textrm{fast}} = ( \mathcal{Z}, \mathcal{A}, \mathcal{O}, f_{\cvec l}^\textrm{fts}, h^\textrm{fts}, \Delta t, \mathcal{L}).$ Here, $\cvec l \in \mathcal{L}$ is a task descriptor that parameterizes the dynamics model of the SSM and is held constant for H steps. We will denote the task descriptor for the $k$th time window of $H$ steps as $\cvec l_k$. The probabilistic dynamics and observation model of the fast time scale for the $t$th time step in the $k$th window can then be described as  
 \begin{align}
  p(\cvec{z}_{k,t}|\cvec{z}_{k,t-1},\cvec{a}_{k,t-1}, \cvec l_k) & = \mathcal{N}(f^\textrm{fts}_{\cvec l}(\cvec{z}_{k,t-1},\cvec{a}_{k,t-1}, \cvec l_k), \cmat{Q}),
   \textrm{ and } \nonumber\\
   p(\cvec{o}_{k,t}|\cvec{z}_{k,t}) & = \mathcal{N}(h^\textrm{fts}(\cvec{z}_{k,t}), \cmat{R}).
 \end{align}
 \paragraph{Task-conditioned marginal transition model.}
Moreover, we have to consider the uncertainty in the task descriptor (which will, in the end, be estimated by the slow time scale model), i.e., instead of considering a single task descriptor $\cvec l_k$, we have to consider a distribution over task descriptors $p(\cvec l_k)$ for inference in the fts-SSM. This distribution will be provided by the slow-time scale SSM for every time window $k$. We can further define the marginal task-conditioned transition model for the time window $k$ that is given by 
\begin{align}p_{\cvec l_k}(\cvec{z}_{k,t}|\cvec{z}_{k,t-1},\cvec{a}_{k,t-1}) & = \int p(\cvec{z}_{k,t}|\cvec{z}_{k,t-1},\cvec{a}_{k,t-1}, \cvec l_k) p(\cvec l_k) d \cvec l_k
\end{align}
\paragraph{Latent observations.}
Following \cite{becker2019recurrent}, we replace the observations by latent observations and their uncertainty, i.e., we use latent observation encoders to obtain $\cvec w_{k,t} = \textrm{enc}_w(\cvec o_{k,t})$ and an uncertainty encoder $\cvec \sigma_{k,t} = \textrm{enc}_\sigma(\cvec o_{k,t})$. The observation model is thus given by $p(\cvec w_{k,t}|\cvec z_{k,t}) = \mathcal{N}(h^\textrm{fts}(\cvec z_{k,t}), \textrm{diag}(\cvec \sigma_{k,t}))$.

\subsubsection{Slow time-scale SSM}
\label{sts}
The slow time scale (sts) SSM only updates every time step H and uses the task parameter $\cvec l$ as a latent state representation. Formally, the SSM is defined as $\mathcal{S}_{\textrm{slow}} = ( \mathcal{L}, \mathcal{E}, \mathcal{T}, f^\textrm{sts}, h^\textrm{sts}, H \Delta t)$. It uses an abstract observation $\cvec \beta \in \mathcal{B}$ and abstract action $\cvec \alpha \in \mathcal{A}$ that summarize the observations and actions, respectively, throughout the current time window. The general dynamics model is hence given by \begin{align}p(\cvec l_{k}|\cvec l_{k-1}, \cvec \alpha_k) = \mathcal{N}(f^{\textrm{sts}}(\cvec l_{k-1}, \cvec \alpha_k), \cvec S).\end{align}

Although there exist many ways to implement the abstraction of observations and actions of time windows, we choose to use a consistent formulation by fusing the information from all $H$ time steps of time window $k$ using Gaussian conditioning. 
 \begin{figure*}[h]
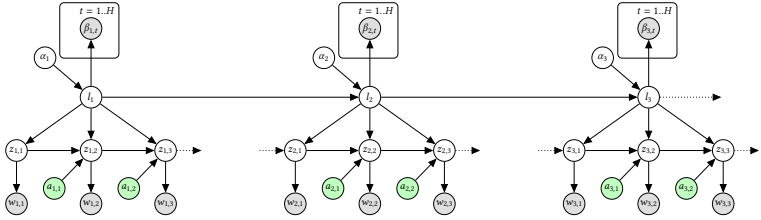

\begin{center}
 \resizebox{\linewidth}{!}{\tikzVaisakhMTS}
\end{center}
\caption{The graphical model corresponding to an MTS3 with 2 timescales. The latent task variable $\cvec l_k$ captures the slow changing dynamics using abstract observation inferred from $\{\cvec \beta_{k,t}\}_{t=1}^H$and abstract action $\cvec \alpha_k$ as described in Section \ref{sts}. The inference in the fast time scale uses primitive observations $\cvec w_{k,t}$, primitive actions $\cvec a_{k,t}$ and the latent task descriptor $l_k$ which parameterizes the fast-changing dynamics of $\cvec z_{k,t}$ for a time window k as discussed in the section \ref{sec: fts-inf}.} 
 \label{fig:mts}
\end{figure*}
\paragraph{Observation abstraction.}
In terms of the abstract observation model, we choose to model $H$ observations $\cvec \beta_{k,t}$, $t \in [1,H]$ for a single slow-scale time step $k$. All these observations can then be straightforwardly integrated into the belief state representation using incremental observation updates. The abstract observation and its uncertainty for time step $t$ is again obtained by an encoder architecture, i.e, 
$$\cvec \beta_{k,t} = \textrm{enc}_{\beta}(\cvec o_{k,t}, t), \quad \cvec \nu_{k,t} = \textrm{enc}_{\nu}(\cvec o_{k,t}, t),$$
and $p(\cvec \beta_{k,t}|\cvec l_k) = \mathcal{N}(h^{\textrm{sts}}(\cvec l_k), \textrm{diag}(\cvec \nu_{k,t})).$
Hence, the abstract observation $\cvec \beta_{k,t}$ contains the actual observation $\cvec o_{k,t}$ at time step $t$ as well as a temporal encoding for the time-step. Although multiple Bayesian observation updates are permutation invariant, the temporal encoding preserves the relative time information between the observations, similar to current transformer architectures.

\paragraph{Action abstraction.} 
\begin{wrapfigure}[13]{r}{.31\linewidth}
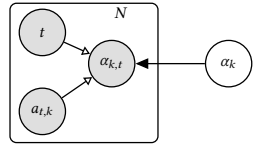

\begin{subfigure}[b]{0.31\textwidth}
   \centering\begin{adjustbox}{width=\textwidth}
         \tikzActAb
     \end{adjustbox}
     \end{subfigure}
\caption{Generative model for the abstract action $\alpha_k$. The hollow arrows are deterministic transformations leading to implicit distribution $\alpha_{k,t}$ using an action set encoder.}
\label{fig:pgmactabs}
\end{wrapfigure}
The abstract action $\cvec \alpha_k$ causes transitions to latent task $\cvec l_{k}$ from $\cvec l_{k-1}$. It should contain relevant information of all primitive actions $\cvec a_{k,t}$, $t \in [1,H]$ executed in the time window $k$. To do so, we again use Bayesian conditioning and latent action encoding. Each control action $\cvec a_{k,t}$ and the encoding of the time step $t$ are encoded in its latent representation and its uncertainty estimate, i.e.,
$$\cvec \alpha_{k,t} = \textrm{enc}_{\alpha}(\cvec a_{k,t}, t), \quad \cvec \rho_{k,t} = \textrm{enc}_{\rho}(\cvec a_{k,t}, t).$$
Single latent actions $\cvec \alpha_{k,t}$ can be aggregated into a consistent representation $\cvec \alpha_{k}$ using Bayesian aggregation \cite{volpp2020bayesian}. To do so, we use the likelihood $p(\cvec \alpha_{k,t}|\cvec \alpha_{k}) = \mathcal{N}(\cvec \alpha_{k}, \textrm{diag}( \cvec \rho_{k,t}))$ and obtain the posterior $p(\cvec \alpha_{k}|\cvec \alpha_{k,1:H}) = \mathcal{N}(\cvec \mu_{ \alpha_{k}}, \cvec \Sigma_{\alpha_k})$, which is obtained by following the standard Bayesian aggregation equations, see Appendix \ref{subsec:sts-infer}. Note that our abstract action representation also contains an uncertainty estimate which can be used to express different effects of the actions on the uncertainty of the prediction. Due to the Gaussian representations, we can compute the marginal transition model 
\begin{align}p_{\cvec \alpha_k}(\cvec l_{k}|\cvec l_{k-1}, \cvec \alpha_{k,1:H}) = \int p_{\cvec \alpha_k}(\cvec l_{k}|\cvec l_{k-1}, \cvec \alpha_k) p(\cvec \alpha_k |\cvec \alpha_{k,1:H}) d\cvec \alpha_k.\end{align}
This transition model is used for inference and its parameters are learned. 

\subsubsection{Connecting both SSMs via inference}
In the upcoming sections, we will devise Bayesian update rules to obtain the prior $p(\cvec l_k| \cvec \beta_{1:k-1},  \cvec \alpha_{1:k})$ and posterior $p(\cvec l_k| \cvec \beta_{1:k},  \cvec \alpha_{1:k})$ belief state for the sts-SSM as well as the belief states for the fts-SSM. The prior belief $p(\cvec l_k| \cvec \beta_{1:k-1},  \cvec \alpha_{1:k})$ contains all information up to time window $k-1$ and serves as a distribution over the task-descriptor of the fts-SSM, which connects both SSMs. This connection allows us to learn both SSMs jointly in an end-to-end manner.  

The probabilistic graphical model of our MTS3 model is depicted in Figure \ref{fig:mts}. In the next section, we will present the detailed realization of each SSM to perform closed-form Gaussian inference and end-to-end learning on both time scales.

\subsection{Inference in the Fast Time-Scale SSM}
\label{sec: fts-inf}
The fts-SSM performs inference for a given time window $k$ of horizon length $H$. To keep the notation uncluttered, we will also omit the time-window index $k$ whenever the context is clear. 
We use a linear Gaussian task conditional transition model, i.e, \begin{align}p(\cvec{z}_{t}|\cvec{z}_{t-1},\cvec{a}_{t-1},\cvec{l}_k) = \mathcal{N}\left(\cmat{A}\cvec{z}_{t-1} + \cmat{B}\cvec{a}_{t-1} + \cmat{C}\cvec{l}_k,  \cmat Q\right),\end{align} where $\cvec A$, $\cvec B$, $\cvec C$ and $\cvec Q$ are state-independent but learnable parameters.
In our formulation, the task descriptor can only linearly modify the dynamics which was sufficient to obtain state-of-the-art performance in our experiments, but more complex parametrizations, such as locally linear models, would also be feasible. Following \cite{becker2019recurrent}, we split the latent state $\cvec z_t = [\cvec p_t, \cvec m_t]^T$ into its observable part $\cvec p_t$ and a part $\cvec m_t$ that needs to be observed over time. 
We also use a linear observation model $p(\cvec w_{t}|\cvec z_{t}) = \mathcal{N}(\cvec H \cvec z_{t}, \textrm{diag}(\cvec \sigma_{t}))$ with $\cvec H = [\cvec I, \cvec 0]$. 

We will assume that the distribution over the task descriptor is also given by a Gaussian distribution, i.e., $p(\cvec l_k) = \mathcal{N}(\cvec \mu_{\cvec l_k}, \cvec \Sigma_{\cvec l_k})$, which will be provided by the slow-time scale (sts) SSM, see Section \ref{sts-inf}. Given these modelling assumptions, the  task variable can now be integrated out in closed form, resulting in the following task-conditioned marginal transition model
\begin{align}p_{\cvec l_k}(\cvec{z}_{t}|\cvec{z}_{t-1},\cvec{a}_{t-1}) 
 = \mathcal{N}\left(\cmat{A}\cvec{z}_{t-1} + \cmat{B}\cvec{a}_{t-1} + \cmat{C}\cvec{\mu}_{\cvec l_k},  \cmat{Q} + \cvec C \cvec \Sigma_{\cvec l_k} \cvec C^T \right), \end{align}
which will be used instead of the standard dynamics equations.
 We follow the same factorization assumptions as in \cite{becker2019recurrent} and only estimate the diagonal elements of the block matrices of the covariance matrix of the belief, see Appendix B. The update equations for the Kalman prediction and observation updates are therefore equivalent to the RKN~\parencite{becker2019recurrent}.

\subsection{Inference in the Slow-Time Scale SSM}
\label{sts-inf}
\paragraph{Prediction Update.}
We follow the same Gaussian inference scheme as for the fts-SSM, i.e., we again employ a linear dynamics model 
$p(\cvec l_{k}|\cvec l_{k-1}, \cvec \alpha_k) = \mathcal{N}(\cvec X \cvec l_{k-1} + \cvec Y \cvec \alpha_{k}, \cvec S),$ where $\cvec X$, $\cvec Y$ and $\cvec S$ are learnable parameters. The marginalized transition model for the abstract actions is then given by 

\begin{align}p_{\cvec \alpha_k}(\cvec{l}_{k}|\cvec{l}_{k-1}) & = \int p(\cvec{l}_{k}|\cvec{l}_{k-1},\cvec{\alpha}_{k}) p(\cvec \alpha_k) d \cvec \alpha_k 
 = \mathcal{N}\left(\cmat{X}\cvec{l}_{k-1} + \cmat{Y}\cvec{\mu}_{\alpha_k}, \cmat{S} + \cvec Y \cvec \Sigma_{\cvec \alpha_k} \cvec Y^T \right). \end{align}

We can directly use this transition model to obtain the Kalman prediction update which computes the prior belief $p_{\cvec \alpha_{1:k}}(\cvec l_k | \cvec \beta_{1:k-1}) = \mathcal{N}(\cvec \mu_{l_k}^-, \cvec \Sigma_{l_k}^-)$ from the posterior belief  $p_{\cvec \alpha_{1:k-1}}(\cvec l_{k-1} | \cvec \beta_{1:k-1}) = \mathcal{N}(\cvec \mu_{l_{k-1}}^+, \cvec \Sigma_{l_{k-1}}^+)$ of the previous time window, see Appendix \ref{subsec:sts-infer}. 

\paragraph{Observation/Task Update.}
\sloppy Similarly, we will use a linear observation model for the abstract observations 
$p(\cvec \beta_{k,t}| \cvec l_k) = \mathcal{N}(\cmat H \cvec l_{k}, \textrm{diag}(\cvec \nu_{k,t}))$ with $\cmat H = [\cmat I, \cmat 0]$.
As can be seen from the definition of the observation matrix $\cmat H$, the latent space is also decomposed into its observable and unobservable part, i.e., $\cvec l_k = [\cvec u_{k}, \cvec v_{k}]$. In difference to the standard factorized Kalman observation update given in Section \ref{subsec: Kalman}, we have to infer with a set of observations $\vec \beta_{k,t}$ with $t = 1 \dots H$ for a single time window $k$. While in principle, the Kalman observation update can be applied incrementally $H$ times to obtain the posterior $p_{\cvec \alpha_{1:k}}(\cvec l_{k} | \cvec \beta_{1:k}) = \mathcal{N}(\cvec \mu_{l_{k}}^+, \cvec \Sigma_{l_{k}}^+)$, such an update would be very slow and also cause numerical inaccuracies. Hence, we devise a new permutation invariant version of the update rule that allows parallel processing with set encoders~\parencite{zaheer2017deep}. We found that this update rule is easier to formalize using precision matrices. Hence, we first transform the prior covariance vectors $\cvec \sigma_{l_{k}}^{u-}$, $\cvec \sigma_{l_{k}}^{l-}$ and $\cvec \sigma_{l_{k}}^{s-}$ to its corresponding precision representation $\cvec \lambda_{l_{k}}^{u-}$, $\cvec \lambda_{l_{k}}^{l-}$ and $\cvec \lambda_{l_{k}}^{s-}$ which can be performed using block-wise matrix inversions of $\cvec \Sigma_{l_{k}}^-$. Due to the factorization of the covariance matrix, this operation can be performed solely by scalar inversions.

\begin{figure}[htb] 
\centering
    \includegraphics[scale=0.45]{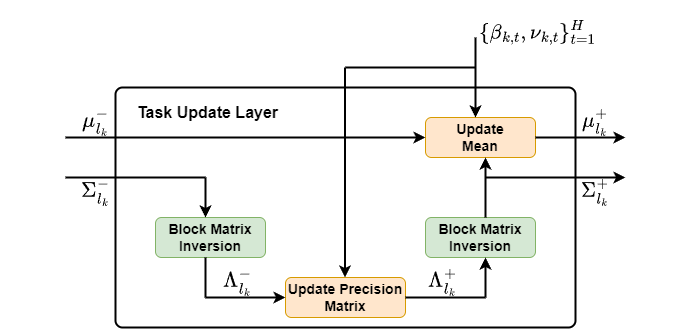}
  \caption{Implementation of task update layer which performs posterior latent task inference in the sts-SSM.}
  \label{fig:schematicTaskU}
\end{figure}

As the update equations are rather lengthy, they are given in Appendix \ref{subsec:sts-infer}, \ref{sec:proofMT}. Subsequently, we compute the posterior precision, where only $\cvec \lambda_{l_{k}}^{u}$ is changed by 
\begin{align}\cvec \lambda_{l_{k}}^{u+} = \cvec \lambda_{l_{k}}^{u-} + \sum_{t=1}^H \cvec 1 \oslash \cvec \nu_{k,t}\end{align}
 while $\cvec \lambda_{l_{k}}^{l+} = \cvec \lambda_{l_{k}}^{l-}$ and $\cvec \lambda_{l_{k}}^{s+} = \cvec \lambda_{l_{k}}^{s-}$ remain constant. The operator $\oslash$ denotes the element-wise division. From the posterior precision, we can again obtain the posterior covariance vectors $\cvec \sigma_{l_{k}}^{u+}$, $\cvec \sigma_{l_{k}}^{l+}$ and $\cvec \sigma_{l_{k}}^{s+}$ using only scalar inversions, see Appendix \ref{subsec:sts-infer}, \ref{sec:proofMT}. The posterior mean $\cvec{\mu}_{l,k}^{+}$ can now be obtained from the prior mean $\cvec{\mu}_{l,k}^{-}$ as
 \begin{equation}
  \begin{aligned}[c]
   \cvec{\mu}_{l,k}^{+}=\cvec{\mu}_{l,k}^{-} + \left[ \begin{array}{l}
\cvec{\sigma}_{l_k}^{u+} \\
\cvec{\sigma}_{l_k}^{s+} \\
\end{array}\right]  \odot \left[ \begin{array}{l}
\sum_{t=1}^{H}  \left(\cvec{\beta}_{k,t}-\cvec{\mu}^{\mathrm{u},-}_{l_k}\right) \oslash \cvec \nu_{k,t} \\
\sum_{t=1}^{H}  \left(\cvec{\beta}_{k,t}-\cvec{\mu}^{\mathrm{u},-}_{l_k}\right) \oslash \cvec \nu_{k,t} \\
\end{array}\right].   \hspace{1.5cm}\\
  \end{aligned}
\end{equation}
\notebox{For $H = 1$, i.e~a single observation, the given equation is equivalent to the factorized Kalman observation update~\parencite{becker2019recurrent}. Furthermore, the given rule constitutes a unification of the batch update rule for Bayesian aggregation \parencite{volpp2020bayesian} and the incremental Kalman update for our factorization of belief state representation \parencite{becker2019recurrent} detailed in Section \ref{sec:prelimrkn} and Appendix \ref{subsec:fts-infer}.} 

\subsection{A General Definition For an  N-level MTS3} 
\label{subsec: N-level}
The human brain's internal representation of the real world is complex, comprising more than just two distinct causal hierarchies. Neuroscience research~\parencite{hohwy2013predictive,friston2008hierarchical,jiang2021predictive} suggests that these systems exhibit multiple hierarchical structures. Consequently, we propose a formal definition for Multi-Hierarchy Theory of Mind (MTS3) that accommodates an arbitrary number (N) of hierarchies.
\begin{definition}
\sloppy An N-level MTS3 can be defined as a family of N-state space models, $\{S_0, S_1, ..., S_{N-1}\}$. Each of the state space models $S_i$ is given by $S_i = (Z_i, A_i, O_i, f_i, h_i, H_i \Delta t, L_i)$, where $Z_i$ is the state space, $A_i$ the action space, and $O_i$ the observation space of the SSM. The parameter $H_i \Delta t$ denotes the discretization time step and $f_i$ and $h_i$ the dynamics and observation models, respectively. Here, $l_i \in L_i$ is a task descriptor that parameterizes the dynamics model of the SSM and is constant for a local window of steps $H_{i+1}$. $l_i$ is a function of the latent state of the SSM one level above it, i.e., $S_{i+1}$. The boundary cases can be defined as follows: for $i=0$, $H_0 = 1$. Similarly, for $i=N-1$, the latent task descriptor $L_i$ is an empty set. For all $i$, $H_i < H_{i+1}$.
\end{definition}

\subsection{Inference In N-Level MTS3}
As seen from the definition, an N-level MTS3 is a set of N SSMs that are strictly feudal~\parencite{dayan1992feudal} from top to bottom. Top-level SSM (managers) makes decisions/predictions independently of the bottom-level SSMs, while the bottom-level SSM (worker) is conditioned on its immediate manager. Each of these SSMs performs inference via two stages at every time-step, i.e, \begin{itemize}
    \item \textbf{Gaussian Conditioning} Updating the posterior belief based on incoming observations.
    \item \textbf{Gaussian Marginalization} Estimating the prior belief for the next step via learned linear dynamics. Note that this stage is conditioned on the current belief of an immediate manager if present.
\end{itemize} 

The efficient Gaussian conditioning and marginalization processes for State-Space Models (SSMs) across all levels in the MTS3 model are specific instances of two broad Gaussian principles derived within the thesis. These principles are formally presented as Gaussian Identity \ref{theo:gc} and Gaussian Identity \ref{theo:gm}, as detailed below:

\begin{restatable}[Gaussian Conditioning]{gi}{gausscond}
    \label{theo:gc} 
    Consider the graphical model given in Figure \ref{fig:gc}, where a set of N conditionally i.i.d observations $\cvec{\Bar{r}} = \{\cvec{r}_i\}_{i=1}^N$ are generated by a latent variable $\cvec{l}$ and the observation model $p(\cvec{r}_i|\cvec{l}) = \mathcal{N}\left(\cvec{r}_i \mid \cmat{H} \cvec{l}, \textrm{diag}(\cmat{\sigma}_i^{obs})\right)$. Assuming an observation model $\cmat{H}=[\cmat{I},\cmat{0}]$, the mean ($\cvec{\mu}$) and precision matrix ($\cvec{\Lambda}$) of the posterior over the latent variable $\cvec{l}$, $p(\cvec{l}|\cvec{\Bar{r}}) = \mathcal{N}\left( \cvec{\mu}_l^{+}, \cmat{\Sigma}_l^{+} \right) = \mathcal{N}\left( \cvec{\mu}_l^{+}, (\cmat{\Lambda}_l^{+})^{-1} \right)$, given the prior $p_0(\cvec{l}) = \mathcal{N}\left( \cvec{\mu}_l^{-}, \cmat{\Sigma}_l^{-} \right) = \mathcal{N}\left( \cvec{\mu}_l^{-}, (\cmat{\Lambda}_l^{-})^{-1} \right)$ have the following permutation invariant closed form updates.
\begin{equation}
\begin{aligned}
\cmat{\Lambda}_l^{+} & = \cmat{\Lambda}_l^{-} + \left[ \begin{array}{ll}
\textrm{diag}(\sum_{i=1}^n\frac{1}{\cvec{\sigma}_i^{obs}}), &\cvec{0 }  \\
\cvec{0 },& \cvec{0} \\
\end{array}\right] \\
   \cvec{\mu}_l^{+}&=\cvec{\mu}_l^{-} + \left[ \begin{array}{l}
\cvec{\sigma}_l^{u+} \\
\cvec{\sigma}_l^{s+} \\
\end{array}\right]  \odot \left[ \begin{array}{l}
\sum_{i=1}^{N}  \left(\cvec{r}_i-\cvec{\mu}^{\mathrm{u},-}_l\right) \odot \frac{1}{\cvec{\sigma}_i^{obs}} \\
\sum_{i=1}^{N}  \left(\cvec{r}_i-\cvec{\mu}^{\mathrm{u},-}_l\right) \odot \frac{1}{\cvec{\sigma}_i^{obs}
} \\
\end{array}\right]   \hspace{1.5cm}\\
  \end{aligned}
\label{eq: ba2}
\end{equation}
\end{restatable}
\begin{proof}
The proof for the above identity is given in Appendix \ref{subsec: condProof}.
\end{proof}
\begin{figure}[H]
\begin{center}
\includegraphics[width=0.95\linewidth]{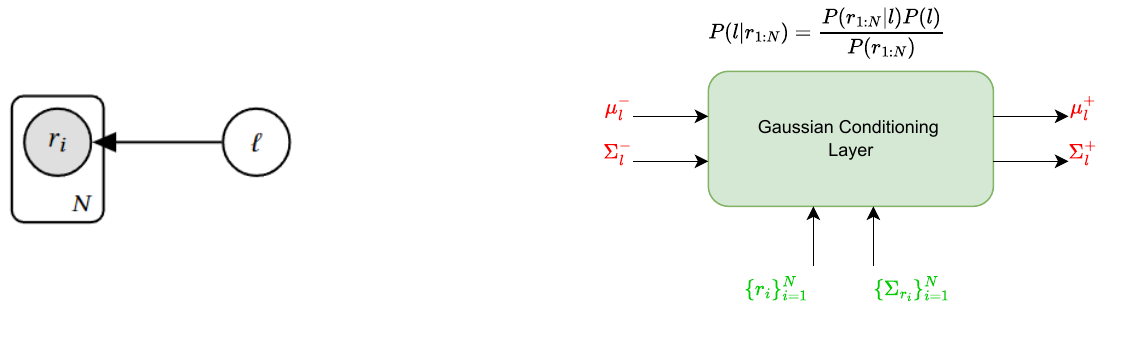}
\end{center}
\caption{Gaussian Conditioning Process and Implementation. \textbf{Left:} A graphical model illustrating the generative process underlying Gaussian conditioning, utilized for Bayesian inversion to infer distributions over model parameters (\(l\)) from observed data (\(\cvec{\Bar{o}} = \{\cvec{o}_i\}_{i=1}^N\)). This model supports various operations including observation updates, task updates, and abstract action updates, each tailored to specific conditions such as different numbers of observations (\(N\)) or observation models (\(H\)). \textbf{Right:} The Gaussian conditioning within neural network architectures across the thesis is implemented as a network layer to perform dynamic parameter updating and end-to-end learning using loss functions of choice.} 
 \label{fig:gc}
\end{figure}

\begin{coro}
\label{cor:1}
The closed form updates for the resulting posterior distribution $p(l|\Bar{r})$ is permutation invariant with respect to the observation set $\Bar{r}$.
\end{coro}
Corollary \ref{cor:1} is also a direct consequence of Theorem 1, which states that the sequential Bayesian update is permutation invariant. The permutation invariance/exchangeability property has both computational and theoretical implications in efficient posterior inference over latent task variables, which is further discussed in the section.

\begin{restatable}[Permutation Invariance Of Bayesian Inversion]{thm}{perminv}
\label{thm:permInv}
For any conditionally i.i.d model where you have a global parameter $\theta$, and a set of observations $X= \{x_i\}_{i=1}^N$ drawn conditionally i.i.d from a distribution $p(X\mid\theta)$, then for any permutation $\pi$,  the posterior $p\left(\theta \mid x_1, \ldots, x_N\right) = p\left(\theta \mid \pi(x_1), \ldots, \pi(x_N)\right)$. Thus the posterior $p\left(\theta \mid X\right)$ is permutation invariant with respect to the set $X$. 
\end{restatable}

\begin{proof}
The proof of the above theorem can be found in Appendix \ref{subsec: permInvProof}.
\end{proof}

The derived update equations can be coded as a layer in the neural network architecture as shown in Figure \ref{fig:gc}.

\begin{restatable}[Linear Combination Gaussian Marginalization]{gi}{gaussmarg}
\label{theo:gm}
\sloppy Consider the graphical model in Figure \ref{fig:gm}, where a set of N normally distributed independent random variables $\Bar{u} = \{ \cvec{u_i} \sim \mathcal{N}(\cvec{\mu}_{\cvec{u_i}},\cmat{\Sigma}_{\cvec{u_i}})\}_{i=0}^N$ forms a ``common effect/V Structure'' with a latent variable $\cvec{y}$. If the conditional distribution $p\left(\cvec{y}|\cvec{u_1}, \cvec{u_2}, . ., \cvec{u_N}\right) = \mathcal{N}( \sum_{i=0}^N \cmat{A_i}\cvec{u_i}, \cmat{\Sigma})$, then marginal $p(\cvec{y}) = \int p(\cvec{y}|\cvec{u_1}, \cvec{u_2}, . ., \cvec{u_N})\prod_{i=1}^Np(\cvec{u_i})d\cvec{u_i} = \mathcal{N}(\sum_{i=1}^N \cmat{A_i}\cvec{\mu}_{\cvec{u_i}} , \cmat{\Sigma} + \sum_{i=1}^N \cmat{A_i}\cmat{\Sigma}_{u_i}\cmat{A_i}^T) $.
\end{restatable}

\begin{proof}
The proof for the above identity is given in Appendix \ref{subsec: margNproof}.
\end{proof}
\begin{figure}[h]
\begin{center}
\includegraphics[width=0.95\linewidth]{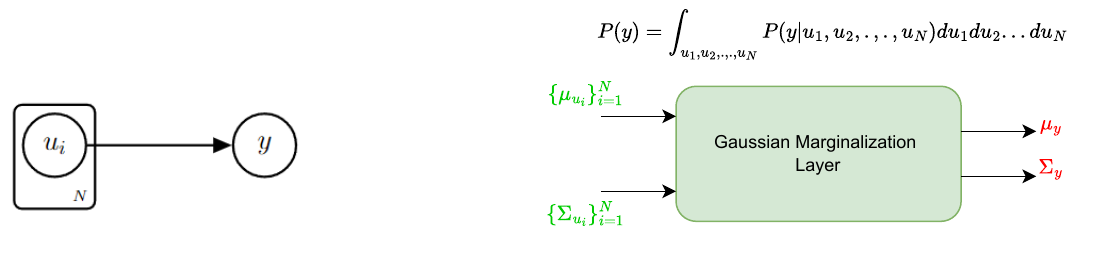}
\end{center}
\caption{Gaussian Marginalization Process and Implementation: \textbf{Left:} A graphical model illustrating the common effect of \(N\) normally distributed independent causes \(\{\cvec{c_i}\}\) on a latent variable \(\cvec{e}\). These causes are marginalized out to obtain \(p(\cvec{e})\). The marginalization can be performed in closed form as per Identity \ref{theo:gm}. The ``predict'' step across time scales in every SSM formalism proposed in the thesis is an instance of this operation. \textbf{Right:} The Gaussian marginalization within neural network architectures across the thesis is implemented as a neural network layer (with learnable parameters) to perform Gaussian marginalization/model averaging and end-to-end learning.} 
 \label{fig:gm}
\end{figure}

Even though our experiments focus on MTS3 models with 2 hierarchies, extensive experimentation with more hierarchies can be taken as future work.

\section{MTS3 as a Hierarchical World Model}

MTS3 allows for a natural way to build world models that can deal with partial observability, nonstationarity, and uncertainty in long-term predictions, properties which are critical for model-based control and planning. Furthermore, introducing several levels of latent variables, each working at a  different time scale allows us to learn world models that can make action conditional predictions/``dreams'' at multiple time scales and multiple levels of state and action abstractions. 


\subsection{Conditional Multi Time Predictions With World Model} 

Conditional multistep ahead predictions involve estimating plausible future states of the world resulting from a sequence of actions. Our principled formalism allows for action-conditional future predictions at multiple levels of temporal abstraction. The prediction update for the sts-SSM makes prior estimates of future latent task variables conditioned on the abstract action representations. However, the task conditional prediction update in the fts-SSM estimates the future prior latent states, conditioned on primitive actions and the inferred latent task priors, which are decoded to reconstruct future observations. 
To initialize the prior belief $p(\cvec z_{k,1})$ for the first time step of the time window $k$, we use the prior belief $p(\cvec z_{k-1, H+1})$ for the last time step of the time window $k-1$.

\begin{figure*}[h]
\begin{center}
\includegraphics[width=0.95\linewidth]{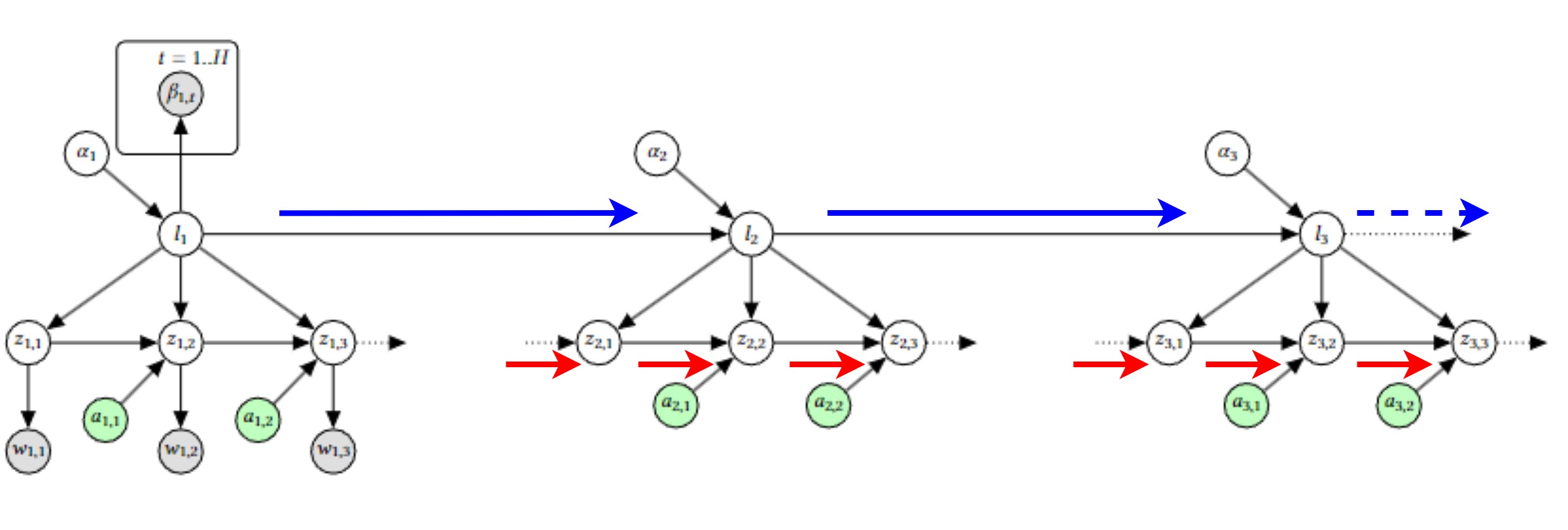}
\end{center}
\caption{Using hierarchical internal world models to envision the future: Our predictions
about future world states rely on \textbf{top-down} predictions at multiple time scales and abstractions. Initially, the model generates abstract predictions at the highest level, denoted $l_k$, using abstract actions ($\alpha_k$): this process is symbolized by \textbf{\textcolor{blue}{blue arrows}}. Subsequently, each higher-level abstract state adjusts the more detailed, lower-level granular states ($z_{k,t}$), represented by \textbf{\textcolor{red}{red arrows}}, over a period determined by the time scale parameter $H$. This mechanism allows the higher level to effectively make "predictions about predictions" concerning the lower level.} 
 \label{fig:mts}
\end{figure*}
\subsection{Optimizing the Predictive Log-Likelihood} 
The training objective for the MTS3 involves maximizing the posterior predictive log-likelihood which is given below for a single trajectory, i.e., 
\begin{align}
\label{eq:objective}
     L & = \sum_{k=1}^N \sum_{t=1}^H \log p(\cvec{o}_{k,t+1}|\cvec{\beta}_{1:k-1},\cvec{\alpha}_{1:k},\cvec{w}_{k,1:t}, \cvec{a}_{k,1:t}) \nonumber \\ & 
     = \sum_{k=1}^N \sum_{t=1}^H \log \iint p(\cvec{o}_{k,t+1}|\cvec{z}_{k,t+1})  p(\cvec{z}_{k,t+1}|\cvec{w}_{k,1:t}, \cvec a_{k,1:t}, \cvec l_k) \\ & \hspace{4cm}p(\cvec l_k|\cvec{\beta}_{1:k-1},\cvec{\alpha}_{1:k})d\cvec{z}_{k,t+1} d \cvec l_k \nonumber \\
     & =  \sum_{k=1}^N \sum_{t=1}^H \log \int p(\cvec{o}_{k,t+1}|\cvec{z}_{k,t+1})  p_{\cvec l_k}(\cvec{z}_{k,t+1}|\cvec{w}_{k,1:t}, \cvec a_{k,1:t}) d\cvec{z}_{k,t+1}.  
\end{align}
The extension to multiple trajectories is straightforward and was omitted to keep the notation uncluttered. Here, $\cvec{o}_{k,t+1}$ is the ground-truth observations at time step $t+1$ and time window $k$ that needs to be predicted from all (latent and abstract) observations up to time step $t$. The prior belief corresponding to the latent state $p_{\cvec l_k}(\cvec{z}_{k,t+1}|\cvec{w}_{k,1:t}, \cvec a_{k,1:t})$ has a closed form solution as discussed in Section \ref{sec: fts-inf}. 
\par
\sloppy We employ a Gaussian approximation of the posterior predictive log-likelihood of the form $ p(\cvec{o}_{k,t+1}|\cvec{\beta}_{1:k-1},\cvec{\alpha}_{1:k},\cvec{w}_{k,1:t}, \cvec{a}_{k,1:t}) \approx \mathcal{N}(\cvec{\mu}_{\cvec{o}_{k,t+1}},\textrm{diag}(\cvec{\sigma}_{\cvec{o}_{k,t+1}}))$ where we use the mean of the prior belief $\cvec{\mu}_{z_{k,t+1}}^-$ to decode the predictive mean, i.e, $\cvec{\mu}_{\cvec{o}_{k,t+1}} =  \textrm{dec}_{\cvec{\mu}}(\cvec{\mu}_{z_{k,t+1}}^{-})$ and the variance estimate of the prior belief to decode the observation variance, i.e., $\cvec{\sigma}_{o_{k,t+1}} = \textrm{dec}_{\sigma}(\cmat{\Sigma}_{z_{k,t+1}}^{-})$. This approximation can be motivated by a moment matching perspective and allows for end-to-end optimization of the logarithmic likelihood without using auxiliary objectives such as the ELBO \cite{becker2019recurrent}.
 \begin{figure*}[h]
\centering
\includegraphics[scale=0.12]{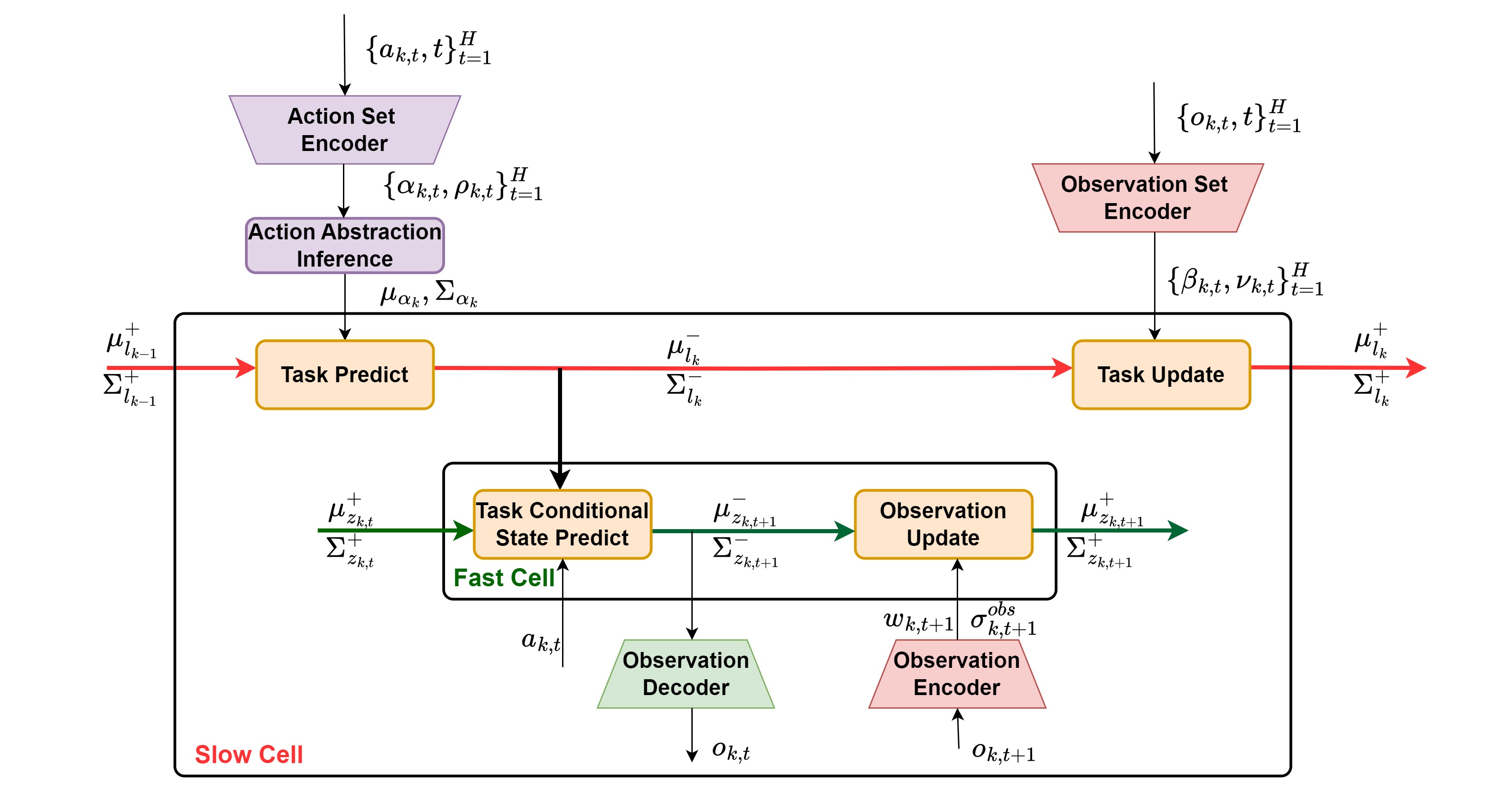}
\caption{Schematic of a 2-Level MTS3 Architecture. Inference in MTS3 takes place via closed-form equations derived using exact inference, spread across two-time scales. For the fast time scale (fts) SSM, these include the task conditional state predict and observation update stages as discussed in Section \ref{subsec:fts-infer} of the main paper. Whereas, for the slow time scale (sts) SSM, these include the task prediction and task update stages which are described in Section \ref{subsec:sts-infer}. More implementation details can also be found in Appendix \ref{sec:mts3-impl}.} 
 \label{fig:mts-schematic}
\end{figure*}
\par
Gradients are computed using (truncated) backpropagation over time (BPTT)~\parencite{werbos1990backpropagation} and clipped.  
We optimize the objective using the Adam~\parencite{kingma2014adam} stochastic gradient descent optimizer with default parameters. A schematic of MTS3 architecture is shown in Figure \ref{fig:mts-schematic}. We refer to Appendix \ref{app:mts3} for more details. For training, we also initialize the prior belief $p(\cvec z_{k,1})$ with the prior belief $p_{\cvec l_{k-1}}(\cvec z_{k-1,H+1}|\cvec w_{k-1,1:H},\cvec a_{k-1,1:H})$ from the previous time window $k-1$. However, we cut the gradients for the fast time scale between time windows as this avoids vanishing gradients, and we observed a more stable learning behavior. Yet, the gradients can still flow between time windows for the fts-SSM via the sts-SSM.
\subsection{Imputation Based Self Supervised Training For Long Term Prediction \label{sec:imputation}} 

\begin{figure}[H]
\centering
\includegraphics[width=0.95\linewidth]{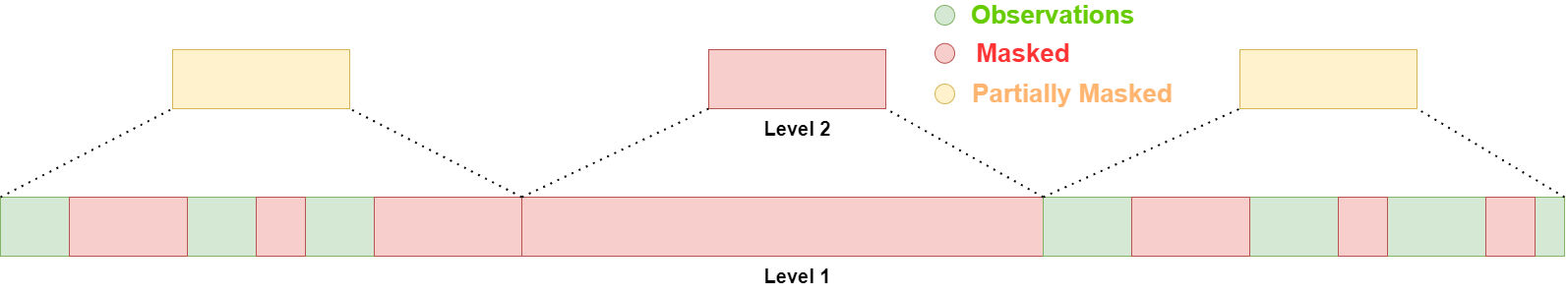}
\caption{Self-supervised training for multi-step ahead predictions by tasking the model to "fill in" the masked observations. Here masking is done at both time scales based on the available observation/set of observations in each time window.}
\label{fig:self-super-mts}
\end{figure}

Using the given training loss results in models that are good in one-time step prediction, but typically perform poorly in long-term predictions as the loss assumes that observations are always available up to time step $t$. To increase the performance of the long-term prediction, we can treat the long-term prediction problem as a case of the problem of ``missing value'', where the missing observations occur in future time steps. Thus, to train our model for long-term prediction, we randomly mask a fraction of observations and explicitly task the network to impute the missing observations, resulting in a strong self-supervised learning signal for long-term prediction with varying prediction horizon length. This imputation scheme is applied at both time scales, masking out single time steps or whole time windows of length H as shown in Figure \ref{fig:self-super-mts}. The imputation mask is also randomly resampled for every minibatch.

\section{Experiments}
In this section, we evaluate our approach to a diverse set of simulated and real-world dynamical systems for long-horizon prediction tasks. Our experiments are designed to answer the following questions. (a) Can MTS3 make accurate long-term deterministic predictions (mean estimates)? (b) Can MTS3 make accurate long-term probabilistic predictions (variance estimates)? (c) How important are the modelling assumptions and training scheme? 

\subsection{Baseline Dynamics Models}
While a full description of our baselines can be found in Appendix \ref{sec:dataMts3}, a brief description of them is given here: (a) \textbf{RNNs} - We compare our method to two widely used recurrent neural network architectures, LSTMs~\parencite{lstm} and GRUs~\parencite{gru}. 
(b) \textbf{RSSMs} - Among several RSSMs from the literature, we chose RKN~\parencite{becker2019recurrent} and HiP-RSSM~\parencite{shaj2022hidden} as these have shown excellent performance for dynamics learning for short-term predictions and rely on exact inference as in our case. (c) \textbf{Transformers} - We also compare with two state-of-the-art Transformer~\parencite{vaswani2017attention} variants. The first variant (AR-Transformer) relies on a GPT-like autoregressive prediction~\parencite{radford2019language,brown2020language}. Whereas the second variant (Multi-Transformer) uses direct multi-step loss~\parencite{zeng2022transformers} from recent literature on long horizon time-series forecasting \parencite{zhou2021informer,liu2022nonstationary,nie2023a}. Here, multistep ahead predictions are performed using a single shot given the action sequences.
\subsection{Environments and Datasets}
\begin{figure*}[b]
\centering
\begin{minipage}{0.85\textwidth}
    \centering
    \includegraphics[scale=0.6]{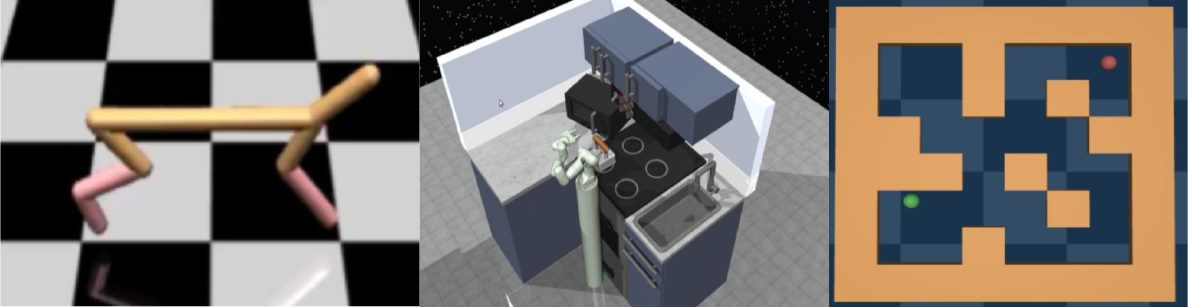}
    \label{fig:d4rl}
\end{minipage}
\\[.1cm] 
\begin{minipage}{\textwidth}
\centering
    \includegraphics[scale=0.55]{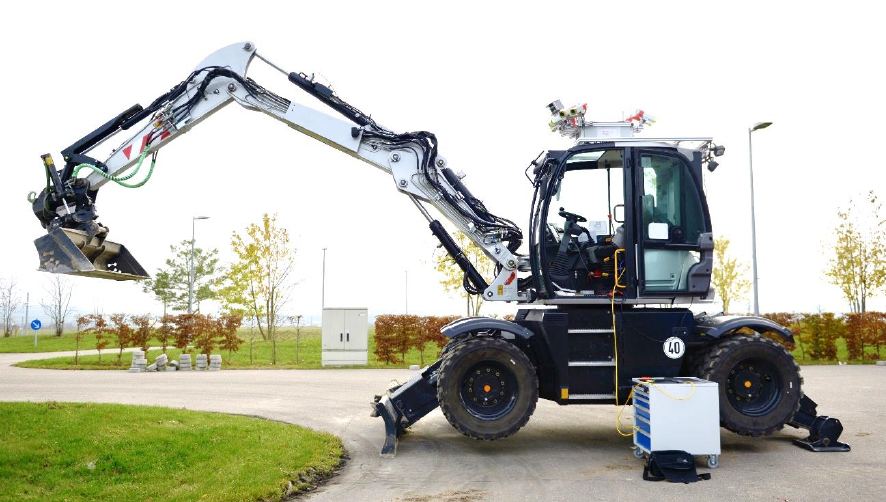}
    \label{fig:additional}
\end{minipage}
\caption{Figures of a subset of the agents used for collecting datasets. (top) D4RL Environments: HalfCheetah, Franka Kitchen, Maze2D-Medium (bottom) JCB Hydradig 110W Excavator } 
\label{fig:composite}
\end{figure*}
We experiment with three broad datasets. In all datasets, we only use information about agent/object positions and we mask out velocities to create a partially observable setting. While full descriptions of these datasets, dataset creation procedure, and overall statistics are given in Appendix \ref{sec:dataAc}, a brief description of them is as follows.
\par{\textbf{D4RL Datasets}} - We use a set of 3 different environments/agents from D4RL dataset~\parencite{fu2020d4rl}, which includes the HalfCheetah, Medium Maze and Franka Kitchen environment. Each of these was chosen because of their distinct properties like sub-optimal trajectories (HalfCheetah), realistic domains / human demonstrations (Kitchen), multi-task trajectories, non-markovian collection policies (Kitchen and Maze) and availability of long horizon episodes (all three). \par{\textbf{Manipulation Datasets}} - We use 2 datasets collected from a real excavator arm and a Panda robot. The highly non-linear non-markovian dynamics due to hydraulic actuators in the former and non-stationary dynamics owing to different payloads in the latter make them challenging benchmarks. Furthermore, accurate modelling of the dynamics of these complex systems is important since learning control policies for automation directly on large excavators is economically infeasible and potentially hazardous. \par{\textbf{Mobile Robotics Dataset}} - We set up a simulated four-wheeled mobile robot traversing a highly uneven terrain of varying steepness generated by a mix of sinusoidal functions. This problem is challenging due to the highly non-linear dynamics involving wheel-terrain interactions and non-stationary dynamics introduced by varying steepness levels. 

\begin{figure}[t]
\begin{subfigure}[b]{1.02\textwidth}
\hspace*{-1.2cm}
         \includegraphics[width=1.1\linewidth]{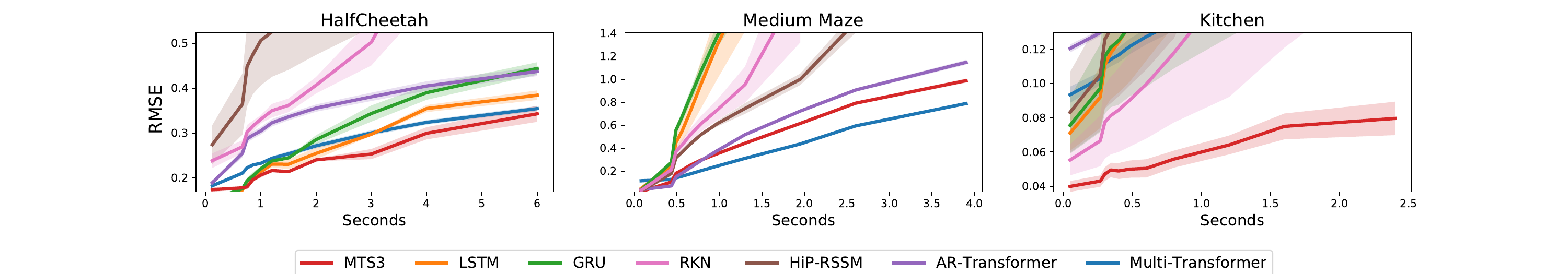}
     \label{fig:m}
     \end{subfigure}
     \\
\begin{subfigure}[b]{1.02\textwidth}
\hspace*{-1.2cm}
         \includegraphics[width=1.1\linewidth]{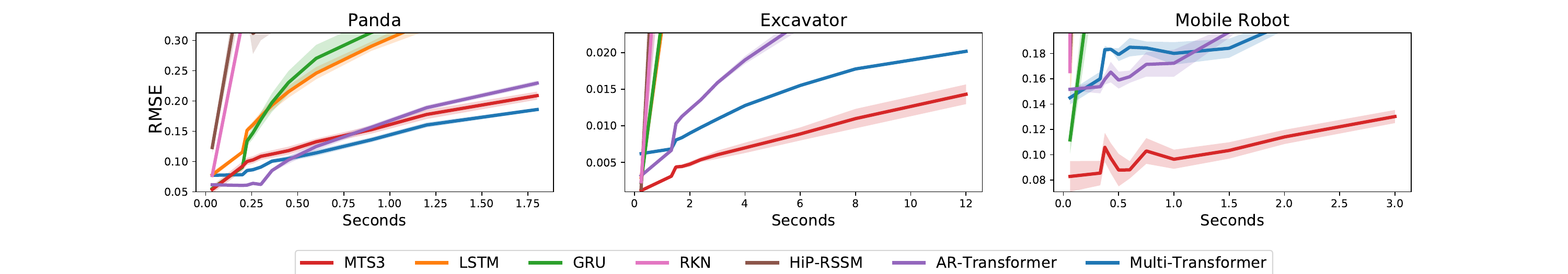}
     \label{fig:m}
     \end{subfigure}
\caption{Comparison with baselines in terms of RMSE for long horizon predictions (in seconds) as discussed in Section \ref{subsec:mean}.}
\label{fig:exps}
\end{figure}

\subsection{Can MTS3 make accurate long-term deterministic predictions (mean estimates)?}
\label{subsec:mean}
Here we evaluate the quality of the mean estimates for long-term prediction using our approach. The results are reported in terms of "sliding window RMSE" in Figure \ref{fig:exps}. We see that MTS3 gives consistently good long-term action conditional future predictions on all 6 datasets. Deep Kalman models~\parencite{becker2019recurrent,shaj2022hidden} which operate on a single time scale fail to give meaningful mean estimates beyond a few milliseconds. Similarly, widely used RNN baselines~\parencite{lstm,gru} which form the backbone of several world models~\parencite{ha2018world,hafner2019learning} give poor action conditional predictions over long horizons. AR-Transformers also fail possibly due to error accumulation caused by the autoregression. However, Multi-Transformers are a strong baseline that outperforms MTS3 in the Medium Maze and Panda dataset by a small margin. However, on more complex tasks like the Kitchen task, which requires modelling multi-object, multi-task interactions~\parencite{gupta2019relay}, MTS3 is the only model that gives meaningful long horizon predictions. A detailed description of the metric "sliding window RMSE" is given in Appendix \ref{sec:mts3-metric}. A visualization of the predicted trajectories vs. ground truth is given in Appendix \ref{sec:mts3-vis}.
\begin{figure*}[t]
\includegraphics[scale=0.45]{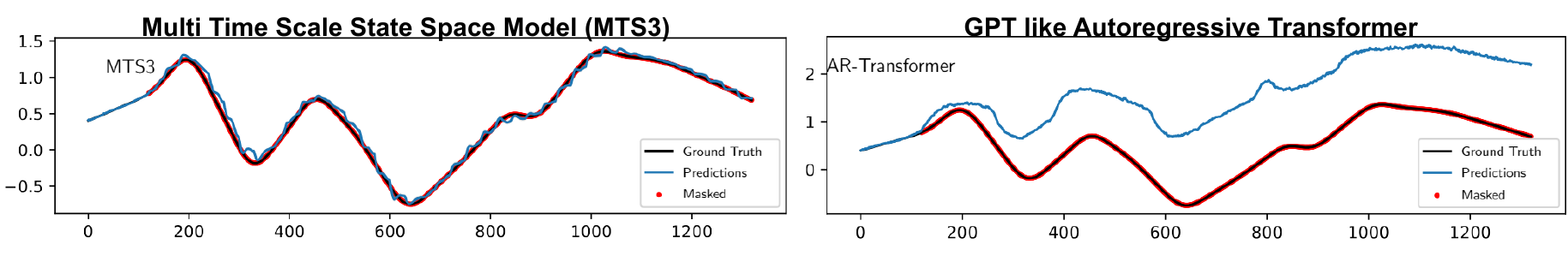}
\caption{Visualizations of the predicted trajectories vs ground truth for MTS3 and GPT like autoregressive transformer (AR-Transformer). More visualizations across algorithms and datasets can be found in Appendix \ref{sec:mts3-vis}.} 
 \label{fig:gptmts3}
\end{figure*}

\subsection{Can MTS3 make accurate long-term probabilistic predictions (variance estimates)?}
Next, we examine the question of whether the principled probabilistic inference translates to accurate uncertainty quantification during long-horizon predictions. We trained all the baselines with a negative
log-likelihood loss and used the same as a metric to quantify the quality of uncertainty estimates. During the evaluation we rely on a sliding window approach (see Appendix \ref{sec:mts3-metric}) and report the results for the last timestep in Table \ref{table:nll}. As seen in Table \ref{table:nll}, MTS3 gives the most accurate uncertainty estimates in all datasets except Medium Maze, where it is outperformed by Multi-Transformer. Also, notably, AR-Transformers and deep Kalman models fail to learn any meaningful uncertainty representation when it comes to long-term predictions. 
\begin{table}[H]
    \centering
    \resizebox{\columnwidth}{!}{
    \begin{tabular}{|c|c|c|c|c|c|c|c|c|c|c|}
        \hline
        {} & \textbf{Prediction} & \multicolumn{7}{c|}{\textbf{Algorithm}} \\ \cline{3-9} 
        & \textbf{Horizon} & \textbf{MTS3} & \textbf{Multi-Trans} & \textbf{AR-Trans} & \textbf{LSTM} & \textbf{GRU} & \textbf{RKN} & \textbf{HiP-RSSM} \\ \hline
        \textbf{Half Cheetah} & 6 s & $\mathbf{-2.80 \pm 0.30}$ & $0.25 \pm 0.05$ & \xmark
        & $7.34 \pm 0.06$ & $7.49 \pm 0.04$ & \xmark & \xmark \\ \cline{2-9}
        \hline
        \textbf{Kitchen} & 2.5 s & $\mathbf{-25.74 \pm 0.12}$ & $-7.3 \pm 0.2$ & \xmark & $32.45 \pm 1.64$ & $32.72 \pm 0.65$ & \xmark & \xmark \\ \cline{2-9}
        \hline
        \textbf{Medium Maze} & 4 s & $-0.21 \pm 0.022$ & $\mathbf{-0.88 \pm 0.02}$ & \xmark & $4.03 \pm 0.32$ & $7.76 \pm 0.07$ & \xmark & \xmark \\ \cline{2-9}
        \hline
        \textbf{Panda} & 1.8 s & $\mathbf{2.79 \pm 0.32}$ & $3.77 \pm 0.33$ & \xmark & $7.94 \pm 0.39$ & $7.91 \pm 0.23$ & \xmark & \xmark \\ \cline{2-9}
        \hline
        \textbf{Hydraulic} & 12 s & $\mathbf{-2.64 \pm 0.12}$ & $-2.46 \pm 0.03$ & \xmark & $7.35 \pm 0.061$ & $7.35 \pm 0.06$ & \xmark & \xmark \\ \cline{2-9}
        \hline
        \textbf{Mobile Robot} & 3 s & $\mathbf{-6.47 \pm 0.71}$ & $-5.17 \pm 0.23$ & \xmark & $11.27 \pm 2.3$ & $14.55 \pm 5.6$ & \xmark & \xmark \\ \cline{2-9}
        \hline
    \end{tabular}
    }
    \caption{Comparison in terms of Negative Log Likelihood (NLL) for long horizon predictions (in seconds). Here bold numbers indicate the top methods and \xmark~denotes very high/nan values resulting from the highly divergent mean/variance long-term predictions.}
    \label{table:nll}
\end{table}

\subsection{How important are the modelling assumptions and training scheme?}
\label{subsec:ablation}
Now, we look at three important modelling and training design choices: (i) splitting the latent states to include an unobservable ``memory'' part using observation model $h^{sts}=h^{fts}=\cmat H = [\cmat I, \cmat 0]$ as discussed in Sections \ref{sts-inf} and \ref{sec: fts-inf}, (ii) action abstractions in Section \ref{sts}, (iii) training by imputation. 
\begin{figure}[H]
\includegraphics[width=\linewidth]{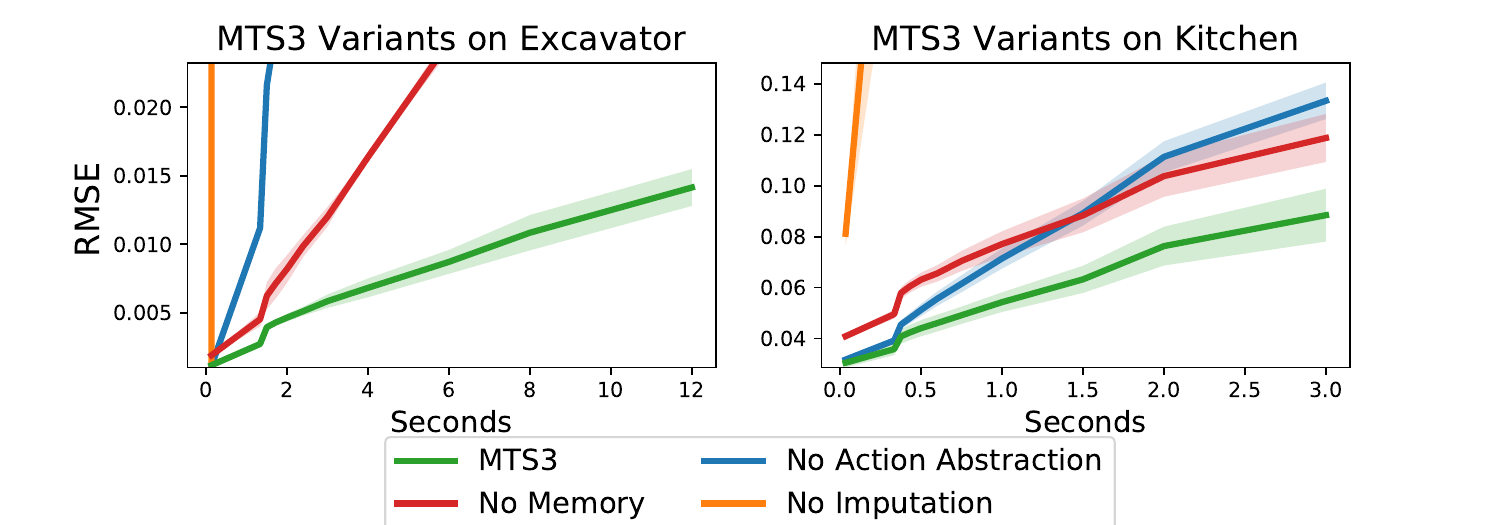}
\caption{ Ablation on discretization step $H. \Delta t$ (a) The long-term prediction results in terms of RMSE, with different $H$ values as discussed in Section \ref{subsec:abl_h} on the hydraulics dataset.} 
 \label{fig:abl}
 \end{figure} 

To analyze the importance of the memory component, we derived and implemented an MTS3 variant with an observation model of $h^{sts}=h^{fts}=\cmat I$ and a pure diagonal matrix representation for the covariance matrices. As seen in Figure \ref{fig:abl}, this results in worse long-term predictions, suggesting that splitting the latent states in its observable and unobservable part in MTS3 is critical for learning models of non-markovian dynamical systems. Regarding (ii), we further devised another variant where MTS3 only had access to observations, primitive actions and observation abstractions, but no action abstractions. As seen in our ablation studies, using the action abstraction is crucial for long-horizon predictions.

Our final ablation (iii) shows the importance of an imputation-based training scheme discussed in Section \ref{sec:imputation}. As seen in Figure \ref{fig:abl} when trained for 1 step ahead predictions without imputation, MTS3 performs significantly worse for long-term prediction suggesting the importance of this training regime.

\begin{figure}
\centering
\begin{minipage}{\textwidth}
  \centering
  \includegraphics[width=.98\linewidth]{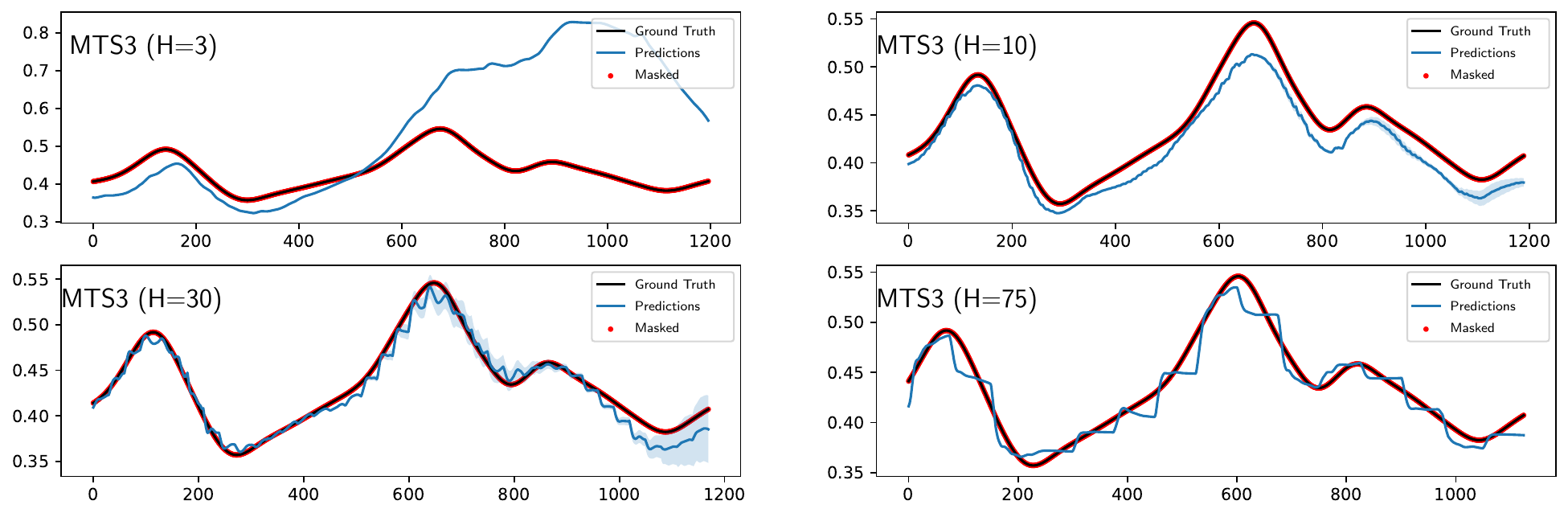}
\end{minipage}
\caption{Ablation on discretization step $H. \Delta t$. The predictions by MTS3 variants with different values of timescale parameter $H.\Delta t$ on a trajectory picked from the hydraulics excavator dataset. The top images are for $H=3$ and 
$H=10$. The bottom images are for $H=30$ and $H=75$. Note that the results reported in the paper are with $H=30$.}
\label{fig:abl2}
\end{figure}

\subsection{What is the role of the discretization step $H.\Delta t$?}
\label{subsec:abl_h}
\begin{figure}[t]
    \centering
    \begin{minipage}{0.45\textwidth}
        \centering
        \includegraphics[width=\linewidth]{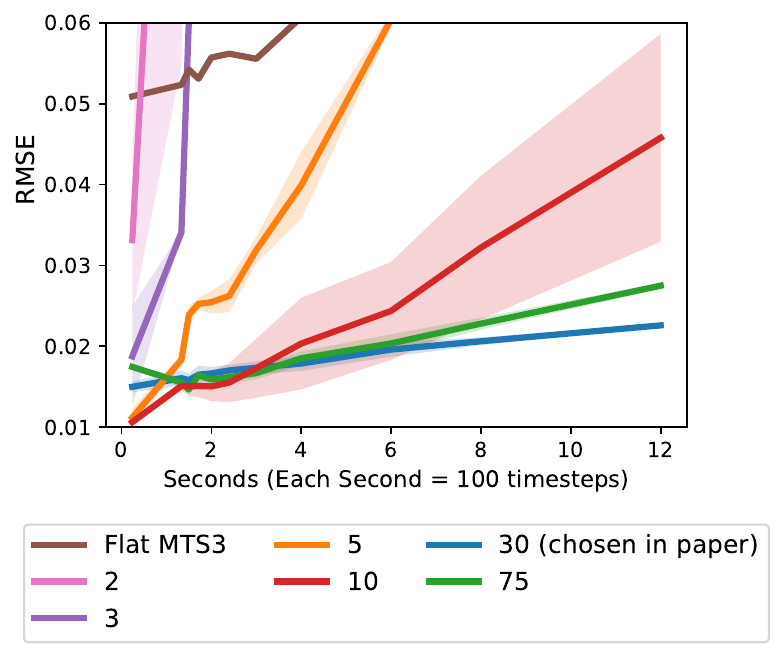} 
    \end{minipage}\hfill
    \begin{minipage}{0.45\textwidth}
        \centering
        \includegraphics[width=\linewidth]{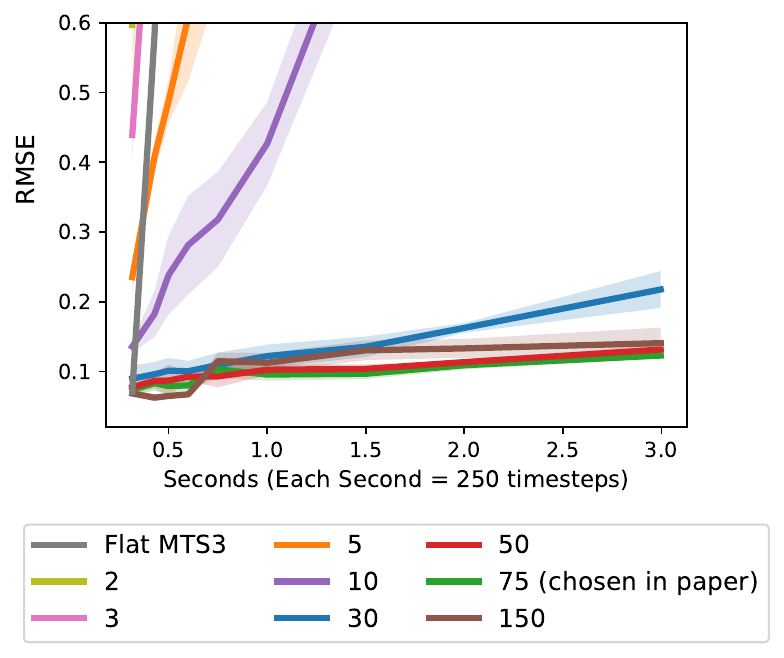} 
    \end{minipage}
    \caption{Ablation on discretization step $H. \Delta t$. The long-term prediction results in terms of RMSE, with different $H$ values as discussed in Section \ref{subsec:abl_h} on (left) the hydraulics dataset and (right) the mobile robot dataset.}
    \label{fig:habl}
\end{figure}

Finally, we perform ablation for different values of $H.\Delta t$, which controls the time scale of the task dynamics. The higher the value of H, the slower the timescale of the task dynamics relative to the state dynamics. As seen in Figure \ref{fig:habl} (left), for the hydraulics data, smaller values of $H$ (2,3,5 and 10) give significantly worse performance. Very large values of $H$ (like 75) also result in degradation of performance. To get an intuitive understanding, we plot the predictions given by MTS3 for different values of $H$ on a trajectory handpicked from the hydraulics excavator dataset. As seen in Figure \ref{fig:abl2} for large values of $H$ like 30 and 75, we notice that the upper level "reconfigures" the lower level every 30 and 75-step window respectively, by conditioning the lower level dynamics with the newly updated task prior. This effect is noticeable as periodic jumps or discontinuities in the predictions, occurring at 30 and 75-step intervals. Also, for a very large $H$ like 75, the fast time scale SSM has to make many more steps in a longer window resulting in error accumulation and poor predictions. A similar trend was observed for the mobile robot data as seen in Figure \ref{fig:habl} (right). For the mobile robot, smaller values of H (like 2,3,5 and 10) and very large values of H (like 150) gave sub-optimal performance. In the paper, we used a value of H=75.

\section{Conclusion and Future Work}
In this work, we introduce MTS3, a probabilistic formalism for learning the dynamics of complex environments at multiple time scales. By modelling the dynamics of the world at multiple levels of temporal abstraction we capture both the slow-changing long-term trends and fast-changing short-term trends in data, leading to highly accurate predictions spanning several seconds into the future. Our experiments demonstrate that simple linear models with principled modelling assumptions can compete with large transformer model variants that require several times more parameters. Furthermore, our inference scheme also allows for principled uncertainty propagation over long horizons across multiple time scales which capture the stochastic nature of environments. We believe our formalism can benefit multiple future applications including hierarchical planning/control. We discuss the limitations and broader impacts of our work in Chapter \ref{chap:conclusion}.

\label{chap:mts3}
\chapter{Outlook}
This thesis questioned the current formalisms for learning-world models due to their inability to capture some critical criteria for a foundational world model. Specifically, we looked at three aspects (i) scalable probabilistic modelling that captures causal relations in our world (Chapter \ref{chap:Acssm}, \ref{chap:Hipssm} and \ref{chap:mts3}), (ii) adaptability to changing tasks (Chapter \ref{chap:Hipssm}) and (iii) hierarchical modelling at multiple time scales and temporal abstractions (Chapter \ref{chap:mts3}).
In this context the thesis proposed two new formalisms Hidden Parameter State Space Models (HiP-SSM in Chapter \ref{chap:Hipssm}) and Multi Time Scale State Space Model (MTS3 in Chapter \ref{chap:mts3}) besides making an existing formalism of SSMs more robust for performing interventions/counterfactuals with action/control signals (Ac-SSM in Chapter \ref{chap:Acssm}).

The proposed formalisms are in line with related theory in computational neuroscience called predictive processing and Bayesian brain hypothesis (Chapter \ref{chap:related}). In the proposed generative models, Bayesian inversion is employed to deduce the underlying causes from the incoming sensory data at various levels of abstraction. The Bayesian inversion results in update rules that use "precision weighting"~\parencite{friston2009free,hohwy2013predictive,seth2014cybernetic} as a mechanism to direct "attention" to relevant sensory information. The idea of "precision estimation" by learned sensory encoders is also in line with discussions in the broader neuroscience community. The world model formalisms proposed are adaptable to changing situations (embodying cognitive flexibility) and can make top-down long-horizon predictions based on nested causal hierarchies. 

Machines that can replicate human intelligence and type 2 reasoning capabilities should be able to reason at multiple
levels of temporal abstraction using internal world models~\parencite{lecun2022path,gupta2024essential}. The work represents a hopeful step towards developing embodied AI systems with a deeper and more intuitive grasp of the world, aligning machine learning processes more closely with natural intelligence. We believe that the findings of this dissertation will contribute to ongoing research in the creation of robust, principled, and scalable world models, while acknowledging the vast scope for further research and development in this area.

\paragraph{Limitations and Future Work} 
The thesis specifically concentrated on the aspect of prediction rather than control or planning. Addressing \textbf{hierarchical planning}, much like hierarchical prediction, remains a significantly underexplored challenge. A logical progression for future research would be to extend the concept of planning within the exact inference framework~\parencite{botvinick2012planning,watson2020stochastic}. This approach necessitates the derivation of backward messages and the use of the Kalman duality principle within the learned latent spaces of a deep encoder. Furthermore, investigating variational approaches to the hierarchical models presented, and considering hierarchical planning/control as a problem of approximate inference~\parencite{toussaint2009probabilistic,friston2009free,millidge2020relationship}, presents another intriguing avenue for research.

A further limitation noted in the thesis is its reliance on proprioceptive sensors for experimentation, without empirical validation using high-dimensional sensory information, such as vision. An important avenue for future research would involve evaluating the usefulness and characteristics of abstractions derived through Bayesian inversion/aggregation when applied to multimodal high-dimensional and \textbf{ sensory inputs}. Conducting experiments with image-based data and exploring "non-reconstruction"-based loss functions, as suggested by \cite{lecun2022path}, represent promising directions for future work.

Though our hierarchical formalism outperform transformer variants on long-horizon predictions on several tasks, a limitation for these to be used as foundational world models is the computational bottleneck due to the sequential nature of dynamics. However, there have been exciting progress recently in deterministic linear SSMs~\parencite{gu2021efficiently,smith2022simplified,mondal2023efficient} that allow efficient \textbf{parallelization} during training. Since our dynamics is linear similar to these approaches, parallelization using similar techniques as employed by deterministic SSMs can be another important direction of future research.\\
Our experimentation was limited to the hierarchical models with just two levels of abstraction, as this configuration proved to be adequate for numerous tasks we carried out. However, there is room for deeper exploration.The intersection of machine learning with broader cognitive theories, particularly the concept of \textbf{machine consciousness}, presents a novel yet underexplored frontier in the field. Traditionally, consciousness has been a topic more familiar to the realms of philosophy and psychology than to machine learning and computer science. This often makes it a delicate subject within our discipline.
\begin{figure}[H]
  \centering
  \includegraphics[width=0.75\linewidth]{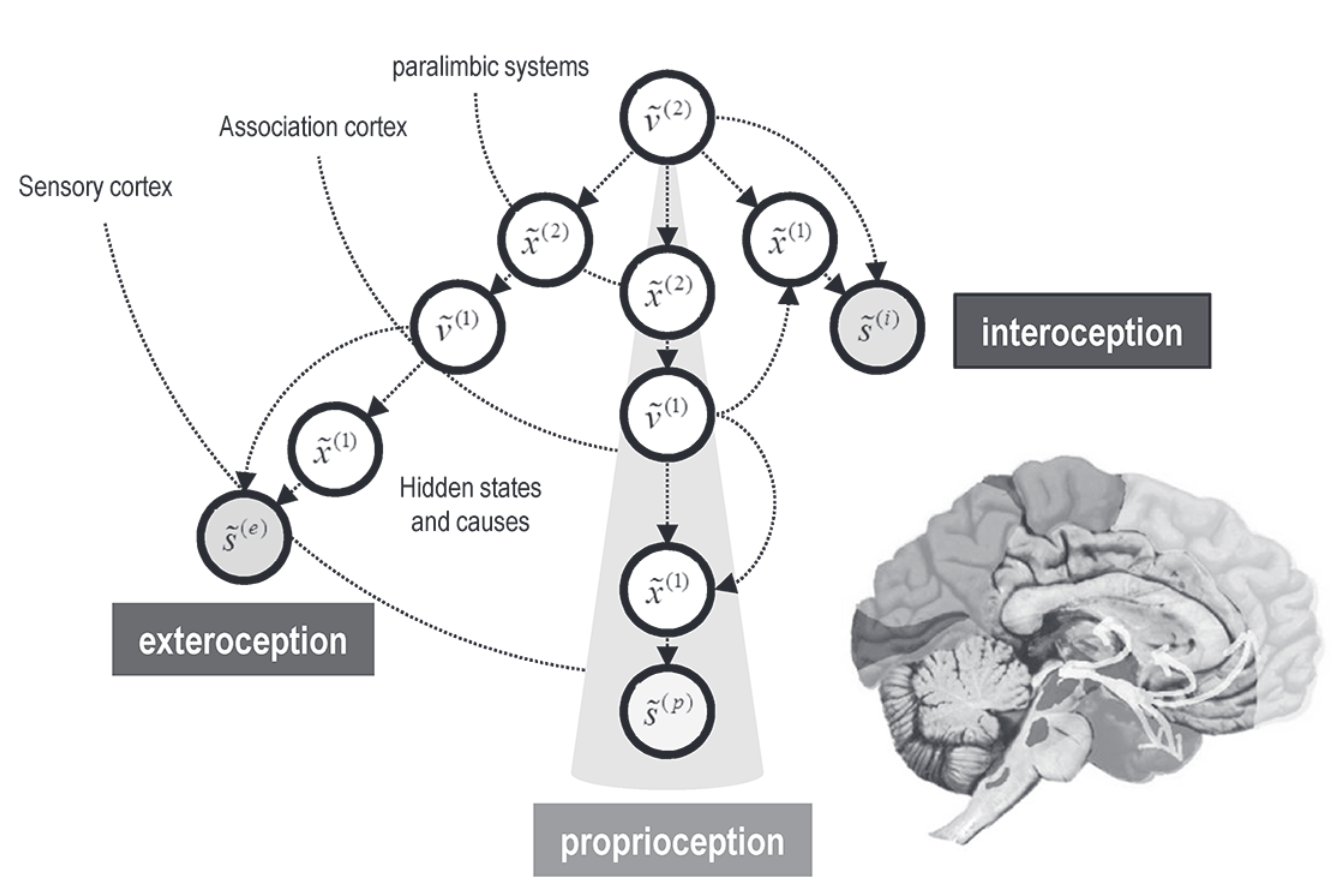} 
  \caption{Perception and consciousness as hierarchical inference from \cite{friston2013consciousness}.}
  \label{fig:consc}
\end{figure}
As I write this thesis in the Spring of 2024, the machine learning field finds itself at a crossroads, significantly influenced by the impressive capabilities of attention/transformer~\parencite{vaswani2017attention} based large language models (LLMs). These advancements suggests a pressing need for mathematical models or analytical frameworks to better understand machine consciousness \parencite{bengio2017consciousness,chalmers2023could}. Numerous theories on consciousness highlight the critical role of recurrent processing~\parencite{chalmers2023could,lamme2010neuroscience}, a feature notably absent in current LLM architectures. In this context, I believe that hierarchical generative models, particularly those extending beyond the 2-level temporal depth as explored in this thesis, represent a promising research direction for embedding consciousness priors~\parencite{friston2013consciousness,bengio2017consciousness} into machine learning systems. This approach could offer profound insights into the mechanisms underpinning consciousness and its potential replication in artificial systems.

\label{chap:conclusion}


\printbibliography[heading=bibintoc]

\appendix
\chapter{Appendix: Action Conditional SSM}
\section{Inverse Dynamics Learning with Action Conditional SSM}
\label{sec:inverseDyn}
\sloppy For the inverse dynamics case we want to learn a model \mbox{$f^{-1}: \vec o_{1:t}, \vec a_{1:{t-1}}, \vec o_{t+1} \mapsto \hat{\vec{a}}_{t}$} where $\vec o_{t+1}$ is the desired next observation for time step $t + 1$ and $\hat{\vec{a}}_{t}$ is the predicted action, i.e., the one to be applied. We introduce an action decoder which decodes the latent posterior $\left(\vec{z}_{t}^+, \vec{\Sigma}_{t}^+ \right)$ and estimates the action required to move to the desired next observation. The action decoder also gets information regarding the next observation as input.
During training, this corresponds to the next observation in the data. 
For control, a desired next observation is used to obtain the action required to reach that observation. The joint angles and velocities are treated as observations in our inverse dynamics experiments.

\begin{figure}[h]
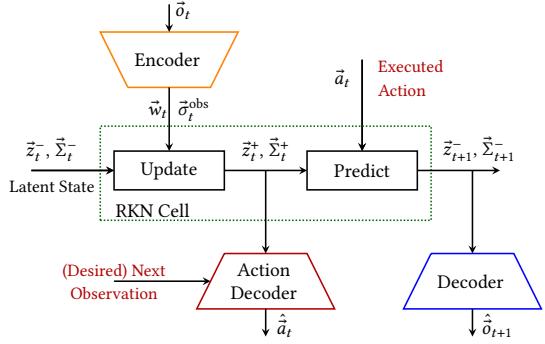

    \centering
    \resizebox{0.7\textwidth}{!}{\tikzACRKNINVDYN}
    \caption{Schematic diagram of the inverse dynamics learning architecture. Here the posterior $\left(\vec{z}_{t}^+, \vec{\Sigma}_{t}^+ \right)$, at the current time step is fed to an action decoder along with the desired target. In the next time-step, the executed action $\vec a_t$ is fed to the predict stage of the RKN cell. 
Further, the next predicted prior $\left(\vec{z}_{t+1}^-, \vec{\Sigma}_{t+1}^- \right)$ is decoded to get the next state $\hat{\vec{o}}_{t + 1}$. \label{fig:inverse}}
\end{figure}

\begin{figure*}[t]
\hspace*{-1cm}
\centering
  \begin{subfigure}{.34\textwidth}
    \includegraphics[width=\linewidth]{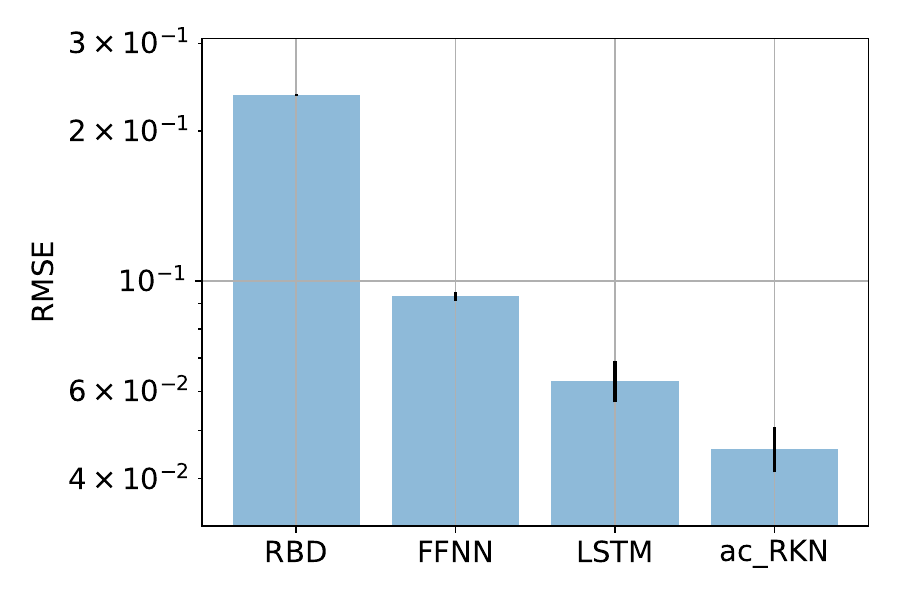}
    \caption{Franka Panda}
    \label{fig:frnaka_inv_torque}
  \end{subfigure} 
\begin{subfigure}{.34\textwidth}
    \includegraphics[width=\linewidth]{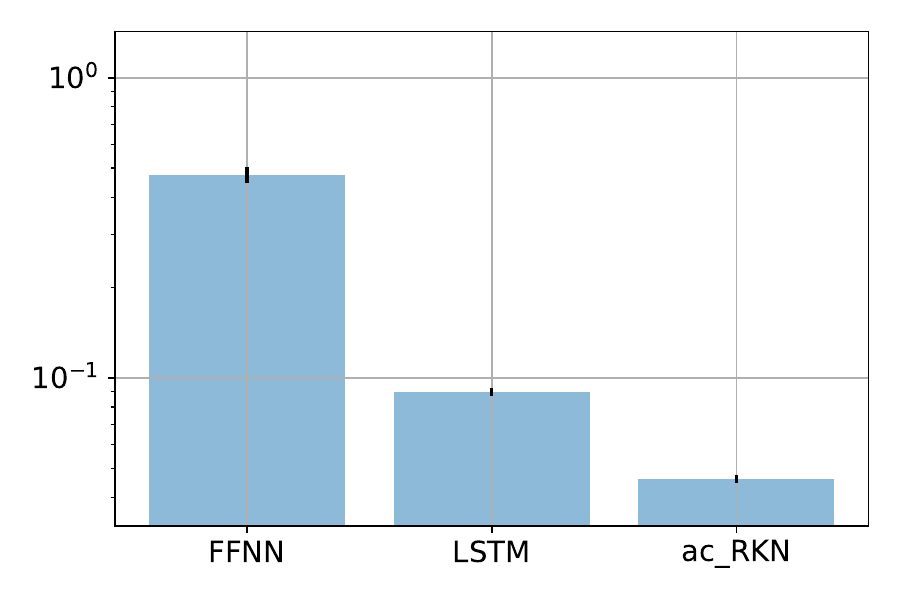}
    \caption{Barret WAM}
    \label{fig:barett_inv}
  \end{subfigure}
    \begin{subfigure}{.23\textwidth}
    \includegraphics[width=\linewidth]{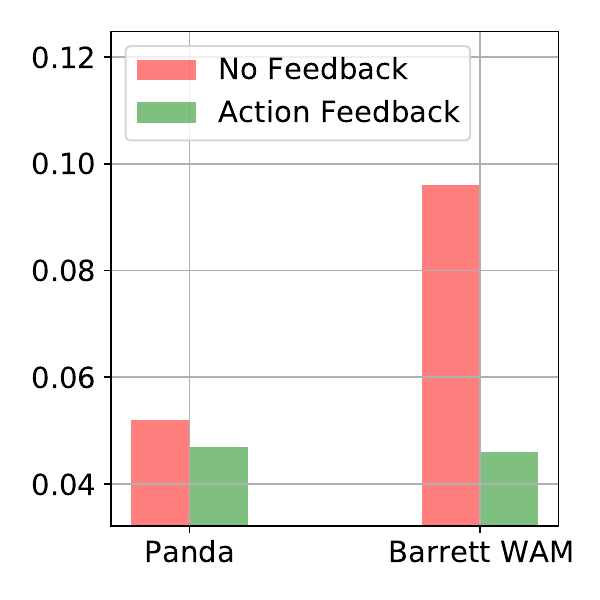}
    \caption{Action Feedback}
    \label{fig:franka_inter}
  \end{subfigure}
\caption{\textbf{(a)} and \textbf{(b)} Joint torque prediction RMSE values in NM of Action-Conditional RKN, LSTM and FFNN for Panda and Barret WAM. A comparison is also provided with the analytical (RBD) model of Panda. \textbf{(c)} Comparison of ac-RKN for learning inverse dynamics with and without action feedback as discussed in Section \ref{sec:inverseDyn}.
}
\vspace{-0.5cm}
\end{figure*}

As seen in Figure \ref{fig:inverse}, instead of merely learning the inverse dynamics, our approach learns inverse and forward dynamics simultaneously by feeding back the executed action to the action-conditional predict stage, which predicts forward in latent space. 
Note that the action decoder also gets the same action as the target during training.
However, the model sees the true action only after the prediction since the action decoder works with the posterior estimate $\left(\vec{z}_t^+, \vec{\Sigma}_t^+ \right)$.  
Hence, the prediction of the inverse dynamics model for the current time step is made independent of this feedback.
We found that enforcing this causal feedback, a necessary structural component of the Bayesian network of the underlying latent dynamical system, improves the performance of the inverse model, as seen in Figure \ref{fig:franka_inter}.


\paragraph{Loss And Training.}
The dual output architecture for our inverse model leads to two different loss functions for the action and observation decoders that are optimized jointly.
Our experiments showed that the loss of the forward model is an excellent auxiliary loss function for the inverse model, and learning this implicit forward model jointly with the inverse model results in much better performance for the inverse model. 
We assume that the reason for this effect is that the forward model loss provides crucial gradient information to form an informative latent state representation that is also useful for the inverse model.
To deal with high-frequency data, we again chose to predict the normalized differences to the previously executed action, that is, $\hat{\vec{a}}_t = \vec{a}_{t-1} + \textrm{dec}_\textrm{action}(\vec{z}_t^+)$. 
Here $\vec{a}_{t-1}$ is the executed action at $t-1$ and $\textrm{dec}_\textrm{action}(\vec{z}_t^+)$ the output of the actual action decoder network. 
The combined loss function for the inverse dynamic case is given by 

 \begin{align*}
     \mathcal{L}_\textrm{inv} = \sqrt{\dfrac{1}{T} \sum_{i=1}^T \parallel \left(\vec{a}_{t+1} - \vec{a}_{t}\right) - \textrm{dec}_\textrm{action}\left(\vec{z}_t^+\right) \parallel^2}    + 
     \lambda \mathcal{L}_\textrm{fwd} 
 \end{align*}
where 
$\lambda$ chooses the trade-off between the inverse model loss and the forward model loss. The value of $\lambda$ is chosen via hyperparameter optimization using GPyOpt~\parencite{gpyopt2016}.
Note that for the inverse model,  the actions are decoded based on the posterior mean of the current time step, while for the forward model, the observations are decoded from the prior mean of the next time step. 

We evaluated the performance of the proposed method for inverse dynamics learning on two real robots, Franka Emika Panda and Barrett WAM. Barret WAM is a robot with direct cable drives. Direct cable drives produce high torques, generating fast and dexterous movements but produce complex dynamics. The rigid body dynamics cannot accurately model these complex dynamics due to the variable stiffness and lengths of the cables~\parencite{nguyen2011incremental}. 

\begin{figure*}[t]
\centering
\hspace*{0.2cm}
\begin{minipage}{.33\textwidth}
\includegraphics[width=\textwidth]{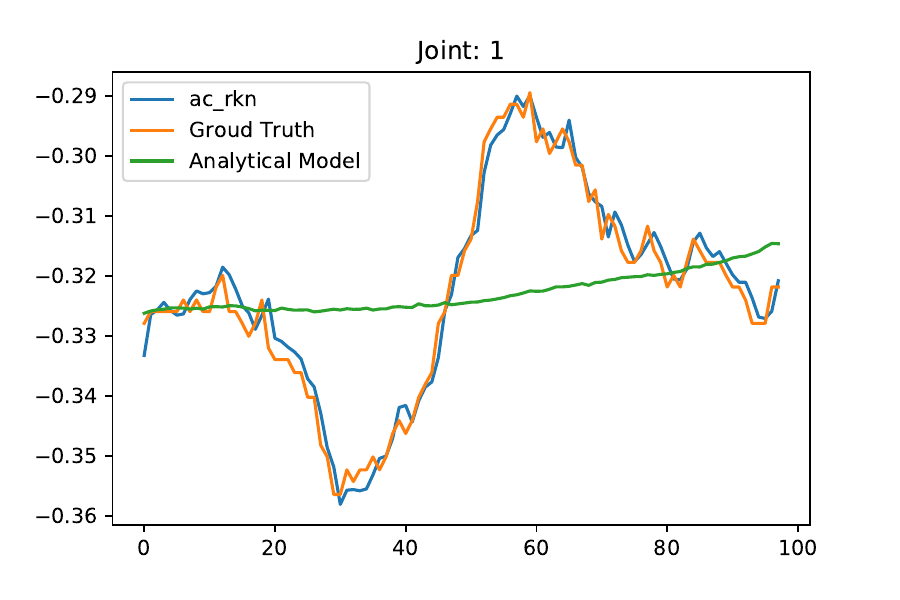}
\end{minipage}%
\begin{minipage}{.33\textwidth}
\includegraphics[width=\textwidth]{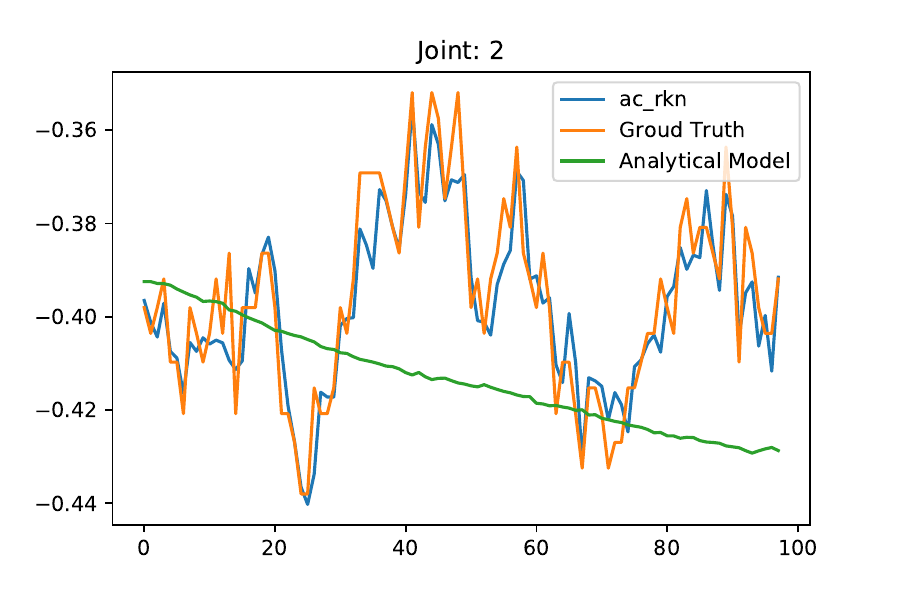}
\end{minipage}%
\begin{minipage}{.33\textwidth}
\includegraphics[width=\textwidth]{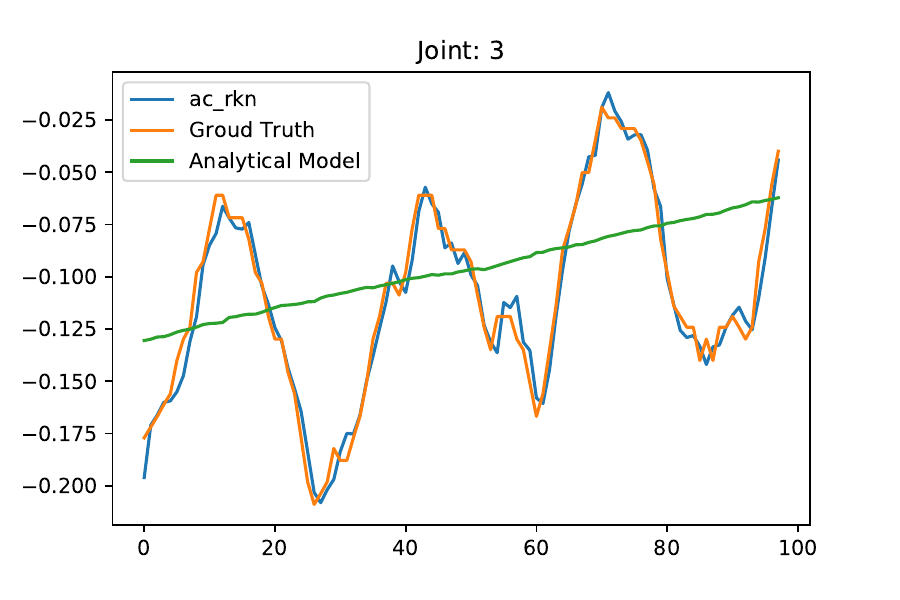}
\end{minipage}

\caption{ Predicted joint torques(normalized) for first 3 joints of the Panda robot arm. The learned inverse dynamics model by ac-RKN matches closely with the ground truth data, while the rigid body dynamics model cannot capture the high-frequency variations in the data. 
}
\label{fig:inv_analytical}
\vspace{-0.5cm}
\end{figure*}

\textbf{Joint Torque Prediction Task}. We benchmark the representational capability of the latent state posterior, $\vec z_t^+$, of ac-RKN in accurately modelling the inverse dynamics of these robots in comparison to deterministic models like LSTMs and FFNN. It is clear from \ref{fig:frnaka_inv_torque} and \ref{fig:barett_inv} that ac-RKN learns highly accurate models of these in contrast with other data-driven methods. This highly precise modelling is often a requirement for high fidelity and compliant robotic control.   


 
\textbf{Impact of Action Feedback}. We also perform an ablation study with and without the action feedback for the prediction step in the latent dynamics. As seen in Figure \ref{fig:franka_inter}, the action feedback always results in better representational capability as this helps the implicit forward model in making better predictions by taking into account its causal effect on the state transitions.

\textbf{Comparison to Analytical Models}. Finally, we make a comparison with the analytical model of the Panda robot for inverse dynamics. Please refer to Appendix C for more details on the analytical model. As evident from Figure \ref{fig:inv_analytical} analytical models gave predictions with much lesser accuracy in comparison to ac-RKN, as it does not consider unmodelled effects such as joint and link flexibilities, backlash, stiction and actuator dynamics.  In such cases, the robot would continuously have to track its current position and compensate the errors with high-gain feedback control thus making it dangerous to interact with the real world and impossible to work in human-centred environments.

\section{Implementation Details}
\label{sec:locally-linear}
\subsection*{Locally Linear Transition Model} The state transitions in the predict stage of the Kalman filter is governed by a locally linear transition model. To obtain a locally linear transition model, the RKN learns $K$ constant transition matrices $\vec{A}^{(k)}$ and combines them using state-dependent coefficients $\alpha^{(k)}(\vec{z_t})$, i.e.,
$ \vec{A}_t = \sum_{k=0}^K \alpha^{(k)}(\vec{z_t}) \vec{A}^{(k)}. $
A small neural network with softmax output is used to learn $\alpha^{(k)}$. Each $\vec{A}^{(k)}$ is designed to consist of four band matrices as in \cite{becker2019recurrent} to reduce the number of parameters without affecting the performance.\\

\section{Details Of Rigid Body Dynamics Model}
The analytical model for Franka Emika Panda is a rigid-body dynamics model that was identified in its so-called base parameters~\parencite{Khalil1991}. Due to the friction compensation in the joints, we observed that the viscous friction is negligible, whereas the observed Coulomb friction is very small yet included in our parameterization. This results in a model with 50 parameters. 
The base parameterization is computed based on the provided kinematic properties of the robotic arm and provides a linear relation between the base parameters and joint torques for a given set of joint positions, velocities and accelerations. 

Due to this linearity, the regression problem can be solved using a linear least-squares method, although additional linear matrix inequality constraints must be fulfilled to ensure that the resulting parameters are physically realizable \parencite{Sousa2019,reuss2022end}.
In order to perform a forward simulation of the robot dynamics, we numerically solve an initial value problem for the implicit set of differential equations defined by the base parameterization of the rigid-body model.
Note that, the model does not parameterize actuator dynamics, nor does it model joint flexibilities, link flexibilities, or stiction. The focus here is to provide a reference baseline to show which effects the acRKN captures in comparison to a text-book robot model.

\chapter{Appendix: Hidden Parameter SSM}
\section{Proof For Gaussian Identity \ref{theo:taskpredict}}

This section provides a proof for the Gaussian identity \ref{theo:taskpredict}. First we derive an expression for the joint distribution $p(\cvec{u},\cvec{v},\cvec{y})$. 

\begin{restatable}[Joint Gaussian Distribution]{gi}{jointgauss2}
\label{identity:joint2}If $\cvec{u} \sim \mathcal{N}(\cvec{\mu}_u + b,\cvec{\Sigma}_u)$ and $v \sim \mathcal{N}(\cvec{\mu}_v,\cvec{\Sigma}_v)$ are normally distributed independent random variables and if conditional distribution $p(\cvec{y}|\cvec{u},\cvec{v}) = \mathcal{N}(\cvec{A}\cvec{u} + b + \cvec{B}v, \cvec{\Sigma})$, the joint distribution has an expression as follows:
 
$\left(\begin{array}{c}\mathbf{u} \\ \mathbf{v} \\ \mathbf{y}\end{array}\right)  \sim \mathcal{N}\left(\left(\begin{array}{c}\cvec{\mu}_{u} \\ \cvec{\mu}_{v} \\ \mathbf{A} \mu_{u}+ b + \mathbf{B} \mu_{v} \end{array}\right),\left(\begin{array}{cccc}\boldsymbol{\Sigma}_{u} & 0& \boldsymbol{\Sigma}_{u} \mathbf{A}^{\top} \\ 0 &\boldsymbol{\Sigma}_{v}  &\boldsymbol{\Sigma}_{v} \mathbf{B}^{\top} \\ \mathbf{A} \boldsymbol{\Sigma}_{u}^{\top} & \mathbf{B} \boldsymbol{\Sigma}_{v}^{\top} & \mathbf{A} \boldsymbol{\Sigma}_{u} \mathbf{A}^{\top}+\mathbf{B} \boldsymbol{\Sigma}_{v} \mathbf{B}^{\top} + \boldsymbol{\Sigma} \end{array}\right)\right)$
\end{restatable}

\textbf{Proof for Identity \ref{identity:joint2}} 

Let the displacement of a variable $\mathbf{u}$ be denoted by $\Delta \mathbf{u}=\mathbf{u}-\langle\mathbf{u}\rangle $.

Since $\mathbf{u}$ and $\mathbf{v}$ are independent, covariances $\left\langle\Delta \mathbf{u} \Delta \mathbf{v}^{\top}\right\rangle=0$.

We can write $\mathbf{y}=\mathbf{A}\cvec{u} + b + \mathbf{B}\cvec{v} + \cvec{\epsilon}$, where $\cvec{\epsilon} \sim \mathcal{N}(\cvec{0},\cvec{\Sigma})$ and $b$ is a constant. Then we have covariance $\left\langle\Delta \mathbf{u} \Delta \mathbf{y}^{\top}\right\rangle=\left\langle\Delta \mathbf{u}(\mathbf{A} \Delta \mathbf{u}+\mathbf{B} \Delta \mathbf{v} + \Delta \epsilon)^{\top}\right\rangle=$ $\left\langle\Delta \mathbf{u} \Delta \mathbf{u}^{\top}\right\rangle \mathbf{A}^{\top}+\left\langle\Delta \mathbf{u} \Delta \mathbf{v}^{\top}\right\rangle \mathbf{B}^{\top} +  \left\langle\Delta \mathbf{u} \Delta \mathbf{\epsilon}^{\top}\right\rangle$. Since $\left\langle\Delta \mathbf{u} \Delta \mathbf{v}^{\top}\right\rangle=\left\langle\Delta \mathbf{u} \Delta \mathbf{\epsilon}^{\top}\right\rangle=0$ we therefore have $\left\langle\Delta \mathbf{u} \Delta \mathbf{y}^{\top}\right\rangle=\boldsymbol{\Sigma}_{u} \mathbf{A}^{\top}.$ 

The derivations for covariances $\left\langle\Delta \mathbf{v} \Delta \mathbf{y}^{\top}\right\rangle$ follow in a similar way, and the corresponding covariance has the expression $\left\langle\Delta \mathbf{v} \Delta \mathbf{y}^{\top}\right\rangle=\boldsymbol{\Sigma}_{v} \mathbf{B}^{\top}$.

\sloppy Similarly, $\left\langle\Delta \mathbf{y} \Delta \mathbf{y}^{\top}\right\rangle=$ $\left\langle(\mathbf{A} \Delta \mathbf{u}+\mathbf{B} \Delta \mathbf{v} + \Delta \mathbf{\epsilon})(\mathbf{A} \Delta \mathbf{u}+\mathbf{B} \Delta \mathbf{v} + \Delta \mathbf{\epsilon})^{\top}\right\rangle=\mathbf{A}\left\langle\Delta \mathbf{u} \Delta \mathbf{u}^{\top}\right\rangle \mathbf{A}^{\top}+\mathbf{B}\left\langle\Delta \mathbf{v} \Delta \mathbf{v}^{\top}\right\rangle \mathbf{B}^{\top} + \left\langle\Delta \mathbf{\epsilon} \Delta \mathbf{\epsilon}^{\top}\right\rangle = \mathbf{A} \boldsymbol{\Sigma}_{u} \mathbf{A}^{\top}+\mathbf{B} \boldsymbol{\Sigma}_{v} \mathbf{B}^{\top} +\Sigma $. The result follows.

\begin{restatable}[Gaussian Marginalization]{gi}{generalgaussmarg}
\label{identity:generalgaussmarg2}
If
$$
\left(\begin{array}{l}
\mathbf{\cvec{u}} \\
\mathbf{\cvec{v}} \\
\mathbf{\cvec{y}}
\end{array}\right) \sim \mathcal{N}\left(\left(\begin{array}{l}
\cvec{\mu}_{u} \\
\cvec{\mu}_{v} \\
\cvec{\mu}_{z}
\end{array}\right),\left(\begin{array}{cccc}
\boldsymbol{\Sigma}_{uu} & \boldsymbol{\Sigma}_{uv} & \boldsymbol{\Sigma}_{uy}\\
\mathbf{\Sigma}_{uv}^{\top} & \boldsymbol{\Sigma}_{vv} & \mathbf{\Sigma}_{v y}\\
 \boldsymbol{\Sigma}_{uy}^{\top}  & \boldsymbol{\Sigma}_{vy}^{\top}   & \boldsymbol{\Sigma}_{yy}
\end{array}\right)\right)
$$
then marginal over y is given as $p(\cvec{y}) = \int_{\cvec{u},\cvec{v}} p(\cvec{y}|\cvec{u},\cvec{v})p(\cvec{u})p(\cvec{v})d\cvec{u}d\cvec{v} = \mathcal{N}\left(\cvec{\mu}_{y}, \mathbf{\Sigma}_{yy}\right)$
\end{restatable}

\textbf{Proof For Identity \ref{identity:generalgaussmarg2}} 

We refer to \cite{bishop2006pattern} for the derivation, which requires calculation of the Schur complement as well as completing the square of the Gaussian p.d.f. to integrate out the variable. The given derivation~\parencite{bishop2006pattern} for two variable multivariate Gaussians can be extended to 3 variable case WLOG.

\textbf{Proof For Identity \ref{theo:taskpredict}} is immediate from Identity \ref{identity:joint2} and Identity \ref{identity:generalgaussmarg2}.

\section{Additional Experiments}
\subsection{Comparison To Soft Switching Baseline}
\label{app:softswitch}
We also implement a soft-switching baseline similar to \cite{fraccaro2017disentangled}, where a soft mixture of dynamics is implemented in the latent transition dynamics (Kalman time update).  Similar to \cite{fraccaro2017disentangled}, we now globally learn $K$ constant transition matrices $\boldsymbol{A}^{(k)}$ and control matrices $\boldsymbol{B}^{(k)}$. An interpolation is done between these using a ``dynamics parameter network"~\parencite{fraccaro2017disentangled} $\alpha_{t}=$ $\boldsymbol{\alpha}_{t}\left(\mathbf{w}_{0: t-1}\right)$. The dynamics parameter network is implemented with a recurrent neural network with LSTM cells that takes at each time step the mean of the encoded observation $w_t$ as input and recurses $\mathbf{d}_{t}=$ $\operatorname{LSTM}\left(\mathbf{w}_{t-1}, \mathbf{d}_{t-1}\right)$ and $\boldsymbol{\alpha}_{t}=\operatorname{softmax}\left(\mathbf{d}_{t}\right)$. The output of the dynamics parameter network is weights that sum to one, $\sum_{k=1}^{K} \alpha_{t}^{(k)}\left(\mathbf{w}_{0: t-1}\right)=1$.
These weights choose and interpolate between $K$ different operating modes:

$$\mathbf{A}_{t}=\sum_{k=1}^{K} \alpha_{t}^{(k)}\left(\mathbf{w}_{0: t-1}\right) \mathbf{A}^{(k)}, \quad \mathbf{B}_{t}=\sum_{k=1}^{K} \alpha_{t}^{(k)}\left(\mathbf{w}_{0: t-1}\right) \mathbf{B}^{(k)}$$

The authors interpret the weighted sum as a soft mixture of $K$ different Linear Gaussian SSMs whose time-invariant matrices are combined using the time-varying weights $\boldsymbol{\alpha}_{t}$. In practice, each of the $K$ sets $\left\{\mathbf{A}^{(k)}, \mathbf{B}^{(k)}\right\}$ models different/changing dynamics, which will dominate when the corresponding $\alpha_{t}^{(k)}$ is high. 

\begin{figure*}
\begin{minipage}{.5\linewidth}
\centering
\scalebox{0.9}{
\includegraphics[width=\linewidth]{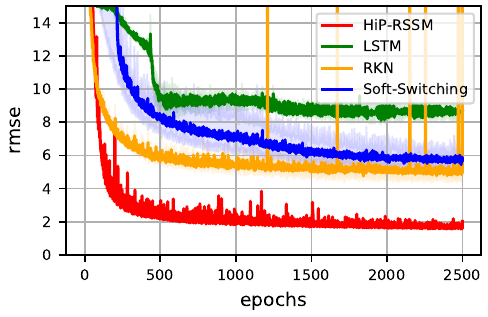}
\label{fig:wheeledtrain}
}%
\subcaption{}
\label{fig:pamtab}
\end{minipage}%
\begin{minipage}{.5\linewidth}
\centering
\scalebox{0.9}{
\includegraphics[width=\linewidth]{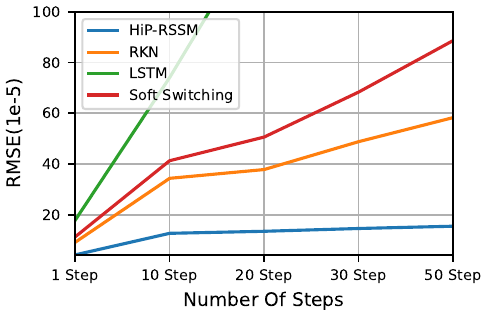}
\label{fig:wheeledsoft}
}%
\subcaption{}
\label{fig:multistep}
\end{minipage}%
\caption{Comparison of different algorithms for Wheeled Mobile Robot in terms of (a) moving average of decoder error in normalized RMSE for the test set, plotted against training epochs  (b) multi-step ahead prediction error in RMSE.} 
\label{fig:mobileab}
\end{figure*}

Figure \ref{fig:mobileab} compares HiP-RSSM with different recurrent architectures, including the soft-switching baseline. As seen in Figure \ref{fig:mobileab}, HiP-RSSM clearly outperforms the soft-switching baseline both in terms of convergence speed and also mult-step ahead predictions. More details on the multistep training procedure can be found in the Appendix \ref{app:multi}.

\subsection{Multi-Step Ahead Predictions}
 \label{app:multi}
In figure \ref{fig:multistep}, we compare the results of multi-step ahead predictions for upto 50 steps of HiP-RSSM with recurrent baselines like RKN\parencite{becker2019recurrent} and LSTM. Inorder to train the recurrent models for multi step ahead predictions, we removed three-quarters of the observations from the temporal sequence and tasked the models with imputing those missing observations, only based on the knowledge of available actions/control commands, i.e., we train the models to perform action conditional future predictions to impute missing observations. The imputation employs the model for multi-step ahead predictions in a convenient way~\parencite{shaj2020action}. One could instead also go for a dedicated multi-step loss function as in approaches like \cite{finn2016unsupervised}.

As seen in figure \ref{fig:multistep}, HiP-RSSM clearly outperforms contemporary recurrent models for multi-step ahead prediction tasks since it takes into account additional causal factors of variation (slopes of the terrain for this robot) in the latent dynamics in an unsupervised manner.

\section{Implementation Details}

\subsection{Context Set Encoder and Latent Task Representation} The HiP-RSSM maps a set of previous interaction histories, $ \{\cvec{o}_{t,n}^l,\cvec{a}_{t,n}^l,\cvec{o}_{t+1,n}^l\}_{n=1}^N$, to a set of latent features and an estimate of uncertainty in these features, $
\{\cvec{r}_{n}^l,(\cvec{\sigma}_n^l)^2\}_{n=1}^N$, using a context encoder. We use a feed-forward neural network as the encoder in all of our experiments, since we deal with high-dimensional vectors as our observations. However, depending upon the nature of the observations, we could use different encoder architectures.\\
The set of latent features and uncertainties, $
\{\cvec{r}_{n}^l,\left(\cvec{\sigma}_n^{\cvec{l}}\right)^2\}_{n=1}^N$, is further aggregated in a probabilistically principled manner using the Bayesian aggregation operator discussed in \ref{subsec:latentTask} to obtain a Gaussian latent task variable, with a mean ($\cvec{\mu}_l$) and diagonal covariance ($\cvec{\sigma}_l$). Intuitively, the context encoder learns to weight the contribution from each observation in the context set based on Bayesian principles and emits a probabilistic representation of the latent task.

 \subsection{Latent Task Transformation Model}
 \label{app:taskmodel}
 To achieve latent task conditioning within the recurrent cell, we include  a task transformation model $(\boldsymbol{c})$, in addition to the locally linear transition model $\boldsymbol{A}_t$ and control model $\boldsymbol{b}$ in the time update stage (section \ref{subsec:timeUpdate}). 
Though in section \ref{subsec:timeUpdate}, we used the notation for a linear task transformation matrix, $\boldsymbol{C}$, to motivate the additive interaction of latent task variables, $\cvec{\mu}_l$ and $\cvec{\sigma}_l$ in the latent space, the task transformation function can be designed in several ways, i.e.:
\begin{itemize}
\item[\textbf{(i)}] \textbf{Linear:} $\boldsymbol{c} =\boldsymbol{C}$, where $\boldsymbol{C} $ is a linear transformation matrix. The corresponding time update equations are given below:
\begin{align*}
\cvec{z}_t^- &=\cmat{A}_{t-1}\cvec{z}_{t-1}^+ + \cmat{b}(\cvec{a}_t) + \cmat{C}\cvec{\mu_l}, \\
\cmat{\Sigma}_t^- &= \cmat{A}_{t-1}\cmat{\Sigma}_{t-1}^+\cmat{A}_{t-1}^T + \cmat{C}(\cmat{I}\cdot\cmat{\sigma}_{\cvec{l}})\cmat{C}^T + \cmat{\Sigma}_{\textrm{trans}}.
\end{align*}

\item[\textbf{(ii)}] \textbf{Locally-Linear:} $\boldsymbol{c} = \boldsymbol{C}_t $, where $\boldsymbol{C}_t = \sum_{k=0}^K \beta^{(k)}(\boldsymbol{z_t}) \boldsymbol{C}^{(k)}$ is a linear combination of k linear control models $\boldsymbol{C}^{(k)}$. A  small  neural  network  with softmax  output  is  used  to  learn $\beta^{(k)}$. The corresponding time update equations are given below: 
\begin{align*}
\cvec{z}_t^- &=\cmat{A}_{t-1}\cvec{z}_{t-1}^+ + \cmat{b}(\cvec{a}_t) + \cmat{C_t}\cvec{\mu_l}, \\
\cmat{\Sigma}_t^- &= \cmat{A}_{t-1}\cmat{\Sigma}_{t-1}^+\cmat{A}_{t-1}^T + \cmat{C_t}(\cmat{I}\cdot\cmat{\sigma}_{\cvec{l}})\cmat{C_t}^T + \cmat{\Sigma}_{\textrm{trans}}.
\end{align*}
\item[\textbf{(iii)}] \textbf{Non-Linear:} $\boldsymbol{c} = \boldsymbol{f}$, where $\boldsymbol{f}(.)$ can be any non-linear function approximator. We use a multi-layer neural network regressor with ReLU activations, which transforms the latent task moments $\cvec{\mu_l}$ and $\cvec{\sigma_l}$ directly into the latent space of the state space model via additive interactions. The corresponding time update equations are given below:
\begin{align*}
\cvec{z}_t^- &=\cmat{A}_{t-1}\cvec{z}_{t-1}^+ + \cmat{b}(\cvec{a}_t) + \cmat{f}(\cvec{\mu}_l), \\
\cmat{\Sigma}_t^- &= \cmat{A}_{t-1}\cmat{\Sigma}_{t-1}^+\cmat{A}_{t-1}^T + \cmat{f}(\cvec{\sigma}_{\cvec{l}})+ \cmat{\Sigma}_{\textrm{trans}}.
\end{align*}
\end{itemize} 

In our ablation study~(Figure \ref{fig:ablTask}), for the linear and locally linear task transformation models, we assume that the dimension of the latent context variable $\cvec{l}$ and the latent state space $\cvec{z}_t$ are equal. This allows us to work with square matrices, which are more convenient. For the non-linear transformation we are free to choose the size of the latent context variable.  However for a fairer comparison we keep the dimension of latent task variable to be similar in all three cases. We choose the non-linear task transformation model in HiP-RSSM architecture as this gave the best performance in practice.

\chapter{Appendix: Multi Time Scale SSM}
\label{app:mts3}
\section{Proofs and Derivations}
\label{sec:proofMT}
\begin{wrapfigure}[7]{r}{.29\linewidth}
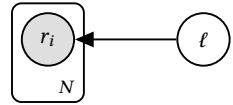

\begin{subfigure}[b]{0.28\textwidth}
   \centering \begin{adjustbox}{width=\textwidth}
            \tikzTaskinfer
     \end{adjustbox}
\end{subfigure}
\caption{Graphical Model For Bayesian conditioning with $N$ observations.}
\label{fig:taskinf}
\end{wrapfigure}

In the following sections, vectors are denoted by a lowercase letter in bold, such as "$\cvec{\mathrm{v}}$", while matrices are denoted by an uppercase letter in bold, such as "$\cmat{M}$". $\cmat{I}$ denotes identity matrix and $\cmat{0}$ represents a matrix filled with zeros. For any matrix $\cmat{M}$, $\cvec{m}$ denotes the corresponding vector of diagonal entries. Also, $\odot$ denotes the elementwise vector product and  $\oslash$ denotes an elementwise vector division.

\subsection{Proof For Bayesian Conditioning As Permutation Invariant Set Operations (Identity \ref{theo:gc})}
\label{subsec: condProof}

\gausscond*

Note that $\cvec \Sigma_l$ is the covariance matrix which is the inverse of the precision matrix $\cvec \Lambda_l$. Due to the observation model assumption $\cmat{H}=[\cmat{I},\cmat{0}]$, they take block diagonal form, $$\cvec \Sigma_l = \left[\begin{array}{cc}\cvec \Sigma_l^u & \cvec \Sigma^{s}_l \\ \cvec \Sigma^{s}_l & \cvec \Sigma^{l}_l  \end{array} \right], \textrm{ with } \cvec \Sigma_u = \textrm{diag}(\cvec \sigma_l^u), \; \cvec \Sigma_l = \textrm{diag}(\cvec \sigma_l^l) \textrm{ and } \cvec \Sigma_{s} = \textrm{diag}(\cvec \sigma_{l}^s).$$  

\paragraph{Proof:}

\paragraph{Case 1 (Single Observation):} Before deriving the update rule for $N$ conditionally iid observations, let us start with a simpler case consisting of a single observation $\cvec r$. If the marginal Gaussian distribution for the latent variable $\cvec l$ takes the form $ p(\mathbf{l}) =\mathcal{N}\left(\mathbf{l} \mid \boldsymbol{\mu}, \boldsymbol{\Lambda}^{-1}\right)$ and the conditional Gaussian distribution for the single observation $\cvec{r}$ given $\cvec{l}$ has the form
, $p(\mathbf{r} \mid \mathbf{l}) =\mathcal{N}\left(\mathbf{r} \mid \mathbf{H} \mathbf{l}+\mathbf{b}, \mathbf{L}^{-1}\right)$. Then the posterior distribution over $\mathbf{l}$ can be obtained in closed form as, 
\begin{equation}
\label{eq:bishop1}
\begin{aligned}
p(\mathbf{l} \mid \mathbf{r}) & =\mathcal{N}\left(\mathbf{l} \mid \mathbf{\Sigma}\left\{\mathbf{H}^{\mathrm{T}} \mathbf{L}(\mathbf{r}-\mathbf{b})+\boldsymbol{\Lambda} \boldsymbol{\mu}\right\}, \boldsymbol{\Lambda}^{-1}\right) ,\textrm{where } \boldsymbol{\Lambda}=\left(\boldsymbol{\Lambda}+\mathbf{H}^{\mathrm{T}} \mathbf{L} \mathbf{H}\right).
\end{aligned}
\end{equation}
We refer to Section 2.3.3 of \cite{bishop2006pattern}, to the proof for this standard result.

\paragraph{Case 2 (Set Of Observations):} Now instead of a single observation, we wish to derive a closed form solution for the posterior over latent variable $\boldsymbol{l}\in \mathbb{R}^{2d}$, given a set of N conditionally i.i.d observations $\Bar{r} = \{r_i\}_{i=1}^N$. Here each element $\boldsymbol{r_i} \in \mathbb{R}^d$ of the set $\Bar{r}$ is assumed to to have an observation model $\cmat{H}=[\cmat{I},\cmat{0}]$. In the derivation, we represent the set of N observations as a random vector $$\Bar{r} = \left[ \begin{array}{ll}
\cvec{r_1}  \\
\cvec{r_2} \\
\cvec{.} \\
\cvec{.} \\
\cvec{r_N}
\end{array}\right]_{Nd \times 1}.$$ 
Since each observation in the set $\Bar{r}$ are conditionally independent, we denote the conditional distribution over the context set as $\Bar{r}  \mid \mathbf{l} \sim \mathcal{N}\left(\bar{H} \mathrm{l},  \cmat{\Sigma}_{r}\right)$, where the diagonal covariance matrix has the following form: $$\cmat{\Sigma}_{r}= \left[ \begin{array}{llllll}
 \textrm{diag}(\cvec \sigma_{r_1}),&0,&0,&..,&0  \\
0,&\textrm{diag}(\cvec \sigma_{r_2}),&0,&..,&0 \\
\boldsymbol{.},&.,&.,&..,&. \\
\boldsymbol{. },&.,&.,&..,&.\\
0,&0,&0,&..,&\textrm{diag}(\cvec \sigma_{r_N})
\end{array}\right]_{Nd \times Nd}.
$$
The corresponding observation model  $\bar{\cmat H}$ is
$$
\bar{\cmat H} =  \left[ \begin{array}{ll}
\cmat{H}  \\
\cmat{H} \\
\cmat{.} \\
\cmat{.} \\
\cmat{H}
\end{array}\right]_{Nd \times 2d} = \left[ \begin{array}{ll}
\boldsymbol{I }, \boldsymbol{0 }  \\
\boldsymbol{I }, \boldsymbol{0} \\
\boldsymbol{.}, \boldsymbol{.} \\
\boldsymbol{. }, \boldsymbol{.} \\
\boldsymbol{I }, \boldsymbol{0}
\end{array}\right]_{Nd \times 2d}.$$

Now given the prior over the latent task variable $\mathrm{l} \sim \mathcal{N}\left(\mu_l^-, \boldsymbol{\Sigma}_l^-\right)$, the parameters of the posterior distribution over the task variable, $p(l|\Bar{r}) \sim \mathcal{N}\left(\mu_l^+, \boldsymbol{\Lambda}_l^+\right)$, can be obtained in closed-form substituting in Equation \eqref{eq:bishop1} as follows.
\begin{equation}
\begin{aligned}
\Lambda_l^+ & =  (\Sigma_l^+)^{-1}  \\ &= \boldsymbol{\Sigma}_l^{-1}+ \bar{\cmat {H}} ^T \cmat{\Sigma}_{r} \bar{\cmat {H}}  \\
& = \boldsymbol{\Sigma}_l^{-1}+ \left[ \begin{array}{ll}
 \textrm{diag}(\cvec \sigma_{r_1}),  \textrm{diag}(\cvec \sigma_{r_2}),  \textrm{diag}(\cvec \sigma_{r_3}), . , . ,   \textrm{diag}(\cvec \sigma_{r_N})  \\
\boldsymbol{0 }, \quad \quad \boldsymbol{0}, \quad \quad \boldsymbol{0}, \quad. , . ,\quad  \boldsymbol{0 }  \\
\end{array}\right]_{2d\times nd} \bar{ \cmat{H}} \\
& =\boldsymbol{\lambda}_l^{-} + \left[ \begin{array}{ll}
\mathrm{diag}(\sum_{i=1}^n\frac{1}{\cvec \sigma_{r_i}}), &\boldsymbol{0 }  \\
\boldsymbol{0 },& \boldsymbol{0} \\
\end{array}\right]_{2d\times 2d} 
\end{aligned}
\end{equation}
\begin{equation}
\begin{aligned}
\mu_l^+ & = \boldsymbol{\mu}_l^{-} + (\Lambda^+)^{-1}\bar{\cmat {H}} ^T \left( \sigma_{\boldsymbol{r}}^{-2} \boldsymbol{I}\right)\left(\boldsymbol{y}-\bar{\cmat {H}}  \boldsymbol{\mu}_{\boldsymbol{x}}\right) \\
& = \boldsymbol{\mu}_l^{-}+ \Sigma^+ \bar{\cmat {H}} \left( \sigma_{\boldsymbol{r}}^{-2} \boldsymbol{I}\right) \left(\boldsymbol{y}- \bar{\cmat {H}}  \boldsymbol{\mu}_{\boldsymbol{x}}\right) \\
& = \boldsymbol{\mu}_l^{-}+ \Sigma^+\left[ \begin{array}{ll}
\sigma_{\boldsymbol{r_1}}^{-2}\boldsymbol{I }, \sigma_{\boldsymbol{r_2}}^{-2}\boldsymbol{I}, \sigma_{\boldsymbol{r_3}}^{-2}\boldsymbol{I }, . , . ,  \sigma_{\boldsymbol{r_n}}^{-2}\boldsymbol{I }  \\
\boldsymbol{0 }, \quad \quad \boldsymbol{0}, \quad \quad \boldsymbol{0}, \quad. , . ,\quad  \boldsymbol{0 }  \\
\end{array}\right] \left(\boldsymbol{y}-\bar{\cmat {H}}  \boldsymbol{\mu}_{\boldsymbol{x}}\right) \\
& =\boldsymbol{\mu}_l^{-} + \left[ \begin{array}{ll}
\boldsymbol{\sigma_l^{u+}}, &\boldsymbol{\sigma_l^{s+}}  \\
\boldsymbol{\sigma_l^{s+}} ,& \boldsymbol{\sigma_l^{l+}}  \\
\end{array}\right]  \left[ \begin{array}{l}
\sum_{n=1}^{N}  \left(\mathbf{r_n}-\mathbf{\mu}^{\mathrm{u},-}_l\right) \odot \frac{1}{\sigma_i} \\
\boldsymbol{0 } \\
\end{array}\right]  \\
& =\boldsymbol{\mu}_l^{-} + \left[ \begin{array}{l}
\boldsymbol{\sigma_l^{u+}} \\
\boldsymbol{\sigma_l^{s+}}  \\
\end{array}\right]  \odot \left[ \begin{array}{l}
\sum_{i=1}^{N}  \left(\mathbf{r_i}-\mathbf{\mu}^{\mathrm{u},-}_l\right) \odot \frac{1}{\cvec \sigma_{r_i}} \\
\sum_{i=1}^{N}  \left(\mathbf{r_n}-\mathbf{\mu}^{\mathrm{u},-}_l\right) \odot \frac{1}{\cvec \sigma_{r_i}} \\
\end{array}\right]  \\
\end{aligned}
\end{equation}

Here $\mu_l^+$ is the posterior mean and $\boldsymbol{\Lambda}_l^+$ is the posterior precision matrix.

\subsection{Derivation For Matrix Inversions as Scalar Operations}

\begin{minv}
Consider a block  matrix of the following form $\cvec A = \left[\begin{array}{cc} \mathrm{diag}(\cvec a^u) & \mathrm{diag}(\cvec a^s) \\  \mathrm{diag}(\cvec a^s) &  \mathrm{diag}(\cvec a^l)  \end{array}\right]$. Then inverse $A^{-1} = \cvec B $ can be calculated using scalar operations and is given as, $\cvec B = \left[\begin{array}{cc} \mathrm{diag}(\cvec b^u) & \mathrm{diag}(\cvec b^s) \\  \mathrm{diag}(\cvec b^s) &  \mathrm{diag}(\cvec b^l)  \end{array}\right]$ where,
\begin{align}
\label{eq: matscalar}
\begin{split}
\cvec b^u & =\cvec a_l \oslash{ ( \cvec a_u \odot \cvec a_l- \cvec a_s \odot \cvec a_s )}  \\
\cvec b^s & = - \cvec a_s \oslash{ ( \cvec a_u \odot \cvec a_l- \cvec a_s \odot \cvec a_s )}  \\
\cvec b^l & =\cvec a_u \oslash{ ( \cvec a_u \odot \cvec a_l- \cvec a_s \odot \cvec a_s )} 
\end{split}
\end{align}.
\end{minv}

\paragraph{Proof:} To prove this we will use the following matrix identity of a partitioned matrix from \cite{bishop2006pattern}, which states
\begin{align}
\label{eq: matide}
\left(\begin{array}{ll}
\mathrm{A} & \mathrm{B} \\
\mathrm{C} & \mathrm{D}
\end{array}\right)^{-1}=\left(\begin{array}{cc}
\mathrm{M} & -\mathrm{MBD}^{-1} \\
-\mathrm{D}^{-1} \mathrm{CM} & \mathrm{D}^{-1}+\mathrm{D}^{-1} \mathrm{CMBD}^{-1}
\end{array}\right)
\end{align}
where M is defined as
$$
\mathrm{M}=\left(\mathrm{A}-\mathrm{BD}^{-1} \mathrm{C}\right)^{-1}.
$$ Here M is called the Schur complement of the Matrix on the left side of Equation \ref{eq: matide}. The algebraic manipulations to arrive at scalar operations in Equation \ref{eq: matscalar} are straightforward.

\subsection{Proof for Permutation Invariance (Theorem \ref{thm:permInv})} 
\label{subsec: permInvProof}

\perminv*

\begin{proof}The posterior distribution $p(\theta \mid X)$, defined by Bayes' theorem as $p(\theta \mid X) = \frac{p(X \mid \theta)p(\theta)}{p(X)}$, relies on the likelihood $p(X \mid \theta) = \prod_{i=1}^N p(x_i \mid \theta)$. Under any permutation $\pi$, the permuted set $X^\pi = \{\pi(x_1), \ldots, \pi(x_N)\}$ preserves the likelihood since $p(X^\pi \mid \theta) = \prod_{i=1}^N p(\pi(x_i) \mid \theta) = \prod_{i=1}^N p(x_i \mid \theta) = p(X \mid \theta)$. The prior $p(\theta)$ and the marginal likelihood $p(X)$ are invariant under permutation. Therefore, $p(\theta \mid X^\pi) = p(\theta \mid X)$, establishing the permutation invariance of the posterior.
\end{proof}

\subsection{Proof for Gaussian Marginalization (Identity \ref{theo:gm})} 
\label{subsec: margNproof}
\gaussmarg*
\begin{proof}  
The proof is a staright forward extension to the Identity \ref{theo:taskpredict} derived in Chapter \ref{chap:Hipssm}.  First, we derive an expression for the joint distribution $p(\cvec{u_1},\cvec{u_2},..,\cvec{u_N},\cvec{y})$.

\begin{restatable}[Joint Distribution]{gi}{gaussjoint}
\label{identity:gaussjoint} If $\cvec{u_1} \sim \mathcal{N}(\cvec{\mu}_1,\cvec{\Sigma}_1)$ and $\cvec{u_2} \sim \mathcal{N}(\cvec{\mu}_2,\cvec{\Sigma}_2)$ are normally distributed independent random variables and if conditional distribution $p(\cvec{y}|\cvec{u_1},\cvec{u_2}) = \mathcal{N}(\cvec{A_1}\cvec{u_1} + \cvec{A_2}\cvec{u_2}, \cvec{\Sigma})$, the joint distribution has an expression as follows:
\begin{equation}
\left(\begin{array}{c}
\mathbf{u_1} \\
\mathbf{u_2} \\
\vdots \\
\mathbf{u_N} \\
\mathbf{y}
\end{array}\right) \sim \mathcal{N}\left(
\left(\begin{array}{c}
\cvec{\mu}_1 \\
\cvec{\mu}_2 \\
\vdots \\
\cvec{\mu}_N \\
\sum_{n=1}^N \mathbf{A}_n \cvec{\mu}_n
\end{array}\right),
\left(\begin{array}{cccc}
\mathbf{\Sigma}_1 & 0 & \cdots & \mathbf{A}_1^\top \mathbf{\Sigma}_1 \\
0 & \mathbf{\Sigma}_2 & \cdots & \mathbf{A}_2^\top \mathbf{\Sigma}_2 \\
\vdots & \vdots & \ddots & \vdots \\
\mathbf{\Sigma}_1 \mathbf{A}_1 & \mathbf{\Sigma}_2 \mathbf{A}_2 & \cdots & \sum_{n=1}^N \mathbf{A}_n \mathbf{\Sigma}_n \mathbf{A}_n^\top + \mathbf{\Sigma}
\end{array}\right)
\right)
\end{equation}
\end{restatable}

We can write $\mathbf{y}= \sum_{n=1}^N \cmat{A_n}\cvec{u_n} + \cvec{\epsilon}$, where $\cvec{\epsilon} \sim \mathcal{N}(\cvec{0},\cvec{\Sigma})$ and $b$ is a constant. 

Let displacement of a variable  $\mathbf{u}$ be denoted by $\Delta \mathbf{u}=\mathbf{u}-\langle\mathbf{u}\rangle $.

Since $\mathbf{u_i}$ and $\mathbf{u_j}$ are independent $\forall i \neq j$, the covariances
\begin{equation}
\label{eq:gu2indep1}
\begin{aligned}
\left\langle\Delta \mathbf{u_i} \Delta \mathbf{u_j}^{\top}\right\rangle=0, \forall i \neq j .
\end{aligned}
\end{equation}

Similarly, \begin{equation}
\label{eq:gu2indep2}
\begin{aligned}
\left\langle\Delta \mathbf{u_i} \Delta \mathbf{\epsilon}^{\top}\right\rangle=0, \forall i.
\end{aligned}
\end{equation}

For any \(i\), we have the covariance
\begin{equation}
   \begin{aligned}
    \left\langle \Delta \mathbf{u}_i \Delta \mathbf{y}^\top \right\rangle & = \left\langle \Delta \mathbf{u}_i \left(\sum_{n=1}^N \mathbf{A}_n \Delta \mathbf{u}_n + \Delta \epsilon \right)^\top \right\rangle\\ & = \left\langle \Delta \mathbf{u}_i \Delta \mathbf{u}_i^\top \right\rangle \mathbf{A}_i^\top + \sum_{j=1, j \neq i}^N \left\langle \Delta \mathbf{u}_i \Delta \mathbf{u}_j^\top \right\rangle \mathbf{A}_j^\top + \left\langle \Delta \mathbf{u}_i \Delta \epsilon^\top \right\rangle.
\end{aligned} 
\end{equation}

Using equations \(\ref{eq:gu2indep1}\) and \(\ref{eq:gu2indep2}\), we therefore derive an expression for the corresponding covariance as:
\[
\left\langle \Delta \mathbf{u}_i \Delta \mathbf{y}^\top \right\rangle = \mathbf{\Sigma}_i \mathbf{A}_i^\top.
\]

Similarly, 
\begin{align*}
\left\langle \Delta \mathbf{y} \Delta \mathbf{y}^\top \right\rangle 
&= \left\langle \left(\sum_{n=1}^N \mathbf{A}_n \Delta \mathbf{u}_n + \Delta \epsilon \right) \left(\sum_{n=1}^N \mathbf{A}_n \Delta \mathbf{u}_n + \Delta \epsilon\right)^\top \right\rangle \\
&= \sum_{n=1}^N \mathbf{A}_n \left\langle \Delta \mathbf{u}_n \Delta \mathbf{u}_n^\top \right\rangle \mathbf{A}_n^\top + \left\langle \Delta \epsilon \Delta \epsilon^\top \right\rangle \\
&= \sum_{n=1}^N \mathbf{A}_n \mathbf{\Sigma}_n \mathbf{A}_n^\top + \mathbf{\Sigma}_\epsilon.
\end{align*}
The result follows.
\begin{restatable}[Marginalization]{gi}{generalgaussmarg}
\label{identity:generalgaussmarg}
If
\[
\left(\begin{array}{c}
\cvec{u}_1 \\
\cvec{u}_2 \\
\vdots \\
\cvec{u}_N \\
\cvec{y}
\end{array}\right) \sim \mathcal{N}\left(\left(\begin{array}{c}
\cvec{\mu}_1 \\
\cvec{\mu}_2 \\
\vdots \\
\cvec{\mu}_N \\
\cvec{\mu}_y
\end{array}\right),\left(\begin{array}{ccccc}
\cmat{\Sigma}_{11} & \cmat{\Sigma}_{12} & \cdots & \cmat{\Sigma}_{1N} & \cmat{\Sigma}_{1y} \\
\cmat{\Sigma}_{21} & \cmat{\Sigma}_{22} & \cdots & \cmat{\Sigma}_{2N} & \cmat{\Sigma}_{2y} \\
\vdots & \vdots & \ddots & \vdots & \vdots \\
\cmat{\Sigma}_{N1} & \cmat{\Sigma}_{N2} & \cdots & \cmat{\Sigma}_{NN} & \cmat{\Sigma}_{Ny} \\
\cmat{\Sigma}_{y1} & \cmat{\Sigma}_{y2} & \cdots & \cmat{\Sigma}_{yN} & \cmat{\Sigma}_{yy}
\end{array}\right)\right)
\]
then marginal over y is given as $p(\cvec{y}) = \int p(\cvec{y}, \cvec{u_1}, \cvec{u_2}, . ., \cvec{u_N})\prod_{i=1}^N \cvec{u_i} = \mathcal{N}\left(\cvec{\mu}_{y}, \mathbf{\Sigma}_{yy}\right)$
\end{restatable}

\textbf{Proof For Identity \ref{identity:generalgaussmarg}} 
We refer to \cite{bishop2006pattern} for the derivation, which requires calculation of the Schur complement as well as completing the square of the Gaussian p.d.f. to integrate out the variable. The given derivation~\parencite{bishop2006pattern} for two variable multivariate Gaussians can be extended to the case of N variables WLOG.

\textbf{Proof For Identity \ref{theo:gm}} is immediate from Identity \ref{identity:gaussjoint} and Identity \ref{identity:generalgaussmarg}.
\end{proof}
\section{Implementation Details}
\label{sec:mts3-impl}

\subsection{Inference In Slow Time Scale SSM}
\label{subsec:sts-infer}

\subsubsection{Inferring Action Abstraction (sts-SSM)}
Given a set of encoded primitive actions and their corresponding variances $ \{ \cvec{\alpha_{k,t}}, \cvec{ \rho_{k,t}} \}_{t=1}^H $, using the prior and observation model assumptions in Section 3.1.2 of main paper, we infer the latent abstract action $p(\cvec \alpha_{k}|\cvec \alpha_{k,1:H}) = \mathcal{N}(\cvec \mu_{ \alpha_{k}}, \cvec \Sigma_{\alpha_k})  = \mathcal{N}(\cvec \mu_{ \alpha_{k}}, \textrm{diag}\cvec(\sigma_{\alpha_k}))$ as a Bayesian aggregation~\cite{volpp2020bayesian} of these using the following closed-form equations:
\begin{wrapfigure}[11]{r}{.31\linewidth}
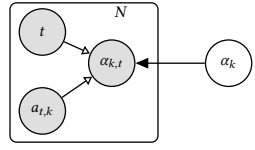

\begin{subfigure}[b]{0.31\textwidth}
   \centering\begin{adjustbox}{width=\textwidth}
         \tikzActAb
     \end{adjustbox}
     \end{subfigure}
\caption{Generative model for the abstract action $\alpha_k$. The hollow arrows are deterministic transformations leading to implicit distribution $\alpha_{k,t}$ using an action set encoder.}
\label{fig:pgmactabs}
\end{wrapfigure}

\begin{align*}
\cvec{\sigma}_{\alpha_{k}} & = \left( \left( \cvec{\sigma}_{0} \right)^\ominus + \sum_{n=1}^N \left(\left(\cvec{\rho_{k,t}}\right)^\ominus\right) \right)^\ominus, \\
\cvec{\mu}_{\alpha_{k}} & = \cvec{\mu}_{0} + \cvec{\sigma}_{\alpha_{k}} \odot \sum_{n=1}^N \left(\cvec{\alpha_{k,t}} - \cvec{\mu}_{0} \right) \oslash \cvec{\rho_{k,t}}
\end{align*}

Here, $\ominus$, $\odot$ and $\oslash$ denote element-wise inversion, product, and division, respectively. The update equation is coded as the ``abstract action inference'' neural network layer as shown in Figure \ref{fig:schematic}. 

\subsubsection{Task Prediction (sts-SSM)}

The goal of this step is to update the prior marginal over the latent task variable $\cvec{l}_k$, $p(\cvec{l}_k|\cvec{\beta}_{1:k-1},\cvec{\alpha}_{1:k})$, given the posterior beliefs from the time window $k-1$ and abstract action $\cvec \alpha_{k}$.

Using the linear dynamics model assumptions from Section 3.3, we can use the following closed-form update equations to compute,
$p(\cvec{l}_k|\cvec{\beta}_{1:k-1},\cvec{\alpha}_{1:k}) = \mathcal{N}(\cvec{\mu}_{l_{k}}^-,\cvec{\Sigma}_{l_{k}}^-)$, where
\begin{equation}
\label{eq:task-predict}
\begin{aligned}[c]
\cvec{\mu}_{l_{k}}^- &=\cmat{X}\cvec{\mu}_{l_{k-1}}^+ + \cmat{Y}\cvec{\alpha}_{k}\\
   \cvec{\Sigma}_{l_{k}}^- &= \cmat{X}\cvec{\Sigma}_{l_{k-1}}^+\cmat{X}^T + \cmat{Y}\cvec{\Sigma}_{\alpha_{k}}\cmat{Y}^T + \cmat S.
  \end{aligned}
\end{equation}

These closed-form equations are coded as the ``task predict'' neural net layer as shown in Figure \ref{fig:schematic}.

\subsubsection{Task Update (sts-SSM)}
\sloppy
In this stage, we update the prior over $l_k$ using an abstract observation set $\{\cvec \beta_{k,t}\}_{t=1}^H$, to obtain the latent task the posterior $\mathcal{N}(\cvec{\mu}_{z_{k,t}}^+,\cmat{\Sigma}_{z_{k,t}}^+) = \mathcal{N}(\left[\begin{array}{c}\cvec{\mu}_{t}^{u+} \\
\cvec{\mu}_{t}^{l+} \end{array}\right],\left[\begin{array}{cc}\cvec \Sigma_t^u & \cvec \Sigma^{s}_t \\ \cvec \Sigma^{s}_t & \cvec \Sigma^{l}_t  \end{array}\right]^+)$, with $\cvec \Sigma_{l_k}^u = \textrm{diag}(\cvec \sigma_{l_k}^u), \; \cvec \Sigma_{l_k}^l = \textrm{diag}(\cvec \sigma_{l_k}^l)$ \textrm{ and } $\cvec \Sigma_{l_k}^{s} = \textrm{diag}(\cvec \sigma_{l_k}^s)$.

To do so we first invert the prior covariance matrix $\left[\begin{array}{cc}\cvec \Sigma_{l_k}^u & \cvec \Sigma_{l_k}^{s} \\ \cvec \Sigma_{l_k}^{s} & \cvec \Sigma_{l_k}^{l}  \end{array}\right]^+$ to the precision matrix $\left[\begin{array}{cc}\cvec \lambda_{l_k}^u & \cvec \lambda_{l_k}^{s} \\ \cvec \lambda_{l_k}^{s} & \cvec \lambda_{l_k}^{l}  \end{array}\right]^+$ for permutation invariant parallel processing. The posterior precision is then computed using scalar operations are follows, where only $\cvec \lambda_{l_{k}}^{u}$ is changed by 
\begin{align}\cvec \lambda_{l_{k}}^{u+} = \cvec \lambda_{l_{k}}^{u-} + \sum_{t=1}^H \cvec 1 \oslash \cvec \nu_{k,t}\end{align}
 while $\cvec \lambda_{l_{k}}^{l+} = \cvec \lambda_{l_{k}}^{l-}$ and $\cvec \lambda_{l_{k}}^{s+} = \cvec \lambda_{l_{k}}^{s-}$ remain constant. The operator $\oslash$ denotes the element-wise division. The posterior precision is inverted back to the posterior covariance vectors $\cvec \sigma_{l_{k}}^{u+}$, $\cvec \sigma_{l_{k}}^{l+}$ and $\cvec \sigma_{l_{k}}^{s+}$. Now, the posterior mean $\cvec{\mu}_{l,k}^{+}$ can be obtained from the prior mean $\cvec{\mu}_{l,k}^{-}$ as
 \begin{equation}
  \begin{aligned}[c]
   \cvec{\mu}_{l,k}^{+}=\cvec{\mu}_{l,k}^{-} + \left[ \begin{array}{l}
\cvec{\sigma}_{l_k}^{u+} \\
\cvec{\sigma}_{l_k}^{s+} \\
\end{array}\right]  \odot \left[ \begin{array}{l}
\sum_{t=1}^{H}  \left(\cvec{\beta}_{k,t}-\cvec{\mu}^{\mathrm{u},-}_{l_k}\right) \oslash \cvec \nu_{k,t} \\
\sum_{t=1}^{H}  \left(\cvec{\beta}_{k,t}-\cvec{\mu}^{\mathrm{u},-}_{l_k}\right) \oslash \cvec \nu_{k,t} \\
\end{array}\right].   \hspace{1.5cm}\\
  \end{aligned}
\end{equation}
\\
The inversion between the covariance matrix and precision matrix can be done via scalar operations leveraging block diagonal structure as derived in Appendix \ref{sec:proofMT}. Figure \ref{fig:schematicTaskU} shows the schematic of the task update layer.
\subsection{Inference In Fast Time Scale SSM}
\label{subsec:fts-infer}
The inference in fts-SSM for a time-window $k$ involves two stages as illustrated in Figure \ref{fig:mts-schematic}, calculating the prior and posterior over the latent state variable $z_t$. To keep the notation uncluttered, we will also omit the time-window index $k$ whenever the context is clear as in Section 3.2.

\subsubsection{Task Conditional State Prediction (fts-SSM)}

\sloppy Following the assumptions of a task conditional linear dynamics as in Section 3.2 of the main paper, we obtain the prior marginal for $p(\cvec{z}_{k,t}|\cvec{w}^k_{1:t-1},\cvec{a}^k_{1:t-1},\cvec{\beta}_{1:k-1},\cvec{\alpha}_{1:k-1}) = \mathcal{N}(\cvec{\mu}_{z_{k,t}}^-,\cmat{\Sigma}_{z_{k,t}}^-)$ in closed form, where
\begin{equation}
\label{eq:cond-predict}
\centering
\begin{aligned}
\cvec{\mu}_{z_{k,t}}^- &=\cmat{A}\cvec{\mu}_{z_{k,t-1}}^- + \cmat{B}\cvec{a}_{k,t-1} + \cmat{C}\cvec{\mu}_{l_{k}}^-, \\
   \cmat{\Sigma}_{k,t}^- &= \cmat{A}\cmat{\Sigma}_{k,t-1}^+\cmat{A}^T + \cmat{C}\cvec{\Sigma}_{l_{k}}^-\cmat{C}^T + \cmat{Q}.
  \end{aligned}
\end{equation}

\subsubsection{Observation Update (fts-SSM)}
In this stage, we compute the posterior belief $p(\cvec{z}_{k,t}|\cvec{w}^k_{1:t},\cvec{a}^k_{1:t},\cvec{\beta}_{1:k},\cvec{\alpha}_{1:k-1}) = \mathcal{N}(\cvec{\mu}_{z_{k,t}}^-,\cmat{\Sigma}_{z_{k,t}}^-)$.  using the same closed-form update as in  \cite{becker2019recurrent}. The choice of the special observation model splits the state into two parts, an upper $\boldsymbol{z}_t^\textrm{u}$ and a lower part $\boldsymbol{z}_t^\textrm{l}$, resulting in the posterior belief $\mathcal{N}(\cvec{\mu}_{z_{k,t}}^-,\cmat{\Sigma}_{z_{k,t}}^-) = \mathcal{N}(\left[\begin{array}{c}\cvec{\mu}_{t}^{u+} \\
\cvec{\mu}_{t}^{l+} \end{array}\right], \left[\begin{array}{cc}\cvec \Sigma_t^u & \cvec \Sigma^{s}_t \\ \cvec \Sigma^{s}_t & \cvec \Sigma^{l}_t  \end{array}\right]^+)$, with $\cvec \Sigma_t^u = \textrm{diag}(\cvec \sigma_t^s), \; \cvec \Sigma_t^l = \textrm{diag}(\cvec \sigma_t^l)$ and $\cvec \Sigma_t^{s} = \textrm{diag}(\cvec \sigma_{t}^s)$. Thus, the factorization allows for only the diagonal and one off-diagonal vector of the covariance to be computed and simplifies the calculation of the mean and posterior to simple scalar operations. 

The closed-form equations for the mean can be expressed as the following scalar equations,
\begin{align*}
\boldsymbol{z}_t^+ = \boldsymbol{z}_t^- +
\left[\begin{array}{c} \boldsymbol{\sigma}^\mathrm{u,-}_t \\
\boldsymbol{\sigma}^\mathrm{l,-}_t \end{array}\right]
\odot
\left[\begin{array}{c}\boldsymbol{w}_t - \boldsymbol{z}^{\mathrm{u},-}_t \\
\boldsymbol{w}_t - \boldsymbol{z}^{\mathrm{u},-}_t  \end{array}\right]
\oslash 
\left[\begin{array}{c}  \boldsymbol{\sigma}_t^{\mathrm{u},-} + \boldsymbol{\sigma}_t^\mathrm{obs} 
 \\  \boldsymbol{\sigma}_t^{\mathrm{u},-} + \boldsymbol{\sigma}_t^\mathrm{obs} \end{array}\right],
\end{align*}

The corresponding equations for the variance update can be expressed as the following scalar operations,
\begin{align*}
\boldsymbol{\sigma}^{\mathrm{u},+}_t &= \boldsymbol{\sigma}^{\mathrm{u},-}_t \odot \boldsymbol{\sigma}^{\mathrm{u},-}_t \oslash \left( \boldsymbol{\sigma}_t^{\mathrm{u},-} + \boldsymbol{\sigma}_t^\mathrm{obs} \right),  \\
\boldsymbol{\sigma}^{\mathrm{s},+}_t &= \boldsymbol{\sigma}^{\mathrm{u},-}_t \odot \boldsymbol{\sigma}^{\mathrm{s},-}_t \oslash \left( \boldsymbol{\sigma}_t^{\mathrm{u},-} + \boldsymbol{\sigma}_t^\mathrm{obs} \right), \\
\boldsymbol{\sigma}^{\mathrm{l},+}_t &= \boldsymbol{\sigma}^{\mathrm{l}, -}_t - \boldsymbol{\sigma}^{\mathrm{s},-}_t \odot \boldsymbol{\sigma}^{\mathrm{s},-}_t \oslash \left( \boldsymbol{\sigma}_t^{\mathrm{u},-} + \boldsymbol{\sigma}_t^\mathrm{obs} \right),
\end{align*},
where $\odot$ denotes the elementwise vector product and  $\oslash$ denotes an elementwise vector division. 

\subsection{Modelling Assumptions}

\subsubsection{Control Model} To achieve action conditioning within the recurrent cell of fts-SMM, we include a control model $b(a_{k,t})$ in addition to the linear transition model $A_t$. $b(a_{k,t}) =f(a_{k,t})$, where $f(.)$ can be any non-linear function approximator.  We use a multi-layer neural network regressor with ReLU activations~\cite{shaj2020action}. 

However, unlike the fts-SSM where actions are assumed to be known and subjected to no noise, in the sts-SSM, the abstract action is an inferred latent variable with an associated uncertainty estimate. Hence we use a linear control model $Y$, for principled uncertainty propagation. 

\subsubsection{Transition Noise}
We assume the covariance of the transition noise $Q$ and $S$ in both timescales to be diagonal. The noise is learned and is independent of the latent state.

\section{Metrics Used For Measuring Long Horizon Predictions}
\label{sec:mts3-metric}
\subsection{Sliding Window RMSE}
The sliding window RMSE (Root Mean Squared Error) metric is computed for a predicted trajectory in comparison to its ground truth. At each time step, the RMSE for each trajectory is determined by taking the root mean square of the differences between the ground truth and predicted values within a sliding window that terminates at the current time step. This sliding window, with a specified size, provides a smoothed localized assessment of prediction accuracy over the entire prediction length.
Mathematically, the sliding window RMSE at time step $t$ is given by:
$$
\operatorname{RMSE}(t)=\sqrt{\frac{1}{W} \sum_{i=t-W+1}^t\left(\mathrm{gt}_i-\operatorname{pred}_i\right)^2}
$$
where $t$ is the current time step, $W$ is the window size, and $\mathrm{gt}_i$ and $\operatorname{pred}_i$ are the ground truth and predicted values at time step $i$, respectively. The extension to multiple trajectories is straightforward and omitted to keep the notation uncluttered.

\subsection{Sliding Window NLL}
The sliding window NLL (Negative Log-Likelihood) metric is computed for a predicted probability distribution against the true distribution. At each time step, the NLL is determined by summing the negative log-likelihood values within a sliding window that terminates at the current time step. This sliding window, with a specified size, provides a smoothed localized evaluation of prediction accuracy across the entire sequence.

Mathematically, the sliding window NLL at time step $t$ is given by:
$$
\mathrm{NLL}(t)=-\frac{1}{W} \sum_{i=t-W+1}^t \log \mathcal{N}\left(\mathrm{gt}_i \mid \mathrm{predMean}_i, \mathrm{predVar}_i\right)
$$
where $t$ is the current time step, $W$ is the window size. $\mathrm{predMean}_i$, $\mathrm{predVar}_i$, and $\mathrm{gt}_i$ represent the predicted mean, predicted variance, and the ground truth at time step $i$.

\section{Visualization of predictions given by different models.}
\label{sec:mts3-vis}
 In this section, we plot the multistep ahead predictions (mean and variance) by different models on 3 datasets on normalized test trajectories. Not that we omit NaN values in predictions while plotting.
 \newpage

 \subsubsection{Franka Kitchen}
  \begin{figure*}[ht!]
\begin{center}
\includegraphics[scale=0.5]{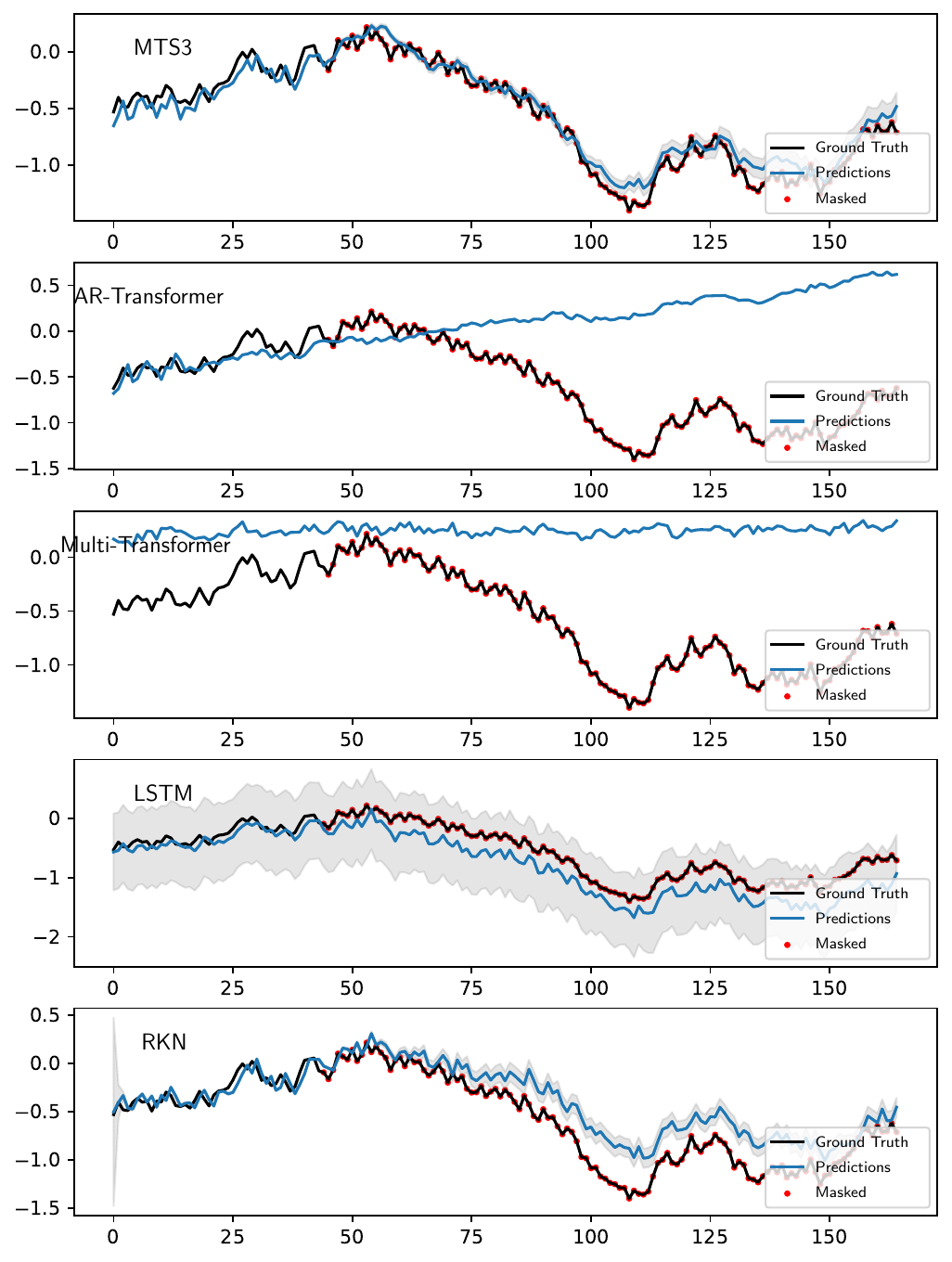}
\end{center}
\caption{Multi-step ahead mean and variance predictions for a particular joint (joint 1) of Franka Kitchen Environment. The multi-step ahead prediction starts from the first red dot, which indicates masked observations. MTS3 gives the most reliable mean and variance estimates.} 
 \label{fig:kitchen}
\end{figure*}
\newpage
 \subsubsection{Hydraulic Excavator}
 
\begin{figure*}[ht!]
\begin{center}
\includegraphics[scale=0.45]{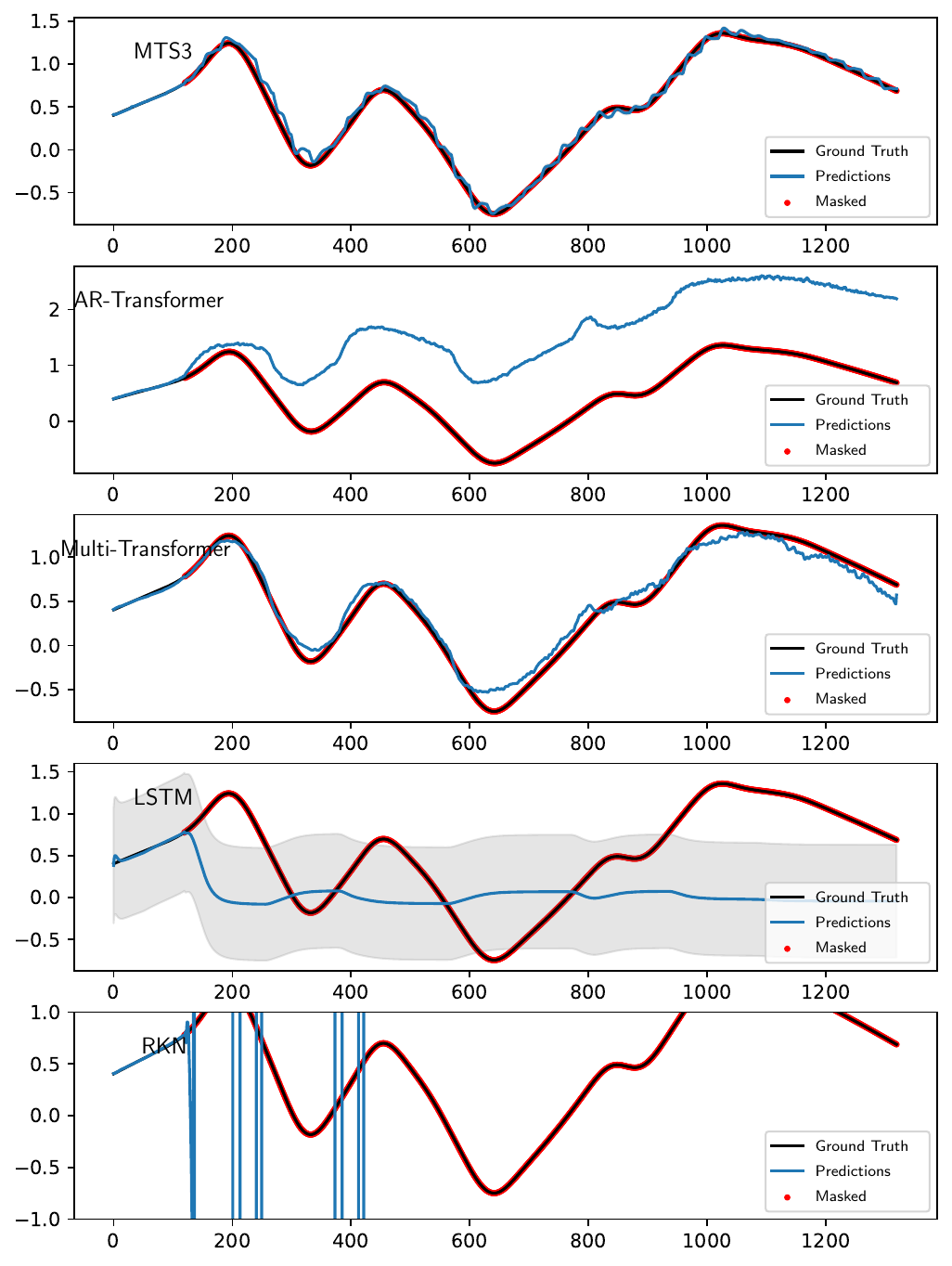}
\end{center}
\caption{Multi-step ahead mean and variance predictions for a particular joint (joint 1) of Excavator Dataset. The multi-step ahead prediction starts from the first red dot, which indicates masked observations. MTS3 gives the most reliable mean and variance estimates even up to 12 seconds into the future. Another interesting observation can also be seen in the predictions for MTS3, where after every window k of sts-SSM, which is 0.3 seconds (30 timesteps) long, the updation of the higher-level abstractions helps in grounding the lower-level predictions thus helping in the long horizon yet fine-grained predictions.} 
 \label{fig:exc}
\end{figure*}
\newpage

\subsubsection{Mobile Robot}
 
\begin{figure*}[ht!]
\begin{center}
\includegraphics[scale=0.5]{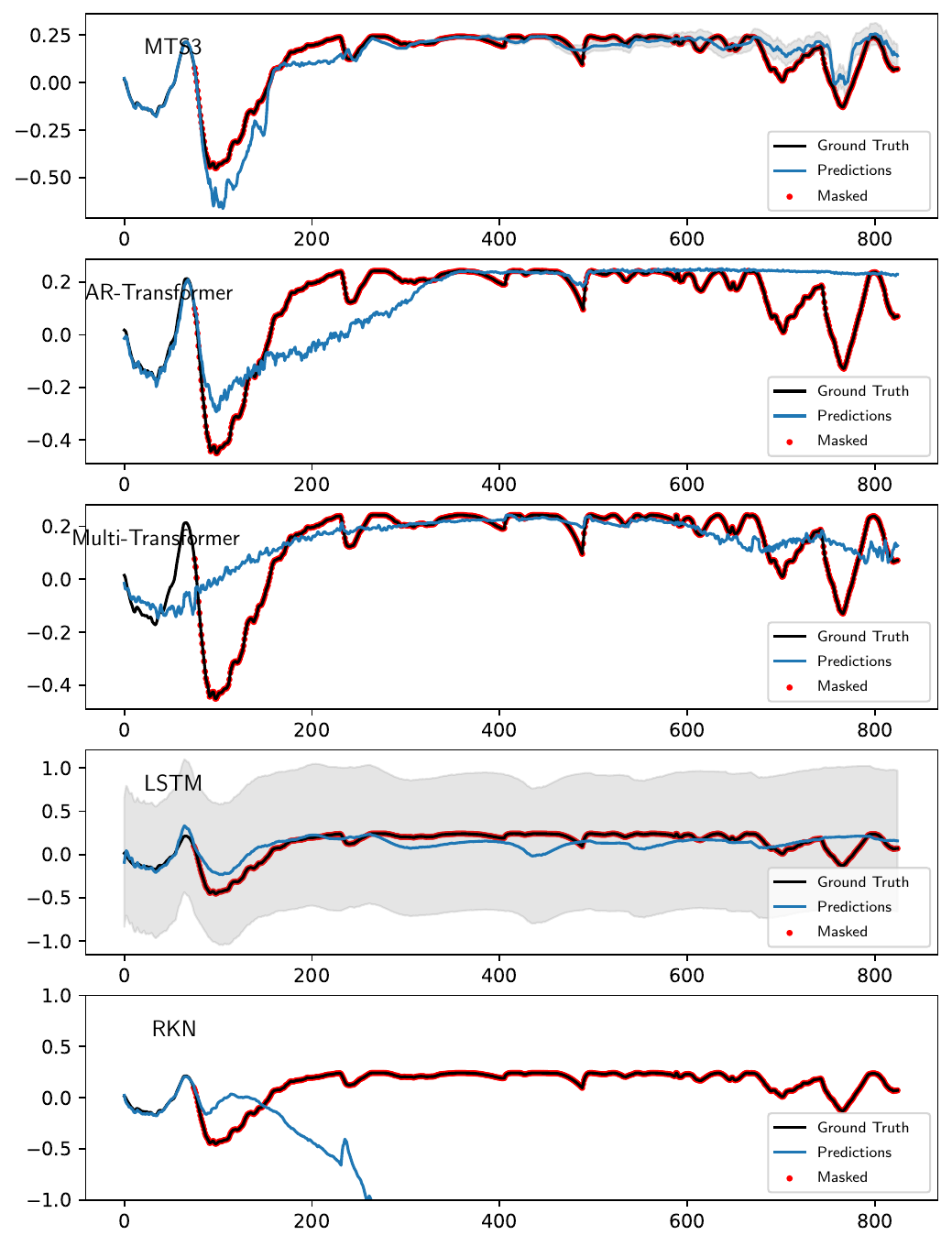}
\end{center}
\caption{Multi-step ahead mean and variance predictions for a particular joint (joint 7) of Mobile Robot Dataset. The multi-step ahead prediction starts from the first red dot, which indicates masked observations. MTS3 gives the most accurate mean and variance estimates among all algorithms. } 
 \label{fig:sin}
\end{figure*}

\chapter{Appendix: Robots and Dataset Details}
\section{Robots and Data Used For In Chapter \ref{chap:Acssm} on Ac-SSM}
\label{sec:dataAc}

The experiments are performed on data from four different robots. The details of robots, data, and data preprocessing are explained below: 
\subsection{Hydraulic Brokk 40 Robot Arm}

\textbf{Observation and Data Set}: The data was obtained from a HYDROLEK–7W 6 degree–of–freedom manipulator with a continuous (360 degree) jaw rotation mechanism. We actuate the joints via hydraulic pistons, which are powered via an auxiliary output from the hydraulic pump. Thus learning the forward model is difficult due to inherent hysteresis associated with hydraulic control. For this robot, only one joint is moved at a time, so we have independent time series per joint. The joint data consists of measured joint positions and the input current to the controller of the joint sampled at 100Hz. \\

\textbf{Training Procedure}: During training, we work with sequences of length 500. For the first 300 time steps those sequences consist of the full observation, i.e., the joint position and current. We give only the current signals in the remaining 200 time steps. The models have to impute the missing joint positions in an uninformed fashion, i.e., we only indicate the absence of a position by unrealistically high values.

\subsection*{Musculoskeletal Robot Arm}
\label{sec:PAMdesc}

\textbf{Observation and Data Set:}  For this soft robot we have 4 dimensional observation inputs(joint angles) and 8 dimensional action inputs(pressures). We collected the data of a four DoF robot actuated by Pneumatic Artificial Muscles
(PAMs). The robot arm has eight PAMs in total with each DoF actuated by an antagonistic pair. The robot arm reaches high joint angle accelerations of up to 28, 000deg/s2 while avoiding dangerous joint limits thanks to the antagonistic actuation and limits on the air pressure ranges. The data
consists of trajectories collected while training with a model-free reinforcement learning algorithm to hit balls while playing table tennis. We sampled the data at 100Hz. The hysteresis associated with the pneumatic actuators used in this robot is challenging to model and is relevant to the soft robotics in general.\\

\textbf{Training Procedure}: During training, we randomly removed three-quarters of the states from the sequences and tasked the models with imputing those missing states, only based on the knowledge of available actions/control commands, i.e., we train the models to perform action conditional future predictions to impute missing states. The imputation employs the model for multi-step ahead predictions in a convenient way. One could instead go for a dedicated loss function as in approaches like \cite{finn2016unsupervised}, \cite{oh2015action} for long term predictions.

\subsubsection*{Franka Emika Panda Robot Arm}

\textbf{Observation and Data Set:} We collected the data from a 7 DoF Franka Emika Panda manipulator during free motion. We chose this task since the robot exhibits different dynamics behaviour due to electric actuators and high frequencies(1kHz). The raw joint positions, velocities and torques were recorded using Franka Interfaces while the joint accelerations were computed by finite differences on filtered velocity data (obtained using a zero-phase 8th-order digital Butterworth filter with a cut-off frequency of 5Hz). The observations for the forward model consist of the seven joint angles in radians, and the corresponding actions were joint Torques in Nm. While the inverse model use both joint angles and velocities as observations. The data was divided into train and test sets in the ratio 4:1. We divide the data into sequences of length 300 while training the recurrent models for forward dynamics and use sequences of length 50 for inverse dynamics.   \\  

\textbf{Training Procedure Forward Dynamics}: Similar to the multi-step ahead training procedure in \ref{sec:PAMdesc}, during training we randomly removed three-quarters of the observations(joint angles) from the sequences and tasked the models with imputing those missing observations, only based on the knowledge of available actions/control commands.

\textbf{Training Procedure Inverse Dynamics}: The recurrent models (LSTM, ac-RKN) uses a similar architecture, as shown in Figure 3 of the main paper, except for the recurrent module. The hyperparameters including learning rate, latent state and observation dimensions, learning rate, control model architecture, action decoder architecture and regularization parameter for the joint forward-inverse dynamics loss function are searched via GpyOpt\cite{gpyopt2016} and is mentioned in Appendix D. The observation encoder and decoder architecture is chosen to be of the same size across the models being compared. For all models, we use the joint positions and velocities as the observation input and differences to the next state as desired observation. The FFNN gets the current observation and desired observation as input and is tasked to predict the joint Torques directly(unlike differences in recurrent models) as in previous regression approaches\cite{nguyen2010using}. 
\label{sec:invtraining}

\subsubsection*{Barrett WAM Robot Arm}

\textbf{Observation and Data Set:} The Barett task is based on a publicly available dataset comprising joint positions, velocities, acceleration and torques of a seven degrees-of-freedom real Barett WAM robot. The original training dataset (12, 000 data points) is split into sequences of length 98. Twenty-four out of the total 119 episodes are utilized for testing, whereas the other 95 are used for training. The direct cable drives which drive this robot produce high torques, generating fast and dexterous movements but yield complex dynamics.  Rigid-body dynamics cannot be model this complex dynamics due to the variable stiffness and lengths of the cables. \\  

\textbf{Training Procedure Inverse Dynamics}: The training procedure is repeated as in \ref{sec:invtraining}

\section{Robots and Datasets Used In Chapter \ref{chap:Hipssm} on HiP-SSM}
\label{sec:dataHip}
\subsection{Franka Emika Panda Robot Arm}

\textbf{Observation and Data Set:} We collected the data from a 7 DoF Franka Emika Panda manipulator during free motion and while manipulating loads with weights 0kg~(free motion), 0.5 kg, 1 kg, 1.5 kg, 2 kg and 2.5 kg. Data is sampled at high frequencies~(1kHz). The training trajectories were motions with loads 0kg(free motion), 1kg, 1.5kg, 2.5 kgs, while the testing trajectories contained motions with loads of 0.5kg and 2 kgs. The observations for the forward model consist of the seven joint angles in radians, and the corresponding actions were joint Torques in Nm. We divide the data into sequences of length 600 while training the recurrent models for forward dynamics, with 300 time-steps~(corresponding to 300 milli-seconds) used as context set and rest 300 is used for the recurrent time-series modelling.   \\  

\textbf{Training Procedure}: For the fully observable case, we trained HiP-RSSM for one-step ahead prediction using an RMSE loss. Similar to the training procedure for partially observable case as in \ref{sec:PAMdesc}, during training we randomly removed half of the observations(joint angles) from the sequences and tasked the models with imputing those missing observations, only based on the knowledge of available actions/control commands.\\
Other recurrent baselines (RKN, LSTM, GRU) are trained in a similar fashion except that, we dont maintain a context set of interaction histories during training/inference.

\subsection{Wheeled Mobile Robot}
\label{app:wheel}
\textbf{Observation and Data Set:} We collected 50 random trajectories from a Pybullet simulator a wheeled mobile robot traversing terrain with sinusoidal slopes. Data is sampled at high frequencies~(500Hz). 40 out of the 50 trajectories were used for training and the rest 10 for testing. The observations consists of parameters which completely describe its location and orientation of the robot. The observation of the robot at any time instance $t$ consists of the following features:.$$
\begin{array}{r}
o_{t}=[x, y, z, \cos (\alpha), \sin (\alpha), \cos (\beta) \\
\sin (\beta), \cos (\gamma), \sin (\gamma)]
\end{array}
$$
where,
$x, y, z$ - denote the global position of the Center of Mass of he robot,
$\alpha, \beta, \gamma-$ Roll, pitch and yaw angles of the robot respectively, in the global frame of reference~\parencite{sonker2020adding}.  We divide the data into sequences of length 300 while training the recurrent models for forward dynamics, with 150 time-steps~(corresponding to 300 milli-seconds) used as context set and rest 150 is used for the recurrent time-series modelling.   \\  

\textbf{Training Procedure}: For the fully observable case, we trained HiP-RSSM for one-step ahead prediction using an RMSE loss. Similar to the training procedure for partially observable case as in \ref{sec:PAMdesc}, during training we randomly removed half of the observations from the sequences and tasked the models with imputing those missing observations, only based on the knowledge of available actions/control commands.\\
Other recurrent baselines (RKN, LSTM, GRU) are trained in a similar fashion except that, we dont maintain a context set of interaction histories during training/inference.

\section{Robots and Data used in Chapter \ref{chap:mts3} on Multi Time Scale SSM}
\label{sec:dataMts3}
In all datasets, we only use information about agent/object positions and we mask out velocities to create a partially observable setting. All datasets are subjected to a mean zero, unit variance normalization during training. During testing, they are denormalized after predictions. The details of the different datasets used are explained below:

\subsection{D4RL Datasets}

\textbf{Details:} We use a set of 3 different environments/agents from D4RL dataset~\cite{fu2020d4rl}, which includes the HalfCheetah, Franka Kitchen and Maze2D (medium) environment. \textbf{(a) HalfCheetah:} We used 1000 suboptimal trajectories collected from a policy trained to approximately 1/3 the performance of the expert. The observation space consists of 8 joint positions and the action space consists of 6 joint torques collected at 50 Hz frequency. 800 trajectories were used for training and 200 for testing. For the long horizon task, we used 1.2 seconds (60 timesteps) as context and tasked the model to predict 6 seconds (300 timesteps) into the future. \textbf{(b) Franka Kitchen:} The goal of the Franka Kitchen environment is to interact with the various objects to reach a desired state configuration. The objects you can interact with include the position of the kettle, flipping the light switch, opening and closing the microwave and cabinet doors, or sliding the other cabinet door. We used the "complete" version of the dataset and collected 1000 trajectories where all four tasks are performed in order. The observation space consists of 30 dimensions (9 joint positions of the robot and 21 object positions). The action space consists of 9 joint velocities clipped between -1 and 1 rad/s. The data was collected at a 50 Hz frequency. 800 trajectories were used for training and 200 for testing. For the long horizon task, we used 0.6 seconds (30 timesteps) as context and tasked the model to predict 2.7 seconds (135 timesteps) into the future. The dataset is complex due to multi-task, multi-object interactions in a single trajectory.
\textbf{(c) Medium Maze:} We used 20000 trajectories from a 2D Maze environment, where each trajectory consists of a force-actuated ball (along the X and Y axis) moving to a fixed target location. The observation consists of as the (x, y) locations and a 2D action space. The data is collected at 100 Hz frequency. 16000 trajectories were used for training and 4000 for testing. For the long horizon task, we used 0.6 seconds (60 timesteps) as context and tasked the model to predict 3.9 seconds (390 timesteps) into the future. Rendering of the three environments is shown in Figure \ref{fig:composite}.


\subsection{Hydraulic Excavator}

\textbf{Details:} We collected the data from a wheeled excavator JCB Hydradig 110W show in Figure \ref{fig:composite}. The data was collected by actuating the boom and arm of the excavator using Multisine and Amplitude-Modulated Pseudo-Random Binary Sequence
(APRBS) joystick signals with safety mechanisms in place. A total of 150 mins of data was collected at a frequency of 100 Hz. of which was used as a training dataset and the rest as testing. The observation space consists of the boom and arm positions, while the joystick
signals are chosen as actions. For the long horizon task we used 1.5 seconds (150 timesteps) as context and tasked the model to predict 12 seconds (1200 timesteps) into the future.  \\

\subsection{Panda Robot With Varying Payloads}

\textbf{Details:} We collected the data from a 7 DoF Franka Emika Panda manipulator during free motion and while manipulating loads with weights 0kg~(free motion), 0.5 kg, 1 kg, 1.5 kg, 2 kg and 2.5 kg. The robot used is shown in Figure \ref{fig:composite}. Data is sampled at a frequency of 100 Hz. The training trajectories were motions with loads of 0kg(free motion), 1kg, 1.5kg, and 2.5 kgs, while the testing trajectories contained motions with loads of 0.5 kg and 2 kg. The observations for the forward model consist of the seven joint angles in radians, and the corresponding actions were joint Torques in Nm. For the long horizon task we used 0.6 seconds (60 timesteps) as context and tasked the model to predict 1.8 seconds (180 timesteps) into the future.  \\  


\subsection{Wheeled Mobile Robot}
\label{app:wheel}
\textbf{Observation and Data Set:} We collected 50 random trajectories from a Pybullet simulator a wheeled mobile robot traversing terrain with slopes generated by a mix of sine waves as opposed to the sine wave terrain for experiments used in Chapter \ref{chap:Hipssm}, making this more challenging. Data is sampled at high frequencies~(500Hz). 40 out of the 50 trajectories were used for training and the rest 10 for testing. The observations consist of parameters which completely describe the location and orientation of the robot. The observation of the robot at any time instance $t$ consists of the following features:
$$
\begin{array}{r}
o_{t}=[x, y, z, \cos (\alpha), \sin (\alpha), \cos (\beta) \\
\sin (\beta), \cos (\gamma), \sin (\gamma)]
\end{array}
$$
where,
$x, y, z$ - denote the global position of the Center of Mass of the robot,
$\alpha, \beta, \gamma-$ Roll, pitch and yaw angles of the robot respectively, in the global frame of reference~\parencite{sonker2020adding}.  For the long horizon task we used 0.6 seconds (150 timesteps) as context and tasked the model to predict 3 seconds (750 timesteps) into the future.   \\

\chapter{Appendix: Hyperparameters}
\section{Hyperparameters: Chapter \ref{chap:Acssm}}
\label{sec:hpyAc}
\subsection*{Pneumatic Musculoskeltal Robot Arm}

\begin{table}[htbp]
\centering
\caption{Forward Dynamics Hyperparameters For Pneumatic Musculoskeletal Robot.}
\label{tab:short}
\resizebox{0.8\textwidth}{!}{\begin{tabular}{|l|l|l|l|}
\hline
\textbf{Hyperparameter} & \textbf{ac-RKN} & \textbf{RKN} & \textbf{LSTM} \\ 
\hline 
Learning Rate & 3.1e-3 & 1.9e-3 & 6.6e-3 \\ 
\hline
Latent Observation Dimension & 60 & 60 & 60  \\ 
\hline
Latent State Dimension & 120 & 120 & 120  \\
\hline
\end{tabular}}
\end{table}

\underline{Encoder} (ac-RKN,RKN,LSTM): 1 fully connected + linear output  (elu + 1) 
\begin{itemize}
	\item Fully Connected 1: 120, ReLU
\end{itemize} 
\underline{Observation Decoder} (ac-RKN,RKN,LSTM): 1 fully connected + linear output:
\begin{itemize}
	\item Fully Connected 1: 120, ReLU
\end{itemize} 
 
\underline{Transition Model} (ac-RKN,RKN):
bandwidth: 3, number of basis: 15
\begin{itemize}
    \item $\alpha(\vec{z}_t)$: No hidden layers - softmax output
\end{itemize}
\underline{Control Model} (ac-RKN):
3 fully connected + linear output
\begin{itemize}
	\item Fully Connected 1: 120, ReLU
	\item Fully Connected 2: 120, ReLU
	\item Fully Connected 3: 120, ReLU
\end{itemize} 

\textbf{Architecture For FFNN Baseline}
2 fully connected + linear output
\begin{itemize}
	\item Fully Connected 1: 6000, ReLU
	\item Fully Connected 2: 3000, ReLU
\end{itemize}

Dropout Regularization - 0.512\\
Learning Rate - 1.39e-2\\
Optimizer Used: Adam Optimizer

\subsection*{Hydraulic Brokk-40 Robot Arm}
\begin{table}[htbp]
\centering
\caption{Forward Dynamics Hyperparameters For Pneumatic Musculoskeletal Robot.}
\label{tab:short}
\resizebox{0.8\textwidth}{!}{
\begin{tabular}{|l|l|l|l|l|}
\hline
\textbf{Hyperparameter} & \textbf{ac-RKN} & \textbf{RKN} & \textbf{LSTM} & \textbf{GRU} \\ 
\hline 
Learning Rate & 5e-4 & 5e-4 & 9.1e-4 & 2.1e-3 \\ 
\hline
Latent Observation Dimension & 30 & 30 & 30 & 30  \\ 
\hline
Latent State Dimension & 60 & 60 & 60 & 60  \\
\hline
\end{tabular}}
\end{table}

\underline{Encoder} (ac-RKN,RKN,LSTM,GRU): 1 fully connected + linear output  (elu + 1) 
\begin{itemize}
	\item Fully Connected 1: 30, ReLU
\end{itemize} 
\underline{Observation Decoder} (ac-RKN,RKN,LSTM,GRU): 1 fully connected + linear output:
\begin{itemize}
	\item Fully Connected 1: 30, ReLU
\end{itemize} 
 
\underline{Transition Model} (ac-RKN,RKN):
bandwidth: 3, number of basis: 32
\begin{itemize}
    \item $\alpha(\vec{z}_t)$: No hidden layers - softmax output
\end{itemize}
\underline{Control Model} (ac-RKN):
1 fully connected + linear output
\begin{itemize}
	\item Fully Connected 1: 120, ReLU
\end{itemize} 

\subsection*{Franka Emika Panda - Forward Dynamics Learning}
\begin{table}[htbp]
\centering
\caption{Forward Dynamics Learning Hyperparameters For Panda.}
\label{tab:short}
\resizebox{0.8\textwidth}{!}{
\begin{tabular}{|l|l|l|l|l|}
\hline
\textbf{Hyperparameter} & \textbf{ac-RKN} & \textbf{RKN} & \textbf{LSTM} & \textbf{GRU} \\ 
\hline 
Learning Rate & 3.1e-3 & 1.7e-3 & 6.6e-3 & 8.72e-3 \\ 
\hline
Latent Observation Dimension & 45 & 30 & 30 & 45  \\ 
\hline
Latent State Dimension & 90 & 60 & 60 & 90  \\
\hline
\end{tabular}}
\end{table}
\underline{Encoder} (ac-RKN,RKN,LSTM,GRU): 1 fully connected + linear output  (elu + 1) 
\begin{itemize}
	\item Fully Connected 1: 120, ReLU
\end{itemize} 
\underline{Observation Decoder} (ac-RKN,RKN,LSTM,GRU): 1 fully connected + linear output:
\begin{itemize}
	\item Fully Connected 1: 240, ReLU
\end{itemize} 
 
\underline{Transition Model} (ac-RKN,RKN):
bandwidth: 3, number of basis: 15
\begin{itemize}
    \item $\alpha(\vec{z}_t)$: No hidden layers - softmax output
\end{itemize}
\underline{Control Model} (ac-RKN):
3 fully connected + linear output
\begin{itemize}
	\item Fully Connected 1: 30, ReLU
	\item Fully Connected 2: 30, ReLU
	\item Fully Connected 3: 30, ReLU
\end{itemize} 

\textbf{Architecture For FFNN Baseline - Forward Dynamics}
3 fully connected + linear output
\begin{itemize}
	\item Fully Connected 1: 1000, ReLU
	\item Fully Connected 2: 1000, ReLU
	\item Fully Connected 3: 1000, ReLU
\end{itemize}

Dropout Regularization - 0.1147\\
Learning Rate - 8.39e-3\\
Optimizer Used: SGD Optimizer

\subsection*{Franka Emika Panda - Inverse Dynamics Learning}
\begin{table}[htbp]
\centering
\caption{Inverse Dynamics Learning Hyperparameters For Panda.}
\label{tab:panda-inverse-dynamics}
\resizebox{0.8\textwidth}{!}{
\begin{tabular}{|l|l|l|l|}
\hline
\textbf{Hyperparameter} & \textbf{ac-RKN} & \textbf{RKN (No Action Feedback)} & \textbf{LSTM} \\ 
\hline 
Learning Rate & 7.62e-3 & 3.5e-3 & 9.89e-3 \\ 
\hline
Latent Observation Dimension & 15 & 30 & 30 \\ 
\hline
Latent State Dimension & 30 & 60 & 60 \\ 
\hline
Regularization Factor ($\lambda$) & 0.158 & 0.179 & 0.196 \\
\hline
\end{tabular}}
\end{table}

\underline{Encoder} (ac-RKN,RKN,LSTM): 1 fully connected + linear output (elu + 1) 
\begin{itemize}
    \item Fully Connected 1: 120, ReLU
\end{itemize} 
\underline{Observation Decoder} (ac-RKN,RKN,LSTM): 1 fully connected + linear output:
\begin{itemize}
    \item Fully Connected 1: 240, ReLU
\end{itemize} 

\underline{Action Decoder} (ac-RKN,RKN,LSTM): 1 fully connected + linear output:
\begin{itemize}
    \item Fully Connected 1: 512, ReLU
\end{itemize} 
 
\underline{Transition Model} (ac-RKN,RKN):
bandwidth: 3, number of basis: 15
\begin{itemize}
    \item $\alpha(\vec{z}_t)$: No hidden layers - softmax output  
\end{itemize}

\underline{Control Model} (ac-RKN):
1 fully connected + linear output
\begin{itemize}
    \item Fully Connected 1: 45, ReLU
\end{itemize} 

\textbf{Architecture For FFNN Baseline - Inverse Dynamics}
3 fully connected + linear output
\begin{itemize}
    \item Fully Connected 1: 500, ReLU
    \item Fully Connected 2: 500, ReLU
    \item Fully Connected 3: 500, ReLU
\end{itemize}

Dropout Regularization - 0.563 \\
Learning Rate - 1.39e-2 \\
Optimizer Used: SGD Optimizer

\subsection*{Barrett WAM - Inverse Dynamics Learning}
\begin{table}[htbp]
\centering
\caption{Inverse Dynamics Learning Hyperparameters Barrett WAM.}
\label{tab:wam-inverse-dynamics}
\resizebox{0.8\textwidth}{!}{
\begin{tabular}{|l|l|l|l|}
\hline
\textbf{Hyperparameter} & \textbf{ac-RKN} & \textbf{RKN (No Action Feedback)} & \textbf{LSTM} \\ 
\hline 
Learning Rate & 7.7e-3 & 1.7e-3 & 9.33e-3 \\ 
\hline
Latent Observation Dimension & 15 & 30 & 45 \\ 
\hline
Latent State Dimension & 30 & 60 & 90 \\ 
\hline
Regularization Factor ($\lambda$) & 0.176 & 0 & 3.42e-3 \\
\hline
\end{tabular}}
\end{table}

\underline{Encoder} (ac-RKN,RKN,LSTM): 1 fully connected + linear output (elu + 1) 
\begin{itemize}
    \item Fully Connected 1: 120, ReLU
\end{itemize} 
\underline{Observation Decoder} (ac-RKN,RKN,LSTM): 1 fully connected + linear output:
\begin{itemize}
    \item Fully Connected 1: 240, ReLU
\end{itemize} 

\underline{Action Decoder} (ac-RKN): 2 fully connected + linear output:
\begin{itemize}
    \item Fully Connected 1: 256, ReLU
    \item Fully Connected 1: 256, ReLU
\end{itemize} 

\underline{Action Decoder} (RKN,LSTM): 1 fully connected + linear output:
\begin{itemize}
    \item Fully Connected 1: 512, ReLU
\end{itemize} 
 
\underline{Transition Model} (ac-RKN,RKN):
bandwidth: 3, number of basis: 15
\begin{itemize}
    \item $\alpha(\vec{z}_t)$: No hidden layers - softmax output
\end{itemize}

\underline{Control Model} (ac-RKN):
1 fully connected + linear output
\begin{itemize}
    \item Fully Connected 1: 45, ReLU
\end{itemize} 

\textbf{Architecture For FFNN Baseline}
3 fully connected + linear output
\begin{itemize}
    \item Fully Connected 1: 500, ReLU
    \item Fully Connected 2: 500, ReLU
    \item Fully Connected 3: 500, ReLU
\end{itemize}

Dropout Regularization - 0.563\\
Learning Rate - 1e-5\\
Optimizer Used: SGD Optimizer

\section{Hyperparameters: Chapter \ref{chap:Hipssm}}
\label{sec:hpyHip}
\subsection{Pneumatic Musculoskeletal Robot Arm}
\underline{\textbf{Recurrent Models}}
\begin{table}[htbp]
\centering
\caption{Forward Dynamics Learning Hyperparameters For Panda.}
\label{tab:panda}
\resizebox{0.8\textwidth}{!}{
\begin{tabular}{|l|l|l|l|l|}
\hline
\textbf{Hyperparameter} & \textbf{HiP-RSSM} & \textbf{RKN} & \textbf{LSTM} & \textbf{GRU} \\ 
\hline 
Learning Rate & 8e-4 & 8e-4 & 1e-3 & 1e-3 \\ 
\hline
Latent Observation Dimension & 15 & 15 & 15 & 15  \\ 
\hline
Latent State Dimension & 30 & 30 & 75 & 75  \\ 
\hline
Latent Task Dimension & 30 & - & - & -  \\
\hline
\end{tabular}}
\end{table}

\underline{Context Encoder} (HiP-RSSM): 1 fully connected + linear output  (elu + 1) 
\begin{itemize}
    \item Fully Connected 1: 240, ReLU
\end{itemize} 
\underline{Observation Encoder} (HiP-RSSM,RKN,LSTM,GRU): 1 fully connected + linear output  (elu + 1) 
\begin{itemize}
    \item Fully Connected 1: 120, ReLU
\end{itemize} 
\underline{Observation Decoder} (HiP-RSSM,RKN,LSTM): 1 fully connected + linear output:
\begin{itemize}
    \item Fully Connected 1: 120, ReLU
\end{itemize} 
 
\underline{Transition Model} (HiP-RSSM,RKN):
number of basis: 15 
\begin{itemize}
    \item $\alpha(\cvec{z}_t)$: No hidden layers - softmax output
\end{itemize}
\underline{Control Model} (HiP-RSSM,RKN):
3 fully connected + linear output
\begin{itemize}
    \item Fully Connected 1: 120, ReLU
    \item Fully Connected 2: 120, ReLU
    \item Fully Connected 3: 120, ReLU
\end{itemize} 

\subsection{Franka Robot Arm With Varying Loads}
\underline{\textbf{Recurrent Models}}
\begin{table}[htbp]
\centering
\caption{Forward Dynamics Learning Hyperparameters For Franka.}
\label{tab:franka}
\resizebox{0.8\textwidth}{!}{
\begin{tabular}{|l|l|l|l|l|}
\hline
\textbf{Hyperparameter} & \textbf{HiP-RSSM} & \textbf{RKN} & \textbf{LSTM} & \textbf{GRU} \\ 
\hline 
Learning Rate & 1e-3 & 1e-3 & 3e-3 & 3e-3 \\ 
\hline
Latent Observation Dimension & 15 & 15 & 15 & 15  \\ 
\hline
Latent State Dimension & 30 & 30 & 75 & 75  \\ 
\hline
Latent Task Dimension & 30 & - & - & -  \\
\hline
\end{tabular}}
\end{table}

\underline{Encoder} (HiP-RSSM,RKN,LSTM,GRU): 1 fully connected + linear output  (elu + 1) 
\begin{itemize}
    \item Fully Connected 1: 30, ReLU
\end{itemize} 
\underline{Observation Decoder} (HiP-RSSM,RKN,LSTM,GRU): 1 fully connected + linear output:
\begin{itemize}
    \item Fully Connected 1: 30, ReLU
\end{itemize} 
 
\underline{Transition Model} (HiP-RSSM,RKN):
number of basis: 32
\begin{itemize}
    \item $\alpha(\cvec{z}_t)$: No hidden layers - softmax output
\end{itemize}
\underline{Control Model} (HiP-RSSM, RKN):
1 fully connected + linear output
\begin{itemize}
    \item Fully Connected 1: 120, ReLU
\end{itemize} 

\subsection{Wheeled Robot Traversing Slopes Of Different Height}
\underline{\textbf{Recurrent Models}}
\begin{table}[htbp]
\centering
\caption{Forward Dynamics Learning Hyperparameters For Wheeled Robot.}
\label{tab:wheeled-robot}
\resizebox{0.8\textwidth}{!}{
\begin{tabular}{|l|l|l|l|l|}
\hline
\textbf{Hyperparameter} & \textbf{HiP-RSSM} & \textbf{RKN} & \textbf{LSTM} & \textbf{GRU} \\ 
\hline 
Learning Rate & 9e-4 & 9e-4 & 1e-2 & 1e-2\\ 
\hline
Latent Observation Dimension & 30 & 30 & 15 & 15 \\ 
\hline
Latent State Dimension & 60 & 60 & 75 & 75 \\ 
\hline
Latent Task Dimension & 60 & - & - & -  \\
\hline
\end{tabular}}
\end{table}

\underline{Encoder} (HiP-RSSM,RKN,LSTM,GRU): 1 fully connected + linear output  (elu + 1) 
\begin{itemize}
    \item Fully Connected 1: 120, ReLU
\end{itemize} 
\underline{Observation Decoder} (HiP-RSSM,RKN,LSTM,GRU): 1 fully connected + linear output:
\begin{itemize}
    \item Fully Connected 1: 240, ReLU
\end{itemize} 
 
\underline{Transition Model} (HiP-RSSM,RKN):
number of basis: 15
\begin{itemize}
    \item $\alpha(\cvec{z}_t)$: No hidden layers - softmax output
\end{itemize}
\underline{Control Model} (HiP-RSSM, RKN):
3 fully connected + linear output
\begin{itemize}
    \item Fully Connected 1: 120, ReLU
\end{itemize}

\section{Hyperparameters: Chapter \ref{chap:mts3}}
\label{sec:hpyMts3}

\paragraph{Compute Resources}For training MTS3, LSTM, GRU and Transformer models we used compute nodes with (i) Nvidia 3090 and (ii) Nvidia 2080 RTX GPUs.
For training more computationally expensive locally linear models like RKN, HiP-RSSM we used
compute nodes with NVIDIA A100-40 GPUs.

\paragraph{Hyperparameters} Hyperparameters were selected via grid search. In general, the performance of MTS3 is not very sensitive to hyperparameters. Among all the baselines, Transformer models were most sensitive to hyperparameters (see Appendix E.5 for details of Transformer architecture). 

\paragraph{Discretization Step:} For MTS3, the discretization step for the slow time scale SSM as discussed in Section 3.1 for all datasets was fixed as $\cmat H \cdot \Delta t = 0.3$ seconds. In our experiments, we found that discretization values between $0.2 \leq\cmat H \cdot \Delta t \leq 0.5$ seconds give similar performance.

\paragraph{Rule Of thumb for choosing discretization step in MTS3:} For any N-level MTS3 as defined in Section 3.4, we recommend searching for discretization factor $H_i$ as a hyperparameter. However, as a general rule of thumb, it can be chosen as $H_i=(\sqrt[N]{T})^i$, where $T$ is the maximum prediction horizon required / episode length. This ensures that very long recurrences are divided between smaller equal-length task-reconfigurable local SSM windows (of length $\sqrt[N]{T}$) spread across several hierarchies.

\paragraph{Encoder Decoder Architecture:} For all recurrent models (MTS3, HiP-RSSM, RKN, LSTM and GRU) we use a similar encoder-decoder architecture across datasets. Small variations from these encoder-decoder architecture hyperparameters can still lead to similar prediction performance as reported in the paper.\\

\underline{Observation Set Encoder} (MTS3): 1 fully connected + linear output:
\begin{itemize}
	\item Fully Connected 1: 240, ReLU
\end{itemize} 

\underline{Action Set Encoder} (MTS3): 1 fully connected + linear output:
\begin{itemize}
	\item Fully Connected 1: 240, ReLU
\end{itemize} 

\underline{Observation Encoder} (MTS3, HiP-RSSM, RKN, LSTM, GRU): 1 fully connected + linear output:
\begin{itemize}
	\item Fully Connected 1: 120, ReLU
\end{itemize} 

\underline{Observation Decoder} (MTS3, HiP-RSSM, RKN, LSTM, GRU): 1 fully connected + linear output:
\begin{itemize}
	\item Fully Connected 1: 120, ReLU
\end{itemize} 

\underline{Control Model (Primitive Action Encoder)} (MTS3, HiP-RSSM, RKN):
1 fully connected + linear output:
\begin{itemize}
	\item Fully Connected 1: 120, ReLU
\end{itemize} 

The rest of the hyperparameters are described below:

\subsection{D4RL Datasets}

\subsubsection{Half Cheetah}
\underline{\textbf{Recurrent Models}}
\begin{table}[htbp]
  \centering
  \resizebox{0.8\textwidth}{!}{
  \begin{tabular}{|c|c|c|c|c|c|}
    \hline
    \textbf{Hyperparameters} & \textbf{MTS3} & \textbf{HiP-RSSM} & \textbf{RKN} & \textbf{LSTM} & \textbf{GRU} \\
    \hline
    Learning Rate & 3e-3 & 1e-3 & 1e-3 &1e-3 &1e-3 \\ \hline
    Latent Observation Dimension &15 & 15 & 15 & 15 & 15  \\ \hline
Observation Set Latent Dimension (sts-SSM) & 15 & - & - & - & - \\ \hline
Latent State Dimension &30 & 30 & 30 & 45 & 45  \\ \hline
Latent Task Dimension &30 & 30 & - & - & -  \\ \hline
Latent Abstract Action Dimension (sts-SSM) & 30 & - & - & - & - \\ 
    \hline
  \end{tabular}}
\end{table}
 
\underline{Transition Model} (HiP-RSSM, RKN):
number of basis: 32 

\begin{itemize}
    \item $\alpha(\cvec{z}_t)$: No hidden layers - softmax output
\end{itemize}

\underline{\textbf{Autoregressive Transformer Baseline}}\\
Learning Rate: 1e-5\\
Optimizer Used: Adam Optimizer \\
Embedding size: 96 \\
Number of Decoder Layers: 4 \\
Number Of Attention Heads: 4

\underline{\textbf{Multistep Transformer Baseline}}\\
Learning Rate: 1e-5\\
Optimizer Used: Adam Optimizer \\
Embedding size: 128 \\
Number Of Encoder Layers: 2\\
Number of Decoder Layers: 1 \\
Number Of Attention Heads: 4

\subsubsection{Franka Kitchen}
\underline{\textbf{Recurrent Models}}
\begin{table}[htbp]
  \centering
  \resizebox{0.8\textwidth}{!}{
  \begin{tabular}{|c|c|c|c|c|c|}
    \hline
    \textbf{Hyperparameters} & \textbf{MTS3} & \textbf{HiP-RSSM} & \textbf{RKN} & \textbf{LSTM} & \textbf{GRU} \\
    \hline
    Learning Rate & 3e-3 & 9e-4 & 9e-4 &1e-3 &1e-3 \\ \hline
    Latent Observation Dimension &30 & 30 & 30 & 30 & 30  \\ \hline
Observation Set Latent Dimension (sts-SSM) & 30 & - & - & - & - \\ \hline
Latent State Dimension &60 & 60 & 60 & 90 & 90  \\ \hline
Latent Task Dimension &60 & 60 & - & - & -  \\ \hline
Latent Abstract Action Dimension (sts-SSM) & 60 & - & - & - & - \\
    \hline
  \end{tabular}}
\end{table}
 
\underline{Transition Model} (HiP-RSSM, RKN):
number of basis: 15 

\begin{itemize}
    \item $\alpha(\cvec{z}_t)$: No hidden layers - softmax output
\end{itemize}

\underline{\textbf{Autoregressive Transformer Baseline}}\\
Learning Rate: 5e-5\\
Optimizer Used: Adam Optimizer \\
Embedding size: 64 \\
Number of Decoder Layers: 4 \\
Number Of Attention Heads: 4


\underline{\textbf{Multistep Transformer Baseline}}\\
Learning Rate: 1e-5\\
Optimizer Used: Adam Optimizer \\
Embedding size: 64 \\
Number Of Encoder Layers: 2\\
Number of Decoder Layers: 1 \\
Number Of Attention Heads: 4


\subsubsection{Maze 2D}
\underline{\textbf{Recurrent Models}}
\begin{table}[htbp]
  \centering
  \resizebox{0.8\textwidth}{!}{
  \begin{tabular}{|c|c|c|c|c|c|}
    \hline
    \textbf{Hyperparameters} & \textbf{MTS3} & \textbf{HiP-RSSM} & \textbf{RKN} & \textbf{LSTM} & \textbf{GRU} \\
    \hline
    Learning Rate & 3e-3 & 9e-4 & 9e-4 &1e-3 &1e-3 \\ \hline
    Latent Observation Dimension &30 & 30 & 30 & 30 & 30  \\ \hline
Observation Set Latent Dimension (sts-SSM) & 30 & - & - & - & - \\ \hline
Latent State Dimension &60 & 60 & 60 & 90 & 90  \\ \hline
Latent Task Dimension &60 & 60 & - & - & -  \\ \hline
Latent Abstract Action Dimension (sts-SSM) & 60 & - & - & - & - \\
    \hline
  \end{tabular}}
\end{table}
 
\underline{Transition Model} (HiP-RSSM, RKN):
number of basis: 15 

\begin{itemize}
    \item $\alpha(\cvec{z}_t)$: No hidden layers - softmax output
\end{itemize}

\underline{\textbf{Autoregressive Transformer Baseline}}\\
Learning Rate: 5e-5\\
Optimizer Used: Adam Optimizer \\
Embedding size: 96 \\
Number of Decoder Layers: 4 \\
Number Of Attention Heads: 4


\underline{\textbf{Multistep Transformer Baseline}}\\
Learning Rate: 1e-5\\
Optimizer Used: Adam Optimizer \\
Embedding size: 128 \\
Number Of Encoder Layers: 2\\
Number of Decoder Layers: 1 \\
Number Of Attention Heads: 4


\subsection{Franka Robot Arm With Varying Loads}
\underline{\textbf{Recurrent Models}}
\begin{table}[htbp]
  \centering
  \resizebox{0.8\textwidth}{!}{
  \begin{tabular}{|c|c|c|c|c|c|}
    \hline
    \textbf{Hyperparameters} & \textbf{MTS3} & \textbf{HiP-RSSM} & \textbf{RKN} & \textbf{LSTM} & \textbf{GRU} \\
    \hline
    Learning Rate & 3e-3 & 9e-4 & 9e-4 &3e-3 &3e-3 \\ \hline
    Latent Observation Dimension &15 & 15 & 15 & 15 & 15  \\ \hline
Observation Set Latent Dimension (sts-SSM) & 15 & - & - & - & - \\ \hline
Latent State Dimension &30 & 30 & 30 & 45 & 45  \\ \hline
Latent Task Dimension &30 & 30 & - & - & -  \\ \hline
Latent Abstract Action Dimension (sts-SSM) & 30 & - & - & - & - \\ 
    \hline
  \end{tabular}}
\end{table}

\underline{Transition Model} (HiP-RSSM,RKN):
number of basis: 32
\begin{itemize}
    \item $\alpha(\cvec{z}_t)$: No hidden layers - softmax output
\end{itemize}

\underline{\textbf{Autoregressive Transformer Baseline}}\\
Learning Rate: 5e-5\\
Optimizer Used: Adam Optimizer \\
Embedding size: 64 \\
Number of Decoder Layers: 4 \\
Number Of Attention Heads: 4


\underline{\textbf{Multistep Transformer Baseline}}\\
Learning Rate: 2e-5\\
Optimizer Used: Adam Optimizer \\
Embedding size: 64 \\
Number Of Encoder Layers: 2\\
Number of Decoder Layers: 1 \\
Number Of Attention Heads: 4


\subsection{Hydraulic Excavator}
\begin{table}[H]
  \centering
  \resizebox{0.8\textwidth}{!}{
  \begin{tabular}{|c|c|c|c|c|c|}
    \hline
    \textbf{Hyperparameters} & \textbf{MTS3} & \textbf{HiP-RSSM} & \textbf{RKN} & \textbf{LSTM} & \textbf{GRU} \\
    \hline
    Learning Rate & 3e-3 & 8e-4 & 8e-4 &1e-3 &1e-3 \\ \hline
    Latent Observation Dimension &15 & 15 & 15 & 15 & 15  \\ \hline
Observation Set Latent Dimension (sts-SSM) & 15 & - & - & - & - \\ \hline
Latent State Dimension &30 & 30 & 30 & 45 & 45  \\ \hline
Latent Task Dimension &30 & 30 & - & - & -  \\ \hline
Latent Abstract Action Dimension (sts-SSM) & 30 & - & - & - & - \\ 
    \hline
  \end{tabular}}
\end{table}

\underline{Transition Model} (HiP-RSSM,RKN):
number of basis: 15
\begin{itemize}
    \item coefficient net $\alpha(\cvec{z}_t)$: No hidden layers - softmax output
\end{itemize}

\underline{\textbf{Autoregressive Transformer Baseline}}\\
Learning Rate: 1e-5\\
Optimizer Used: Adam Optimizer \\
Embedding size: 96 \\
Number of Decoder Layers: 4 \\
Number Of Attention Heads: 4


\underline{\textbf{Multistep Transformer Baseline}}\\
Learning Rate: 5e-5\\
Optimizer Used: Adam Optimizer \\
Embedding size: 64 \\
Number Of Encoder Layers: 2\\
Number of Decoder Layers: 1 \\
Number Of Attention Heads: 4


\subsection{Wheeled Robot Traversing Uneven Terrain}
\begin{table}[htbp]
  \centering
  \resizebox{0.8\textwidth}{!}{
  \begin{tabular}{|c|c|c|c|c|c|}
    \hline
    \textbf{Hyperparameters} & \textbf{MTS3} & \textbf{HiP-RSSM} & \textbf{RKN} & \textbf{LSTM} & \textbf{GRU} \\
    \hline
    Learning Rate & 3e-3 & 8e-4 & 8e-4 &1e-3 &1e-3 \\ \hline
    Latent Observation Dimension &30 & 30 & 30 & 30 & 30  \\ \hline
Observation Set Latent Dimension (sts-SSM) & 30 & - & - & - & - \\ \hline
Latent State Dimension &60 & 60 & 60 & 90 & 90  \\ \hline
Latent Task Dimension &60 & 60 & - & - & -  \\ \hline
Latent Abstract Action Dimension (sts-SSM) & 60 & - & - & - & - \\ 
    \hline
  \end{tabular}}
\end{table}
 
\underline{Transition Model} (HiP-RSSM,RKN):
number of basis: 15
\begin{itemize}
    \item coefficient net $\alpha(\cvec{z}_t)$: No hidden layers - softmax output
\end{itemize}

\underline{\textbf{Autoregressive Transformer Baseline}}\\
Learning Rate: 5e-5\\
Optimizer Used: Adam Optimizer \\
Embedding size: 128 \\
Number of Decoder Layers: 4 \\
Number Of Attention Heads: 4


\underline{\textbf{Multistep Transformer Baseline}}\\
Learning Rate: 5e-5\\
Optimizer Used: Adam Optimizer \\
Embedding size: 64 \\
Number Of Encoder Layers: 4\\
Number of Decoder Layers: 2 \\
Number Of Attention Heads: 4


\subsection{Transformer Architecture Details}
For the AR-Transformer Baseline, we use a GPT-like autoregressive version of transformers except that for the autoregressive input we also concatenate the actions to make action conditional predictions.

For Multi-Transformer we use the same direct multistep prediction and loss as in recent Transformer time-series forecasting literature~\cite{zhou2021informer,liu2022nonstationary,nie2023a,zeng2022transformers}. A description of the action conditional direct multi-step version of the transformer is given in Algorithm \ref{algo:trans}.

\begin{algorithm}
\caption{MultiStep Transformer}
\textbf{Require:} Input past observations $\mathbf{o_{inp}} \in \mathbb{R}^{S \times C}$; Input Past Actions $\mathbf{a_{inp}} \in \mathbb{R}^{S \times A}$;  Future Actions $\mathbf{a_{pred}} \in \mathbb{R}^{O \times A}$;Input Length $S$; Predict length $O$; Observation Dimension $C$; Action Dimension $A$; Feature dimension $d_k$; Encoder layers number $N$; Decoder layers number $M$. \\
\vspace{0.5cm}
1: $ \mathbf{o_{inp}} \in \mathbb{R}^{S \times C}, \cmat a_{inp} \in \mathbb{R}^{S \times A} , \cmat a_{pred} \in \mathbb{R}^{O \times A} $   \\
2: $\cmat X_{inp} = \mathrm{ConCatFeatureWise
}\left(\cmat o_{inp}, \cmat a_{inp}\right)$ \hspace{3.4cm} $\triangleright \cmat X_{inp} \in \mathbb{R}^{S \times (C+A)}$ \\
3: $\cmat X_{pred} = \mathrm{ConCatFeatureWise
}\left(\operatorname{Zeros}(O, C), \cmat a_{pred}\right)$ \hspace{2cm} $\triangleright \cmat X_{pred} \in \mathbb{R}^{O \times (C+A)}$\\
4: $\mathbf{X}_{\mathrm{enc}}, \mathbf{X}_{\mathrm{dec}}=\mathbf{X}_{\mathrm{inp}}, \operatorname{ConCat}\left(\cmat X_{inp}, \cmat X_{pred}\right)  \quad \quad \triangleright \mathbf{X}_{\mathrm{enc}} \in \mathbb{R}^{S \times (C+A)}, \mathbf{X}_{\mathrm{dec}} \in \mathbb{R}^{\left(S+O\right) \times (C+A)}$ \\
5: $\cmat X_{\mathrm{enc}}^{0}=\operatorname{Embed}\left(\mathrm{X}_{\mathrm{enc}}\right)$ 
\hspace{7cm} $\triangleright \mathbf{X}_{\mathrm{enc}}^{0} \in \mathbb{R}^{S \times d_k}$ \\
6: \For{$l$ in $\{1, \cdots, N\}$}{
    7: $\quad \cmat X_{\mathrm{enc}}^{l-1}=\operatorname{LayerNorm}\left(\cmat X_{\mathrm{enc}}^{l-1}+\operatorname{Attn}\left(\cmat X_{\mathrm{enc}}^{l-1}\right)\right)$ 
    \hspace{3.1cm} $\triangleright \cmat X_{\mathrm{enc}}^{l-1} \in \mathbb{R}^{S \times d_k}$ \\
    8: $\quad \cmat X_{\text {enc }}^{l }=\operatorname{LayerNorm}\left(\cmat X_{\text {enc }}^{l-1 }+\operatorname{FFN}\left(\cmat X_{\text {enc }}^{l-1 }\right)\right)$ 
    \hspace{3cm} $ \triangleright \cmat X_{\mathrm{enc}}^{l } \in \mathbb{R}^{S \times d_k}$
}
9: $\cmat X_{\mathrm{dec}}^{0}=\operatorname{Embed}\left(\cmat X_{\mathrm{dec}}\right)$
\hspace{3cm} $\triangleright \cmat X_{\mathrm{dec}}^{0} \in \mathbb{R}^{\left(S+O\right) \times d_k}$\\
10: \For{\textbf{for} $l$ in $\{1, \cdots, M\}$}{
    11: $\quad \cmat X_{\operatorname{dec}}^{l-1}=\operatorname{LayerNorm}\left(\cmat X_{\operatorname{dec}}^{l-1}+\operatorname{Attn}\left(\cmat X_{\operatorname{dec}}^{l-1 }\right)\right)$ 
    \hspace{2cm} $\triangleright$ Decoder \\
    12: $\quad \cmat X_{\mathrm{dec}}^{l-1 }=\operatorname{LayerNorm}\left(\cmat X_{\mathrm{dec}}^{l-1 }+\operatorname{Attn}\left(\cmat X_{\mathrm{dec}}^{l-1 \prime}, \cmat X_{\mathrm{enc}}^{N }\right)\right)$ 
    $\triangleright \cmat X_{\mathrm{dec}}^{l-1 } \in \mathbb{R}^{\left(S+O\right) \times d_k}$ \\
    13:$\quad \cmat X_{\mathrm{dec}}^{l }=\operatorname{LayerNorm}\left(\cmat X_{\mathrm{dec}}^{l-1 }+\operatorname{FFN}\left(\cmat X_{\mathrm{dec}}^{l-1 }\right)\right)$ 
    \hspace{1cm} $\triangleright \cmat X_{\mathrm{dec}}^{l } \in \mathbb{R}^{\left(S+O\right) \times d_k}$
}
14: $\mathbf{y}=\operatorname{MLP}\left(\cmat X_{\mathrm{dec}}^{M}\right)$
\hspace{0.5cm} $\triangleright \mathbf{y} \in \mathbb{R}^{(S+O) \times C}$ \\
15: Return y \hspace{7cm}$\triangleright$ Return the prediction results
\label{algo:trans}
\end{algorithm}

\chapter{List Of Publications}
\includepdf[pages=-]{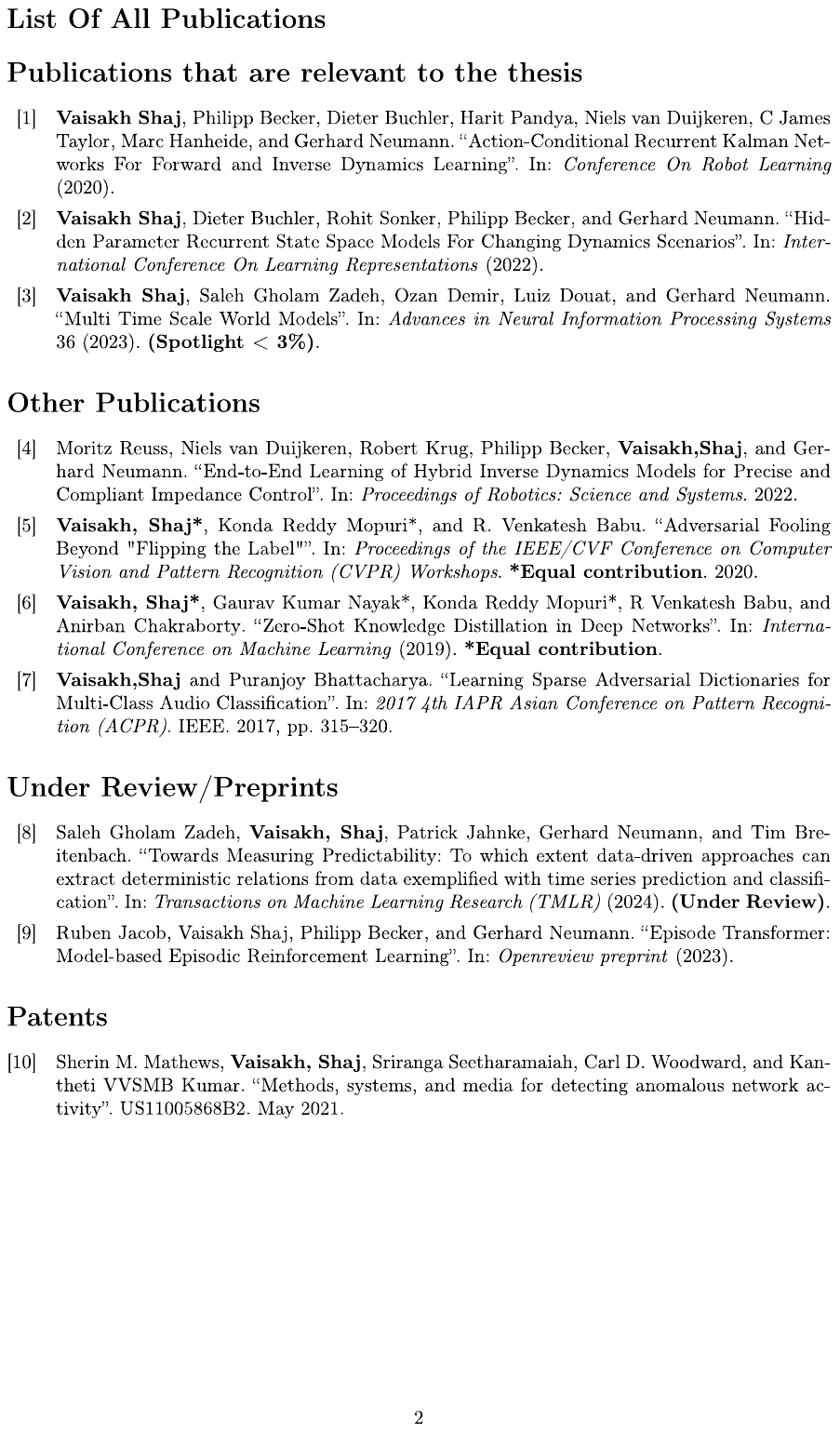}  

\end{document}